# Contrastive Topographic Models: Energy-based density models applied to the understanding of sensory coding and cortical topography

Simon Kayode Osindero

Gatsby Computational Neuroscience Unit
University College London

A thesis submitted to the University of London in the
Faculty of Science for the degree of Doctor of Philosophy

February 2004

# Abstract

We address the problem of building theoretical models that help elucidate the function of the visual brain at computational/algorithmic and structural/mechanistic levels. We seek to understand how the receptive fields and topographic maps found in visual cortical areas relate to underlying computational desiderata. We view the development of sensory systems from the popular perspective of probability density estimation; this is motivated by the notion that an effective internal representational scheme is likely to reflect the statistical structure of the environment in which an organism lives. We apply biologically based constraints on elements of the model.

The thesis begins by surveying the relevant literature from the fields of neurobiology, theoretical neuroscience and machine learning. After this review we present our main theoretical and algorithmic developments: we propose a class of probabilistic models, which we refer to as 'energy-based models', and show equivalences between this framework and various other types of probabilistic model such as Markov random fields and factor graphs; we also develop and discuss approximate algorithms for performing maximum likelihood learning and inference in our energy based models. The rest of the thesis is then concerned with exploring specific instantiations of such models. By performing constrained optimisation of model parameters to maximise the likelihood of appropriate, naturalistic data-sets we are able to qualitatively reproduce many of the receptive field and map properties found *in vivo*, whilst simultaneously learning about statistical regularities in the data.

# Acknowledgements

I would like to take this opportunity to thank the people who have made my studies at UCL so rewarding.

In particular I am extremely grateful my advisors, Peter Dayan and Geoff Hinton for providing peerless guidance and inspiration. I am also indebted to Max Welling, Yee Whye Teh and the members of the Gatsby Unit for wonderful interactions, discussions and collaborations.

Lastly, but most importantly, a heartfelt thanks goes to my family and friends for their love, support and encouragement.

This work was generously supported by the Wellcome Trust and the Gatsby Charitable Foundation.

# Contents

# List of Figures

# List of Tables

# List Of Abbreviations & Acronyms

| | | |
|---|---|---|
| BCM | — | Bienenstock, Cooper & Munroe |
| BPTT | — | Back-Propagation Through Time |
| CD | — | Contrastive Divergence |
| CEBM | — | Conditional Energy Based Model |
| DoG | — | Difference of Gaussians |
| EBM | — | Energy Based Model |
| EM | — | Expectation Maximisation |
| GSM | — | Gaussian Scale Mixture |
| HMC | — | Hybrid Monte Carlo |
| ICA | — | Independent Components Analysis |
| IWF | — | Iterative Wiener Filter |
| KL | — | Kullback-Liebler |
| LGN | — | Lateral Geniculate Nucleus |
| MAP | — | Maximum a Posteriori |
| MCMC | — | Markov Chain Monte Carlo |
| MEM | — | Maximum Entropy Model |
| MRF | — | Markov Random Field |
| OD | — | Ocular Dominance |
| PCA | — | Principal Components Analysis |
| PoT | — | Product of Student-t |
| RF | — | Receptive Field |
| SOM | — | Self-Organising Map |
| ZCA | — | Zero-phase Components Analysis |

# Chapter 1

# Introduction

## 1.1 Computational neuroscience & machine learning

The brain is an exquisitely complicated and beautiful piece of biological machinery, and the enigma of its *modus operandi* has intrigued scientists and philosophers for millennia. Whilst we still cannot claim anything close to a full understanding of the mystery, parts of the puzzle are starting to fall into place and great progress has been made in neuroscience, particularly over the latter half of the last century. Part of the reason for this has been the appreciation that the brain is essentially a computational device, and can therefore be approached from a computational and information theoretic perspective as well as from a biophysical one — thus leading to the field of computational neuroscience.

A particularly important set of issues for computational neuroscience centres on the nature and development of cortical representations. What do these representations say about stimuli, and how do they say this? What computational operations do they facilitate? How does their development and maintenance depend on genetic and environmental influences ('nature' *vs.* 'nurture')? How are the units of representation arranged physically and spatially within the cortex, and what role might this organisation serve? In this thesis, we choose to focus on the areas which inform answers to these problems, with a particular emphasis on the visual system.

Our approach to this task draws heavily on the field of machine learning. Specifically, we *begin* by developing new unsupervised learning paradigms and algorithms drawing inspiration from neurobiology, and then we seek to apply these methods to elucidate aspects of neuroscience.

## 1.2   Sensory processing & the visual system

The study of sensory processing, and in particular the processing of visual information, is one of the areas in which real progress, in terms of improving our understanding, seems tantalisingly close. Although much remains to be discovered and clarified, there is a wealth of experimental data available about the structure and mechanisms within the visual system — particularly at the earlier stages.

We concentrate on several facets of this data. There is abundant data on the response properties of cells within the early visual system in terms of the stimuli and conditions that best excite them. There is also a large amount of evidence regarding the spatial arrangement of response properties within different areas of cortex. By making additional assumptions based from an understanding of information processing, computation and learning in general, and integrating our knowledge of the developmental processes involved, we can use these lines of evidence to help us make hypotheses which address important questions such as: "What computational function might be instantiated by the underlying neural circuitry?"

## 1.3   Levels of modelling

As with many disciplines, there are multiple routes by which 'modelling' in neuroscience can be approached and there are also many possible taxonomies for classifying different levels of study. Borrowing from and integrating the methodological classifications put forward by Marr [1982] and also Dayan and Abbott [2001] we consider the following sub-divisions: (i) Interpretive; (ii) Descriptive/Phenomenological; (iii) Algorithmic and; (iv) Mechanistic/Implementational.

At the most panoramic level, operationally speaking, is the interpretive approach. This encompasses attempts to formulate an understanding of what a system does or achieves conceptually, and *why* this function might be important, useful or valuable. For example, viewed at this level we might wish to describe the retina as (amongst other things) a spatio-temporal decorrelation device that senses then pre-processes visual information in order to maximise the efficiency of information transfer under the bandwidth constraints of the optic nerve.

The descriptive/phenomenological level would typically be a parsimonious mathematical characterisation of a set of observations about a system's behaviour; perhaps an abstract characterisation of the functional operation that it performs. For instance, theta neuron description of type I/II pyramidal cells [Gutkin and Ermentrout, 1998].

Closely allied to the descriptive level in spirit, but somewhat more specific in terms of information manipulation, is what is termed the algorithmic level (which itself could be easily subdivided). Algorithmic level descriptions entail a complete account of the steps required to perform a particular computational function or sub-function that a system is hypothesised to carry out. For instance, the perceived motion of a noisy edge is well described by an algorithm implementing approximate Bayesian inference [Weiss et al., 2002]. Note that there may be very many possible algorithms for achieving the same computational goal.

Finally, at the mechanistic level we have considerations about the biological/biophysical implementation of the previous schemes. At this level, an account of action potential generation might consider the actual ion channels involved that cause the membrane conductance changes that lead to a spike.

Obviously these classes are neither exhaustive nor mutually exclusive, and to a certain extent they are nested and recursive — for instance the components of a mechanistic approach are often themselves phenomenological descriptions of underlying processes. Also there are other important distinctions that one could make such as the overall degree of abstraction and the scale of focus. These issues aside, such schemes can provide a useful and clarifying structure when approaching complicated systems.

### 1.3.1 Density estimation approaches to representational learning

At an interpretive level, it has been postulated that a 'goal' of visual development is to adapt neural connectivity in order to facilitate the formation of 'good' or 'useful' internal representations; this is a view that we largely share. An appropriate measure of 'goodness' might be how efficiently a representational scheme captures and utilises the statistical structure within the ensemble of inputs it is required to represent; a 'good' representational scheme in some way embodies an accurate internal probabilistic model of the external world. Whilst admittedly, this relationship between structured density modelling and learning 'good' representations is not entirely perspicuous, it does provide a useful starting point when trying to address problems in sensory organisation and neural computation.

Causal generative models are one type of probabilistic approach that have recently been successfully applied in the context of representational learning. Mathematically, these models aim to describe an observed data distribution as having arisen from the interaction of a set of internal 'causes'; the distribution of these causes within the model is given in a parameter specified way. Learning consists of adjusting the parameters of the model such that the generated data matches the observed data as well as possible. Representation is implemented by taking the statistical inverse of the generative model — that is to say inferring the causes that, consistent with the model generative mechanisms, likely gave rise to a particular input. These inferred causes are the semantic elements of the representation and if the generative model accurately describes the distribution of inputs to be rep-resented, and the causes are suitably structured[1], then the representation is some sense meaningful.

## 1.4 Thesis focus and goals

A main goal of this thesis is to develop new machine learning models and algorithms for unsupervised density estimation, and then to use these novel computational methods to broaden our set of tools for understanding the development of receptive fields and cortical topography. Specifically, rather than use causal generative models we propose a paradigm of unsupervised representational learning in an 'energy-based' framework to help us understand the receptive fields and topographic maps in visual cortex. This framework assigns energies to states of a network, and then uses those energies to define a probability distribution — as discussed in Chapter 4.

A key objective is to produce models that can take naturalistic inputs and, through well-founded unsupervised learning algorithms, form statistically efficient representations whilst at the same time developing structure that is biologically informative with respect to receptive fields and topographic maps. This type of combined model has received relatively little attention so far, particularly in the context of high-dimensional inputs. A main reason for this has been the apparent intractability of such models.

In terms of the aforementioned levels of modelling, we hope to span a range that primarily covers interpretational, descriptive and algorithmic levels whilst being informed, but not necessarily constrained, by mechanistic considerations. We believe that such multi-level models are a necessary step on the long road to gaining a better understanding of neural computation in sensory systems and our aim is to build on existing approaches both in terms of model sophistication and in terms of biological realism.

There will be many aspects of the data that we will not attempt (nor be able) to capture or explain; as is the case with most models. Many of the finer biological details, such as cortical micro-circuitry or even observations such as Dale's law will be neglected or abstracted. Also, we remain aware of the distinction between an interpretation of biological outcomes in a particular model setting and the 'actual' developmental processes and algorithms. Nonetheless, we suggest that this work makes a useful contribution by exploring novel computational approaches to understanding sensory processing and topographic map development, and because it expands the repertoire of techniques that can be brought to bear on such problems.

In addition, this work also makes a useful contribution to the general field of unsupervised learning, with implications and potential utility outside of the field of computational neuroscience. Some of the tools to whose development we have contributed hold great promise, and may have fruitful application in other machine learning and data modelling domains such as pattern classification, bioinformatics or natural language processing.

---

[1] As opposed to, for instance, a trivial look-up table approach with one cause per datum.

## 1.5   Chapter overviews

A summary of the remaining thesis chapters follows:

### Chapter 2

We begin by reviewing sections of the biological literature that are most relevant to the study of receptive field and topographic map development in visual cortex, focussing on properties of primary visual cortex. The wealth of observations is vast, and so our coverage of many issues is necessarily brief and serves to provide a background for some of the interesting aspects which subsequent chapters will try to develop.

### Chapter 3

We then move on to consider existing theoretical approaches to the problems of learning representations and modelling receptive field and topographic map formation. Again, the full body of work is extremely large and our treatment inevitably covers some topics very briefly. We divide our focus into several sections, considering machine learning and statistical approaches alongside Hebbian learning methods and more descriptive procedures. A broad spectrum of approaches is discussed and comprehensively referenced. The chapter concludes with a discussion summarising present methods with respect to their strengths, weaknesses, and lacunae.

### Chapter 4

In this chapter we lay the foundations of our theoretical approach. We set up the formalism of what we term 'energy-based Models' (EBM's); these use a parameterised energy function to define probability distributions. The general concept is not new, and indeed close links can be made with some existing approaches — in particular undirected graphical models and factor graphs. However, our integrated approach endows our framework with additional levels of interpretation that are different from existing methods. One of the ways in which this is done is through the concept of 'deterministic latent variables', in addition to the more familiar stochastic ones.

This can be thought of as a process in which the factors in a factor graphical representation are endowed with representative power, and may be constructed recursively from other factors. This can be useful because, if the model is being learned from data and uses an appropriate parametric family, these functions and the values induced upon them by data can take useful interpretations and thereby provide a statistically motivated representational scheme.

Additionally, we utilise and develop upon the contrastive divergence learning algorithm recently proposed by Hinton [2000, 2002] by augmenting it with more advanced Monte Carlo techniques such as Hybrid Monte Carlo.

Finally, our paradigm allows for potentially useful re-interpretations of extant models. In particular we discuss ICA and infomax models as they relate to our energy-based approach, and suggest a novel generalisation which we later explore in Chapter 6.

### Chapter 5

In this chapter we present a Boltzmann machine based account for the development of retinotopy and ocular dominance. By suitably constraining the weight parameters and using naturalistic stereo input patterns we are able to show refinement of retinotopy and the emergence of ocular preferring regions as a consequence of performing unsupervised learning of a density model.

Various manipulations to the model produce changes which qualitatively replicate the changes observed under comparable manipulations *in vivo*.

In addition to learning the feedforward weights of our model we are also able, at least to an extent, to learn patterns of lateral connectivity.

### Chapter 6

In this chapter we develop a model that overcomes some of the difficulties we found inherent in the Boltzmann machine; namely the awkwardness with which continuous valued quantities are represented. We put forward an energy-based model which we name the Product of t's (PoT). This can be considered to be a product of experts in which each expert is a Student's t-distribution; alternatively, it can be viewed as a novel extension of Student-t prior ICA.

We present an efficient Gibbs sampling scheme for the PoT model and use this, in combination with contrastive divergence, to train density models of digitised patches of natural scenes. The 'feature detectors' that develop bear a strong relationship to simple cells found in V1, as one obtains through ICA. However, our approach is significantly different in that we are able to deal trivially with learning overcomplete representations and, unlike overcomplete ICA, do not require any iterative sampling to produce a representation.

By extending the model hierarchically we are able to learn feature detectors that have properties qualitatively similar to complex cells. By constraining this hierarchical model we are able to learn representations that also display a topographic mapping of response properties reminiscent of the map structures observed in V1.

This chapter also explicitly relates our approach to the topographic ICA framework [Hyvarinen et al., 2001], Gaussian scale mixtures [Wainwright and Simoncelli, 2000], and to ideas about divisive normalisation in visual processing [Simoncelli and Schwartz, 1999].

### Chapter 7

In this chapter we explore several applications of the energy-based models that we have proposed. Primarily, we explore the use of the PoT model for natural images as a prior in a MAP denoising algorithm, and show our results to be superior to simple wiener filtering and competitive with other more advanced methods. Additionally, we demonstrate the potential of our framework for general unsupervised feature extraction by applying EBM's to a dataset of handwritten digits and to a dataset of human faces.

### Chapter 8

This chapter is the conclusion of thesis in which we re-iterate our main contributions, discuss general extensions and outstanding issues, and present some final ideas for future work.

# Chapter 2

# Review of Biological Literature

## 2.1  Introduction

In this chapter we review some of the neurobiological literature that is salient with respect to this thesis. The considerable body of knowledge that has been acquired about the visual system comes from many different sources and the arsenal of techniques available to the modern neuroscientist makes it possible to study neural processes over a wide range of spatio-temporal scales and resolutions. Even though there is still a great deal of technical progress to be made in terms of recording from and characterising the behaviour of neural systems, the conjunction of current methods gives us an impressively detailed view of how the brain operates. In many ways, the data already outstrips the predictive capabilities of existing models.

Notable experimental techniques include: anatomical, histological and molecular biological methods for determining gross and fine structure; *in vitro* electrophysiological studies of cortical slices and neural cultures; *in vivo* electrophysiological recordings from anaesthetised and awake animals; *in vivo* optical imaging; and noninvasive methods such as (f)MRI, PET, MEG and ERP measurements.

The chapter begins with a general overview of visual pathways and circuits, a discussion of receptive fields and tuning functions in general, and a discussion of neural representations and population coding. The remainder of the chapter is then split to cover the three linked themes: (i) receptive fields in primary visual cortex (section 2.5); (ii) the topographic organisation of receptive field properties in primary visual cortex (section 2.6); and (iii) the development of structure in primary visual cortex (section 2.7).

The coverage in this chapter is necessarily brief, but pointers to more comprehensive sources are given where appropriate.

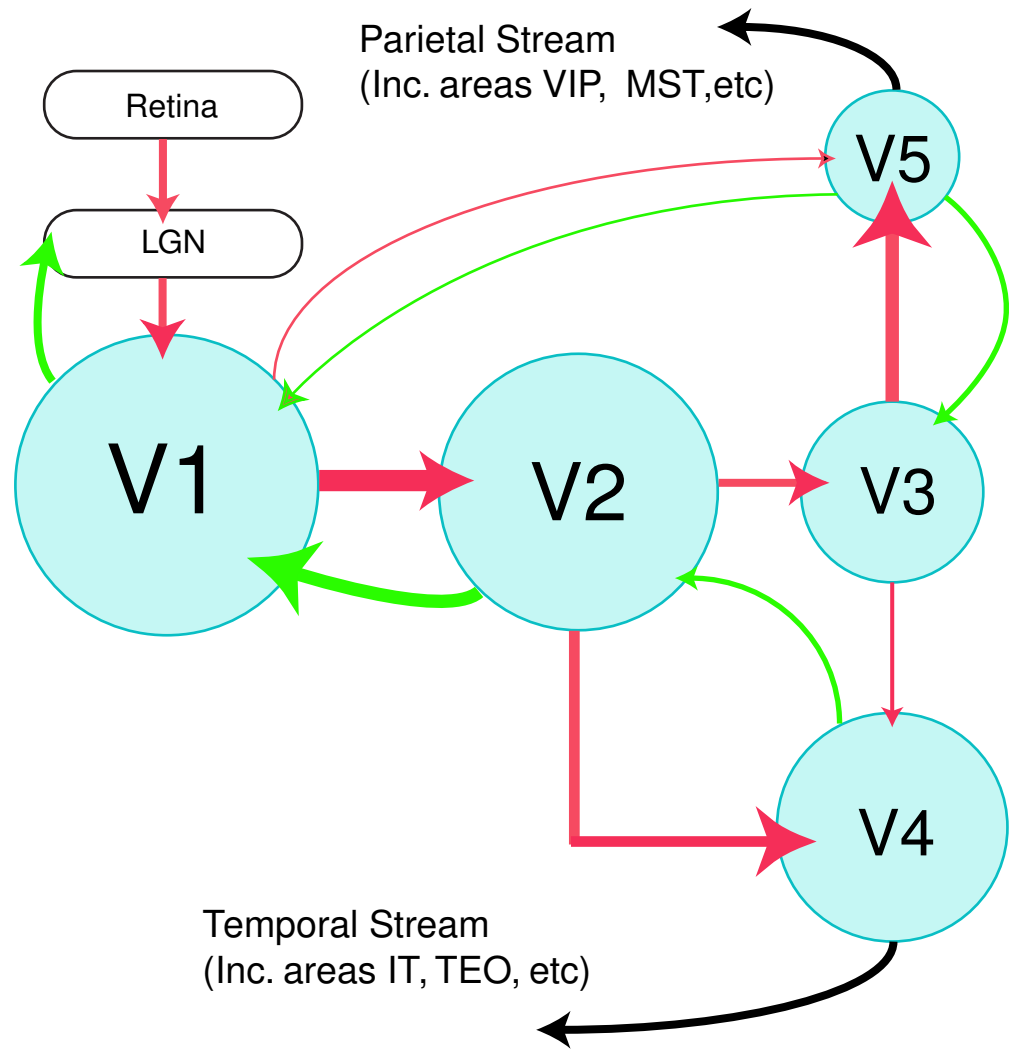

**Figure 2.1:** Simplified illustration showing the organisation of processing streams in the mammalian visual system. Red arrows indicate feedforward pathways, green arrows indicate feedback. The thickness of the arrow is roughly proportional to the strength of the connection and the area of the cortical regions (blue circles) is roughly in proportion to their size. (The diagram is based major areas identified in the visual system of the macaque; however the general organisation is conserved across species.) Tovee [1996], Zigmond et al. [2000], Lennie [1998] were used as the main sources in the construction of this figure.

## 2.2 Visual Pathways

### 2.2.1 Overview

The mammalian visual system is organised as a loose parallel hierarchy of processing areas. A schematic representation of this organisation is given in figure 2.1. As illustrated by the arrows in the diagram, there are substantial feed-back projections in addition to the feedforward pathways.

The figure also helps to illustrate two main divisions within the visual pathways, the temporal stream (V1→V2→V4→IT) and the parietal stream (V1→V2→V3→V5). These two streams are commonly referred to as the 'what' and 'where' pathways [Ungerleider and Mishkin, 1982]; with the temporal stream thought to be largely involved in shape detection and object recognition, and the parietal stream more heavily involved with motion processing and spatial localisation.

The development of these visual pathways is governed both by genetic factors, and by the statistics of activity within the pathways, which are themselves strongly governed by the statistics of external inputs. A major component of our work will be concerned with trying to understand how the circuits in the mature animal might relate to these statistics from a computational perspective. In terms of anatomical focus, this thesis mainly addresses the temporal stream and in particular the primary visual cortex.

### 2.2.2   Primary Visual Cortex: Circuits

We now summarise some observations about the circuitry and anatomy of primary visual cortex. The laminar architecture of V1 is shown schematically in figure 2.2. In addition to the laminar structure shown by this figure, primary visual cortex (in common with many cortical areas) also has a columnar organisation in which cells within columns perpendicular to the cortical surface tend to display similar response properties. Furthermore, nearby columns also tend have similar response properties, whilst more distant columns tend to show different selectivities. These receptive field and topographic properties are elaborated in later in sections 2.5 and 2.6.

There are three main pathways from the retina via the LGN to primary visual cortex, each having different signal-response properties [Callaway, 1998, Casagrande, 1994, Shapley and Lennie, 1985]. In the macaque they are called the M (magnocellular), P (parvocellular) and K (koniocellular) pathways and correspond to the three main classes of retinal ganglion cell. Of these, the M and P paths are the main thalamocortical inputs to V1 and have their major terminations in layers $4C\alpha$ and $4C\beta$. A notable point about the thalamocortical projections is that they are all glutamatergic; the thalamus cannot provide direct inhibition to the cortex. In terms of functional effects, however, the thalamus can deliver effective and rapid inhibition to V1 via excitation of inhibitory interneurons [Zigmond et al., 2000].

A further point of note is the large expansion factor in population size of the neural representation in going from thalamus to V1. After a 'compressive coding' from the retina to the optic tract [Atick and Redlich, 1992, Atick et al., 1992] there is a large expansion, by a factor of 100 or so, in terms of neuron number at the level of V1[1]. The V1 representation is, therefore, formally highly overcomplete.

Inputs from other cortical areas, consisting mostly of feedback projections from

---

[1]The factor of 100 here describes the total number of neurons. Arguably, in a statement such as this, it is the number of outputs from V1 that counts, but this number still gives an expansion factor of 10 to 20 for most primates.

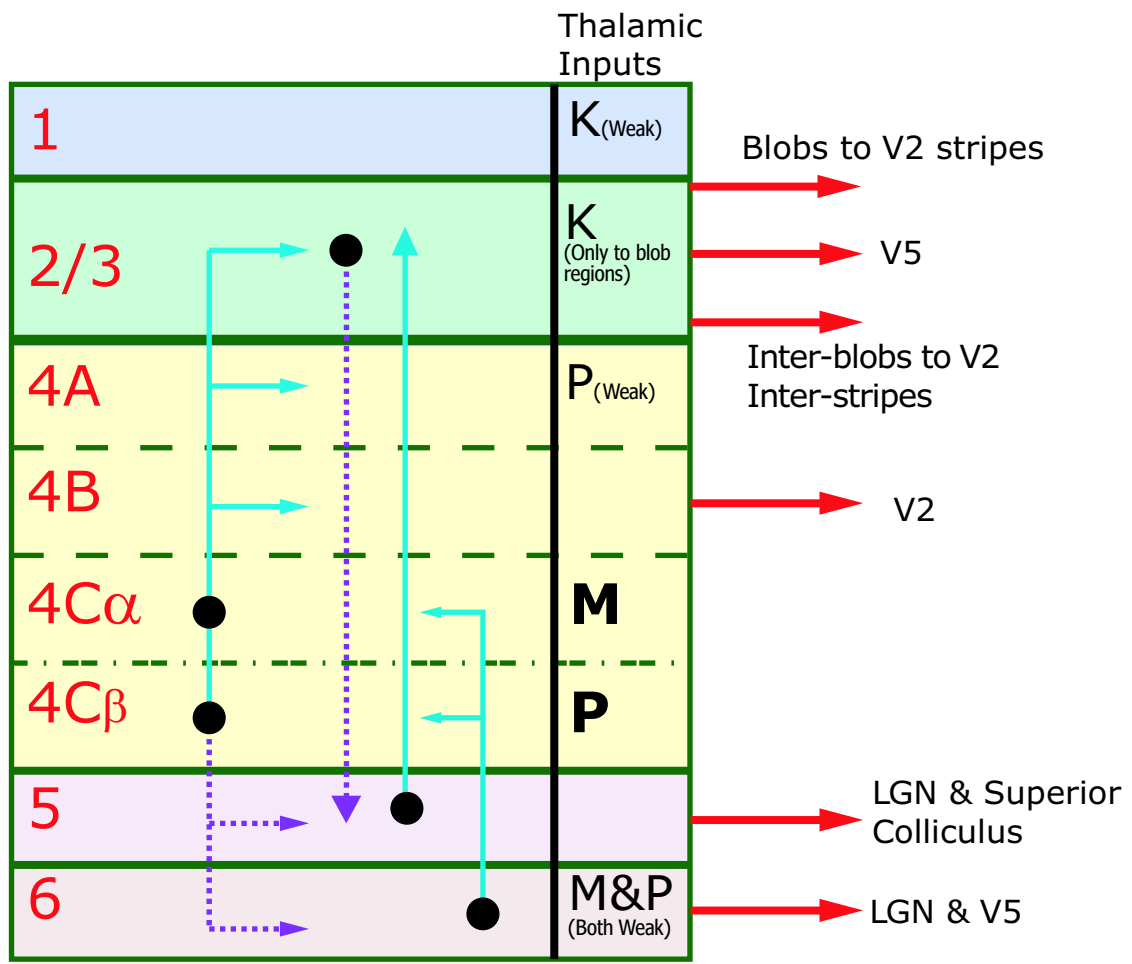

**Figure 2.2:** Simplified layout of LGN to V1 contacts, and laminar architecture of primate V1. The right hand column shows the dominant thalamic inputs to the different layers of V1. In the left section of the figure the turquoise and purple arrows illustrate the major excitatory 'vertical' circuits between different laminae. Also important, although not shown on the diagram, are 'horizontal' circuits which provide lateral recurrent connectivity within V1.

Layer 1 has very few neurons but is tightly packed with axons, dendrites and synapses. Layers 2 and 3 consist of a dense arrangement of cell bodies and many local dendritic interconnections from neurons in other layers. This layer (and also layer 1) appears to receive inputs from the K pathway. The outputs from layers 2 and 3 are primarily sent to other cortical areas. Layer 4 of visual cortex is notable in its size – it is much larger in V1 than it is in other cortical areas, and has been further subdivided into separate sections. Layer $4C$ is the site of the major thalamic input to V1, with the M pathway sending inputs the upper section, layer $4C\alpha$, and the P pathway sending its inputs to the lower section, layer $4c\beta$. Layer $4B$ receives a large input from layer $4C\alpha$ and sends its outputs to other cortical areas, predominantly V2. The neurons in layer $4B$ also receive, through their distal dendrites which extend up into the superficial layers, lateral inputs from other parts of V1 and also other cortical areas. Layer 5 has relatively few cell bodies compared to the surrounding areas. It sends local connections to layers 2/3 and to layer 4A/B, it also sends significant extra-cortical outputs to the superior colliculus and the LGN. Layer 6, like layer 4, is dense with cells and it sends feedback projections to layer 4C, as well as outputs to extra-striate areas and a large feedback output to the LGN. [Callaway, 1998, Zigmond et al., 2000, Tovee, 1996] extra-striate visual areas, terminate mainly in the superficial layers where they make contact with the distal dendrites of cells from other layers. In addition to these feedback connections, V1 also has considerable internal recurrent connectivity both 'vertically' between laminae within a column, and 'horizontally' linking regions up to millimetres away along the cortical surface. Anatomical studies show that number of feedback and recurrent connections is actually greater than that of the feedforward connections. However, it should be noted that number of synapses does not necessarily directly relate to physiological efficacy; distal contacts often have a much more limited effect than those close to the soma [Zigmond et al., 2000]. Consequently the principal input source for V1 does appear to be thalamic.

In terms of cortical output V1 sends projections to several areas, most of which are identified in the visual pathways shown in figure 2.1. The bulk of this cortical output goes to the neighbouring area, V2. In addition to these cortical projections, V1 also sends a substantial feedback connection to the thalamus. Our understanding of these feedback connections (and indeed neural feedback in general) is currently quite limited, and they have been the subjects of much speculation and interest, for example see [Sillito and Jones, 2002, Webb et al., 2002] and related work.

## 2.3   Receptive Fields and Tuning Functions

The receptive field and the tuning function are two closely linked, fundamental, but ill-defined concepts in sensory neuroscience. A simple and generic definition of the term receptive field might be given as follows: "The receptive field (RF) of a neuron within a neural system is a characterisation of how the inputs to the system map onto that neuron's response." However, this definition clearly leaves ambiguities — for instance what should we assume the response to be? Mean firing rate, inter-spike intervals, precise spike times or more elaborate measures linked to the activity of other neurons are all possibilities. Additionally, this simple concept could be made richer by considering internal state variables, such as arousal or attention, although such factors are generally difficult to account for properly.

In the idealised case, the RF would be sufficient to determine the expected response of a neuron to an arbitrary stimulus. However, even if we ignore issues such as internal state, it is seldom practical to characterise the behaviour of a neuron fully. Consequently, the term receptive field is often applied to rather approximate descriptions of the mapping from input space to outputs. Figure 2.5 gives an example of simple characterisations of some V1 receptive fields.

The term tuning function is sometimes used interchangeably with the term receptive field, although it is useful to make a distinction. The tuning function for a neuron should be thought of as a low-dimensional representation of its receptive field. Whereas the receptive field is usually considered to be a characterisation over the input space itself, tuning functions take as their arguments some parametric specification of a particular stimulus ensemble. For example, considering an ensemble of sinusoidal gratings we might describe the output as a function of the orientation and spatial frequency parameters of the patterns. To state that a cell is tuned to a parametric aspect of an input ensemble implies that the response varies significantly as we alter that parameter. In contradistinction to those aspects to which a neuron is tuned, we also find aspects to which neurons display an *invariant* response, i.e.: the (non-zero) response does not vary significantly over a range of parameter settings.

Empirically, elucidating a receptive field or a tuning function usually reduces to probing the system with a range of inputs, recording responses and then looking for a compact, approximately complete description of the stimulus-response mapping. The choice of stimuli used in probing, what exactly we decide the 'response' to be (rates, spike times, etc), and the methods applied in analysis are all very important here. Different methodologies can yield significantly different results, and making good choices can be especially difficult if one is dealing with highly non-linear units. Broadly speaking, two or three phases of analysis are usually performed in practice: (i) *ad hoc* probing with spots of light or bar stimuli; (ii) systematic probing using parametrically controlled stimulus ensembles, such as gratings or m-sequences [Sutter, 1987, Reid et al., 1997]; and occasionally (iii) *ad hoc* probing with natural scenes or natural movies.

With acknowledgement of these methodological caveats, much of the work in profiling receptive fields, particularly those early on in the visual pathways, can be understood using the tools of linear systems theory augmented with some simple non-linear mechanisms. In this framework the response of a neuron is given by the inner product of a filter and the input pattern, perhaps followed by some static non-linearity. For example, the following equation gives a good descriptive level account of many (quasi-)linear receptive fields,

$$F(t) = G\left(\iiint \mathrm{d}x\,\mathrm{d}y\,\mathrm{d}\tau\, S(x, y, t - \tau)D(x, y, \tau)\right) \tag{2.1}$$

where $S(x, y, t)$ is a spatio-temporal stimulus pattern, $D(x, y, \tau)$ is the first order spatio-temporal kernel that characterises the cell's receptive field, and $G(\cdot)$ is some static non-linearity – perhaps implementing rectification and saturation. The nature of the response, $F(t)$, is usually taken as an expected or trial averaged firing rate, or possibly as a rate parameter in some point process model of spiking. Such equations, however, say nothing about the implementational details nor the computational desiderata that they may serve.

The full literature on receptive field profiling techniques is large, and it would be inappropriate to discuss them all in detail here. General overviews and discussions of these methods can be found in Wandell [1995], Rieke et al. [1997], and Dayan and Abbott [2001]. Additionally, the following references give a more detailed coverage of specific advanced approaches: Ringach et al. [1997], Stanley [2002], Sahani and Linden [2002], Schwartz et al. [2002], Smyth et al. [2003], Pillow et al. [2003], and Paninski [2003].

## 2.4  Population Coding

Complementary to the issue of what a single neuron reports about its inputs is the broader issue of how a population of neurons combines synergistically to encode information. The notion of distributed neural representations [Hinton, 1984] is a key concept in modern computational neuroscience, although much remains to be clarified — both experimentally and theoretically. It is often the case that the population level is more appropriate than the single neuron level if we wish to fully understand the way in which information is represented and computations are performed. Stalwart citations for population codes include the representation of a global motion direction in V5/MT [Newsome et al., 1989] and the encoding of place in the hippocampus [O'Keefe and Dostrovsky, 1971]. However, there is a sense in which *all* neural representations employ some sort of distributed code.

At present, the precise way in which populations represent information is not clear. In addition to the information carried at the individual neuron level it has been speculated that properties such as inter-population timing or phase may encode additional properties, and there is some experimental evidence for such hypotheses (e.g. see Huxter et al. [2003]).

In terms of modelling there is a fairly broad range of different viewpoints. Many authors simply consider any distributed representation produced by a model of a neural system as a candidate 'population code'. Another style of approach is more specific and its primary focus *is* the population code. Work from this perspective often considers whole populations as encoding single, low-dimensional abstracted feature values, or more recently uncertainties and multiplicities of such feature values [Zemel and Hinton, 1995, Pouget et al., 1998, Zemel et al., 1998, Pouget et al., 2000, Sahani and Dayan, 2003].

Regarding the computational significance of population coding, several hypotheses have been put forward. One key property is that such an encoding scheme provides robustness to noise as well as to or cell death/damage. They have also been used as key components in models of multi-sensory integration [Zhang, 1996] and short-term/working memory [Seung, 1996]. Distributed representations that are sparse with respect to the typical number of 'active' members of the population have been suggested to be particularly well suited to perceptual inference and as a code for robust memory storage [Willshaw et al., 1969, Tsodyks and Feigel'man,

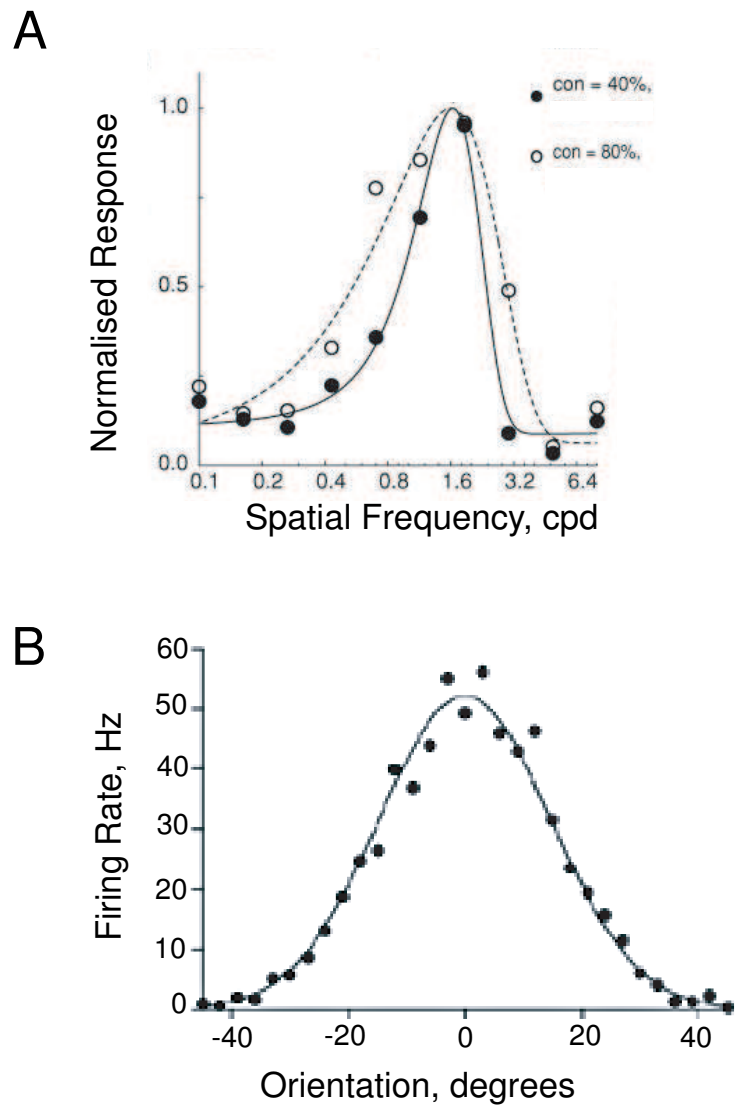

**Figure 2.3:** Example tuning curves for V1 simple cells to sine wave grating stimuli. (A) Normalised response as a function of spatial frequency at optimal orientation. Open and closed circles represent the tuning function at two different contrasts. (B) Mean firing rate as a function of orientation at optimal spatial frequency. (A) adapted from Sceniak et al. [2002] (B) adapted from a draft of Dayan and Abbott [2001], data originally from Henry et al. [1974].

1988, Rolls and Treves, 1998].

In this thesis we approach the problem of representational learning (both for neurons and populations) from the viewpoint of statistical density estimation. This perspective brings to light several additional interesting aspects about population codes. For instance, in order to capture higher-order correlational structure we require that there be response elements in the code which are sufficiently non-linear; linear units are only able to specify pairwise statistical relations.

## 2.5   V1 Receptive Fields

We now move to a more detailed consideration of the input-output properties of
V1 neurons. Initially a mostly classical view of these properties is presented; then,
in section 2.5.3, some observations and results that do not sit so neatly within
the classical perspective are discussed. Finally in section 2.5.4 quantitative data
summarising the distribution of some V1 receptive tuning functions are provided.

In their experiments on the visual cortex of the cat in the 1950's Hubel and
Wiesel initially made visual stimuli by making spots on glass slides and projecting
them on a screen. Although these stimuli had previously been found effective in
driving retinal ganglion cells, they found that few cortical neurons were particularly
responsive to them. One day they made the serendipitous observation that one cell
responded vigorously when the slide was being removed from their projection system
– it happened that the edge of the slide projected a thin black line onto the screen
and this turned out to excite the cell. And, so the story goes, this was the start of
a beautiful set of experiments probing the visual system [Hubel, 1982].

Hubel, Wiesel and other pioneering experimenters discovered that most cells in
primary visual cortex can be driven strongly by the right kind of gratings, edges or
elongated bars. More specifically, cells well tuned for some or all of the following
parameters have been found:

- spatial location;

- orientation;

- spatial frequency;

- temporal frequency;

- direction of motion;

- eye of origin;

- stereo disparity; and

- color.

Figure 2.3 illustrates tuning curves for some of these properties.

In their investigations, Hubel and Wiesel grouped the cells they recorded from
into three main classes – 'simple cells', 'complex cells' and 'hyper-complex cells'
[Hubel and Wiesel, 1962, 1968]. The different classes shared the general set of tuning
properties outlined above but differed in the precise structure of their receptive fields.
The simple cells were found to have elongated receptive field sub-regions that could
be segmented into on- and off-regions — in the on-regions a response could be evoked
by turning on a suitably oriented light stimulus, in the off-regions a response would be evoked when a light was turned off.[2] Complex cells on the other hand could be excited by a suitably oriented light turned either or off anywhere within their spatial region of sensitivity, i.e.: a response that is somewhat invariant to both the position of a localised stimulus and its polarity and phase. Hyper-complex cells were defined as those possessing inhibitory zones at one or both ends of the oriented excitatory regions (and hence are sometimes known as 'end-stopped' cells). This type of cell is less prevalent than the others, and has since been re-classified as a subset of simple and complex cells. We will not give much detailed consideration of end-stopping effects, other than to note that they may arise a simple consequence of (unusually strong) non-classical phenomena of the sort discussed in section 2.5.3.

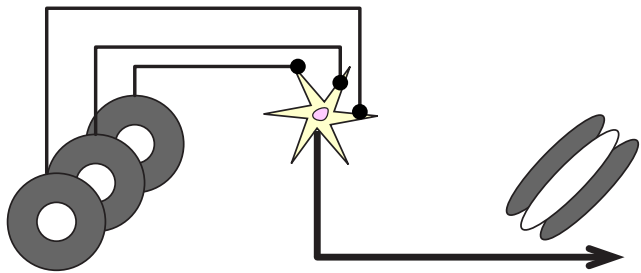

**Figure 2.4:** Mechanisms proposed by Hubel and Wiesel to explain oriented receptive field structure of simple cells. White, on response; gray, off response. Centre-surround LGN cells with appropriately aligned receptive field locations converge upon a simple cell, giving rise to orientation selectivity through feedforward excitation. Adapted from Hubel and Wiesel [1962].

### 2.5.1 V1 Simple Cells

Hubel and Wiesel defined a simple cell as one that: (i) had spatially segregated ON and OFF subregions; (ii) exhibited summation within each such subregion; (iii) had ON and OFF subregions that were antagonistic; and (iv) was such that its response to any stimulus could be predicted from the arrangement of excitatory and inhibitory subregions. If a neuron failed to pass one of these tests it was defined as complex [Hubel and Wiesel, 1962, 1968]. Mechanistically, it was originally postulated that simple cells might be formed by the summation of outputs from appropriately configured LGN neurons [Hubel and Wiesel, 1962]

Since those early experiments other researchers have benefited from advancing technologies and analytic tools, and have designed increasingly sophisticated experiments and analyses. These achieve their most spectacular successes when applied to simple cells with relatively linear response properties. Figure 2.5 illustrates receptive field maps (or more precisely a set of first-order spatial kernels) for four simple cells from the macaque, retrieved using a subspace reverse correlation technique [Ringach et al., 1997, Ringach, 2002].

---

[2]In addition to the oriented simple cells, investigators have also noted the presence of simple cells lacking orientation tuning. These receptive fields are much more like those seen in LGN, and are primarily found in layer 4.

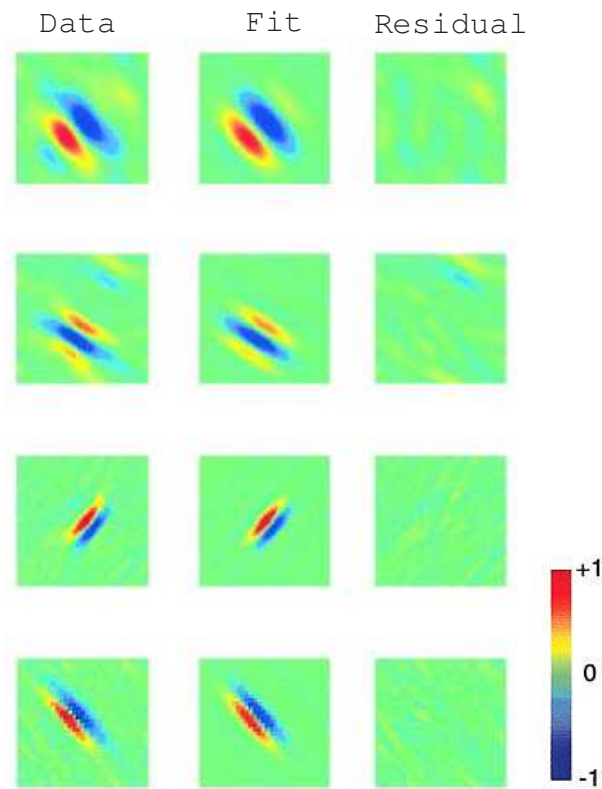

**Figure 2.5:** Empirically determined receptive fields. The left column shows the linear (spatial) kernels for 4 different V1 simple cells, retrieved by a subspace reverse correlation method. The central column shows a Gabor function fit to the kernel, and the left column shows the residual between this fit and the actual response. Values are color coded and in the range -1 to 1, mapped by blue and red respectively. Adapted from Ringach [2002]

Computationally, simple cells have been traditionally proposed as edge detectors or localised, oriented spatial[3] frequency detectors. More recently they have also been identified as being very similar to the linear 'independent' components of natural images and to the elements of a sparse coding basis for such images — we discuss this further in our review of models in chapter 3.

Good parametric fits to the spatiotemporal receptive field of simple cells are given by a combination of appropriate first order spatial and temporal kernels. For instance, a standard spatial kernel $D_s(x, y)$ is given by the Gabor function

$$D_s(x, y) = \frac{1}{2\pi\sigma_w\sigma_l} \exp\left(-\frac{x^2}{2\sigma_w^2} - \frac{y^2}{2\sigma_l^2}\right) \cos\left(2\pi fx - \phi\right) \qquad (2.2)$$

where $\sigma_w$ and $\sigma_l$ are, respectively, the minor and major axes of the elliptical gaussian envelope, $x$ and $y$ are displacements along these directions, $f$ is the spatial frequency and $\phi$ is the spatial phase. A standard temporal kernel $D_t(\tau)$ is the following

---

[3]or spatio-temporal

proposed by Adelson and Bergen [1985],

$$D_t(\tau) = \alpha \exp\left(-\alpha\tau\right)\left(\frac{(\alpha\tau)^5}{5!} - \frac{(\alpha\tau)^7}{7!}\right) \text{ for } \tau \geq 0 \text{ and } D_t(\tau) = 0 \text{ otherwise,}  \quad (2.3)$$

where the constant $\alpha$ is a shape parameter. By combining the above spatial and temporal kernels we can construct an approximate parametric expression for the spatio-temporal receptive field of a non-directionally tuned cell. The first order response to a stimulus $S(x, y, t)$ could then be given in the form of equation 2.1, as

$$F(t) = \iiint \mathrm{d}x\,\mathrm{d}y\,\mathrm{d}\tau\, S(x, y, t - \tau) D_s(x, y) D_t(\tau) \quad (2.4)$$

We may also wish to make slightly more sophisticated descriptions by applying a static non-linearity to the linear response $F(t)$, for instance by adding a lower threshold and an upper saturation level.

In order to describe direction sensitive cells we need to use a slightly different formulation. The spatio-temporal description of a direction sensitive cell is not separable into the 'usual' space and time co-ordinates. A simple way to get around this is by forming an alternative set of space-time axes, appropriately rotated (in four-dimensional space-time) from the original ones. Equally straightforward modifications can be applied to capture other aspects of stimulus dependence. For instance, to describe binocular response properties we could simply combine two kernels (such as the type in equation 2.2), one for each eye.

Although it generally appears to be a good fit to the data for the linear kernels, several authors have argued against the choice of a Gabor function as a parametric description of V1 simple cell kernels [Stork and Wilson, 1990, Wallis, 2001]. Whilst these arguments are interesting it is unclear whether the issue is an important one given that we know the true situation is much more baroque anyway. Also, whilst this simple feed-forward idea of Hubel and Wiesel is appealing and does have substantial experimental support [Alonso et al., 1996, 2001, Alonso, 2002, Ferster, 1994, Ferster et al., 1996, Lampl et al., 2001] it seems unlikely to be the complete story (for instance see Ben-Yishai et al. [1995], Somers et al. [1995]). Some of these subtleties will be touched upon in section 2.5.3.

## 2.5.2   V1 Complex Cells

Complex cells generally show the same range of specificities as seen in simple cells (orientation, direction, spatial frequency, etc selectivity), but differ in that they have slightly larger receptive field areas for a given eccentricity, are invariant to contrast polarity, and display a degree of translation invariance in their responses. This fact alone already poses a fatal challenge for purely linear models of response.

Regarding their computational role, complex cells have been suggested to be an important initial component for invariant object recognition (for example see Fukushima et al. [1983], and Riesenhuber and Poggio [1999]), and some of the most successful practical object recognition systems employ hand-crafted, complex-cell-like components (for example see Lecun et al. [1989], Lowe [2000]).

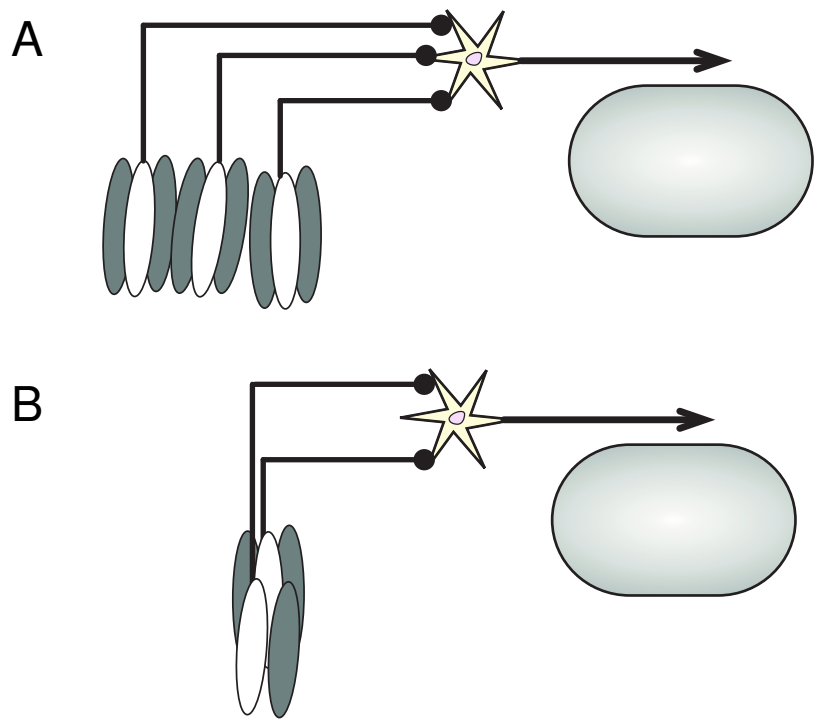

**Figure 2.6:** Mechanisms proposed by Hubel and Wiesel to explain the receptive fields of complex cells. The outputs of simple cells with (A) slightly translated receptive fields, or similarly (B) quadrature differences in phase, are summed together to give a complex cell which responds to oriented stimuli in a localised spatial region regardless of contrast polarity/spatial phase and with a degree of invariance to location.

Mechanistically, it was conjectured that complex cells were formed by pooling the output responses from simple cells with quadrature differences in phase or slight shifts in location, as illustrated in figure 2.6. As with the basic mechanism for simple cells outlined above, this construction does have some experimental support [Martinez and Alonso, 2001] but again there is likely much more to the story.

Nevertheless, mathematical accounts consistent with this mechanistic description do indeed perform reasonably well in predicting the responses of complex cells to simple stimuli. Adelson and Bergen [1985] suggest taking the squared outputs of an array of linear-filter simple cells (such as given by equation 2.4) and linearly summing them to give the complex cell output; other authors, such as Riesenhuber and Poggio [1999], suggest taking the maximum of the simple cell array. We generalise a little to give an expression such as,

$$F(t) = G\left(\sum_i g\left(\iiint \mathrm{d}x\,\mathrm{d}y\,\mathrm{d}\tau\, S(x,y,t-\tau)D_s^{(i)}(x,y)D_t^{(i)}(\tau)\right)\right) \qquad (2.5)$$

where $D_s^{(i)}(x,y)$ and $D_t^{(i)}(\tau)$ are the spatial and temporal kernels describing the $i^{\mathrm{th}}$

simple cell components, and $g$ and $G$ are static non-linearities. Employing $G(\lambda) = \lambda$ and $g(\lambda) = \lambda^2$ gives the Adelson-Bergen description, whilst for large $n$, $G(\lambda) = \lambda^{\frac{1}{n}}$ and $g(\lambda) = \lambda^n$ gives the Riesenhuber-Poggio description. This type of mathematical characterisation is able to capture many basic aspects of the data on complex cell responses.

### 2.5.3   Non-classical Results

Sections 2.5.1 and 2.5.2, whilst being accurate in their simplified context, clearly do not cover the full complexity found in V1. This section introduces some of the so-called 'non-classical' effects that have been observed, and presents a brief overview of more complicated issues.

#### The Simple vs. Complex Cell Distinction

As mentioned in section 2.5.1, Hubel and Wiesel initially classified cells as being simple or complex based on a number of partly subjective tests [Hubel and Wiesel, 1962, 1968, Mechler and Ringach, 2002]. Subsequently, it was recognised that a key property of the tests was linearity of summation within the receptive field and a number of groups proposed a quantitative test to assign neurons to the two classes [Skottun et al., 1991]. One such test of linearity is the ratio of the amplitude of the first harmonic of the response and the mean spike rate when the neuron is stimulated with drifting sinusoidal gratings – the $F_1/F_0$ ratio. (Another test is to look at the periodicity of response when stimulated with counter-phase gratings – complex cells would be expected to show clear show period doubling.) Empirically, this $F_1/F_0$ ratio is bimodally distributed over the V1 population and the classification of cells based upon it corresponds closely to the classical definitions of Hubel and Wiesel.

Recently, the standard methods of classifying simple and complex cells has been called into question [Mechler and Ringach, 2002]. These authors argue that the bimodal distribution in $F_1/F_0$ ratio (or other similar quantities) need not reflect two distinct subpopulation of cells. They show that such a distribution could equally well arise as a consequence of non-linearities acting on a single population of cells with a unimodal distribution of parameters. Whether or not this is the case there clearly *are* qualitative differences between some subsets of neurons in V1 in terms of the linearity of their response. However, Mechler and Ringach's proposition does have interesting implications when considering the mechanisms by which the identified simple and complex cell types might arise. Along such lines as this, the proposal that complex cells arise by pooling the outputs of phase quadrature simple cells has been questioned by several authors. Alternative models for the origin of complex cell responses have been suggested, such as the work by Chance et al. [1999] which postulates that simple and complex cells arise as just the low- and high-gain limits of the same network circuitry, or the work by Mel et al. [1998] which suggests that complex cell properties arise as a consequence of intra-dendritic processing.

In defence of the more classical picture, however, some anatomical and physiological support for the hierarchical description has been discovered in recent years. Alonso and Martinez [1998] and Martinez and Alonso [2001] present results which suggest that many complex cells are primarily driven by monosynaptic inputs from simple cells.

As with many such arguments, it is highly likely that both sides are correct to some extent – that the observed distribution of properties is due to *both* feedforward and network effects. Indeed from a purely interpretive computational perspective it is more important to simply acknowledge the existence of cells with this range of qualitatively different properties rather than to resolve the exact mechanisms by which they are formed.

**Extra-Classical Centre and Surround**

In addition to a 'central' area of sensitivity (i.e.: the 'classical' receptive field of Hubel and Wiesel) cells also receive inputs from a much larger area, often termed the 'surround'. The surround is unable to induce a response alone, but modulates activity that is generated by stimuli at the centre. Different groups have reported various types of modulatory phenomena, with some reporting suppression, others facilitation, and still others a mixture of the two; furthermore the extent of these effects has been reported to show contrast dependence. Additionally, estimates of the spatial extents of these processes vary between different reports.

Cavanaugh et al. [2002a,b] seem to reconcile some of the discrepancies and put some of the disparate observations onto a common ground. However, the full story is still far from being clear. One of the key points to come from their results is that if we wish to think of separate centre and surround mechanisms then we must take care in defining exactly what constitutes the regions of the centre and surround. Another key point from these papers (and also clearly highlighted in the work of Jones et al. [2002], Jones et al. [2001] and Sillito and Jones [1996]) is that the effect which a stimulus has depends strongly on its context in a non-linear way.

Several authors (e.g.: [Carandini and Heeger, 1994, Cavanaugh et al., 2002a,b]) suggest phenomenological models of these effects by adding divisive interactions to standard linear models. For example Cavanaugh et al. [2002a,b] consider separately constructing (rectified) linear responses for a centre mechanism, $L_c(x, y, \tau)$, and a surround mechanism, $L_s(x, y, \tau)$, and then combining them to give an overall response $L_{\text{norm}}(x, y, \tau)$ as:

$$L_{\text{norm}}(x, y, \tau) = G\left(\frac{k_c L_c(x, y, \tau)}{1 + k_s L_s(x, y, \tau)}\right) \quad (2.6)$$

where $k_c$ and $k_s$ are appropriate constants, and $G(\cdot)$ is a non-linearity. This kind of divisive interaction extends the capability of many basic descriptive approaches and allows a wider range of data to be reproduced, for instance see Simoncelli and Schwartz [1999].

Mechanistically, the origin of these surround effects is still rather unclear – lateral interactions within V1 and/or cortical feedback from other visual areas have been postulated. Similarly, the computational role of centre-surround interactions is also unclear. It has been variously suggested that they might produce an enhancement in signal contrast, that they might help in producing invariant responses or that they might help to decorrelate V1 outputs.

### Responses to more natural stimuli

Most experiments that are performed use artificial stimuli – largely out of the need to have tight parametric control over the experimental variables and to allow for reasonably simple interpretations. (Although see, for example, Baddeley et al. [1997], Vinje and Gallant [2000, 2002], Weliky et al. [2003], and Smyth et al. [2003]) However, despite the fact that strong responses can be elicited by artificial stimuli, the overall nature of most experimental test sequences is quite *unlike* that which the animal would be expected to encounter under natural conditions or even when simply viewing freely in a laboratory. As a consequence, whilst it is perhaps reasonable to have confidence about receptive field measurements that are restricted to small regions of visual space, we may wish to exercise caution in extrapolating too much from results which appear to give a sense of global processing until we are able to deal more adequately with richer stimuli.[4]

### Higher-order Analyses

Often a neuron's response is considered as being some 'instantaneous' firing probability or some time averaged rate, but there is an issue of the time scale over which it is appropriate to average — especially since many cells adapt their responses over time when under prolonged stimulation. Furthermore, variables other than rate may be important. For example the temporal pattern of individual spikes, or the phase of responses across several units can be shown to carry significant information.

Enormous volumes of data are required in order to account for such 'higher-order' effects properly, and the necessary analyses are far from trivial. For instance, consideration of higher-order functional dependencies in a Wiener or Volterra expansion, or consideration of spike timing patterns typically require orders of magnitude more trials than a simple linear rate analysis.

---

[4]Although even then, there still remains the question of what, exactly, is truly 'natural'.

### 2.5.4   Distribution of some RF Properties

Figures 2.7, 2.8 2.9 and 2.10 collect information about the receptive fields found in mammalian primary visual cortex and summarise how the parameters of Gabor function fits are distributed.

Figure 2.7 shows a collection of histograms of properties in macaque V1. The orientation and spatial frequency bandwidth histogram shows that most cells are reasonably well tuned for these properties. The aspect ratio histogram shows that most cells are reasonably circular with a tendency to elongation perpendicular to the direction of periodicity. We note that, in terms of spatial frequency, the coverage spans about one order of magnitude.

Figure 2.8 gives a clearer indication of the receptive field shape by plotting a normalised measure of the envelope of a fitted Gabor. We see that the distribution of shapes observed lies, mostly, along in a one parameter family in this space and that most cells have between 2 and 5 subfields.

Figure 2.9 shows the distribution of the phase of the Gabor fit for a population of neurons. This has been plotted modulo $\frac{\pi}{2}$ and ignores the polarity of the receptive field. We see that there is a tendency to cluster into even- and odd-symmetric receptive fields.

Lastly, in figure 2.10, we show a histogram of the orientation tuning properties. The data we show is close to being uniform, although a slight over-representation of the cardinal orientations (vertical and horizontal) seems apparent. These results are still regarded as being somewhat controversial.

## 2.6   Topographic Organisation in Primary Visual Cortex

A striking feature of many cortical areas, and particularly those involved with sensation, is the presence of topographic order. If we flatten out the cortical surface and consider it as a two-dimensional[5] sheet then, for most locations, small changes in position on this surface correspond to small changes in some or all of the tuning properties of the underlying cells — there is a topographic map of tuning properties. However, cells are tuned to many dimensions, and consequently discontinuities and/or singularities in the maps for individual tuning properties are inevitable since the cortex is effectively a two-dimensional sheet.

Section 2.6.1 presents a summary of types of topographic organisation found in mammalian primary visual cortex, while sections 2.7 and 2.7.1 discuss some issues relating to development.

---

[5]This is often reasonable due to the columnar structure of many cortical areas.

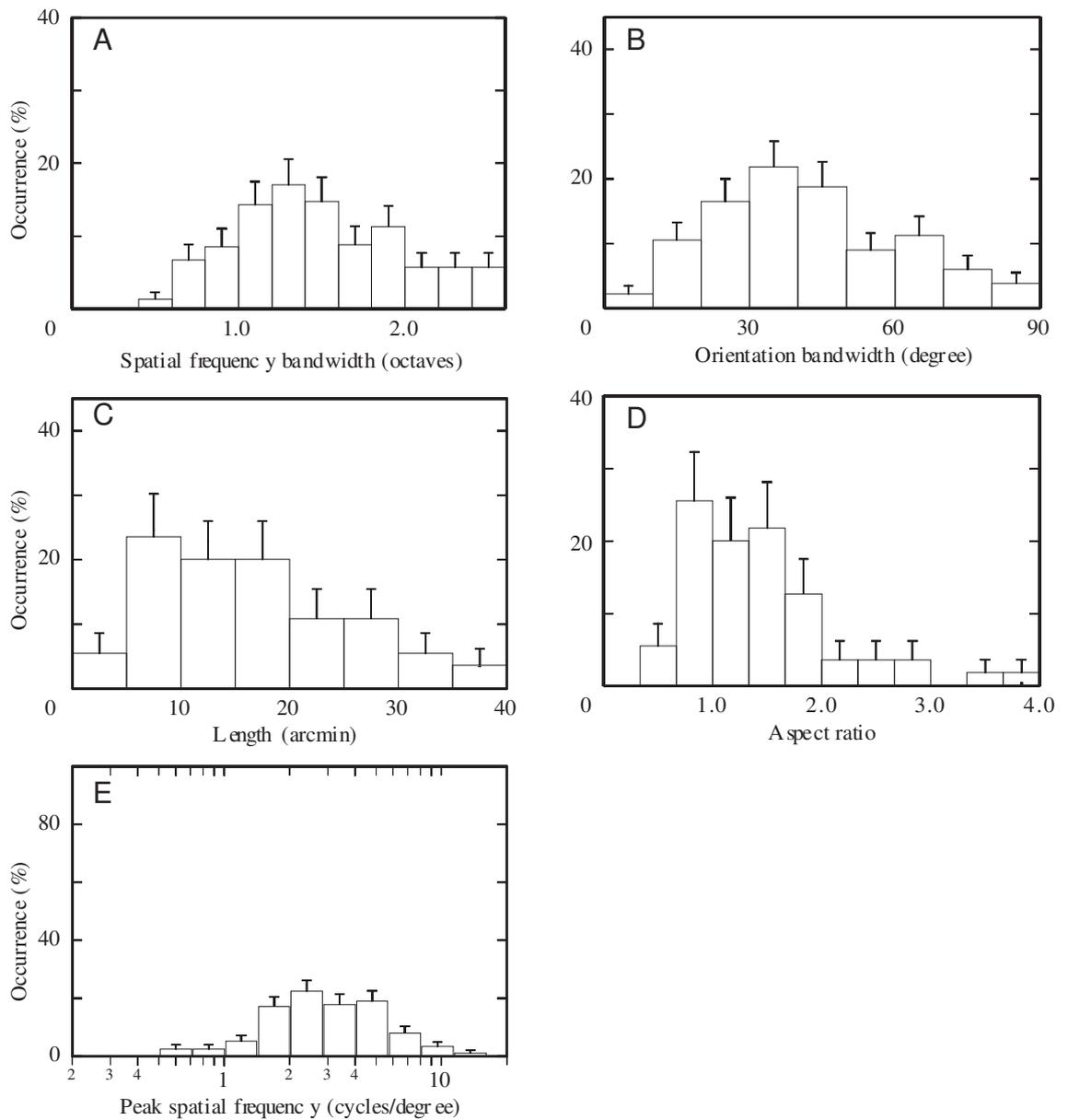

**Figure 2.7:** Histograms summarising some receptive field properties from macaque V1. (A) Distribution of spatial frequency bandwidths. (B) Distribution of orientation bandwidths. The definition of bandwidth for panels (A) and (B) is full width at half maximum (FWHM). (C) Distribution of receptive field lengths ($\sigma_l$). (D) Distribution of aspect rations ($\sigma_l/\sigma_w$). (E) Distribution of peak in spatial frequency tuning curves. Adapted from van Hateren and van der Schaaf [1998], with original data from De Valois et al. [1982].

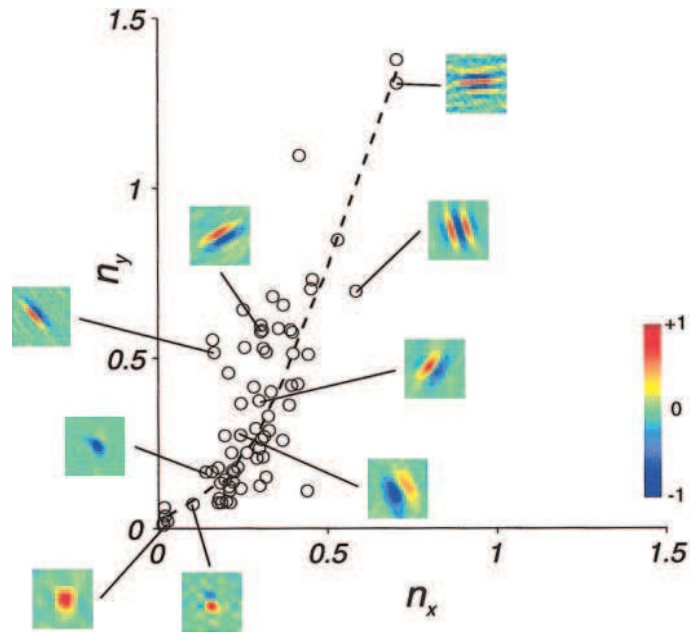

**Figure 2.8:** Distribution of receptive field shapes in the $(n_x, n_y)$ plane. ($n_x$ and $n_y$ are normalised measures of the receptive field envelope size. $n_x = \sigma_w f$ and $n_y = \sigma_l f$ in the notation of equation 2.2.) A number of receptive fields are shown along the distribution. Dashed line, a smooth version of the scatterplot. Blob-like receptive fields are mapped to points near the origin. Neurons with several subfields are mapped to points away from the origin. Adapted from Ringach [2002].

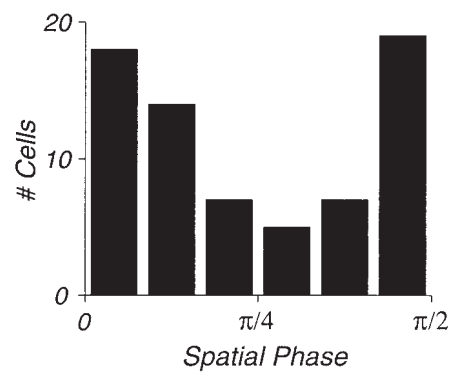

**Figure 2.9:** Distribution of Gabor function fit phases across macaque V1. Note the clustering at 0 and $\pi/2$ radians. Adapted from Ringach [2002].

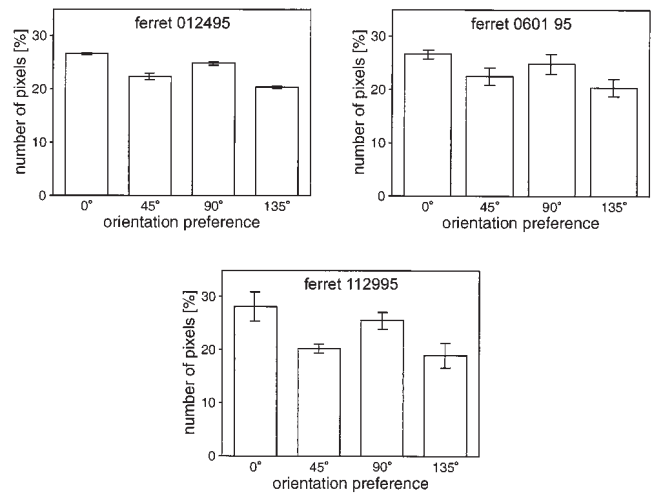

**Figure 2.10:** Histograms illustrating the slight bias in orientation representation; data shown from three animals. The representation of different orientations in mammalian had been thought to uniform and it certainly is close to being so. However, recent results (from developing ferrets) suggest that there is a slight over-representation of cardinal orientations (i.e. vertical and horizontal.) Adapted from Chapman and Bonhoeffer [1998].

### 2.6.1   Types of cortical map and their properties

A strong initial indication that mammalian cortex might have an orderly structure in terms of the spatial layout of tuning function and receptive field properties came (again) from pioneering work in electrophysiology [Hubel and Wiesel, 1977]. Diagonal electrode penetrations were made through the cortical surface, thus sampling different cortical locations at different depths. Cells encountered at equally spaced locations were characterised and it was noted that stimulus preferences changed gradually for most of the penetration with occasional, brief regions of rapid change.

Subsequently, experiments involving the intra-ocular injection of radioactive tracers gave insights into how properties varied on a cortical scale for example, Tootell et al. [1988a,b,c,d,e]. More recently, high resolution optical imaging [Kalatsky and Stryker, 2003, Blasdel and Salama, 1986, Grinvald et al., 1986] and high resolution fMRI [Chen and Ugurbil, 1999, Logothetis et al., 2002, Brewer et al., 2002] have allowed experimenters to record responses over a larger cortical area, and to record maps of sensitivity to several different variables simultaneously. This has given a more complete view of how receptive field properties change as one moves across the cortical surface, and also how the changes in feature preferences might be related to each other. Maps of joint tuning to different stimulus parameters are now available from a number of species and show very interesting properties. Currently the best characterised maps are probably those for tuning to:

- retinotopic position (e.g.: Tootell et al. [1988e]),

- ocular dominance (OD) (e.g.: Grinvald et al. [1991]),

- orientation (e.g.: Blasdel [1992]),

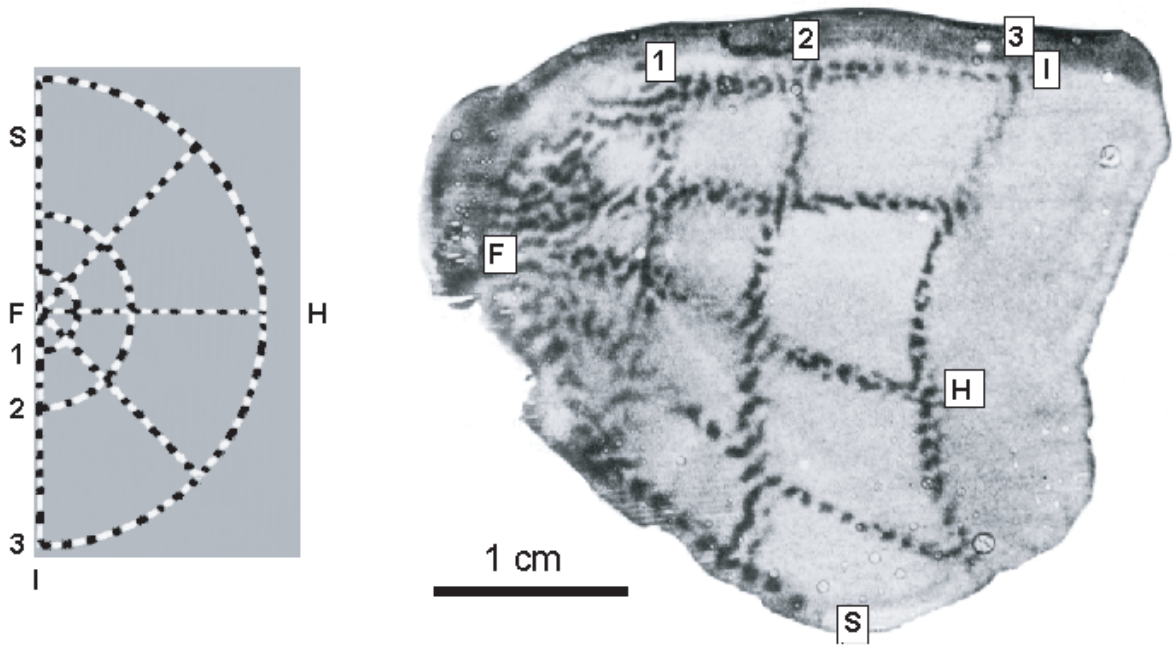

**Figure 2.11:** Retinotopic mapping in macaque. The pattern of spokes in the left panel was presented to the animal and was imaged using a 2-deoxyglucose auto-radiographic technique. The form of the mapping can be approximately be described as a complex-log transformation (see Dayan and Abbott [2001] for details). The magnification factor is largest at the fovea and decreases towards the periphery. In conjunction with the variable density of retinal ganglion cell sampling this results in an approximately constant ratio of representation size between V1 and the retinal ganglion cells.

Whilst on a coarse scale the retinotopic maps is smooth, on a finer scale it can be rather more irregular. This is primarily because the map for ocular preference (see figure 2.12) interrupts retinotopy, necessitating discontinuities and 'switchbacks', but it is also because of effects such as receptive field scatter. Figure adapted from Swindale [1996], originally from Tootell et al. [1988e].

- direction of motion (e.g.: Weliky et al. [1996], Kim et al. [1999]) and,

- spatial frequency (e.g.: Issa et al. [2000]).

In addition to these variables there is a somewhat less data available for maps of other properties, such as tuning to colour (e.g: Roe and Ts'o [1999]). Erwin et al. [1995] and Swindale [1996] provide good reviews of much of the literature.

Typical examples of the maps measured for retinotopy, ocular dominance and orientation are presented and discussed in figures 2.11, 2.12 and 2.13.

On a local scale we see considerable order within a map as well as consistent relationships between different types of map. Using more global measures, we see both disorder within a map and co-ordination between maps. The properties of, and relationships between, different kinds of map have been studied quite extensively.

The following list aims to summarise some of the main observations about V1 cortical maps that have been noted in the literature. This list is given for completeness; not all properties are key and neither our models nor other existing approaches address all these issues. For general references or where no specific references are given see Erwin et al. [1995] and Swindale [1996].

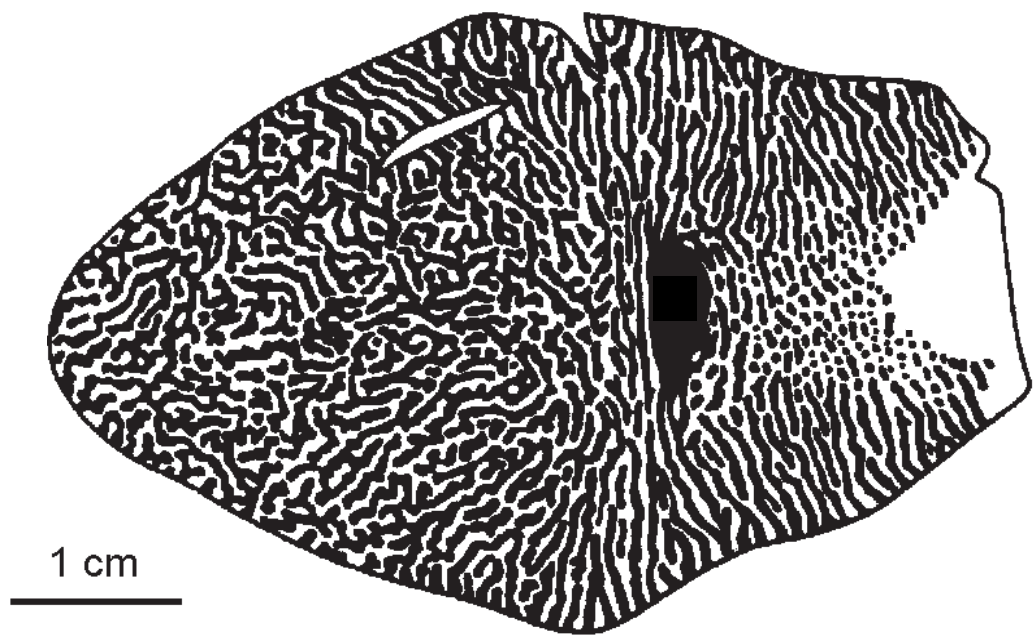

**Figure 2.12:** This figure schematically illustrates the pattern of ocular dominance columns across macaque V1, with black and white indicating the ipsi- and contralateral bias. Within a column the thalamic target cells of layer 4C show the strongest bias, with many cells being exclusively monocular. Cells outside of layer 4C typically show a more binocular response but usually reflect the underlying layer 4C bias with preference for one eye or the other.

The ocular dominance pattern found in mature animals develops from one in which all layer 4 cells receive significant inputs from both eyes (usually with a somewhat stronger input from ipsi-lateral side). The adult pattern overcomes this initial bias (in the regions that end up favouring the contra-lateral eye), and arises through a process of pruning (i.e.: synaptic depression or loss) as well as strengthening.

Also note that the ocular dominance bands tend to meet the boundaries of V1 at right angles. Figure adapted from Swindale [1996], originally from Florence and Kaas [1992].

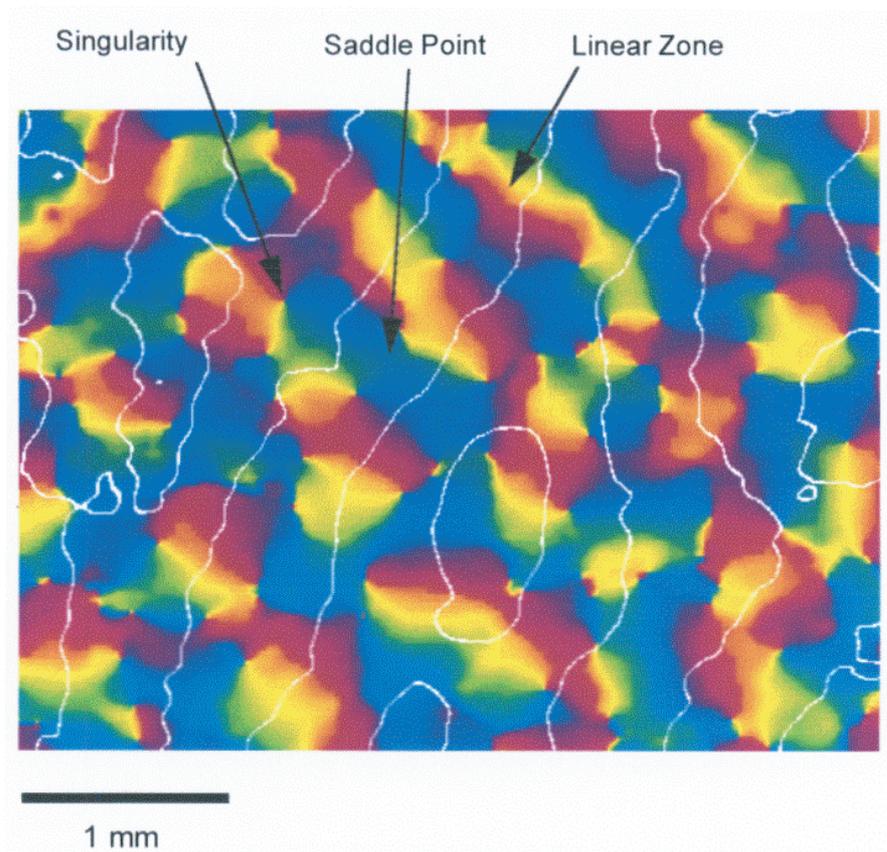

**Figure 2.13:** Composite figure showing the arrangement of orientation domains (a single colour represents a unique range or orientation preferences) and their relationship with ocular dominance column boundaries, which are shown as thin white lines. Note that the iso-orientation domains tend to intersect ocular dominance column borders at right angles.

Also note the highlighted features. *Linear Zones:* Regions in which the iso-orientation domains are approximately parallel slabs of uniform width. *Saddle points* Regions which define a local peak in orientation in one direction and a local trough in the orthogonal direction. *Singularities:* Also known, somewhat colloquially, as pinwheels. Points about which there is a half-rotation and all orientations are represented. Figure adapted from Swindale [1996].

- The power spectra of the maps for both ocular dominance and orientation are dominated by an elliptical annulus. This is a consequence of the fact that the maps have a general overall periodicity that is somewhat anisotropic. (The exact details of the power spectra vary amongst species.) A related observation is that the maps for ocular dominance and orientation tend to be 'globally orthogonal' — the major axes of the ellipses in the respective power spectra are at right angles. Furthermore, the periodicity of the orientation pattern perpendicular to the stripes of ocular dominance is approximately the same as for the stripes themselves, and is larger than the period in the orthogonal direction.

- The auto-correlation function for single maps in macaque have a 'Mexican-hat' shape. This implies that whilst there is considerable local structure, there is little structure on a global scale other than the coarse periodicity noted above.

- Ocular dominance band borders and iso-orientation contours have a tendency to meet at right angles. (This orthogonality property is particularly pronounced in macaques and is also present, albeit to a lesser extent, in cats.) [Muller et al., 2000, Obermayer and Blasdel, 1997, Kim et al., 1999]. A complementary observation is that the magnitude of the gradient for ocular dominance and orientation tuning tend to be anti-correlated i.e.: we tend to see rapid changes of one of the variables accompanied by slow changes in the other.

- Ocular dominance borders and motion iso-direction contours also tend to meet at right angles. This is in agreement with a tendency for gradient vectors of preferred orientation and directions to be parallel to each other in conjunction with the previous observation about orientation contours [Kim et al., 1999].

- The rate of change of orientation appears to be positively correlated with the rate of change of retinal position. This observation, made by Das and Gilbert [1997], is somewhat puzzling since it implies an uneven sampling density of different orientations over space.

- Pinwheels and saddle points of the orientation map tend to occur at the centres of ocular dominance bands, and furthermore the pinwheel locations within the band tend to correspond to local maxima in ocularity [Crair et al., 1997].

- Fractures in the direction map sub-divide iso-orientation regions into domains selective for opposite directions of motion [Weliky et al., 1996].

- Pinwheel singularities in the orientation map tend to be connected by an odd number of fractures (lines of discontinuity) in the direction map [Kim et al., 1999].

- A large majority ($\sim 80\%$) of pinwheels are nearest neighbour to a pinwheel of *opposite* sign.

- A continuous range of spatial frequencies seem to be mapped in a locally continuous fashion (although with fractures) across the cortex [Issa et al., 2000] (although others have reported finding a binary mapping of low and high frequency regions, such as Shoham et al. [1997]). In comparison to ocular dominance and orientation domains, the layout of spatial frequency domains seems more irregularly spaced.

- The regions corresponding to preference for the highest and lowest spatial frequencies have a tendency to co-localise with the pinwheels of the orientation maps [Issa et al., 2000].

- There seems to be no structure, i.e. an essentially random pattern, in the 'map' for spatial phase.

- Recent studies suggest that, in addition to maps for 'classical' properties, there may also be continuity and 'maps' for extra-classical receptive field properties, for example Yao and Li [2002].

### 2.6.1.1   Inter- and Intra-species Variability of Map Structure

Whilst there is a considerable amount of conserved structure between the topographic organisation in the primary visual cortices of widely different mammals, there are also significant differences. For instance several species completely lack ocular dominance columns, and the much-beloved tree shrew has a *vertical* segregation of its ipsi- and contra-lateral inputs[6] [Muly and Fitzpatrick, 1992]. Another example is orientation selectivity in rat, where the 'map' for orientation seems to be a random 'salt-and-pepper' layout[Girman et al., 1999]. As well as these rather qualitative differences there are also a quantitative differences between the maps of different species, for example the ocular dominance structure of in cat area 17 is considerably 'patchier' than in macaque V1.

## 2.7   Development of Structure in Primary Visual Cortex

Development is clearly a process which depends on both genetic *and* environmental factors. However, neither the extent to which the two factors affect the development of different parts of the brain, nor how this might change during the course of development are at all clear. Related to this, and also controversial, are the issues of: (i) to what extent, and (ii) in what way, neural activity is necessary for the development of various cortical structures (both individual receptive fields and overall maps) found in mature animals?

---

[6]The contralateral eye dominates the central section of layer 4 whilst the upper and lower sections tend to be more binocular and also receive innervation from the ipsilateral eye.

Activity independent mechanisms, such as axon guidance via molecular cues (e.g.: Tessier-Lavigne and Goodman [1996]) clearly play an important role in establishing the coarse structural order neural connectivity, such as the gross ordering of topographic maps and the underlying arborisation of projective fields . It is much less clear how important such factors are in setting up finer details of receptive fields and feature maps though, and the debate is complicated by the possibility that this degree of importance is different across species.

The main activity-dependent mechanism is activity-dependent synaptic plasticity. The strength of the synaptic connection between two neurons can be modified as a consequence of the patterns of activity which they undergo. Many of the details of this process seem to be captured by variants of the Hebbian learning rule [Hebb, 1949, Bi and Poo, 1998, 2001, Dayan and Abbott, 2001], which we discuss in chapter 3, although there is much that remains to be understood.

At present, the literature addressing these questions is diverse and often seemingly contradictory — although the rapidly developing applications of molecular biology promise to bring greater clarity. The coverage presented here cannot do justice to a complete survey and we make just a few points related to receptive field and topographic map formation in higher vertebrates. For a more general and comprehensive introduction to mechanisms of cortical development see Price and Willshaw [2000].

### 2.7.1   Developmental Manipulations and Abnormalities

Thus far, one of the principal approaches to understanding the significance of activity independent mechanisms relative to activity-dependent mechanisms (and within this the significance of experience dependent and experience independent activity) has been to make various manipulations to the normal developmental process.

We now summarises some of the main experimental manipulations that have been carried out. These abnormal developmental states provide us with some insight into the normal developmental processes. As a general point, it is interesting to note that most of the interventions performed have a limited window or 'critical period' in which they have effects that persist to adulthood (for example see Horton and Hocking [1997], although see Feldman [2003] and Godde et al. [2002] for discussions of plasticity in the maps of adult animals).

### Dark rearing and binocular lid suture

This involves animals being raised in total darkness from birth, or having eyelids in each eye sutured together. The results of this manipulation differ quite significantly between different species. Normal patterns of ocular dominance have been reported in macaque V1 [Horton and Hocking, 1996], but this does not seem to hold for cat area 17 in which we see abnormal or reduced segregation of the OD columns. Curiously, however, cat area 18 *does* seem to develop normal ocular dominance structure [Swindale, 1981, 1988, Mower et al., 1985]. The influence of dark rearing on orientation maps has been less closely studied, perhaps because both orientation selectivity and structured orientation maps can be found in macaque from birth [Wiesel and Hubel, 1974]. However, in cat area 17 and 18 the normal receptive field structure of a large proportion of orientation selective cells is disrupted by dark rearing [Blakemore and Van Sluyters, 1975, Blakemore and Price, 1987].

### Monocular deprivation

In this manipulation, one of the animal's eyes is deprived of normal input, usually by eyelid suture. If vision through one eye is prevented for a period of a few days during, or shortly after, the period when segregation of ocular preference occurs, the ocular dominance stripes or patches formed by the deprived eyes inputs shrink, while those from the normal eye expand. This has been shown for both monkeys [Hubel et al., 1977, Horton and Hocking, 1997] and kittens [Shatz and Stryker, 1978]. These observations are among the strongest pieces of evidence that segregation is a competitive process in which the final outcome is determined, at least in part, by visually evoked neural activity.

### Artificially induced strabismus

Artificially induced strabismus (usually produced by cutting eye muscles) in the cat results in a more sharply segregated pattern of left- and right-eye inputs (i.e. a lower degree of binocularity at the boundaries) [Lowel, 1994]. Similar results are also obtained with alternating monocular exposure [Tieman and Tumosa, 1997]. In monkeys made artificially anisometropic (by rearing with a -10D lens in front of one eye) it seems that OD stripes are more widely spaced and more irregularly organized than normal. These observations, in conjunction with the results of monocular deprivation, suggest abnormal postnatal visual inputs can cause changes in subsequent columnar organization. In particular, both results suggest that ocular dominance column spacing is not solely determined by intrinsic intra-cortical interactions but is also affected by the degree of correlation between the inputs to the two eyes.

### TTX impulse blocking

Stryker and Harris [1986] eliminated spontaneous retinal activity by giving intra-vitreal injections of TTX to kittens from the age of 14 days onwards. Subsequent testing at age four to six weeks failed to show any evidence for ocular dominance columns. Since OD columns are normally detectable in kittens by 14 days it appears that the removal of spontaneous activity causes a degradation of this early structure.

### Orientation biased rearing

Whilst there is relative consensus on the effects of visual experience on the development of ocular dominance maps, it remains controversial whether orientation preference maps are similarly affected by restricting the range of orientations experienced early in life. A recent study by Sengpiel et al. [1999] using optical imaging techniques suggests that there is indeed a shift in the orientations represented although representations for orientations that the animal has never been exposed to remain. Their study found that up to twice as much cortical surface area was devoted to the experienced orientation as to the orthogonal one.

## 2.8   Summary

### V1 Receptive Field Properties

To first order, a linear spatio-temporal filter followed by a static non-linearity provides a reasonable mathematical description of the behaviour of simple cells when presented with test stimuli. The character of the filter appropriate for most cells is localised in orientation, spatial position, spatial frequency and temporal frequency, and is well described by taking the temporally modulated Gabor function in equation 2.4. Similarly, complex cells are quite well described by a sum of such (rectified) simple-cell outputs with appropriate spatial displacements and/or shifts in phase.

However, it is also clear that these descriptions have some major lacunae and there is considerable room for more sophisticated treatment. One fairly straightforward extension of these models is the addition of some sort of divisive interactions; doing so allows one to capture some, but by no means all, of the non-classical effects that have been observed.

### Topographic Maps

The spatial organisation of receptive field properties in primary visual cortex has a characteristic topographic map structure for many dimensions of tuning. These maps generally show considerable local order and often exhibit global periodicity. However they generally lack rigid global structure. Statistically consistent relationships between of the maps for different tuning dimensions can be discovered.

**Development of Structure**

The development of the map properties, and the underlying receptive field organisation which these properties express, seems to be a process depending upon *both* 'activity independent' and 'activity-dependent' factors. Furthermore, in the case of activity dependence it seems that visual experience is required for the proper development of some aspects of the normal adult organisation, and that abnormal experience can significantly alter the pattern of development.

**Summary Discussion**

A question which naturally arises is 'what role, if any, is served by the structural organisation of the receptive fields and topographic maps that we see?'. If one neglects the possible importance of volume signalling effects (e.g.: diffusion of retrograde messengers or $Ca^{2+}$), then it appears that topographic organisation can have no *computational* significance — the topology of the connections is all that is operationally important. Diffusive signalling may indeed be important in sculpting the form of organisation actually observed. However, probably much more important are the costs in terms of the volume of 'wiring' used to construct a particular network topology and the costs in terms of signal latency. Consequently, one might reasonably hypothesise that the topographic configuration of different cortical areas arises as a result of having to perform particular computations whilst trying to minimise these costs — in terms of tissue volume and processing time.

This suggests that we might be able to use the known properties of receptive fields and the observed topographic structure, along with other connectivity information and some simple assumptions, to inform our deductions about cortical function. Specifically, by constructing and optimising models that incorporate both specific computational desiderata and aspects of spatial organisation, we might hope to gain new insights and predict new observations or relationships with respect to receptive fields, topographic maps and neural information processing,

# Chapter 3

# Review of Current Models

## 3.1 Introduction

In this chapter we will present some of the main contributions that have been made to modelling the formation of receptive fields and cortical maps in V1. Excellent and detailed reviews of some aspects of the literature have been provided in Dayan and Abbott [2001], Swindale [1996] and Erwin et al. [1995], and rather than duplicate their work we focus on more recent developments; a full and in-depth treatment of the whole field is, however, beyond the scope of this thesis

We begin with an overview of modelling styles, before moving on to discuss at a general level the issue of representation in topographically ordered systems. Subsequently, we move on to review some of the existing methods in the literature. Finally, we discuss issues raised by our consideration of these extant models, and in particular those that pertain particularly to the density estimation paradigm developed in later chapters.

## 3.2 Modelling Overview

A broad spectrum of approaches has been applied to the modelling of receptive fields and topographic maps. Some models mainly focus on receptive fields, some models mainly focus on topographic maps; although with appropriate manipulation and additional assumptions, most receptive field based models can be coerced into saying at least something about topography.

Two conceptually useful dimensions for delineating different methods are: (i) the level or style of modelling applied; and (ii) the types of abstraction utilised. The different approaches that have been taken are not antagonistic, rather they all try to illuminate the same difficult problem from a range of perspectives. Furthermore, the different types of model are able to provide us with different constraints, and this can be very useful when trying to piece together larger parts of the puzzle as whole.

**Modelling Styles**

A main dimension of differentiation is the degree of 'top-down' versus 'bottom-up' design in the model formulation. Generally speaking, 'top-down' models postulate some sort of objective or computational function which they seek to optimise and therefore, in the terminology, of Chapter 1, usually sit mainly at an interpretational level. On the other hand, 'bottom-up' approaches generally make some data-based assumptions about the neurobiological processes involved and then set out to create a description using these elements; this tends to put them mainly at the phenomenological or mechanistic level of Chapter 1. Although clear in their computational function, top-down models are generally looser on biological realism; conversely, bottom-up approaches tend to have greater biological fidelity but are not as perspicuous with respect to computational function. Ideally, one would like to have models that combine the strengths of the two approaches; however, currently no model is wholly satisfying on both counts.

**Types of Abstraction**

A second useful dimension for characterising models is the nature of abstractions applied by a given model, and also the degrees to which they are present. Regarding the nature of abstraction, we might consider the linked sub-themes of: (i) biological abstraction; (ii) computational/algorithmic abstraction; and (iii) abstraction of 'representational semantics'.

In terms of the biological sub-theme, for instance, few models consider cortical micro-circuitry with much accuracy. Rather, the atomic elements of different approaches are be variously thought of as idealised single-neurons, clusters of neurons, or something on the scale of micro-columns or larger — the finer functional details being abstracted within the model. Similarly, properties such as receptive field scatter are normally abstracted away in considerations of maps, and spiking is often abstracted away in considerations of neural response.

Regarding abstraction within computational/algorithmic sub-theme, some models postulate that a particular computational process occurs without giving any details of how it might actually be instantiated.

Abstraction of 'representational semantics' refers to the way in which we interpret elements of the model as corresponding to elements of the world. For example, an important distinction within this sub-theme is the one that can be made between so called 'feature based' or 'low-dimensional' approaches, and 'high-dimensional' approaches. Feature based models abstract away many details of input and receptive field structure, thereby allowing them to concentrate on the topographic structure. High-dimensional approaches are able to entertain richer inputs and representations, but only at greater computational cost and so are often more limited in scale.

## 3.3 General Issues

Before moving to consider any specific models, we first step back and discuss some broadly relevant high-level issues.

Sensory systems give an organism a representation of its external world so we begin by touching upon some aspects of 'representation' in general. This is then followed by a consideration of topographically organised representational systems.

### 3.3.1 Representation

Although seemingly common-sensical, a thorough consideration of many issues regarding representation is rather complex. For instance, answering questions such as "What does it mean for system $A$ to represent system $B$?" at a *deep* level is non-trivial. Works such as that by Bennet et al. [1989] have attempted to formalise a framework in this area, but for our purposes a simpler treatment is more appropriate. Given a system with a set of inputs and a set of internal states, we will take the mappings between these inputs to internal states to be our loose definition of 'representation'. Furthermore, we will take the internal states to be described by a pattern of 'activity' on a set of neural 'units'; the terms used being deliberately vague in order to cover a range of approaches.

#### 'Good' representations

In trying to understand sensory neurobiology, particularly from a top-down perspective, it is commonly postulated that the observed neural developmental processes lead to systems that form 'good' or 'effective' representations. However, these notions are somewhat ill-posed unless augmented by other factors to specify what our metric of 'goodness' should be, and also what it is that should be represented.

Arguably, the most relevant measure of 'goodness' is linked to the evolutionary fitness that a sensory system confers upon an organism in its natural environment, but a complete consideration along these lines is clearly intractable at present. Consequently, simpler augmenting factors have been argued as relevant proxies for neurobiological systems, and these factors are often a key element in 'top-down' approaches. Such objective functions include measures based around: reconstruction fidelity; redundancy reduction; mutual information; Fisher information; temporal smoothness/stationarity & invariance; minimum description length; and probability density estimation.[1] It is also worth pointing out that we can often manipulate and rephrase measures of goodness to translate between the different objectives listed previously; the same 'goal' can often be specified in multiple ways.

Many objectives are also linked in their outcome through a further important issue — the consideration of 'what' is to be represented. We would ideally like to consider the 'natural' environment to which the organism of study is adapted; but again this is difficult and not trivial to define. Some approaches employ simplified or abstracted inputs, whilst others seek to use more realistic inputs. In particular the statistics of so-called 'natural scenes' have been the subject of considerable enquiry (e.g.: Field [1987], Ruderman [1994], Ruderman and Bialek [1994]), and play an essential role in several models. However, on this point we note that there is no *necessary* equation between learning and development and natural scene statistics; it could be evolutionarily programmed. Nevertheless, it does seem highly likely that such stimulus statistics are important — particularly for higher areas.

---

[1]These are simply some examples, this list is by no means comprehensive.

### Representation by neural populations

It is useful to distinguish between the concept of a population of neurons sharing the burden of representing inputs, and the more specific concept of them co-operating (in a non-trivial way) to represent a single instantiation. We will reserve the terms population code or distributed representation to refer to the latter case.

One of the key ways in which a population of neurons can share the burden of representing the world is by showing a *diversity* of responses to different stimuli; put simply different neurons responding in different ways to different inputs. This concept is directly motivated by the experimental data, as well as by considerations of numerous plausible objective functions, and also by basic intuition.

There have been many suggestions concerning the use of population codes in the brain, and Pouget et al. [2000], Dayan and Abbott [2001], and Sanger [2003] provide reviews of some of the literature. However, much of the work that has been done tends to be concerned with 'low-dimensional' representations by populations, i.e.: populations that collectively encode the value of one or a few scalar variables that can be mapped to abstract 'features' of the world. We are generally more interested in what might be termed 'high-dimensional' population codes in which such features are implicitly represented by the activity of a network rather than being explicitly defined. Furthermore, we are interested in sparse population codes — that is population codes in which a small fraction of the total number of units play a significant role in representing a given instantiation. This is precisely the sort of population code that the brain appears to utilise [Baddeley et al., 1997, Vinje and Gallant, 2000, Weliky et al., 2003].

### Overcompleteness

Representation by overcomplete populations is also a particularly interesting and potentially important aspect in understanding sensory systems. As noted in chapter 2, the transformation from the retina/LGN to V1 is overcomplete; furthermore, the total number of 'visual pathway' neurons beyond V1 is greater still. None of the objective functions mentioned previously immediately demand an overcomplete representation, although by formulating the details of the objectives in the right way we can create frameworks which favour such schemes.

One might imagine that, regardless of the precise nature of the computations carried out at various stages of the visual system, it would be useful to be able to perform the same computations at different orientations and scales using the same basic circuitry since, and such a feat seems easier to accomplish with an overcomplete representation than with a minimally complete one. Another motivating factor for overcomplete representations is that they can lead to increased robustness to noise. Finally, there is an intuition that overcomplete representations might facilitate the discovery of higher-order structure by helping to transform higher-order complicated dependencies within the input into simpler dependencies between the elements of the re-representation.

**Hierarchies**

Hierarchies of representation are also strongly indicated by the neurobiological data. As with overcompleteness, there are few objectives that demand hierarchical representations although again some frameworks can be adjusted to readily admit them. However, there is a real need for more principally motivated accounts of how such structure might be learnt, and how the underlying computational goals might be described.

### 3.3.2   Topographic Organisation

Experimental data on topographic maps was reviewed in Chapter 2, and the two key questions that this data prompts are: (i) how does this structure arise; and (ii) why should it be this way?

The study of pattern formation in dynamic (biological) systems is an interesting and beautiful field of study in its own right, and many relatively simple processes can lead to surprisingly intricate outcomes (for example Turing [1952]). This suggests that an understanding of some basic principles of neural development might be sufficient to explain much of the structure that we see.

Viewed from a purely computational perspective, there are few good reason to prefer any particular spatial layout. However, in a reality, physical constraints are clearly important and will affect the survival value of a particular system. One presumes that the organisation we see reflects a subtle interplay between computational desiderata on the one hand, and biophysical constraints on the other hand. Neglecting the potential role of volume diffusion in neural computation, there remains a number of possible restricting factors from which one could motivate particular spatial arrangements. Such factors include the metabolic cost of wiring volume and action potential propagation, and time taken for signal transduction.

By studying mechanisms, computational desiderata, and biophysical constraints in conjunction with functional aspects of representation one can hope to extract more information from the data than by studying representation or maps alone.

### 3.3.3   Topographic Representation: Continuity & Diversity

There appear to be two key factors involved in producing realistic reproductions of the sorts of topographic maps seen *in vivo*. These are (i) generation of spatially local (approximate) continuity of the representational elements; and (ii) global diversity of these representational elements. There are many ways in which these requirements can be met, and this is reflected by the fact that many different approaches are able to account for at least some of the data.

Diversity of representational elements tends to arise naturally in many top-down approaches as a consequence of the choice of objective function. Bottom-up approaches, on the other hand, are often built with explicit mechanisms to cause differentiation in the responses of units, usually through some sort of competitive process.

Arguments motivated from a desire to have population codes and overcomplete representations can be made for the existence of 'population continuity' of response preferences — that is to say the existence of many units with similar representational properties within a population with overall diversity. However, most of the objective functions that have been considered for top-down models have no intrinsic notion of space, so in this sense *spatial* continuity is difficult to motivate from a purely computational perspective.

Top-down models that are concerned with reproducing topographic maps must usually be constrained or coerced in some way to deliver the desired results — for instance by 'spatially' restricting their parameters or by adding spatial considerations to the objective function. Topography in bottom-up models is typically generated by including co-operative mechanisms that spread activity, plasticity effects, or both to nearby units.

Sections 3.4 and 3.5 briefly review some of the proposed models for learning representations and for producing topographic maps. This discussion is structured to roughly cluster: (i) bottom-up models, sub-divided into feature-based approaches and Hebbian learning approaches; and (ii) top-down models employing various objective functions. Our coverage of bottom-up models will be rather brief and we mostly cover activity-dependent approaches; this is to allow us to concentrate on the top-down models that have greater relevance to our work in later chapters.

## 3.4   Bottom-Up Approaches

### 3.4.1   Feature Based Models

**Structural Primitives**

Many of the earliest accounts of topographic organisation simply sought a compact description of the structural forms observed experimentally — prominent examples include the ice-cube model [Hubel and Wiesel, 1974, 1977] and the pinwheel model [Braitenberg and Braitenberg, 1979]. These descriptive accounts are now largely superseded by other more advanced treatments.

**'Noise' Models**

So called 'noise models' offer a slightly more mechanistic approach to feature based modelling [Swindale, 1980, 1982, Rojer and Schwartz, 1990, Grossberg and Olson, 1994]. The basic idea is to start from an array of noisy 'feature-preference vectors' arranged on a two-dimensional cortical sheet. A band pass filter is then applied, which encourages an overall periodicity with neighbouring units having similar feature preferences. This approach has been tried in both the fourier domain and in the spatial domain. More sophisticated spatial domain approaches employ successive convolution operations, and such methods are able to dynamically incorporate the effects of non-linearities such as saturation.

   Despite their abstraction, many of these methods are able to produce very realistic patterns for cortical maps. We interpret the band-pass nature of the filtering (combined in some cases with stabilisation and saturation mechanisms) as providing the impetus for both diversity and continuity of representation. This is interesting since it reinforces our notion that although the patterns we see appear to be very elaborate, they may actually be generated by very simple underlying mechanisms. However, noise models can license little speculation as to what those mechanisms might be, or what purpose they may serve.

**Elastic Net & Self-Organising Map**

Both the elastic net [Durbin and Willshaw, 1987] and the self-organising map (SOM) [Kohonen, 1982, 1983] are competitive feature-based models that have use an input-dependent learning rule to produce topographic maps of feature preferences. They are also both closely related to the 'tea-trade' model[2] [von der Malsburg and Willshaw, 1977, Willshaw and von der Malsburg, 1979], which preceded them and provided the inspiration for the elastic net algorithm [Dayan, personal communica-tion]. We can also view them as competitive Hebbian models that work in a low-dimensional feature space characterising both inputs and response preferences.

---

[2]So called because of a fantastic analogy between neurons, molecular markers, and importers and buyers of tea!

In both models each unit has a 'prototypical' feature vector associated with its receptive field and, when presented with an input, representation in these models is implemented via a two stage process. First, a measure of similarity between the input and the each prototype is computed. Then these similarity measures interact through a (soft-)max function[3], which essentially implements a (stochastic) 'winner-takes-all' competition. It is these competitive mechanisms that are ultimately responsible for endowing the models with response diversity.

Learning in both models is based on these post-competition activities, although the way in which this is done to encourage feature continuity differs slightly between the two models. In the SOM the activity from the 'winning' unit spreads into a local neighbourhood, and this spread of activity is then used as the basis for a Hebbian update rule. In the elastic net, the results of the soft-max competition are directly used in a Hebbian update, but there is an additional term which moves the feature-preference of a unit to be closer to a weighted sum of its neighbours. Although seemingly slight, this difference between the two models actually has important consequences. The elastic net algorithm results from the minimisation of a cost function[4], whereas conversely it can be shown that the self-organising map algorithm does *not* result from a from minimisation of a cost function (although it can be viewed as *approximating* the minimisation of a cost function) [Erwin et al., 1992].

Elastic net and SOM feature-based models have been extremely successful in reproducing realistic looking topographic maps of feature preference, and both have successful been used to make predictions about experimental results; for example the elastic net models by Goodhill and Willshaw [1990], Goodhill and Willshaw [1994], Erwin et al. [1995], Goodhill et al. [1997], and [Goodhill and Cimponeriu, 2000], and the SOM models by Obermayer et al. [1991a] Obermayer et al. [1991b] and Obermayer et al. [1992]. Once again this demonstrates that the underlying principles behind map formation can be captured by elegantly simple methods.

A further advantage of these feature based approaches over the 'high-dimensional' methods that we shall move on to discuss is the fact that it is computationally feasible to implement them on a very large scale. This makes it practical to study the effects of boundaries and different areal geometry on map patterns, as for instance in Goodhill et al. [1997]. Such effects are likely to have a significant influence on the final map patterns seen *in vivo*, and so understanding their influence is important.

---

[3]The elastic net uses a soft-max, the SOM typically uses a hard-max.

[4]As such, we might have place the elastic net in the 'top-down' models section, however we chose to present it here due to the similarity between other feature based approaches and between the Hebbian models we subsequently discuss.

### 3.4.2   Bottom-Up Hebbian Learning

There are many models based around Hebb's seminal hypothesis on how the connections between two neurons adapt as a consequence of correlated activity patterns [Hebb, 1949]. Generally, these models aspire to be closely linked with plausible neurobiology although inevitably speculation and simplification is made; for instance, the level of focus in both time and space is often ambiguous and the 'units' of these models are variously interpreted as cells, clusters of cells or micro-columns.

We have sub-divided our treatment of bottom-up Hebbian models into three classes in a (somewhat artificial) attempt to group together similar approaches based on the complexity of lateral interactions which they entertain. Generally speaking, these lateral interactions are responsible both for providing global competition between the units and thereby generating response diversity, and also for providing local co-operation and thereby generating spatial continuity and topographic organisation of responses.

Lastly, we note that many of the 'top-down' models discussed in section 3.5 result in parameter update rules that have a Hebbian character, in the sense that the objective function gradients for 'weight' updates have a functional form that involves a product of (a function of) 'pre-synaptic' activity and (a function of) 'post-synaptic' activity.

#### Correlational Hebbian Models

In an influential set of papers, Linsker [1986c,b,a] demonstrated a 'modular self-adaptive network' based upon sequential layers of linear processing[5], trained using a Hebbian learning rule, stabilised by weight saturation. Due to the linearity of the network, it was possible to base learning simply on correlation functions for the inputs rather than explicitly compute updates to the unit activities. Each layer of the model was trained separately and sequentially.

The first layer of the model consisted of cells with uncorrelated, 'spontaneous' activity. Interestingly, under appropriate initialisation conditions and settings of free parameters, subsequent layers of the model developed both centre-surround and oriented 'receptive fields'. Also, by adding lateral connections in the final layer, he was able to obtain a representation of orientation which showed some of properties (pinwheels and continuity) seen in biological maps.

One of the more significant aspects of Linsker's model was the fact that it demonstrated Hebbian learning rules could lead to symmetry-breaking and the formation of oriented structure in the absence of correlated inputs. This speaks to the (still ongoing) controversy about whether and what kind of inputs are required for the development of early cortical map and receptive field properties.

---

[5]except for the final layer

Miller and colleagues have also implemented a number of 'correlation-based' Hebbian models that have been able to reproduce many aspects of receptive field development and of the joint maps for ocular dominance and orientation specificity. (See Miller et al. [1989], Miller [1992, 1994], Erwin and Miller [1998], and related work.) An appealing feature of their approach is that its relative simplicity and linearity make it amenable to analytical dissection. Consequently, they have been able to stipulate some of the general conditions under which their model will lead to various types of behaviour; such general analysis lends itself more easily to, for instance, experimental comparison.

### Competitive Hebbian Models

There are several models that are, essentially, *high-dimensional* implementations of the self organising map[Kohonen, 1982, 1983]. Obermayer et al. [1990] uses simplified inputs consisting of elliptical activity patches to produce a reproduction of orientation selectivity in a topographic map, and Goodhill [1993] uses a high-dimensional SOM model to give an account for ocular dominance using simple, correlated random dot patterns. In more recent work, Piepenbrock and Obermayer [2001] use inputs consisting of digitised natural scene images to produce simple-cell like receptive fields and a crude map of orientation preference.

### Recurrent Dynamic Hebbian Models

In pioneering work by von der Malsburg [1973], a simple model of a cortical orientation processing consisting of a hexagonal array of excitatory ($E$) and inhibitory ($I$) units was proposed. The units were arranged such that $E$ cells excited neighbouring $E$ cells and, via the $I$ cells, inhibited those units at larger distances. As such this was one of the first simulations to incorporate the now common 'Mexican-hat' style of interaction with short range excitation/co-operation and long range inhibition/competition.

This model was initially applied to inputs consisting of oriented patches of activity on a hexagonal array and successfully delivered a simple map of orientation preference. Subsequent modifications [von der Malsburg and Willshaw, 1976, Willshaw and von der Malsburg, 1976] treated (separately) retinotopic refinement using inputs consisting of localised clusters, and ocular dominance using two 'retinae' that had anti-correlated activity.

This early approach was very influential and incorporates the key elements seen in many subsequent models — namely feedforward excitation, competitive and co-operative network interaction through structured lateral connections, and a stabilised Hebbian learning rule.

A more recent collection of models embodying similar principles of dynamic lateral interaction are those by Miikulainen et al. [Sirosh and Miikkulainen, 1993, 1995, Miikkulainen et al., 1997, Choe and Miikkulainen, 1998, Bednar and Miikkulainen, 2003]. These authors have managed to achieve impressively realistic looking results that give some of the most comprehensive reproductions to date of a wide range of receptive field and topographic map phenomena. Also, significantly, their work is one of the few to include explicit learning of lateral interactions as well as feed-forward ones.

### Sliding Threshold Hebbian Models

There have been several models based around the self-stabilising Hebbian learning framework originally proposed by Bienenstock, Cooper and Monroe [Bienenstock et al., 1982]. It was originally formulated in a somewhat *ad hoc* bottom-up manner based on speculations about biological plasticity mechanism, hence its inclusion in this section. However, BCM-like learning rules can also be shown to arise as a consequence of optimising an objective function [Intrator and Cooper, 1992, Intrator et al., 1993]. Furthermore, this objective can be shown to have close connections to the statistical techniques such as projection pursuit [Friedman and Stuetzle, 1981, Friedman, 1987] which seek to find 'interesting' (eg: non-gaussian) directions in a data set. (Projection pursuit itself can, in some cases, be related to Independent Components Analysis, an approach which we discuss in section 3.5.4.)

Shouval et al. [1997] demonstrated that a network of BCM units with 'Mexican-hat' recurrent interactions trained on digitised patches of natural scenes could develop receptive fields that show a qualitative similarity to V1 simple cells. Furthermore, the spatial layout of receptive field properties shows some evidence of ocular dominance structure and of patterned orientation maps qualitatively resembling those seen *in vivo*.

### Stabilisation of Hebbian learning

Many Hebbian learning models would suffer from unstable growth due to intrinsic positive feedback were it not for additional stabilisation mechanisms, which are somewhat arbitrarily imposed. (Notable exceptions are those Hebbian models that can be related to an objective function, such as the elastic net or BCM approaches.)

Hebbian models are usually stabilised by the imposition of some kind of constraint on the synaptic weight vectors, perhaps in conjunction with upper and/or lower saturating bounds for each individual weight element. This introduces coupling and competition between different 'synaptic weights' and can have a profound influence on the outcome of learning. In particular, we can identify the following four factors as being of vital importance.

- Whether the weight constraint is applied to the pre-synaptic cell (i.e. constraining a function of efferent synapses from cell), the post-synaptic cell (i.e. constraining a function of the afferent synapses to a cell), or both.

- The nature of the weight vector constraint. For example, constraining the sum of weights, the sum of squares of weights, or the sum of absolute values of weights. All these constraints are somewhat abstracted from the neurobiological data [Turrigiano, 1999, Turrigiano and Nelson, 2000]

- The way in which the constraint is dynamically enforced. The most common methods are subtractive and multiplicative normalisation. In subtractive normalisation the constraint is enforced by the subtraction of an appropriate amount of a fixed vector; in multiplicative normalisation the constraint is enforced by modifying each weight by an amount proportional to its present value.

  Subtractive normalisation tends to generate 'stronger' competition than multiplicative normalisation. Also, whereas multiplicative leaves the direction of the synaptic weight vector unchanged, subtractive normalisation alters it.

- The presence of upper or lower limits for the values that each weight may attain, and the relative values of such limits to any constraints on the entire weight vector.

Miller and Mackay [1994] and Goodhill and Barrow [1994] have analysed the precise effects of some of these factors in simple cases.

## 3.5 Top-Down Approaches

### 3.5.1 Wire-length Minimisation

As noted earlier in this chapter, and also in chapter 2, a natural motivation for the observed topographic organisation would be to minimise or reduce the metabolic costs associated with wiring length/volume, and signal propagation. Several authors have attempted to address this issue directly (e.g.: Mitchison and Durbin [1986], Mitchison [1991, 1995], Todorov et al. [1995], and Koulakov and Chklovskii [2001]); however, the task is fraught with difficulty. Taking into account full details of cellular branching and morphology over the scales required would take inordinate amounts of computation, and approximations are clearly necessary. Moreover, without a principled understanding of what computational operations might be taking place we are on loose ground when trying to minimise wiring costs — we have no sense of what should wire to what. Again, of course simplifications can be made and a reasonable assumption might be that cells with similar receptive fields should have tight connectivity. In this light, we can actually view feature based models such as the elastic net as trying to minimise some approximate wiring cost[6], subject to the demand that diverse coverage of the feature space also be provided [Mitchison, 1995].

---

[6]Indeed the elastic net is an efficient and effective algorithm for finding good solutions to travelling salesman problems.

## 3.5.2 Mutual Information

Maximisation of the mutual information [Shannon, 1948] between the inputs and representational elements of a sensory system is another objective function that has been proposed as being relevant when considering sensory development. Many of these ideas have their roots in the seminal concept of 'redundancy reduction' by Barlow [1961, 1989] (although see Barlow [2001] for some updated thoughts).

In an influential series of papers, Linsker [Linsker, 1987, 1988b,a, 1989] explicitly proposed the concept of 'infomax' as a possible objective for sensory organisation. Linsker [1989] presented a model trained by maximising the mutual information between the inputs patterns and the 'winning' unit from a set of outputs engaged in locally co-operative, stochastic competition. The activations and interactions in his model are such that all the gradient information necessary to maximise the objective is locally available to the units in the network, and furthermore the step-wise updates required for learning have a Hebbian form. With appropriate interactions this algorithm is able to deliver ordered, 'neighbourhood preserving' maps.

This work lead the way for other information-theoretically inspired approaches such as the infomax framework put forward by Bell and Sejnowski [1995], which we discuss in section 3.5.4.

## 3.5.3 Temporal Invariance-Seeking Models

There are a range of models that depend upon the *temporal* as well as the spatial properties of inputs. Broadly speaking, many of these are informed by the hypothesis that interesting features in the world vary on a timescale that is much longer than the timescale on which individual sensory detectors vary. (For example, consider an object moving across the visual field. The object's identity remains fixed, the object's co-ordinates may vary quite slowly, whilst elements of the retinal representation may vary on a much more rapidly.)

We do not consider such models in detail here since our focus in this thesis does not entertain spatio-temporal inputs. Instead we provide a brief, annotated bibliography and direct the reader to the following references for more information.

Inspired by the work of Fukushima et al. [1983] and Lecun et al. [1989], Foldiak [1991] was one of the early proponents that learning invariant, or slowly changing, aspects of a spatiotemporal sequence could lead to a useful representational system and explain properties such as complex cell receptive fields. Stone [1996] proposed that perceptually salient aspects of the world vary smoothly over time, and proposed seeking representations that have small short term variance (whilst maximising long term variance.) Similar proposals are also raised by the work of Kayser et al. [2001], Einhauser et al. [2002], and also Wiskott and Sejnowski [2002]. Hurri and Hyvarinen [2003a,b] propose an object that maximises 'temporal coherence', which essentially amounts to looking for representations in which the changes in the representational elements is sparsely distributed over time; this also implies stability. Along slightly different lines are ideas about 'predictive coding', which explicitly seek to model temporal sequences such as the work of Rao and Ballard [Rao and Ballard, 1995, Rao, 1999]. Many of the aforementioned ideas seem to be promisingly tied together, along with considerations of topography and the ICA density models that we discuss next, in the recent 'bubbles' framework proposed by Hyvarinen et al. [2003].

### 3.5.4   Density Estimation: Distributed Causal Models

We now consider a collection of models that posit density estimation as an objective function. Explicit density estimation as a means of learning sensory representations is a relatively new idea (for good introductory discussions see chapter 10 of Dayan and Abbott [2001]). The basic idea is as follows. We assume that there is a well defined distribution over the inputs that the given sensory system will encounter.[7] In many domains there is considerable statistical structure in this distribution of inputs. A 'good' representational system in the density estimation paradigm is one which is able to give a compact but accurate account of this statistical structure. The internal elements combine to specify the probability of a particular input with the goal being that these probabilities match the empirical distribution as well as possible. Furthermore, a given input induces some state on these internal elements — it is this induction that forms the basis of the mapping from 'input' to 'representation'. We can then consider this representation process as an analogue of representation formation in biological systems. For instance, we can consider the 'receptive fields' of the internal elements in a model and then compare these to experimental observations.

Later in this thesis we will develop a framework centered around energy-based or undirected models, however most of the approaches explored in the literature so far have used causal density models. In most of these approaches a distributed set of factors combine to cause an input. Distributed causal models have mainly been applied to general representational learning without consideration for topographic organisation, although more recent years have seen their application to cortical maps.

---

[7]For instance, in the case of vision this might consist of the distribution of pixellated images of the world — clearly not all scenes are equally likely, and many are extremely unlikely. Consider, for example the patterns produced when a TV set is tuned to 'static'.

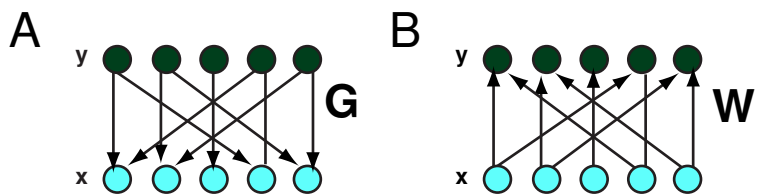

**Figure 3.1:** ICA graphical model and infomax schematic. (A) The network for square, noiseless infomax. (B) Causal graphical model for ICA/sparse coding. In a square noiseless case there exists an infomax model isomorphic to every causal model, in terms of optima.

## Independent Components Analysis (ICA)

Several convergent themes, when applied to 'natural' images and sounds, lead to essentially the same set of unsupervised representational learning models. (In addition to our coverage, other unifying reviews of these issues can be found, for example, in Olshausen [1996] and Lee et al. [2000]).

Both the information maximisation approach of Bell and Sejnowski [1995, 1997] and the sparse coding approach of Olshausen and Field [1996, 1997] discussed can be subsumed within the statistical framework of independent components analysis (ICA)[8], which we now discuss. ICA is schematically illustrated by the causal graphical model [Jordan, 1998] in figure 3.1 (A).

Figure 3.1 (A) is meant to imply the following statistical relationships,

$$p(\mathbf{s}) = \prod_i p(s_i) \tag{3.1}$$

$$p(\mathbf{x}|\mathbf{s}) = \mathcal{N}(\mathbf{Gs}; \Sigma) \tag{3.2}$$

The $\{s_i\}$ are the statistically independent causes which are linearly combined, and have Gaussian noise added, to give the observed inputs $\mathbf{x}$. $\mathcal{N}(\mathbf{Gy}; \Sigma)$ defines a gaussian distribution with mean $\mathbf{Gy}$ and covariance $\Sigma$.[9] In the general case the dimensionality of $\mathbf{s}$ need not be the same as the dimensionality of $\mathbf{x}$.

## ICA: Infomax

The Infomax framework proposed by Bell and Sejnowski was subsequently recognised as implementing a particular type of ICA. Bell and Sejnowski [1995, 1997] maximise, with respect to feedforward weight parameters, $\mathbf{W}$, the mutual information between a set of inputs, $\mathbf{x}$, and a set of outputs, $\mathbf{y}$, which are linear combinations of the inputs followed by a bounded, monotonic point-wise non-linearity $g(\cdot)$ (typi-cally the tanh or sigmoid function).

---

[8]The term ICA was originally proposed by Herrault and Jutten [1986]. Subsequently, Comon [1994] played an important role in developing and 're-introducing' the technique. The causal graphical viewpoint that we express here (and which is pretty much standard at present) is largely popularised by Pearlmutter and Parra [1996], Mackay [1996], Cardoso [1997].

[9]It is also common to consider the 'noise free' case, i.e. $\Sigma = 0$.

$$\mathbf{y} = g(\mathbf{W}\mathbf{x}) \tag{3.3}$$

$$\mathbf{W}_{\text{opt}} = \arg\max_{\mathbf{W}} I(\mathbf{x}; \mathbf{y})$$

$$= \arg\max_{\mathbf{W}} \left( H(\mathbf{y}) - H(\mathbf{y}|\mathbf{x}) \right)$$

$$= \arg\max_{\mathbf{W}} H(\mathbf{y}) \tag{3.4}$$

$I(\mathbf{x}; \mathbf{y})$ is the mutual information between $\mathbf{x}$ and $\mathbf{y}$, $H(\mathbf{y})$ is the entropy of $\mathbf{y}$, and $H(\mathbf{y}|\mathbf{x})$ is the conditional entropy of $\mathbf{y}$ give $\mathbf{x}$. Their model is shown schematically in figure 3.1 (B).

In Bell and Sejnowski's original framework, $\dim(\mathbf{y}) \leq \dim(\mathbf{x})$; equality was normally assumed, i.e. a 'complete' or 'square' model. Recently, Shriki et al. [2001] proposed an overcomplete version of Infomax, (i.e. $\dim(\mathbf{y}) > \dim(\mathbf{x})$) which we will cover in Chapter 4.

The last re-writing in equation 3.4 is obtained as a consequence of the deterministic mapping between $\mathbf{x}$ and $\mathbf{y}$. Thus we see that information maximisation in this noise free setting is equivalent to maximising the entropy of the output units. This entropy is maximised if every output unit is as independent as possible and uniformly distributed over its range. Consequently, performing infomax in such a model is equivalent to searching for statistically independent directions in input space whose cumulative marginal distributions match the non-linearity used in the output neurons. In the case of a tanh-like function, this corresponds to looking for independent (as far as possible) directions that have sparse, heavy tailed distributions.

Bell and Sejnowski [1997] show that when applied to digitised patches of natural images such an algorithm delivers receptive fields that are remarkably reminiscent of simple cell receptive fields. We show examples of their results in figure 3.2 (A).

**ICA: Sparse Coding**

Field [1987, 1994], and related work, argue directly for sparse, statistically independent representations as being useful in characterising the structure of 'natural' inputs. Olshausen and Field [1996] demonstrate that a basis set for natural image patches optimised for a particular measure of sparseness delivers a set of vectors that resemble simple cell receptive fields, as shown in figure 3.2 (B). Their sparseness measure translates directly into the (log of) a sparse source prior for ICA, whilst their squared-norm reconstruction cost corresponds to Gaussian noise.

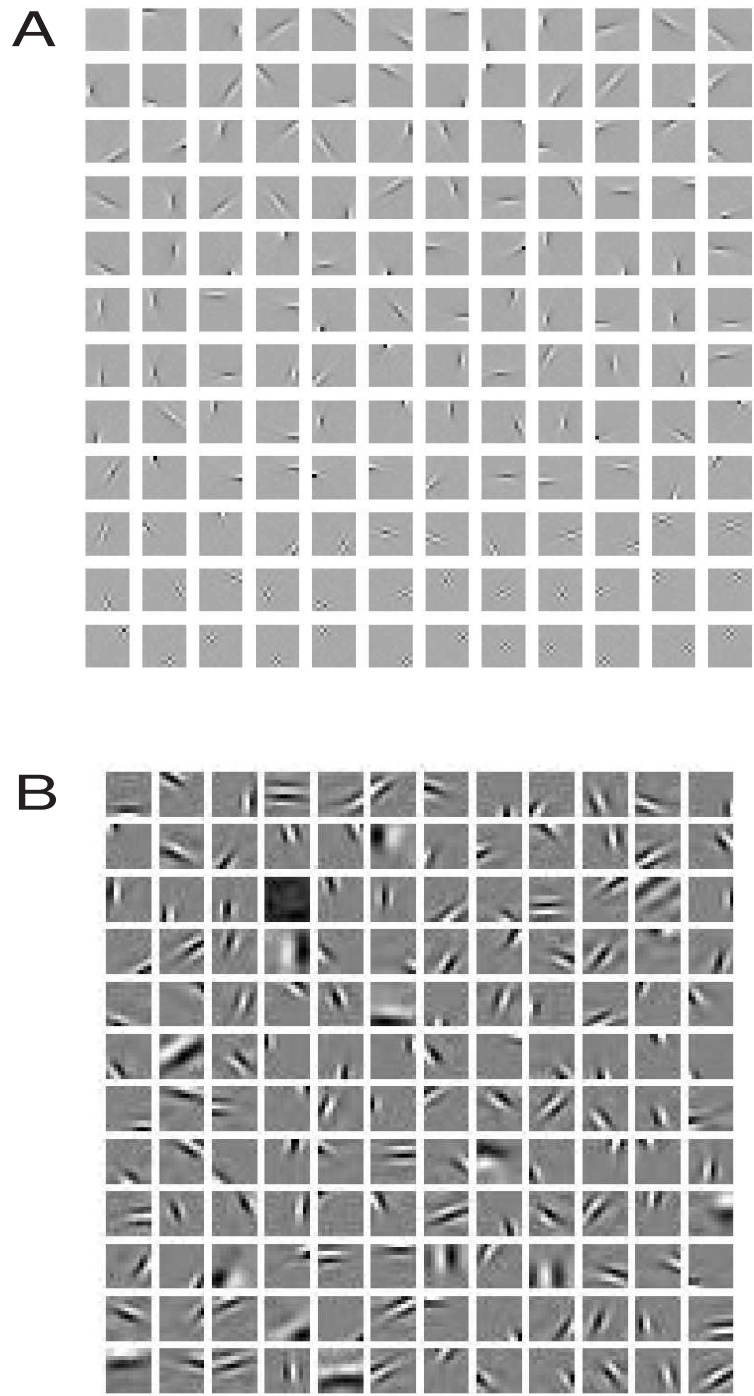

**Figure 3.2:** ICA applied to natural image patches. (A) Example filters from infomax-ICA model [Bell and Sejnowski, 1997]. (B) Example basis functions from a sparse coding model [Olshausen and Field, 1997].

**Maximum Likelihood Parameter Estimation**

Maximum likelihood parameter estimation in ICA models consists of adapting the mixing matrix $\mathbf{G}$, and also perhaps the parameters of the priors $p(y_i)$, so as to maximise the likelihood of an observed data-set. In the square noiseless case the likelihood can be directly estimated. In such a case, we can also relate the optimal settings of generative weights $\mathbf{G}$ to the optimal settings of the feedforward weights $\mathbf{W}$ in a square, noiseless infomax network in which the non-linearity on the outputs is given by the cumulative density function for the priors in the causal ICA model. In such cases $\mathbf{G}_{\text{opt}} = \mathbf{W}_{\text{opt}}$, [Lee et al., 2000].

In the general case in which we have noise and/or overcompleteness we have the problem that the system is not fully determined, and the mapping between sources and outputs need not be one-to-one. Exact inference in these situations is often intractable, so approximations are required. Learning in noisy and/or overcomplete models is usually carried out using Monte Carlo methods or some variant of the EM algorithm [Baum et al., 1970, Dempster et al., 1977, Neal and Hinton, 1998].

### 3.5.5 Density Estimation: Hierarchical, Distributed Causal Models

Learning in hierarchical models has proved to be a much more challenging task than learning a single representational layer. As with the overcomplete cases of a single layer representation exact inference is often intractable, however the problem is usually far worse in hierarchical models and effective approximation methods can be difficult to construct. Understanding learning, inference and representations in hierarchies is an important problem, not least because the brain's sensory systems appear to be organised in this way, and developing better tools to further this understanding is an active area of research.

**RGBN: Non-linear sparse distributed representations**

Hinton and Ghahramani [1997] present a hierarchical model called a Rectified Gaussian Belief Network (RGBN). This is a multi-layer causal generative model in which the unit states are either positive real values or zero — the probability mass at zero being finite, and therefore potentially leading to very sparse distributed representations. Several approximate schemes for learning are proposed and the model is shown to perform well on a selection of toy examples. Hinton and Ghahramani [1998] present a modification that introduces (undirected) Mexican-hat lateral interactions to the generative model, and show that when trained on a toy stereo disparity problem the model is able to form a topographically ordered representation with disparity tuned units.

### Hierarchical extensions to ICA: complex cells and topography

A number of papers have built on the success of the ICA framework in reproducing the properties of V1 simple cells based. In particular, Aapo Hyvarinen and colleagues have made several important contributions.

Hyvarinen et al. [2001] and Hyvarinen and Hoyer [2001] develop the ICA model further by adding an additional 'second layer'. In doing so they are able to capture aspects not only of simple cell receptive field formation, but also of complex cell properties and topographic maps. We describe this model more completely in section 6.10 when we compare it to our own work. In brief, they consider a generative model in which the variances of the 'independent' components are coupled, being generated themselves by a stochastic process. These variance components turn out to have interpretations as complex cell responses. Further, by constraining the parameters of the model they are able to induce topographic orderings of the learnt receptive fields.

In related work, Hoyer and Hyvarinen [2001] apply a non-negative ICA model to the 'complex cell' outputs of the type obtainable from a (pre-trained) two-layer ICA model. Their results show that the non-negative features that develop appear to code for elongated 'contours', with additional evidence for end-stopping and other phenomena.

Karklin and Lewicki [2003] also recently presented a slightly different, sequentially constructed, hierarchical extension to ICA. For a fixed set of ICA basis functions they train a generative model for the log variances of each source; this generative model is itself an ICA model. Once again, this allows dependence amongst the sources to be introduced by coupling their variances. (In this respect, their work is similar to models proposed by Hyvarinen and also to approaches that we propose in Chapter 6).

### Helmholtz Machines

Dayan et al. [1995] propose an interesting twist to the graphical modelling paradigm. The Helmholtz machine employs, effectively, two different sets of model parameters acting on the same[10] set of nodes — those of a causal generative model and also those of a corresponding recognition model. The causal generative model specifies a generative distribution over the input space; the recognition model aims to perform the statistical inverse of this model by taking inputs and (probabilistically) mapping them to a posterior distribution of causes in the generative. This separation of parameters was proposed in an attempt to make tractable the otherwise problematic inference of latent causes given an input.

Learning in Helmholtz machine models normally follows an *approximate* version of the expectation maximisation (EM) algorithm [Dempster et al., 1977, Neal and Hinton, 1998]. The particular form of approximation used makes learning more tractable, although it does remove the guarantees of improvement that the EM algorithm normally gives. Nevertheless, the Helmholtz machine yields an attractive framework in which perceptual inference can be performed rapidly (using a separate recognition model), and in which a hierarchical, non-linear distributed representation can be learned using local rules. For more details about the Helmholtz machine and the learning algorithm (known as 'wake-sleep'), see Dayan et al. [1995], Hinton et al. [1995], Dayan and Hinton [1996], Dayan [1999] and Dayan and Abbott [2001].

---

[10]Although it is possible to have nodes in one network that do no have counterparts in the other; in particular extra nodes in the recognition network have been suggested.

## 3.6 Discussion

### Feature Based Models

Feature based models are capable of explaining much of the experimental data, and a particular strength is their (relatively) low computational complexity — this allows them to be implemented on much larger scale than high-dimensional models and therefore allow us to consider more global effects. However, there are several aspects of these models that are less desirable. For instance one has to decide, *a priori*, which features are important and also upon the functional form of the feature matching process. Also, despite their ability to give insight into the formation of topographic maps they are able to say very little about the processes of representation within those maps.

### Hebbian Learning

Many different Hebbian based models seem capable of explaining sizeable portions of the basic data on V1 simple cell receptive fields and for the gross structure of topographic maps. However, although models such as this are extremely useful in demonstrating the type of self-organisation that can be achieved using simple principles, the purely bottom-up approach has its drawbacks. In particular, whilst we can give a statistical characterisation of the basic Hebb rule and we can sometimes analyse the dynamic properties of network models [Miller and Mackay, 1994, Goodhill and Barrow, 1994, Dayan and Abbott, 2001], it is generally difficult to get a complete understanding of the computational functions such networks carry out — especially in more complicated cases. For instance, consider the family of LISSOM models (Miikkulainen et al. [1997] and related work). These arguably present some of the most comprehensive Hebbian approaches to describing receptive field and map properties thus far. However, a proper analysis of their development and computational function is lacking, and it is unclear exactly what their plasticity rules and the resulting network actually achieve *computationally*.

### Mutual Information & Redundancy Reduction

The concept of maximising information transfer and the related ideas of redundancy reduction are intuitively appealing — that a system should try to encode as much information as possible about its inputs makes a lot of sense. However, we believe that there are several issues that information maximisation alone does not address. One of the main problem with efficient coding hypotheses is that they ignore *computation*. The brain does not simply re-represent information in an efficient form, nor does it seem to simply remove redundancy. Rather, as suggested in Barlow [2001], we might look for ways in which redundancy in inputs is *transformed* — perhaps making more explicit the complicated underlying structure of an input ensemble by a series of transformations. Computations need not preserve information at each step, indeed they might be *expected* to be lossy. Such ideas on 'redundancy transformation' seem to fit rather naturally within the density estimation paradigm, since these methods can be viewed as performing statistically justified structural transformations and decompositions.

### Causal Density Models

The paradigm of *causal* density estimation, and ICA and its descendants in particular, has been remarkably successful in its ability to describe and explain many of the features seen in primary visual cortex — not only simple cell receptive fields, but also with most recent developments starting to capture complex cells and other higher order structures, as well as maps for orientation and retinotopy.

However, despite these successes there are several problems with the causal graphical model approach. In many models that might be of interest, the process of statistically inferring causes for an input is intractable. Such inference is necessary for proper learning in causal models, as well as for what we interpret as representation-formation in the 'developed' model. The construction of efficient, effective and rapid approximations schemes is a challenging issue and remains an ongoing problem in both computational neuroscience and machine learning.

An example of current techniques to overcome these difficulties is highlighted by Olshausen and Field [1997], which uses an iterative network settling scheme to arrive at a MAP estimate for the posterior distribution. However, the problems of intractable inference become further exacerbated as we move to greater degrees of overcompleteness or to deeper hierarchies, and good approximations become harder to achieve.

One way of dealing with the problem of learning in overcomplete hierarchies is to learn each layer sequentially, and many models do resort to this. However, such sequential learning can be somewhat unsatisfactory when it comes to understanding the operation of the full model as a whole. An alternative and inspirational approach is suggested by the Helmholtz machine, which learns a separate model to invert the whole generative process and provide the necessary inferences. This kind of 'direct recognition' seems very appealing, and fits nicely with the fact that perceptual inference can be very rapid. However, thus far, the Helmholtz Machine has proved to be disappointingly ineffective in real-world implementations.

Lastly, we note that there are some statistical relationships that one might expect to find in the world that are not especially amenable to description by a causal model. For example, constraint based relationships or contextual effects can be difficult to specify using a purely causal framework

## Summary

Whilst there have been many modelling triumphs with regard to describing receptive field and topographic map structure in the visual brain, no model is able to describe all the current data and there is clearly scope for considerable improvements and exploration of new methods. In particular there is a need for a principled computational approach that is able to deal with overcomplete and hierarchical representations in such a way as to make the process of inference relatively fast, accurate and tractable. There is also room to create a richer statistical description of the relationships between elements of a representation.

In this thesis we will favour the framework of probability density estimation as a paradigm since we feel it is one of the more promising candidates for helping us to understand neural computation and hierarchical, non-linear organisation in sensory systems — the ability to take us beyond V1, to V2 and to other higher sensory areas.

However, rather than pursue an approach employing causal generative models we develop machine learning methods that allow us to explore a non-causal probabilistic framework for modelling receptive fields and topographic maps. Relative to the causal framework, we find this has some advantages (but also some disadvantages). In particular, we will suggest non-causal models in which inference can be rapid and direct, which easily produce overcomplete population codes, and which in principle can be extended to deep hierarchies.

# Chapter 4

# Energy-Based Models

## 4.1 Introduction

In this chapter we put forward a framework for unsupervised learning in energy-based density models. Essentially this combines an energy-based viewpoint of probability distributions with a general and versatile way of constructing energy functions, and a set of tools for optimising model parameters.

Any probability distribution over the configurations, $\mathbf{x}$, of a set of variables $\{x_i \in X\}$ can be written as a 'Boltzmann distribution' by constructing an appropriate 'energy function', $E(\mathbf{x})$. Similarly, subject to rather mild conditions[1] we can use arbitrary energy functions to define probability distributions *de novo*. The functional form of the Boltzmann distribution is

$$p^\infty(\mathbf{x}) = \frac{e^{-E(\mathbf{x})}}{Z} \qquad (4.1)$$

where the denominator, $Z$, is a functional of $E(\mathbf{x})$ and acts as a normalising constant to ensure the distribution integrates to 1. (We will explain the choice of the $\infty$ superscript in $p^\infty(\mathbf{x})$ later.)

The form of expression in equation 4.1 may be considered somewhat axiomatic, since simply taking $E(\mathbf{x})$ to be the negative log of the configuration probabilities parameterised in any other way would satisfy the requirements. However, there are a large number of distributions for which the Boltzmann is a particularly natural means of expression. These include the Exponential Family, Markov Random Fields (MRF's), Maximum Entropy models (MEM's or maxent models), and Products of Experts (POE's) to name but a few. Indeed all models that can be conveniently represented as a factor graph [Kschischang et al., 2001] (which contains as a subset those models that can be expressed as an undirected graphical model) can be neatly expressed as a Boltzmann distribution.

---

[1]If $X$ has compact support, we simply require $E(\mathbf{x}) > -\infty$. If $X$ has unbounded support then we also require that $\int e^{-E(\mathbf{x})} d\mathbf{x} < \infty$. These conditions ensure that the distribution is finite everywhere and can be normalised.

The normalising constant, $Z$, is commonly referred to as the *partition function*, and equation 4.1 is often also called the 'equilibrium distribution' — both acknowledging the origins of the Boltzmann distribution in statistical physics. In case of continuous variables, Z is given by the integral

$$Z = \int e^{-E(\mathbf{x}')} d\mathbf{x}' \tag{4.2}$$

and similarly by a sum in the case of discrete variables. In the general case, and also for many models of interest, it is analytically and computationally intractable to compute $Z$ — the necessary sums and integrals often do not have closed form solutions, and in the discrete case the number of terms in the sum grows exponentially with the number of variables. This fact can make energy-based models difficult to work with and may present problems with respect to learning or inference.

This chapter elaborates a framework for energy-based Models (EBM's), presents an approximate algorithm for fitting them to data and explores the relationships between EBM's and particular examples from other model frameworks.

**Original Contributions**

The main original contributions in this chapter are: (i) the proposal of the flexible energy-based model framework, highlighting the novel concept of *deterministic*, as well as more familiar stochastic latent variables; (ii) the clarification of the mathematical connections between this perspective and other modelling approaches; (iii) the proposal of the use of the Hybrid Monte Carlo technique in conjunction with the contrastive divergence algorithm; and (iv) an exposition of possible problems with the contrastive divergence procedure. Parts of this chapter appear, in earlier forms, as an article in the Journal of Machine Learning: Special Issue on Independent Component Analysis [Teh et al., 2003], co-authored with Yee-Whye Teh, Max Welling and Geoffrey Hinton.

## 4.2   Mathematical Preliminaries

**General Framework**

Although we may write *any* probability distribution as an energy-based model, for the most part in this thesis we will consider particular types of distribution. Specifically those for which the energy function can be expressed as a sum of fairly simple terms, such as

$$E(\mathbf{x}) = \sum_{i=1}^{M} f_i(\mathbf{x}; \theta_i) \tag{4.3}$$

where we have a set of functions, $f_i$, which may depend on some or all of the variables. Each function, $f_i$, is parameterised by a parameter set $\{\theta_{ij} \in \theta_i\}$. Additionally, we will concentrate upon models in which the number of terms, $M$, in the sum 4.3 is comparable to or greater than the number of random variables, i.e.: $M \geq \dim(\mathbf{x})$. The enormous power and flexibility of the energy-based framework comes from the fact that we are able to cope with (almost) arbitrary choices for the functional form of the $f_i$ — the only restriction is that the corresponding Boltzmann distribution be normalisable.

### Two types of latent variable

We note here that some of the variables, $x_i$, may be considered latent (a.k.a. hidden). Latent variables are usually considered as being stochastically related to observables and/or one another, however we will find it useful to consider the concept of 'deterministic latent variables' i.e.: latent variables that are given by deterministic functions of other variables. The existence of latent variables is easily incorporated within the energy-based framework and can be used to increase the expressive power of a model. Note that in the case of stochastic latent variables, whilst the overall model would have a description in terms of energy functions such as 4.3, the marginal distribution over visible variables may no longer be easily expressed as a sum of simple terms.

If one so wishes, deterministic latent variables can be viewed in an equivalent way to stochastic ones by considering adding terms to the energy function that are infinite if particular relationships are not satisfied. For instance, consider that we partition the variables $\{x_i\}$ into a set of visible variables $\mathbf{x}_v$ and a set of latent or hidden variables $\mathbf{x}_h$. Now consider a function $G(\mathbf{x}_v, \mathbf{x}_h)$ which is non-negative everywhere, and which is zero if and only if $\mathbf{x}_h = g(\mathbf{x}_v)$ where $g(\cdot)$ may be an arbitrary mapping from $\mathbf{x}_v$ to $\mathbf{x}_h$. Now consider the energy function given by,

$$E'(\mathbf{x}) = \beta G(\mathbf{x}_v, \mathbf{x}_h) + \sum_{i=1}^{M} f_i(\mathbf{x}; \theta_i) \qquad (4.4)$$

where $\beta$ is a positive constant. For finite $\beta$ we have a model with stochastic latent variables. However, if we let $\beta$ go to infinity then we have a related, but subtly different model whose latent variables are deterministic and are described by the functional relationship $\mathbf{x}_h = g(\mathbf{x}_v)$. Deterministic latent variable models arguably have some advantage over models with stochastic latent variables because representational inference in deterministic models is trivial. Whereas inferring stochastic latent variables usually requires us to have access to a partition function, or alternatively to take a variational or Monte Carlo approach, inferring deterministic latent variables is direct and simply requires us to compute the specified functional mapping. As an aside, we also note that the procedure used to infer deterministic variables, i.e. taking a functional mapping to a single value, is actually not so different to what is commonly performed as an approximation with stochastic variables

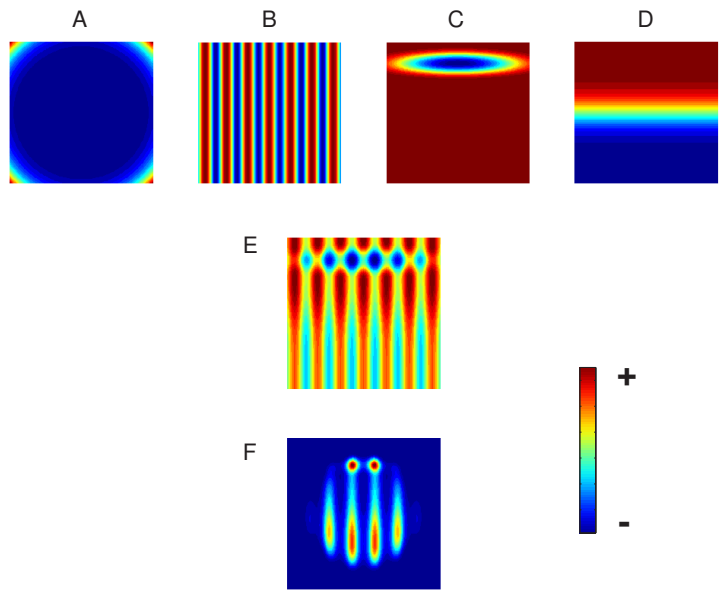

**Figure 4.1:** Figure illustrating the piecewise construction of an energy function and the corresponding probability distribution. The top panels A-D depict 4 energy function components, shown in the $(x, y)$ plane with the energy level shown by a color scale. The centre figure, E, shows the sum of functions B-D (function A is essentially a confinement function, and if made part of the sum the very large values at the edge of the plot would obscure the picture.) Finally, panel F shows the resulting probability distribution. The function forms used were: (A) $(x^2 + y^2)^2$ (B) $\sin x$ (C) $1 + e^{-\frac{1}{2}\left(\left(\frac{x}{\sigma_x}\right)^2 + \left(\frac{y - \mu_y}{\sigma_y}\right)\right)}$ (D) $\frac{1}{1 + e^{-y}}$. Also, note that much of the structure in the high energy regions is much less visible when we exponentiate and move to probability space.

when we make a MAP estimate for the posterior.

Stochastic latent variable models are discussed in more depth and put into practice in Chapter 5, whilst Chapter 6 considers a model with deterministic latent variables. For the remainder of this chapter we will assume that the variables in our energy-based models are all either deterministic latent variables or observable.

Finally to give a feel for the sorts of things that are possible with the general approach, figure 4.1 illustrates a simple example with a set of energy functions and the corresponding probability distribution. These energy functions have been arbitrarily chosen and are meant only to illustrate the flexibility of our framework.

## 4.3 Connecting EBM's With Other Approaches

The energy-based models put forward in this chapter can be related to several other approaches in the machine learning literature. On a general level, a close connection can be made between our energy-based viewpoint and the factor graph framework. (Translations between factor graphs and many other graphical models can then easily be made [Yedidia et al., 2002].) An example of a factor graph is given in 4.2. The circles represent variable nodes and the squares represent 'factor' nodes. The distribution over the configurations is given by the product of the factors associated with the variables, where the association by functional dependence is shown graphically by a connected edge. We can trivially translate such graphs to and from the energy-based framework by associating an energy function with each factor node. The function, $E_n(\mathbf{x})$, associated with a node having factor $\phi_n(\mathbf{x})$ is given by

$$E_n(\mathbf{x}) = -\log \phi_n(\mathbf{x}) \tag{4.5}$$

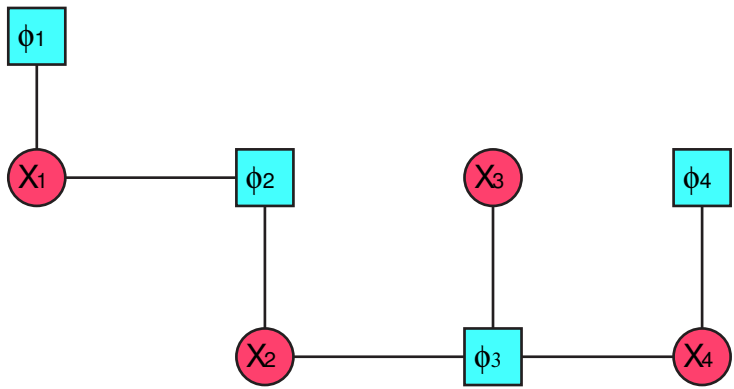

**Figure 4.2:** Example factor graph. The distributions over the configurations of $\mathbf{x} = [x_1, x_2, x_3, x_4]^T$ are given by $p(\mathbf{x}) = \frac{1}{Z} \prod_{i=1}^{4} \phi_n(\mathbf{x})$. The functional dependency of the factors $\phi_n$ are illustrated graphical by connected edges in the graph.

A similar relationship allows us to transform a general undirected graphical model or Markov random field into an energy-based model. The negative log of the clique potentials take the role of energy functions. Another fruitful interpretation can be to think of deterministic latent variables (and their associated energy functions) as factor nodes that are endowed with clear representational semantics and which also may be constructed recursively or hierarchically.

In addition to these general relationships, we can also make more detailed connections between the energy-based approach and specific examples of other model frameworks. The following sections discuss several classes of models which have been popular in modelling visual stimuli.

### 4.3.1   Products of Experts

We begin our tour of models with Products of Experts which are actually developmental antecedents of the current energy-based framework[2]. Hinton [1999] proposed combining $n$ individual 'expert' models, $p_m(\mathbf{x}; \theta_m)$ as follows

$$p(\mathbf{x}; \theta_1 \ldots \theta_n) = \frac{\prod_m p_m(\mathbf{x}; \theta_m)}{\sum_{\mathbf{x}'} \prod_m p_m(\mathbf{x}'; \theta_m)} \tag{4.6}$$

The individual models can afford to concentrate on a small subset of input space, provided that the other models do not cause interference and and that they model

---

[2]The CD algorithm was originally developed to deal specifically with PoE models.

other regions of space well.

Note that there is an important difference between combining experts as a product, and combining experts as a sum or a mixture. A mixture-of-experts distribution always has greater variance than its components; on the other hand, a product-of-experts distribution can combine many vague experts and yield an overall distribution that is sharp and well defined. This fact ties in rather nicely with some ideas of coarse coding in neural populations.

Products of experts fit snugly within the EBM framework; the energy functions are simply given by

$$E_m(\mathbf{x}) = -\log p_m(\mathbf{x}; \theta_m) \tag{4.7}$$

Indeed, the EBM framework arose as a natural generalisation of our work done with products of experts.

### 4.3.2 Square Noiseless ICA and Infomax

As discussed in chapter 3, we can form an equivalence relationship between performing maximum likelihood in an ICA causal generative model and maximising the mutual information between inputs and outputs of a linear model with monotonic, bounded outputs. The requirement for this equivalence to hold is that we have as many input dimensions as output dimensions (i.e. a 'square' model) and that we have no noise.

In this square and noiseless regime we can also easily show an additional equivalence between these two approaches and our energy-based models. We simply take our additive contributions to the total energy to be appropriate linear combinations of the input variables. These energy functions will have a simple relationship to the source priors of an ICA model or the output non-linearity in an Infomax approach.

For example, consider the case where a causal ICA model has sources $\mathbf{s}$ distributed according to $p(\mathbf{s})$ and (square, invertible) generative weight matrix, $\mathbf{G}$

$$p(\mathbf{s}) = \prod_i p_i(s_i) \tag{4.8}$$

$$\mathbf{G}^{-1} = \mathbf{W} \tag{4.9}$$

Inputs are generated in the model by linearly mixing the sources, i.e.

$$\mathbf{x} = \mathbf{G}\mathbf{s} \tag{4.10}$$

We may write this as a probability distribution using the following semantics.

$$p(\mathbf{x}, \mathbf{s}) = \delta(\mathbf{x} - \mathbf{G}\mathbf{s}) \prod_i p_i(s_i) \tag{4.11}$$

$$p(\mathbf{x}|\mathbf{s}) = \delta(\mathbf{x} - \mathbf{G}\mathbf{s}) \tag{4.12}$$

A maximum likelihood objective function for such a model would be equivalent to performing information maximisation using a non-linearity on the outputs of the form

$$y_i = g_i(\mathbf{w}_i^T \mathbf{x}) \tag{4.13}$$

where the functions $g_i$ are the cumulative distribution functions of the corresponding source prior distributions, $p_i(\mathbf{s})$, and $\mathbf{w}_i$ is the $i^{\text{th}}$ row of $\mathbf{W}$ (arranged as a column vector).

In our energy-based framework the objective would also be the same if we used energy functions of the form

$$E(\mathbf{x}) = \sum_{i=1}^{M} E_i(y_i; \theta_i) \tag{4.14}$$

$$y_i(\mathbf{x}) = \mathbf{w}_i^T \mathbf{x} \tag{4.15}$$

$$E_i(y_i) = -\log p_i(s_i = y_i) \tag{4.16}$$

In this simple case the partition function for the energy-based model *can* be computed and is given by

$$Z = |\det(\mathbf{W})|^{-1} \tag{4.17}$$

The origin of this can be most easily seen if we consider the problem as a change of variables using equation 4.10, and employ the Jacobian of the linear transformation. Then we may write

$$
\begin{aligned}
p(\mathbf{x}) &= \prod_{i=1}^{M} p_i(s_i = \mathbf{w}_i^T \mathbf{x}) \left| \det \left[ \frac{\partial \mathbf{s}}{\partial \mathbf{x}} \right] \right| \\
&= \prod_{i=1}^{M} p_i(s_i = \mathbf{w}_i^T \mathbf{x}) |\det \mathbf{W}|
\end{aligned}
\tag{4.18}
$$

Thus the energy-based perspective allows us to interpret square ICA as a filtering model in which energies are associated with a set of linear filter outputs, $y_i$. We will refer to these outputs as 'features' and can think of them as deterministic latent variables.

### 4.3.3   Overcomplete Generalisations of ICA and Infomax

The exact equivalence between ICA, Infomax and EBM's breaks down when we generalise to overcomplete models, i.e. when we have more features/sources than input dimensions. However, it is still possible to derive intimate connections between the models.

### Overcomplete energy-based models

The energy-based model described by equations 4.14-4.16 is trivially extended to the overcomplete case; we simply take more features/energy functions than we have input dimensions. This yields a density model which is different to the usual generalisation of ICA to the overcomplete case, and retains the previously expressed interpretation as a linear filtering approach. This property is attractive since the features remain simple and deterministic functions of the input. This is in contradistinction to the generalisation of ICA within the causal generative model framework. With causal generative ICA, conditioned on an input, the posterior distribution over possible source configurations (features in the previous terminology) is often analytically intractable, and methods such as MCMC, variational approximations, or gradient ascent to MAP points are usually required.

In the overcomplete EBM another property arises; not all combinations of feature values are allowed. Since we have a deterministic mapping to a space of higher dimension than that in which the inputs lie, the valid feature set is a low-dimensional manifold in the space of all possible feature values. (This also arises when the information maximisation approach is made overcomplete.)

A further point, which might be seen as something of a disadvantage, is that we are no longer able to compute the partition function since it is no longer given by a simple expression. However, this and related issues (such as the difficulty of computing the density for a given input) are also common to the overcomplete causal generative model framework. Whilst it remains simple to draw samples from overcomplete causal generative models, the marginal distribution on the inputs now requires the computation of a difficult integral since we must sum over all possible source/feature combinations that might have produced a given input configuration.

EBM's in which energies are assigned to linear filter outputs are explored in more detail in chapter 6. In the meantime, we have seen that density models corresponding to square ICA can be extended to the overcomplete case in a novel way and that furthermore this extension renders the problem of inference (or representation) trivial; the posterior distribution of features given an input effectively reduces to a point, and this point is easily computed.

We now discuss how we can mathematically relate the previously described overcomplete energy-based model to overcomplete causal-generative ICA. Consider an initially overcomplete ICA model to which we have added 'auxiliary' input dimensions $\mathbf{a}$ to make the model square. We will denote the total input space by $\mathbf{v} = [\mathbf{x}^T, \mathbf{a}^T]^T$. We will also augment the filter matrix by appending extra columns, $\mathbf{F}$, connecting the auxiliary inputs to the features, i.e. the total filter matrix is now $\mathbf{H} = [\mathbf{W} \ \mathbf{F}]$. We will assume that the new filters are chosen such that $\mathbf{H}$ is invertible, i.e. that the new enlarged space is fully spanned. For this enlarged ICA model we can again write the probability distribution in a form similar to equation 4.18, given

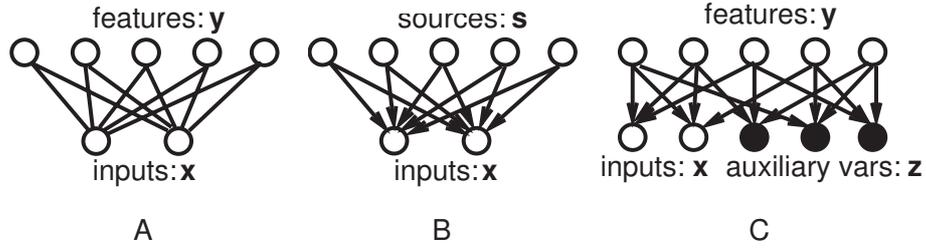

**Figure 4.3:** Graphical representation showing relationships between overcomplete ICA and overcomplete energy-based models. (A) depicts the overcomplete EBM as an undirected graphical model. (B) Directed graphical model for overcomplete causal ICA. (C) Directed graphical model depicting the causal interpretation of the overcomplete EBM, with the filled circles denoting that the auxiliary variables are observed. These observations introduce dependencies between the features/sources.

here by

$$p(\mathbf{x}, \mathbf{a}) = \prod_{i=1}^{M} p_i(\mathbf{w}_i^T \mathbf{x} + \mathbf{f}_i^T \mathbf{a}) |\det \mathbf{H}| \tag{4.19}$$

where $\mathbf{f}_i$ are the rows of $\mathbf{F}$. Next, we write the probability density for the conditional distribution,

$$
\begin{aligned}
p(\mathbf{x}|\mathbf{a}) &= \frac{p(\mathbf{x}, \mathbf{a})}{p(\mathbf{a})} \\
&= \frac{\prod_{i=1}^{M} p_i(\mathbf{w}_i^T \mathbf{x} + \mathbf{f}_i^T \mathbf{a}) |\det \mathbf{H}|}{\int \prod_{i=1}^{M} p_i(\mathbf{w}_i^T \mathbf{x} + \mathbf{f}_i^T \mathbf{a}) |\det \mathbf{H}| d\mathbf{x}'}
\end{aligned}
\tag{4.20}
$$

Notice that the $|\det \mathbf{H}|$ terms will cancel. If we choose the auxiliary variables $\mathbf{a} = 0$ then equation 4.20 can reformulated as

$$p(\mathbf{x}|\mathbf{a} = 0) = \frac{\prod_i p_i(\mathbf{w}_i^T \mathbf{x})}{\int \prod_i p_i(\mathbf{w}_i^T \mathbf{x}') \, d\mathbf{x}'} \tag{4.21}$$

This equation above can easily be interpreted as an energy-based model. Indeed, it is precisely the type of overcomplete energy-based model discussed in the previous section. We see that an expression for the partition function in this case can be given (intractably) as

$$Z = \int \prod_i p_i(\mathbf{w}_i^T \mathbf{x}') \, d\mathbf{x}' \tag{4.22}$$

The above derivation is independent of the precise choice of the filters $\mathbf{F}$; as long as they span the extra dimensions of the 'full' space, the derivation is valid. Figure 4.3 illustrates the relationship between the model classes.

### Relating EBM's to Information Maximisation

The original infomax approach [Bell and Sejnowski, 1995] was proposed for square or undercomplete models. Shriki et al. [2001] proposed an extension to the case of overcomplete representations (with the additional possibility of recurrent inter-actions at the outputs). As in the original formulation, there is a parameterised, deterministic, non-linear mapping between inputs and outputs; maximising mutual information is once again equivalent to maximising the entropy of the outputs.[3] Shriki et al. [2001] showed that the expression for the entropy in the overcomplete case can be expressed as:

$$H(\mathbf{y}) = -\int d\mathbf{x}\, p^0(\mathbf{x}) \log \frac{p^0(\mathbf{x})}{\sqrt{\det(\mathcal{J}(\mathbf{x})^T \mathcal{J}(\mathbf{x}))}} \qquad (4.23)$$

$$\mathcal{J}_{ij}(\mathbf{x}) = \frac{\partial y_i(\mathbf{x})}{\partial x_j} \qquad (4.24)$$

where as usual $p^0(\mathbf{x})$ is the data distribution, and $\mathcal{J}(\mathbf{x})$ is the Jacobian for the transformation between inputs, $\mathbf{x}$, and outputs, $\mathbf{y}$. Note that, as in the case of the overcomplete EBM, we have a deterministically mapped, overcomplete representation in which the inputs are mapped one-to-one onto a *subset* of all possible outputs, i.e. the image of that mapping forms a lower dimensional manifold in output space.

In the general overcomplete case, the quantity $\sqrt{\det(\mathcal{J}(\mathbf{x})^T \mathcal{J}(\mathbf{x}))}$ in equation 4.23 does not evaluate to 1 when integrated over the space of inputs. This is important because it means that we cannot interpret the overcomplete infomax objective function as minimising a KL-divergence as we can in the complete case. However, let us consider the probability density defined as,

$$p_{\text{info}}(\mathbf{x}) = \frac{1}{Z} \sqrt{\det(\mathcal{J}(\mathbf{x})^T \mathcal{J}(\mathbf{x}))} \qquad (4.25)$$

where $Z$, as usual, is a normalization constant. $p_{\text{info}}(\mathbf{x})$ *is* consistent with a probability distribution in the EBM framework if we choose the following function for our energy,

$$\begin{aligned} E_{\text{info}}(\mathbf{x}) &= -\log\left(\sqrt{\det(\mathcal{J}(\mathbf{x})^T \mathcal{J}(\mathbf{x}))}\right) \\ &= -\frac{1}{2}\operatorname{Tr}\left[\log_M\left(\mathcal{J}(\mathbf{x})^T \mathcal{J}(\mathbf{x})\right)\right] \end{aligned} \qquad (4.26)$$

however, in the general case this energy does not easily decompose into a sum of simple terms.[4] Although the energy function is rather complicated, this energy-based density model actually has a simple interpretation in terms of the mapping from inputs to representational space. The distribution $p_{\text{info}}(\mathbf{x})$ is transformed precisely to a uniform distribution $p_{\text{info}}(\mathbf{y}) = 1/Z$ on the manifold in $\mathbf{y}$-space. The normalization constant $Z$ may thus be interpreted as the *volume* of this manifold. Minimizing the KL divergence $KL[p^0(\mathbf{x})||p_{\text{info}}(\mathbf{x}))]$ can therefore be interpreted as seeking to map the data to a manifold in a higher dimensional embedding space, in which the data are distributed as uniformly as possible. The relation between information maximization and the above energy-based approach is summarized by the following expression,

$$H(\mathbf{y}) = -KL(p^0(\mathbf{x})||p_{\text{info}}(\mathbf{x})) + \log(\text{Manifold-Volume}) \tag{4.27}$$

---

[3]There is a slight technical issue in that, since the feasible output manifold is actually set of measure zero (since we are deterministic and overcomplete), there are complications in defining quantities such as entropy. This is, however, ignored.

[4]Note that the $\log_M$ in equation 4.26 denotes the *matrix* logarithm function.

The first term describes the 'fit' of the model $p^0(\mathbf{x})$ to data, while the second term is simply the entropy of the uniform distribution $p(\mathbf{y})$ on the manifold. Relative to the energy-based approach, maximizing the mutual information will thus have a stronger preference to increase the volume of the manifold, since this is directly related to the entropy of $p(\mathbf{y})$. Note that in the square case the manifold is exactly the whole image space $[0,1]^M$, hence its volume is always fixed at 1, and equation 4.27 reduces exactly to the KL divergence $KL(p^0(\mathbf{x})||p_{\text{info}}(\mathbf{x}))$.

### 4.3.4 Maximum Entropy

Maximum entropy (or maxent) models [Zhu et al., 1998, 1997] are also naturally expressed as a Boltzmann distribution. In the basic maximum entropy framework we consider that we wish to focus only on certain aspects of a distribution, or alternatively that we only have knowledge about a certain aspects of a distribution. Then, consistent with these aspects, we express our remaining ignorance by choosing the distribution that has the maximum entropy. (For a good, though somewhat tendentious, discussion of the philosophy behind this approach see Jaynes [1982] and related work.)

In particular, if we have a set of functions $c_i(\mathbf{x})$ and we know the expected value of those functions over the distribution we seek to approximate, i.e.

$$\langle c_i(\mathbf{x}) \rangle_P = \gamma_i \tag{4.28}$$

then it is easy to show that the maximum entropy distribution consistent with this information takes the form

$$p^0(\mathbf{x}) = \frac{1}{Z} e^{-\sum_i^M \lambda_i c_i(\mathbf{x})} \tag{4.29}$$

where the $\lambda_i$ are Lagrange multipliers (commonly called 'maxent weights') set up to satisfy 4.28. The distribution in equation 4.29 has an obvious formulation as an energy-based model with the energy function

$$E(\mathbf{x}) = -\sum_i^M \lambda_i c_i(\mathbf{x}) \tag{4.30}$$

In the maximum entropy framework the features $c_i(\mathbf{x})$ must be pre-specified by some means (e.g. prior knowledge of the problem) and then the $\lambda_i$ are the free parameters to be fit based on data. In this sense we can interpret EBM's as maximum entropy models the parameters of the feature functions, $c_i(\mathbf{x})$, as well as the weights $\lambda_i$ are learned from data.

## 4.4   Maximum Likelihood Parameter Estimation in EBM's

This section describes procedures for performing maximum likelihood parameter fitting in energy-based models. Extending these procedures to perform maximum a posteriori (MAP) parameter estimation would require only minor modifications, however performing full Bayesian learning in the general case remains a difficult and largely unsolved problem.

Let $p^0(\mathbf{x})$ be the distribution of configurations in the data, and $p^\infty(\mathbf{x})$ be the model distribution given by 4.1.[5] We would like $p^\infty$ to approximate $p^0$ as well as possible, and the standard measure of the difference between two probability distributions is the Kullback-Liebler (KL) divergence. The KL divergence is a non-negative, asymmetric functional of two distributions and is zero iff the two distributions are identical. It is defined as:

$$KL(p^0\|p^\infty) = \int p^0(\mathbf{x}) \log \frac{p^0(\mathbf{x})}{p^\infty(\mathbf{x})} d\mathbf{x} \qquad (4.31)$$

Since $p^0$ is fixed, minimising the KL divergence with respect to the parameters is equivalent to maximising the likelihood of the data under the model with respect to those parameters. For energy-based models as defined by equations 4.1, 4.2 and 4.3 the derivative of the KL divergence with respect to a parameter $\theta_{ij}$ is given by

$$
\begin{aligned}
\frac{\partial KL(p^0\|p^\infty)}{\partial \theta_{ij}} &= \frac{\partial}{\partial \theta_{ij}} \left\{ \int \left[ p^0 \log p^0 + p^0 E(\mathbf{x}) \right] d\mathbf{x} + \log Z \right\} \\
&= 0 + \int p^0 \frac{\partial E(\mathbf{x})}{\partial \theta_{ij}} d\mathbf{x} + \frac{1}{Z} \int \frac{\partial}{\partial \theta_{ij}} e^{-E(\mathbf{x})} d\mathbf{x} \\
&= \left\langle \frac{\partial E(\mathbf{x})}{\partial \theta_{ij}} \right\rangle_{p^0} - \left\langle \frac{\partial E(\mathbf{x})}{\partial \theta_{ij}} \right\rangle_{p^\infty}
\end{aligned}
\qquad (4.32)
$$

where $\langle \cdot \rangle_q$ is the expectation operator under distribution $q$. The first term is simply the expected gradient of the energy function with respect to the parameters under the data distribution and is easily obtained using samples from the data distribution. The second term, which comes from the partition function, is the expected gradient of the energy with respect to the parameters under the distribution cur-rently specified by the model. (For historical reasons the first term is often called the 'positive' or 'wake' phase, whilst the second term is called the 'negative' phase or the 'sleep' phase.) Like the partition function itself, the expectation in the second phase is analytically intractable for many interesting distributions. Consequently, direct optimisation is often impossible.

---

[5]We will assume for now that any latent variables are stochastic rather than deterministic. Stochastic latent variables simply add an extra layer of summation/integration, and will be considered in Chapter 5.

One way of overcoming this problem is to use Markov Chain Monte Carlo (MCMC) methods to approximate the expectation by obtaining samples from $p^\infty$, the distribution specified by the model. (For an excellent overview and general introduction to Monte Carlo techniques, see Neal [1993].) Explicitly, we can recursively apply an appropriate stochastic transition operator to obtain the samples needed to compute an approximation to the expectations in equation 4.32.

Let $T(\mathbf{x}|\mathbf{x}')$ represent the probabilities of going from $\mathbf{x}'$ to $\mathbf{x}$, under the action of the stochastic transition of operator $\mathcal{T}$, where $\mathcal{T}$ is the transition operator for a Markov chain whose (unique) stationary distribution is that specified by the model, i.e. an 'equilibrium invariant Markov Chain' for the model or model class. Let $\mathcal{M}^1(q(\mathbf{x}))$ represent the distribution obtained when the distribution $q(\mathbf{x})$ is transformed under the action of operator $\mathcal{T}$, and let $\mathcal{M}^n(q(\mathbf{x}))$ represent the distribution obtained after $n$ recursive operations.

$$\mathcal{M}^1(q(\mathbf{x})) = \int T(\mathbf{x}|\mathbf{x}')q(\mathbf{x}')d\mathbf{x}' \tag{4.33}$$

$$\mathcal{M}^n(q(\mathbf{x})) = \mathcal{M}^1\left(\mathcal{M}^{n-1}(q(\mathbf{x}))\right) \tag{4.34}$$

$$\mathcal{M}^0\left(q(\mathbf{x})\right) = q(\mathbf{x}) \tag{4.35}$$

A procedure for obtaining samples from (an approximation to) the equilibrium distribution is to initialise chains at a very broad initial distribution $q(\mathbf{x})$, (say a broad gaussian, or a uniform distribution if the space is discrete or bounded), and then to run the Markov Chain operator for a very large number of steps — ideally until we achieve convergence.

If we can obtain such 'equilibrium samples' then we may obtain a consistent, unbiased estimate of the expectation of the derivative in the term of equation 4.32, and optimise the parameters by performing gradient descent in the KL divergence (equivalently gradient *ascent* in the likelihood function). The empirical stepwise updates for the parameters are given by

$$\triangle\theta_{ij} \propto -\sum_{\text{data } \mathbf{x}_d}\frac{\partial E(\mathbf{x}_d)}{\partial\theta_{ij}} + \sum_{\text{'equilibrium' samples } \mathbf{x}_s}\frac{\partial E(\mathbf{x}_s)}{\partial\theta_{ij}} \tag{4.36}$$

This update rule can be intuitively understood as reducing the energy at locations where data is observed (the first term in equation 4.36) and at the same time increasing the energy at locations where the model currently predicts data with high probability (the second term in 4.36). At the end of learning these terms will eventually balance out, and will result in a landscape with low energy (thus high probability) in regions where data is present, and high energy (thus low probability) elsewhere.

Whilst, in principle, the procedure described above allows us to obtain the terms required to perform gradient based optimisation there are several drawbacks. One main problem is that the method is extremely computationally expensive — the Markov chain has to be run for very many steps before it approaches the equilibrium distribution $p^{\infty}$. This problem is compounded by the fact that it is hard to estimate how many steps are required, or whether we have actually converged upon the intended distribution. Another problem is the fact that the equilibrium distribution of a model can have a very large variance. This is problematic not only because it makes accurate gradients difficult to obtain, but also because of the phenomenon of 'variance aversion'. This is essentially an interaction between the learning algorithm and the parameter-gradient variance as a function of position in model space. The effect is that learning will tend to avoid areas in which the parameter gradient estimates have high variance, possibly even if on average these regions of parameter space are favourable.[6] Very many independent samples are needed in order to overcome these variance related problems, thus incurring even greater computational cost. Therefore, estimating the derivatives 4.32 is often extremely slow and can be very unreliable.

## 4.5   Contrastive Divergence

This section argues that instead of trying to compute the expression for the likelihood gradient at each point we should seek to optimise a different objective function that nevertheless improves the quality of our model. The basic premise is that it is unnecessary to estimate derivatives averaged over the equilibrium distribution in order to train an energy-based model from data. Rather, we will replace the second term in 4.36 with an estimate of the derivatives averaged over another distribution; one that is easier to sample from and has lower intrinsic variance. This technique was proposed by Hinton [Hinton, 2000, 2002] and is called Contrastive Divergence learning.

There are two main ideas involved in contrastive divergence learning. The first is to start the Markov chain used to provide samples at the *data* distribution $p^0$ rather than initialising at some vague, broad distribution (e.g.: a Gaussian with large variances) as is conventionally done. The second idea is to run the Markov chain for just a few iterations rather than until (an approximation to) equilibrium.

---

[6]This phenomenon is actually a general problem with stochastic, gradient-based optimisation. Some intuition can be gained by considering sand spread out on a horizontal sheet that is undergoing a natural mode of vibration in the vertical direction — the sand particles will collect at the stationary nodes even though, on average, the sheet is the same height everywhere.

We define $p^n(\mathbf{x})$ to be the distribution obtained after $n$ iterations of an equilibrium invariant Markov chain initialised at the data distribution $p^0(\mathbf{x})$,

$$p^n(\mathbf{x}) = \mathcal{M}^n(p^0(\mathbf{x})) \qquad (4.37)$$

The contrastive divergence algorithm performs stepwise parameter updates based upon

$$\triangle\theta_{ij} \propto -\sum_{\text{data } \mathbf{x}_d} \frac{\partial E(\mathbf{x}_d)}{\partial\theta_{ij}} + \sum_{\text{samples } \mathbf{x}_s \sim p^n} \frac{\partial E(\mathbf{x}_s)}{\partial\theta_{ij}} \qquad (4.38)$$

Relative to maximum likelihood learning we have simply replaced $p^\infty$ with $p^n$. As will be discussed later, there is some dependence of the performance of the algorithm on $n$ but very often a value of $n = 1$ is effective and is chosen as the 'default' setting. Similarly, the results may depend subtly on the choice of Markov chain transition operator $\mathcal{T}$.

There are several general intuitive arguments that can be put forward in support of contrastive divergence. One intuitive argument for the contrastive divergence algorithm is that it alters the parameters in ways which suppress consistent tendencies for an equilibrium-invariant Markov chain to deviate away from regions where we have data. This is clearly desirable since, if the model were a good fit to the data, an equilibrium-invariant Markov chain would spend most of its time on the data manifold. Another intuitive argument goes as follows: at the start of learning, assuming we start with small parameter values and thus a relatively 'flat' energy landscape, the influence of the negative phase term is likely to be minor relative to the positive phase term. Then as learning proceeds, the model distribution becomes more like the data distribution and thus an MCMC sampler started at the data is should already be quite close to equilibrium. Finally, it is also easy to see that the algorithm has a fixed point if the data distribution and the model distribution exactly coincide, since $p^\infty = p^n = p^0$ and by inspection the two terms in 4.38 cancel out.

In addition to the rather qualitative reasoning above, we can also provide a somewhat firmer theoretical and mathematical motivation for contrastive divergence. We will now show that the algorithm performs approximate-gradient descent on a cost function. This is a slightly weaker statement than actually performing gradient descent on a cost function and does not *guarantee* convergence; nevertheless it does give us reason to believe that our algorithm should be well behaved.

The $n$-step contrastive divergence cost function is defined [Hinton, 2002] as

$$CD^n = KL(p^0\|p^\infty) - KL(p^n\|p^\infty) \qquad (4.39)$$

Note that this consists of the usual KL divergence between the data distribution and the model distribution, from which the KL divergence between the $n$-step dis-tribution, $p^n$, and the model distribution is subtracted. We see that in the limit that $n \to \infty$ the contrastive divergence objective is the same as the maximum likelihood objective function.

Taking derivatives of the cost function 4.39 with respect to a parameter $\theta_{ij}$ gives the following expression for the gradient,

$$\frac{\partial CD^n}{\partial \theta_{ij}} = \left\langle \frac{\partial E(\mathbf{x})}{\partial \theta_{ij}} \right\rangle_{p^0} - \left\langle \frac{\partial E(\mathbf{x})}{\partial \theta_{ij}} \right\rangle_{p^n}$$
$$- \int \left( E(\mathbf{x}) - \log p^n(\mathbf{x}) \right) \frac{\partial p^n(\mathbf{x})}{\partial \theta_{ij}} d\mathbf{x} \qquad (4.40)$$

The first two terms in equation 4.40 are, under the exchange of empirical samples for expectations, identical to the ones proposed for the learning algorithm in equation 4.38. The last term represents the effect that changes in $\theta_{ij}$ have upon the contrastive divergence objective as a consequence of alterations to the $n$-step distribution reached by the Markov chain. These effects depend on the way in which the parameters interact with the particular choice of Markov chain transition operator. This dependency is rather complicated and the effects are very hard to compute. Fortunately, simulations by Hinton [2002] suggest that this term is usually small and of the same sign as the sum of the other terms, and that it can be safely ignored; results later in this thesis and elsewhere [Hinton et al., 2001, Teh et al., 2003] further support this claim .

The pseudo-code in Algorithm 1 summarises the procedure for contrastive divergence learning in batch mode. Note that the MCMC samplers for each batch start at the data points used in that batch. This data matching further helps reduce the variance of the estimates in the algorithm.

---

**Algorithm 1** Contrastive Divergence Learning for Energy-Based Models

1. Compute the gradient of the total energy with respect to the parameters and average over the data cases $\mathbf{x}_d$.
2. Run MCMC samplers for $n$ steps, starting at every data-vector $\mathbf{x}_d$, keeping only the last sample $\mathbf{x}_s$ of each chain.
3. Compute the gradient of the total energy with respect to the parameters and average over the samples $\mathbf{x}_s$.
4. Update the parameters using,

$$\triangle \theta_{ij} = \frac{\eta}{N} \left( - \sum_{\text{data } \mathbf{x}_d} \frac{\partial E(\mathbf{x}_d)}{\partial \theta_{ij}} + \sum_{\text{samples } \mathbf{x}_s \sim p^n} \frac{\partial E(\mathbf{x}_s)}{\partial \theta_{ij}} \right) \qquad (4.41)$$

where $\eta$ is the learning rate and $N$ the number of samples in each mini-batch.

## 4.6   Sampling Methods for Contrastive Divergence

In principle, any valid MCMC method can be used to obtain the 'negative phases' samples (i.e.: $x \sim p^n(\mathbf{x})$) required for contrastive divergence learning. However, the particular choice of method in a given model may affect the efficiency and overall performance of the algorithm. The two main techniques considered in this thesis are Gibbs sampling and a stochastic dynamics method known as Hybrid Monte Carlo (HMC). The following section discusses the broadly applicable HMC method whilst Gibbs sampling techniques, with respect to different models, are discussed in more detail in other chapters.

### 4.6.1   Hybrid Monte Carlo

Hybrid Monte Carlo (HMC) and related dynamical techniques are Markov chain sampling methods that derive from some computational approaches which were developed concurrently with the Metropolis algorithm as a way of simulating physical systems. The methodology has subsequently found application in several interesting machine learning problems, for instance Bayesian learning of neural networks [Neal, 1994], and has an advantage over many simple MCMC methods (at least in some problems) because it avoids much of the random walk behaviour that can slow down sampling methods. An excellent and comprehensive discussion of the technique is given in Neal [1993]; here we present only the main points and describe how we have applied HMC the context of contrastive divergence.

The basic idea behind HMC is that we can generate samples from a probability distribution by simulating the stochastic dynamics of an appropriate physical system. Consider a system of particles, each having a total energy (the Hamiltonian), $\mathcal{H}$, given by

$$\mathcal{H}(\mathbf{p}, \mathbf{q}) = \mathcal{V}(\mathbf{q}) + \mathcal{K}(\mathbf{p}) \qquad (4.42)$$

where $\mathcal{V}$ is a potential energy, which is dependent on a co-ordinate vector $\mathbf{q}$, and $\mathcal{K}$ is a kinetic energy which is dependent upon a momentum vector $\mathbf{p}$ with a component $p_i$ along each co-ordinate axis $q_i$.

The probability of finding a particle with position $\mathbf{q}$ and momentum $\mathbf{p}$ is given by

$$
\begin{aligned}
P(\mathbf{q}, \mathbf{p}) &= \frac{1}{Z_H} e^{-\mathcal{H}(\mathbf{q}, \mathbf{p})} \\
&= \left[ \frac{1}{Z_V} e^{-\mathcal{V}(\mathbf{q})} \right] \left[ \frac{1}{Z_T} e^{-\mathcal{K}(\mathbf{q})} \right]
\end{aligned} \qquad (4.43)
$$

where we have assumed a temperature of '1' in 'natural units.'

It is clear from the way that the joint distribution 4.43 factors, that the spatial distribution of particles in such a system, i.e. the marginal distribution of the $\mathbf{q}$, is just the Boltzmann distribution of the potential energy. If the potential energy, $\mathcal{V}$ corresponded exactly with the energy, $E$, defining an EBM, then the equilibrium distribution of particle locations in this system would exactly match our probability desired distribution.

So, one way to sample from a given distribution, $p^\infty = \frac{1}{Z}e^{-E(\mathbf{x})}$, is to simulate a system of particles at thermal equilibrium in potential energy given by $\mathcal{V}(\mathbf{x}) = E(\mathbf{x})$. This can be done by following a two stage procedure in which we simulate (i) Hamiltonian dynamics that (up to discretisation error) conserve energy, interspersed with (ii) stochastic transitions that represent contact with a thermal reservoir and allow the energy to equilibrate.

In actual physical systems, we can describe the Hamiltonian dynamics using the following pair of coupled differential equations,

$$\frac{dq_i}{d\tau} = +\frac{\partial \mathcal{H}}{\partial p_i} = p_i \tag{4.44}$$

$$\frac{dp_i}{d\tau} = -\frac{\partial \mathcal{H}}{\partial q_i} = -\frac{\partial E}{\partial q_i} \tag{4.45}$$

In our simulations we replace these differential equations with corresponding difference equations and employ a small step size. The stochastic transitions only involve the momentum variables, and in the simplest scheme we simply re-sample the momentum from the marginal distribution – an isotropic, multivariate Gaussian.

Due to the inevitable presence of discretisation error in computational simulation some sophisticated tricks are required to ensure everything works correctly. The two main devices are: (i) a Metropolis-Hastings acceptance step at the end of each trajectory of deterministic Hamiltonian dynamics; and (ii) a 'leap-frog' scheme to set the order of the discrete approximations to 4.44 and 4.45. These modifications eliminate the potential bias introduced by inexact simulation and ensure detailed balance and reversibility. The reader is referred to Neal [1993] for more details. The HMC procedure for obtaining samples from a distribution $p^\infty = \frac{1}{Z}e^{-E(\mathbf{x})}$ is summarised in algorithm 2.

We have adapted this procedure in several ways in order to use HMC as part of our contrastive learning procedure for continuous-valued energy-based models. Specifically we initialise the dynamics at the data and run for just a small number of 'outer-loop' iterations. Within each 'outer-loop', we may choose to run for quite a few – say 20 or 30 – 'inner-loop' leapfrog iterations. The precise numbers for both inner and outer loops are somewhat heuristic. Additionally, we adapt the leap-frog step size, $\epsilon$, slowly throughout learning (but between sampling runs) to maintain a Metropolis acceptance rate between 90% and 95%. It is possible that this online adaptation may bias the samples obtained and the dynamics of learning, however we do not believe this to be a serious problem in the cases we explore. Figure 4.4

---

**Algorithm 2** Basic Hybrid Monte Carlo Procedure

---

1. Initialise the sampler using an appropriate broad distribution for position and an isotropic unit variance Gaussian for momentum.

   **Outer Loop: Loop until convergence or boredom**

2. Randomly choose a direction $\lambda$ for the trajectory, with the two values $\lambda = 1$ (forwards) and $\lambda = -1$ (backwards) being equally likely.

   **Inner Loop: Repeat for $n_l$ Leapfrog Iterations**

   3. Perform the following ordered updates to the momenta and position variables.

$$p_i \leftarrow p_i - \frac{\lambda \epsilon}{2} \frac{\partial E}{\partial q_i} (\mathbf{q})$$

$$q_i \leftarrow q_i + \lambda \epsilon p_i$$

$$p_i \leftarrow p_i - \frac{\lambda \epsilon}{2} \frac{\partial E}{\partial q_i} (\mathbf{q})$$

   4. Repeat step 3 for $(n_l - 1)$ iterations.

5. Regard the endpoint of the leapfrog iterations as a candidate for the next state and apply to Metropolis procedure to decide whether or not to accept. I.e. we accept with probability $A\left((\mathbf{q}, \mathbf{p}), (\mathbf{q}^*, \mathbf{p}^*)\right) = \min(1, e^{\mathcal{H}(\mathbf{q}^*, \mathbf{p}^*) - \mathcal{H}(\mathbf{q}, \mathbf{p})})$ where $(\mathbf{q}, \mathbf{p})$ denotes the state from the previous outer loop iteration, and $(\mathbf{q}^*, \mathbf{p}^*)$ denotes the proposal obtained from the last run of the inner loop. If we do not accept the proposed state, we make the new state the same as the old state.

6. If we have converged or have reached the previously specified number of outer loop iterations, then exit. Otherwise, take the new state and return to step 2.

---

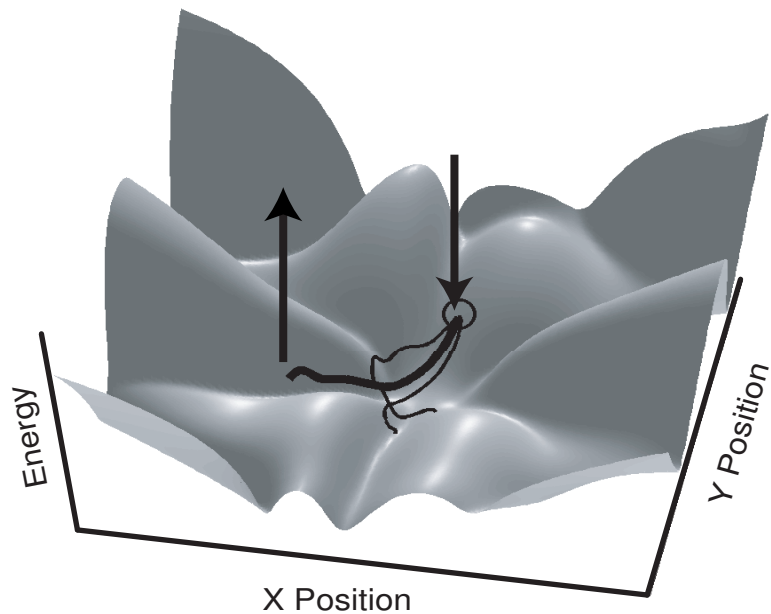

**Figure 4.4:** Illustration of an energy landscape and HMC trajectories near the end of learning of a toy data set. The lines indicate three sample trajectories (each initialised with a different momentum) emanating from a data point, marked with a circle. The arrows show the action of the CD algorithm on the energy landscape — lowering the energy at the start of this trajectory and raising the energy at an endpoint.

illustrates a set of HMC trajectories at the end of learning in a toy example.

Lastly we mention that, in more elaborate models, we employ back-propagation [Rumelhart et al., 1986] to compute the necessary energy gradients and therefore sometimes dub this method of learning with HMC 'Contrastive Backpropagation.' The combination of back-propagation, hybrid Monte Carlo and contrastive divergence is potentially extremely powerful and general. We believe that it could be profitably applied to a very broad class of models within the energy-based framework; all that is required is a continuous, differentiable energy function as might be described, for instance, by an arbitrary multi-layer neural network with energy functions associated with some or all the nodes. However, in the context of this thesis we restrict ourselves to a subset of such models that can be reasonably related to neural circuitry.

## 4.7 Contrastive Divergence: Issues and Complications

The contrastive divergence method seems to perform very well in practice. Although, to qualify this statement, we note that it is hard to assess absolute performance for the very same reasons that make the learning of many energy-based models difficult in the first place — namely the intractability of the partition function and related quantities. However, despite good empirical performance we can identify several concerns that may potentially afflict the contrastive divergence learning procedure.

### Unnoticed Empty Modes

One obvious potential problem with the learning procedure is the possible existence of regions of low energy that are far away (in terms of moves made by the MCMC sampler) from any data points. Since the sampler is started at the data and run for just a few iterations, it is unlikely that those distant regions would be explored; consequently it is possible that they could go completely unnoticed during learning. This would result in the final model assigning high probability to regions that are far away from the actual data, as illustrated schematically in figure 4.5 (A,B). One can conceive schemes that might provide protection against such an eventuality. For instance, one could try to intersperse the contrastive divergence parameter updates with parameter changes that use much more computation and try to sample from an approximation to the equilibrium distribution (as one might hope to get from very long runs of Monte Carlo sampling started from a uniform distribution or a broad Gaussian, perhaps in conjunction with simulated annealing [Kirkpatrick et al., 1983]). This lengthy sampling might be hoped to find deep modes regardless of location and would thus help to correct the problem. Alternatively, as a diagnostic rather than a preventative measure, one could measure the distribution of data energies at the end of learning and compare this with the distribution of energies that arise after lengthy and thorough sampling.

In practice, modest forms of complexity control such as 'weight decay'[7] seem sufficient to prevent pathological situations.

### Local vs Global Landscaping

An issue related to the one outlined above is the possibility that contrastive divergence may end up providing a good 'local' description of the probability landscape, but may be inaccurate for 'global' properties. As noted previously, one of the appealing features the contrastive divergence as given in algorithm 1 is that it reduces variance by pairing $n$-step samples with the data points from which those sampling chains were started with. However, this may also mean that the learning procedure is rather insensitive to differences in energy at distantly separated modes within the data. This scenario is illustrated schematically in figure 4.5 (C,D).

The behaviour of many MCMC samplers depends only on the local and relative aspects of structure of the distribution (for instance local energy gradients or the difference in energy between two points) rather than the global properties such as the absolute energy. Consequently, one could imagine a pathological scenario in which two identical modes in the data at very distant locations were assigned the same local structure but very different global energies. Technically speaking this would not be stable, however the chance of getting sufficient samples in the intervening regions to disrupt such a pathology could be arbitrarily low.

---

[7]Which in some circumstances is essentially equivalent to performing MAP estimates of the parameters. For instance, a weight decay term proportional to the parameter in question is equivalent to using a Gaussian prior; a fixed magnitude weight decay term is equivalent to using a Laplacian prior.

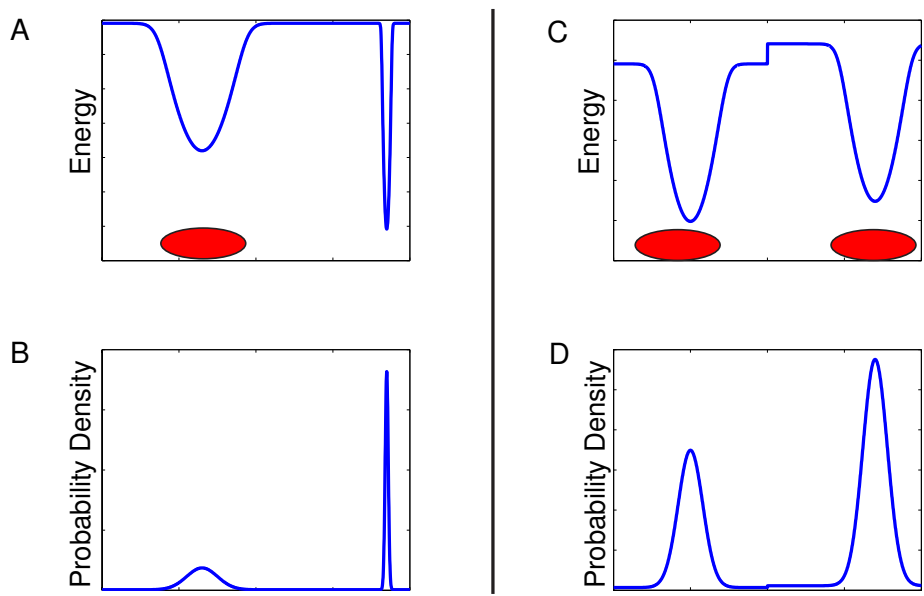

**Figure 4.5:** Figure illustrating potential complications with contrastive divergence learning in a simple 1 dimensional problem. The red ellipses represent the locations of clusters of data. (A,B) Panel A illustrates the possible effect of a deep, energy well far from the data. Panel B shows the corresponding probability distribution. (C,D) Panel C shows the possible effects of having correct local structure but incorrect global structure. The energy landscape (albeit a little contrived) has two identical wells, but the left well is offset from the right by a fixed energy difference. Panel D shows the corresponding distribution.

Again one can conceive schemes to overcome this potential difficulty, such as methods related to those in Hinton et al. [2004]. However as with the possibility of empty modes, in practice, we are not aware of this problem arising in our work so far.[8] This is partly because many of the applications of energy-based models have been made on data that is more or less distributed in a connected fashion (albeit often along some complicated manifold), consequently situations with many widely separated modes in the data have not often been encountered. If faced with such a situation, the best solution would perhaps be to alter the form of the MCMC sampler to adjust for the structure of the problem.

### Dependency on MCMC Methods

The results from contrastive divergence can depend on both the nature of the MCMC chain used, and on the number of steps for which it is run for after starting at the data. Both properties are clearly somewhat undesirable.

An analysis of some toy examples by Mackay [2001] has shown that the fixed points of the contrastive divergence algorithm can depend on the choice of the transition operator used in the Markov Chain. Furthermore, this can be true even if the operators are all ergodic and respect the detailed balance conditions. These findings highlight the fact that fixed points of the algorithm are, in some cases, *not* fixed points of the likelihood and that there can be systematic biases.

### Bias-Variance Trade-off

We have performed investigations on toy examples and studied the dependency of the gradient updates on the number of applications of a given Markov chain (similar work has been performed independently by Williams and Agakov [2002]). In such cases we observe what may be interpreted as a 'bias-variance trade-off'. The difference between the mean direction of the contrastive divergence updates and the 'true' maximum likelihood update generally seems to decrease with the number of MCMC steps, $n$. However, as the number of MCMC steps increases the variance of the samples obtained also increases, and so if we are using a fixed number of samples to compute the gradient there should be an optimal setting for $n$. However, this setting is rather dependent upon both the problem at hand, and the particulars of the Markov chain sampler.

---

[8]Although, arguably, it might be hard to detect.

It is difficult to fully evaluate the importance of the previous issues and complications. Practical experience suggests that the potential problems or biases are often minor or unimportant. When data is relatively abundant and the MCMC sampler mixes well throughout learning, the parameter estimates prove to be close to the maximum likelihood solutions[9]. However, potential pitfalls should always be borne in mind when designing the details of a particular instantiation of the contrastive divergence algorithm. Also, it may be prudent to design validation studies based upon understanding of the algorithm's foibles. As a final point we re-iterate that the contrastive divergence algorithm shares the deficits that can hinder all gradient based algorithms, such as sub-optimal local minima and the fact that (without modification) we do not have covariant[10] updates.

## 4.8   Experiments

This chapter draws to a close with two simple examples. Firstly, we present an exposition of learning in toy example designed to illustrate the framework and learning procedure, and also to demonstrate the flexibility of the approach. Secondly, we compare the results from contrastive divergence learning in an 'ICA-equivalent' EBM to the results obtained using exact sampling and also to those from the Bell and Sejnowski infomax algorithm. Chapters 5–7 present specific models applied to the development of receptive fields and topography in the visual system.

### 4.8.1   Toy Example

We generated a simple data set consisting $3,000$ points from a mixture of three Gaussian distributions, each contributing 1000 points. A sample of 250 points from this distribution is given in figure 4.7 (A). The construction of our model is illustrated in figure 4.6 — we took a simple, single-layer sigmoidal neural network (with 20 hidden units – the deterministic latent variables) and made the component energy functions equal to the weighted output of each unit.

$$s_i = \sigma(\mathbf{J}_i^T \mathbf{x} + b_i) \tag{4.46}$$

$$E_i(s_i) = a_i s_i \tag{4.47}$$

The parameters $(\{J_{ij}\}, \{b_i\}, \{a_i\})$ were then learned using a Hybrid Monte Carlo implementation of contrastive divergence (5 outer loops iterations, 10 inner loop iterations, and adaptive HMC step size, $\epsilon$, to maintain acceptance in the range 60% – 80%, parameter learning rate 0.05), with training being performed on randomly selected batches of size 250 for 500 complete passes through the data.

---

[9]At least in those situations where the ML solutions can be calculated 'exactly' using other methods.

[10]Basically, the gradient update equation is not dimensionally correct. For more detail see, for example, Mackay [1996] and Amari et al. [1996b].

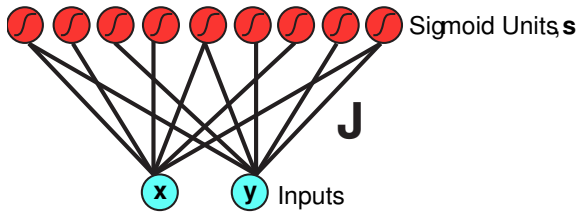

**Figure 4.6:** Model architecture for toy example. We have a single layer of sigmoid units. The energy function associated with each unit, $s_i$, is simply the activity of that unit multiplied by a scalar, $a_i$; the total energy is just a sum of these weighted sigmoid outputs.

Figure 4.7 (E) depicts the energy landscape at the end of learning, and figure shows 4.7 (F) the corresponding (un-normalised) probability density. For comparison we show the (log) probability maps for the training data in figure 4.7 (B) and (E). Also, figure 4.7 (D) shows a sample of 250 data points from the model, obtained by performing HMC for 10,000 outer loop iterations.

This example was chosen somewhat arbitrarily and is, admittedly, not especially taxing. Also, the choice of our network and energy functions was made to highlight the potential flexibility of our approach rather than because it is a good match to the data. A well known property of feedforward networks with sigmoidal units is that, with sufficient units and layers, it is possible to approximate any function to arbitrary accuracy [Bishop, 1995]. We are, however, aware of the gulf that can arise between possibility and reality; issues such as over-fitting should be bourne in mind. In a real density modelling task, of course, one would be well advised to build prior knowledge about the problem domain into the construction of the energy function and any latent variables.

### 4.8.2 Blind Source Separation

To assess the performance of contrastive divergence as a learning algorithm, we compared a hybrid Monte Carlo implementation of contrastive divergence with an exact sampling algorithm, as well as with the Bell & Sejnowski algorithm [Bell and Sejnowski, 1995] on a standard 'blind source separation' problem. The model is complete,[11] and the energy is defined through

$$s_i = \sigma(\mathbf{J}_i^T \mathbf{x}) \tag{4.48}$$

$$E_i(s_i) = -\log((1 - s_i)s_i) \tag{4.49}$$

---

[11]Recovering more sound sources than input dimensions (sensors) is not possible with our energy-based model, since the features are not marginally independent.

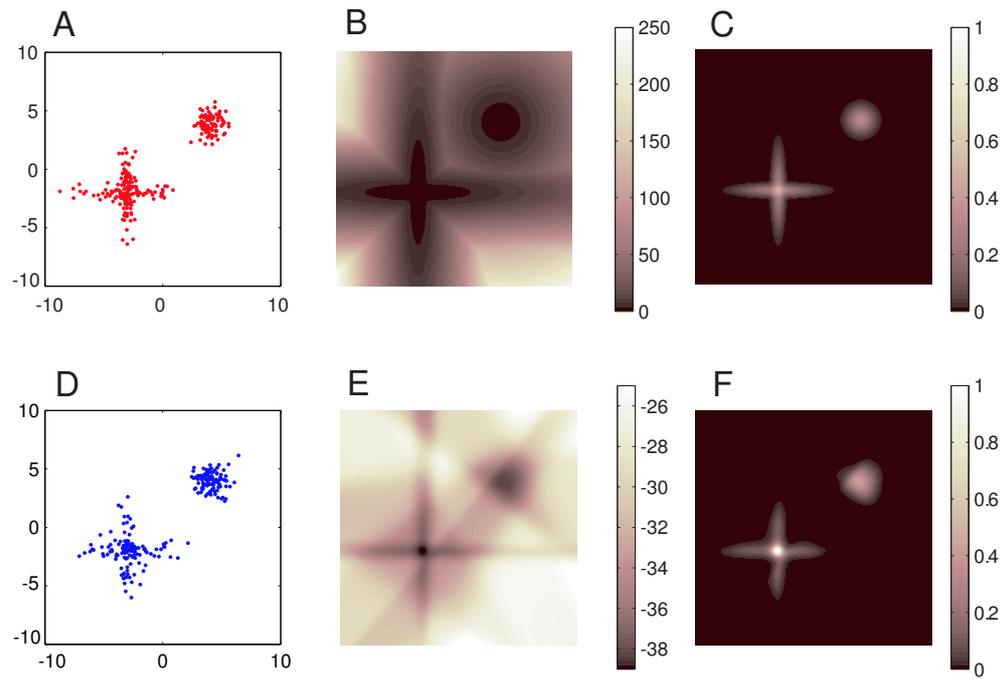

**Figure 4.7:** (A) Sample of 250 points from training data set. The full set of 3,000 was generated as an equal mixture of the following three Gaussians: $\mathcal{N}_1\left([4\ 4],[.5\ 0;0\ .5]\right)$ ;$\mathcal{N}_1\left([-3\ 2],[4\ 0;0\ .1]\right)$ ;$\mathcal{N}_1\left([-3\ 2],[.1\ 0;0\ 4]\right)$. (B) Log probability of distribution from which data was generated. (C) Probability distribution from which data was generated. (D) Sample of 250 points from an approximation to the model's equilibrium distribution after training. (E) Energy landscape learnt by model. (F) Corresponding (un-normalised) probability learnt by mode. density. (N.b. this figure uses a different color scale to many others in this thesis; this is for purposes of visibility in the probability maps.)

where $\sigma(s_i) = 1/(1 + \exp(-s_i))$ is the sigmoid function. This model is strictly equivalent to the noiseless infomax/ICA model with sigmoidal outputs used by Bell and Sejnowski [1995].

The data consisted of 16, 5-second stereo CD recordings of music, sampled at 44.1 kHz.[12] Each recording was monoized, down-sampled by a factor of 5, randomly permuted over the time-index and re-scaled to unit variance. The resulting 88436 samples in 16 channels were linearly mixed using the standard *instamix* routine with $b = 0.5$ (1 on the diagonal and 1/9 off the diagonal),[13] and whitened before presentation to the various learning algorithms.

The parameters to be learnt were the $\{J_{ij}\}$ the elements of the inverse mixing matrix; the estimated mixing matrix itself being given by $\mathbf{A} = \mathbf{J}^{-1}$. We compared three different ways of computing or estimating the gradient for the parameter updates:

Algorithm: **HMC** We used the hybrid Monte Carlo implementation of contrastive divergence. This implementation uses 1 step of hybrid Monte Carlo simulation to sample from $p^1(\mathbf{x})$; this single iteration of the outer loop consisted of 30 leap frog steps, with the step sizes adapted at the end of each simulation so that the acceptance rate is about 90%.

Algorithm: **Equil** For noiseless ICA, it is possible to sample efficiently from the true equilibrium distribution using the causal generative view. These samples can then be used to estimate the second term of 4.36. To be fair, we used a number of samples equal to the number of data vectors in each mini-batch since this is the number of samples used in the contrastive divergence approach.

Algorithm: **Exact** As previously mentioned, in the square case we can compute the partition function exactly using equation 4.17, and thus evaluate the second term of equation 4.32 exactly. This is precisely Bell and Sejnowski's algorithm.

Parameter updates were performed on mini-batches of 100 data vectors. The learning rate was annealed from 0.05 down to 0.0005 in 10000 iterations of learning,[14] while a momentum factor of 0.9 was used to speed up convergence. The initial weights were sampled from a Gaussian with a standard deviation of 0.1.

During learning we monitored the Amari distance[15] to the true un-mixing matrix. In Figures 4.8 and 4.9 we show the results of the various algorithms on the sound separation task. The main conclusion of this experiment is that contrastive divergence is able to deliver solutions that correspond to good maximum likelihood estimates, thereby helping to validate the ideas presented in this chapter.

---

[12] Prepared by Barak Pearlmutter and available at `http://sweat.cs.unm.edu/~bap/demos.html`.

[13] Available at `http://sound.media.mit.edu/ica-bench/`.

[14] 2000 iterations each at 0.05, 0.025, 0.005, 0.0025 and 0.0005.

[15] The Amari distance [Amari et al., 1996a] measures a distance between two matrices $A$ and $B$ up to permutations and scalings: $\left( \sum_{i=1}^{N} \sum_{j=1}^{N} \frac{|(AB^{-1})_{ij}|}{\max_k |(AB^{-1})_{ik}|} + \frac{|(AB^{-1})_{ij}|}{\max_k |(AB^{-1})_{kj}|} \right) - 2N^2$.

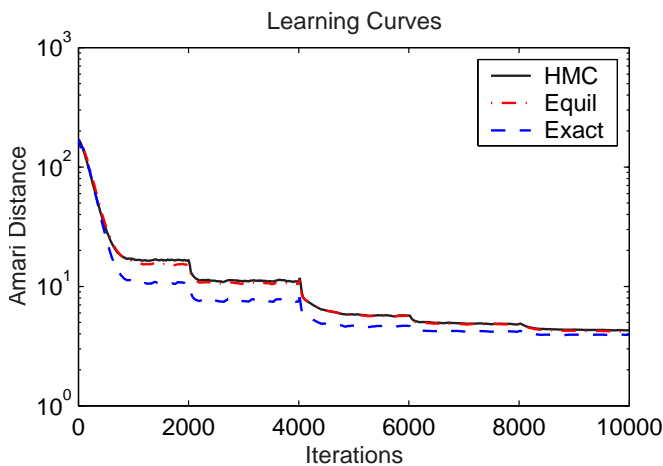

**Figure 4.8:** Evolution of the Amari distance for the various algorithms, averaged over 100 runs. Note that HMC converged just as fast as the exact sampling algorithm Equil, while the exact algorithm Exact is only slightly faster. The sudden changes in Amari distance are due to the annealing schedule.

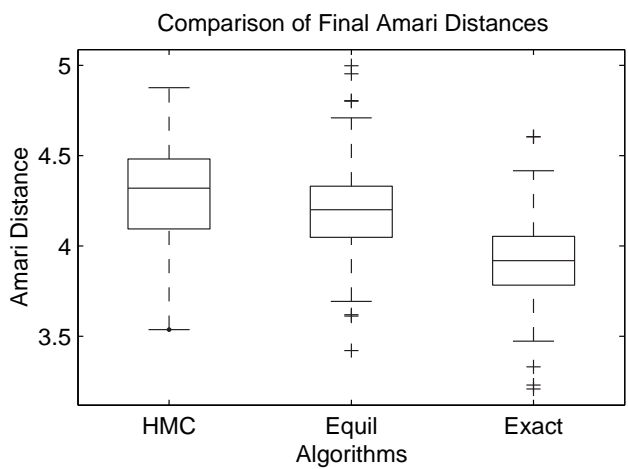

**Figure 4.9:** Final Amari distances for the various algorithms, averaged over 100 runs. The boxes have lines at the lower quartile, median, and upper quartile values. The whiskers show the extent of the rest of the data. Outliers are denoted by "+". This plot shows that the deterministic method Exact performs slightly better than the sampling methods HMC and Equil, probably due to the variance induced by the sampling. More importantly, it shows that learning with brief sampling (HMC) performs equally well as learning with samples from the equilibrium distribution (Equil).

## 4.9 Discussion

This chapter has presented the framework of energy-based models (EBM's) that will be used extensively throughout the rest of this thesis. Traditionally many models that we consider to fall within the category of EBM's (e.g. Boltzmann Machines) have been of limited practical use because of the great difficulty in performing parameter estimation. However, through the introduction of the Contrastive Divergence algorithm there is now an effective means of training such models, thus allowing their potential to be explored. The algorithm is derived from a principled objective function and has a very strong tendency to deliver solutions which are at or close to local optima of the likelihood function. (Recall that the algorithm is approximate however, and the convergence to such local optima is not guaranteed.)

We have also seen that several other types of model can be subsumed within the energy-based framework and this, combined with the flexibility of EBM's, offers the possibility of developing novel extensions to existing approaches. In conclusion we believe that EBM's provide a powerful and general modelling tool which may have great utility in many areas; in particular they may be of use in understanding neural information processing, as presented in the following chapters.

## Acknowledgements

The figures in section 4.8.2 are from experiments carried out by Max Welling; I have performed similar studies on a synthetic BSS task and obtained essentially the same results in terms of performance.

# Chapter 5

# Boltzmann Machines

## 5.1 Introduction

In chapter 4 we introduced our framework for energy-based models and the associated techniques for learning within that framework. We now build upon those theoretical foundations and use a simple, but powerful model — the Boltzmann machine — as a tool to help understand receptive field and topographic map development from the computational perspective of unsupervised, representational learning from image statistics.

The Boltzmann machine [Hinton and Sejnowski, 1986] is an energy-based architecture for unsupervised representational learning and seems like an appealing candidate model for helping understand the visual system. In particular its strengths of finding population code representations and extensibility to a hierarchy lie in the areas of weakness for many current models activity-dependent neural development.

In this chapter we present a model of ocular dominance and retinotopy development, and demonstrate successes in learning to represent inputs effectively and statistically efficiently. Our models are also able to reproduce some of the characteristic patterns observed experimentally under conditions of abnormal rearing.

**Original Contributions**

The main original contributions in this chapter are: (i) the combination of the contrastive divergence algorithm with a Boltzmann machine model having interconnected hidden units; (ii) the derivations of a variational approximation to the contrastive divergence algorithm, to make learning in such models more efficient; (iii) the use of this model and algorithm to simultaneously capture the statistical structure of simple, naturalistic inputs and to model aspects of topographic map and receptive field structure; (iv) the reproduction in this model of the phenomena of ocular dominance and retinotopic refinement, as well as the effects of strabismus and monocular deprivation; and (v) the use of this model to make basic predictions about the pattern of lateral neural connectivity between regions of different ocularity.

## 5.2   Boltzmann Machine: Model Formulation

The original formulation of the Boltzmann machine [Hinton and Sejnowski, 1986] defines probability distributions over nodes which can take on binary activities, $[0, 1]$ or "off" and "on", and owes its inspirational origins to descriptions of physical systems of particles with spin $\frac{1}{2}$, such as magnets. In such systems, we can think of each element as being able to exist in one of two states, each having a (possibly) different energy that may depend on the state of other parts of the system. As a result of this dependency between the components of the system, different configurations have different energies. The Boltzmann machine considers only simple bilinear interactions between the units which give rise to quadratic energy terms. The general overall energy function is given by,

$$E(\mathbf{x}) = -\left(\frac{1}{2}\mathbf{x}^T\mathbf{W}\mathbf{x} + \mathbf{x}^T\mathbf{b}\right) \tag{5.1}$$

where $\mathbf{x}$ in this expression denotes the binary state vector describing the configuration of the system. (The matrix $\mathbf{W}$ represents the coupling between units, and typically has zero diagonal — self-connections are forbidden. The vector $\mathbf{b}$ represents an 'external bias' term. In practice this bias can be implemented by adding to the network a suitably weighted unit whose activity is always 1; for notational convenience we will often absorb the bias parameters into the weight matrix.) The probability of finding the system in a particular configuration is given by the Boltzmann distribution based upon the this expression for the total energy,

$$p^\infty(\mathbf{x}) = \frac{1}{Z}e^{-E(\mathbf{x})} \tag{5.2}$$

$$Z = \sum_{\forall \mathbf{x}'} e^{-E(\mathbf{x}')} \tag{5.3}$$

Viewed from the perspective of a single unit experiencing the effects of all the other units, we obtain the following conditional relationship,

$$p(x_i = 1|\{x_{j\neq i}\}) = \sigma\left(\sum_{j\neq i} x_j W_{ij}\right) \tag{5.4}$$

where $\sigma(\cdot)$ denotes the sigmoid function. The probability of a given unit, $x_i$, being in the "on" state is a weighted sigmoid function of the states of other units, $\{x_{j\neq i}\}$. For our subsequent purposes we might like to think of our synthetic neurons as having a sigmoidal activation function, and the probability of being in the 1-state (or equivalently the state mean) as a mean level of neural activation. We consider

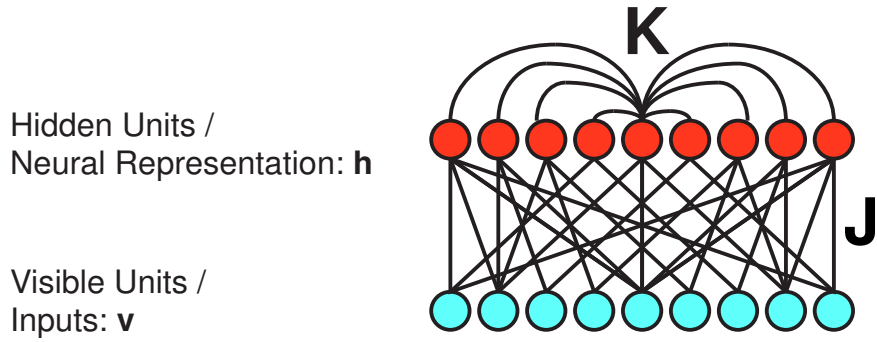

**Figure 5.1:** Schematic illustration of a general network with hidden units.

the couplings between units as embodying neural connections and synaptic inputs.

In order to build a representational system from this framework we will consider some of the $x_i$ to be 'observable' or 'visible' variables, $\mathbf{v}$, and some to be 'latent' or 'hidden' variables, $\mathbf{h}$. The visible variables will comprise the 'inputs' to the system whilst the hidden variables will comprise the internal representational elements. Using the terminology of Chapter 4, these hidden variables are stochastic rather than deterministic in nature with respect to the underlying model. Explicitly distinguishing between the two types of variable and setting up our model parameters, we will re-write the energy function of equation 5.1 as,

$$E(\mathbf{v}, \mathbf{h}) = -\left( \mathbf{v}^T \mathbf{J} \mathbf{h} + \frac{1}{2} \mathbf{h}^T \mathbf{K} \mathbf{h} \right) \tag{5.5}$$

and the corresponding probability distribution as,

$$p^\infty(\mathbf{v}, \mathbf{h}) = \frac{1}{Z} e^{-E(\mathbf{v}, \mathbf{h})} \tag{5.6}$$

We have separated the connection weight matrix into visible-to-hidden connections, $\mathbf{J}$, and hidden-to-hidden connections, $\mathbf{K}$, and we do not consider visible-to-visible connection weights. These concepts are illustrated schematically in figure 5.1.

Our goal will be to adapt $\mathbf{J}$, and possibly $\mathbf{K}$, such that the marginal distribution over the visible variables, given by

$$p^\infty(\mathbf{v}) = \frac{1}{Z} \sum_h e^{-E(\mathbf{v}, \mathbf{h})} \tag{5.7}$$

matches the distribution of our chosen input ensemble as well as possible.

We shall relate this approach to neurobiology by considering the weight parameters to be analogues of synaptic weights and the states of the hidden variables to be the analogues of neural activity in the internal representation.

# 5.3   Contrastive Divergence Learning in Boltzmann Machines

We first present the standard contrastive divergence learning rule for our Boltzmann machines, this differs slightly from the presentation in chapter 4 due to the presence of *stochastic* latent variables. Subsequently, we introduce a modified form of contrastive divergence that uses a free energy approximation [Welling and Hinton, 2002].

## 5.3.1   Preliminaries

Before beginning our exposition it will be useful to firstly define the term 'free energy', and secondly to present Jensen's inequality, upon which several derivations will rely.

### Free Energies

The free energy functional commonly arises in the physical sciences, and has the general form of an expected energy term minus an entropy term. Free energy like quantities also commonly arise in probabilistic modelling, if we relate log probabilities to energies.

To define a free energy we generally need a probability distribution, $\mathcal{P}$, over the elements $s$ of a domain $\mathcal{S}$ and a function, $E(s)$, which associates energies to these elements. The free energy of the distribution, $\mathcal{P}$, with respect to the underlying energy function, $E$, is then defined as the following functional.

$$F[\mathcal{P}] = \langle E(s) \rangle_{\mathcal{P}} - H[\mathcal{P}] \tag{5.8}$$

where $F[\mathcal{P}]$ is the free energy of the distribution $\mathcal{P}$, $\langle E(s) \rangle_{\mathcal{P}}$ is the expected energy under distribution $\mathcal{P}$, and $H[\mathcal{P}]$ is the entropy of $\mathcal{P}$.

### Jensen's Inequality

Consider a concave function[1], $f(\cdot)$, variables $s_i$ and variables $\lambda_i$ such that $0 \leq \lambda_i \leq 1$ and $\sum_i \lambda_i = 1$. Jensen's inequality states

$$\sum_i \lambda_i f(s_i) \leq f(\sum_i \lambda_i s_i) \tag{5.9}$$

with equality iff $s_i = s_j \ \forall j$ for which $\lambda_i, \lambda_j \neq 0$. Jensen's inequality also extends naturally to integrals, taken as the limit of sums. Interpreting the $\lambda_i$ as probabilities it can be expressed concisely as $\langle f(s) \rangle \leq f(\langle s \rangle)$. Figure 5.2 illustrates Jensen's inequality graphically.

---

[1] A similar inequality, with the sign reversed, holds for convex functions

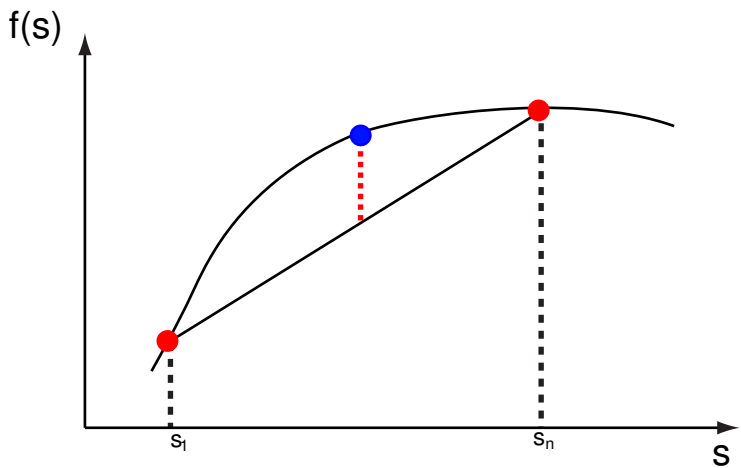

**Figure 5.2:** Graphical illustration of Jensen's inequality. Consider a convex combination of $s_i$, with bounding values denoted $s_1$ and $s_n$. We see by simple inspection of the geometry that the function evaluated at the average of the $s_i$ will always be greater than or equal to the average of the function evaluations at the $s_i$.

### 5.3.2   Standard CD Algorithm

Recapping from chapter 4, for a model over variables $\{x_i\}$ where the probability of a configuration $\mathbf{x}$ is given by,

$$p^\infty(\mathbf{x}) = \frac{1}{Z} e^{-E(\mathbf{x})} \tag{5.10}$$

the $n$-step contrastive divergence update, for a parameter $\theta_{ij}$,

$$\triangle \theta_{ij} \propto -\left\langle \frac{\partial E(\mathbf{x})}{\partial \theta_{ij}} \right\rangle_{p^0} + \left\langle \frac{\partial E(\mathbf{x})}{\partial \theta_{ij}} \right\rangle_{p^n} \tag{5.11}$$

where $p^n$ is the distribution achieved by $n$ iterations of a suitable MCMC sampler.

In the case that we partition our variables into visible variables, $\mathbf{v}$, and *stochastic* hidden variables, $\mathbf{h}$, the parameter updates in equation 5.11 are slightly modified to give us

$$\triangle \theta \propto -\left\langle \frac{\partial E(\mathbf{v}, \mathbf{h})}{\partial \theta_{ij}} \right\rangle_{p(\mathbf{h}|\mathbf{v})p^0(\mathbf{v})} + \left\langle \frac{\partial E(\mathbf{v}, \mathbf{h})}{\partial \theta_{ij}} \right\rangle_{p(\mathbf{h}|\mathbf{v})p^n(\mathbf{v})} \tag{5.12}$$

The expectations are now over the distributions for $(\mathbf{v}, \mathbf{h})$ given by $p(\mathbf{h}|\mathbf{v})p^0(\mathbf{v})$ and $p(\mathbf{h}|\mathbf{v})p^n(\mathbf{v})$, and reflect the fact that we must take into account all the possible configurations of hidden variables compatible with an input configuration, weighted by their probability of co-occurrence with that input. (This process is clearly unnecessary with *deterministic* latent variables since, in this case, there is only one compatible configuration.)

The form of the update in equation 5.12 is easily derived. Consider re-writing equation 5.7 using a modified energy function which only explicitly involves terms in $\mathbf{v}$.

$$p^{\infty}(\mathbf{v}) = \frac{1}{Z} \sum_h e^{-E(\mathbf{v},\mathbf{h})}$$
$$= \frac{1}{Z'} e^{-E'(\mathbf{v})} \tag{5.13}$$

where we have defined a new energy function[2],

$$E'(\mathbf{v}) = -\log \left( \sum_h e^{-E(\mathbf{v},\mathbf{h})} \right) \tag{5.14}$$

Substituting $E'(\mathbf{v})$ for $E(\mathbf{v})$ in equation 5.11 we obtain equation 5.12.

Applying equation 5.12 to the Boltzmann machine described in 5.5 and 5.6 we obtain the following plasticity rules for the connection matrix elements

$$\triangle J_{ij} \quad \propto \quad \langle v_i h_j \rangle_0 - \langle v_i h_j \rangle_n \tag{5.15}$$
$$\triangle K_{kl} \quad \propto \quad \langle h_k h_l \rangle_0 - \langle h_k h_l \rangle_n \tag{5.16}$$

where $\langle \cdot \rangle_0$ is now shorthand for the expectation with respect the distribution $p(\mathbf{h}|\mathbf{v})p^0(\mathbf{v})$ and $\langle \cdot \rangle_n$ is the expectation with respect the distribution $p(\mathbf{h}|\mathbf{v})p^n(\mathbf{v})$. These learning rules only require local information, with respect to the parameter that is being changed, and we can think of them as combining a Hebbian term during one phase of learning ('wakefulness') and an anti-Hebbian term during a separate phase ('sleep').

Putting all this together yields the contrastive divergence method given in Algorithm 3.

One difficulty with the updates in equations 5.17 and 5.18 is that we must take expectations over the posterior distribution of hidden variables for each input configuration. Unless we have $\mathbf{K} = 0$ — a 'Restricted Boltzmann Machine (RBM)' — the expectation of the hidden units given visibles is not generally analytically tractable, and acquiring sufficient samples from the correct posterior can be rather difficult and time consuming. This motivates us to investigate alternative approximations.

### 5.3.3 Free Energies and Variational Approximations

Consider the KL divergence between a data distribution, $p^0(\mathbf{v})$, and an energy-based model distribution $p^{\infty}(\mathbf{v})$ for a model with stochastic hidden variables. We can write

---

[2]Note this is actually the free energy of the (model specified) conditional distribution $p(\mathbf{h}|\mathbf{v})$.

---

**Algorithm 3** Standard CD Learning With Gibbs Sampling In A Boltzmann Machine

---

1. For each data case $\mathbf{v}_d$ clamp the visible units at the data vector and perform Gibbs sampling on the hidden units until they are at, or at least very close to, thermodynamic equilibrium — i.e. the posterior distribution of the model in this case.

2. Use a set of samples of the hidden units, conditioned on the corresponding data, to approximate the gradient of the energy with respect to the parameters — the positive phase gradient.

3. Fix the hidden units at one of the posterior samples, and then re-sample the visible units conditioned upon this.

    3b.   Sequentially repeat a process of sampling the hidden's given the visibles and the visibles given the hiddens for $n - 1$ times to implement a $\text{CD}^n$ learning procedure.

4. Given a set of $n$-step samples for the visible units, once again settle the hidden units.

5. Use a set of samples of the hidden units, conditioned on the corresponding data, to approximate the gradient of the energy with respect to the parameters — the negative phase gradient.

6. Update the parameters using,

$$\triangle J_{ij} = \frac{\eta_J}{N} \left( \langle v_i h_i \rangle_0 - \langle v_i h_i \rangle_n \right) \tag{5.17}$$

$$\triangle K_{kl} = \frac{\eta_K}{N} \left( \langle h_k h_l \rangle_0 - \langle h_k h_l \rangle_n \right) \tag{5.18}$$

where $\eta_J$ and $\eta_K$ are the learning rates for the visible-hidden (i.e. thalamocortical) weights and hidden-hidden (i.e. lateral or cortico-cortical) weights respectively, and $N$ is the number of samples in each mini-batch of data.

---

this as follows,

$$KL(p^0(\mathbf{v})|p^\infty(\mathbf{v})) = \int p^0(\mathbf{v}) \log \frac{p^0(\mathbf{v})}{p^\infty(\mathbf{v})} d\mathbf{v} \tag{5.19}$$

$$= \int p^0(\mathbf{v}) \log p^0(\mathbf{v}) d\mathbf{v} - \int p^0(\mathbf{v}) \log \left( \int p^\infty(\mathbf{v}, \mathbf{h}) d\mathbf{h} \right) d\mathbf{v} \tag{5.20}$$

$$= -H_0(\mathbf{v}) - \int p^0(\mathbf{v}) \log \left( \int \frac{q(\mathbf{h}; \mathbf{v}) p^\infty(\mathbf{v}, \mathbf{h})}{q(\mathbf{h}; \mathbf{v})} d\mathbf{h} \right) d\mathbf{v} \tag{5.21}$$

Where $H_0(\mathbf{v})$ is the entropy of the visible units under the data distribution, and $q(\mathbf{h}; \mathbf{v})$ is some distribution over the hidden variables which may depend upon the configuration of the visible variables. (We will actually want to think of this $q(\mathbf{h}; \mathbf{v})$ as an approximate conditional or posterior distribution for $\mathbf{h}$ given $\mathbf{v}$.) We have introduced this distribution, $q(\mathbf{h}; \mathbf{v})$, so that we can use Jensen's inequality to take the integral outside of the logarithm.

Continuing from equation 5.21 and applying inequality 5.9 we can obtain the following upper bound on the KL divergence

$$KL(p^0(\mathbf{v})|p^\infty(\mathbf{v})) \leq -H_0(\mathbf{v})$$
$$- \int p^0(\mathbf{v}) \left[ \int q(\mathbf{h}; \mathbf{v}) \log p^\infty(\mathbf{v}, \mathbf{h}) d\mathbf{h} - \int q(\mathbf{h}; \mathbf{v}) \log q(\mathbf{h}; \mathbf{v}) d\mathbf{h} \right] d\mathbf{v} \tag{5.22}$$

Substituting $p^\infty(\mathbf{v}, \mathbf{h}) = \frac{1}{Z} e^{-E(\mathbf{v}, \mathbf{h})}$ and re-ordering the terms we obtain

$$KL(p^0(\mathbf{v})|p^\infty(\mathbf{v})) \leq \underbrace{\langle E(\mathbf{v}, \mathbf{h}) \rangle_{0,q} - H_{0,q}(\mathbf{v}, \mathbf{h})}_{F_q^0} + \log Z \tag{5.23}$$

Where $H_{0,q}(\mathbf{v}, \mathbf{h})$ is the entropy of the distribution $p^0(\mathbf{v}) q(\mathbf{h}; \mathbf{v})$ and $\langle \cdot \rangle_{0,q}$ is an expectation with respect to this distribution. These two terms together define a free energy which we will refer to as $F_q^0$. As we shall see, $F_q^0$ is an upper bound on the 'true' free energy of the data distribution which we denote $F^0$. (I.e. $F^0$ is the free energy of the distribution $(\mathbf{v}, \mathbf{h}) \sim p^0(\mathbf{v}) p(\mathbf{h}|\mathbf{v})$ with respect to the energy function of the underlying model.)

It turns out that the log partition function can also be expressed as a (negative)

free energy, which we shall denote $F^\infty$, as follows

$$
\begin{aligned}
\log Z &= \log \left( \sum_{\mathbf{v}, \mathbf{h}} e^{-E(\mathbf{v}, \mathbf{h})} \right) \\
&= \log \left( \sum_{\mathbf{v}, \mathbf{h}} \frac{p^\infty(\mathbf{v}, \mathbf{h}) e^{-E(\mathbf{v}, \mathbf{h})}}{p^\infty(\mathbf{v}, \mathbf{h})} \right) \\
&= \sum_{\mathbf{v}, \mathbf{h}} p^\infty(\mathbf{v}, \mathbf{h}) \log \left( \frac{e^{-E(\mathbf{v}, \mathbf{h})}}{p^\infty(\mathbf{v}, \mathbf{h})} \right) \\
&= -\langle E(\mathbf{v}, \mathbf{h}) \rangle_\infty + H^\infty(\mathbf{v}, \mathbf{h}) \\
&= -F^\infty
\end{aligned}
\tag{5.24}
$$

Where the $H^\infty(\mathbf{v}, \mathbf{h})$ is the entropy of the distribution $p^\infty(\mathbf{v}, \mathbf{h})$, and $\langle \cdot \rangle_\infty$ is an expectation with respect to this distribution. (Jensen's inequality was used to transform the log-sum, with the equality condition holding since the ratio $\frac{e^{-E(\mathbf{v}, \mathbf{h})}}{p^\infty(\mathbf{v}, \mathbf{h})}$ is a constant.) $F^\infty$ is the free energy of the equilibrium distribution.

Putting all this together we obtain the following expression which is a difference of two free energies, the free energy of the data distribution combined with the posterior $q(\mathbf{h}; \mathbf{v})$ and the free energy of the equilibrium distribution.

$$
KL(p^0(\mathbf{v}) | p^\infty(\mathbf{v})) \leq F_q^0 - F^\infty
\tag{5.25}
$$

The bounding inequality is tight and $F_q^0 = F^0$ if we have $q(\mathbf{h}; \mathbf{v}) = \frac{p^\infty(\mathbf{v}, \mathbf{h})}{p^\infty(\mathbf{v})}$ i.e. if $q(\mathbf{h}; \mathbf{v})$ is the conditional distribution consistent with our current model.

It is easy to show that the difference between the left and right hand sides of equation 5.25 (i.e. the slackness of the bound) is the KL divergence between our approximate distribution and the true distribution.

$$
KL(p^0(\mathbf{v}) | p^\infty(\mathbf{v})) - \left( F_q^0 - F^\infty \right) = KL(q(\mathbf{h}; \mathbf{v}) | p(\mathbf{h} | \mathbf{v}))
\tag{5.26}
$$

Therefore we see that, for the purpose of learning, the best choice of posterior distribution from a given family minimises $KL\left(q(\mathbf{h}; \mathbf{v}) | p(\mathbf{h} | \mathbf{v})\right)$. Naturally this results in the 'true' posterior if the class of models over which we consider the minimisation is unconstrained. It is also easy to see that the equilibrium distribution has the lowest free energy of all possible distributions; this follows from the non-negativity of the KL divergence.

So, we can interpret minimising a KL-divergence (and thus equivalently maximising likelihood) as minimising the difference between the equilibrium free energy and the data free energy. Also, since the bound holds for any choice of conditional distribution $q(\mathbf{h}; \mathbf{v})$, we can consider approximate methods of minimising the KL divergence that do no act directly but rather seek to reduce the upper bound. This then allows us to use approximate, and more tractable, posterior distributions. As pointed out in Neal and Hinton [1998] this free energy perspective can be used to understand the EM algorithm and variants thereof that use partial maximisations and/or approximate updates.

**Contrastive Free Energies**

We will now see how we can view contrastive divergence using free energies, and then consider a modification of the contrastive divergence procedure that uses approximate free energies. We first, however, introduce some additional notation. We will use the symbol $F^n$ to refer to the 'true' free energy of the $n$-step distribution, where true implies that we are using the conditional distribution consistent with our model.

The n-step contrastive divergence was defined as

$$CD^n = KL(p^0 \| p^\infty) - KL(p^n \| p^\infty) \tag{5.27}$$

which using equation 5.26 can also be re-written as the free energy difference

$$CD^n = (F^0 - F^\infty) - (F^n - F^\infty)$$
$$= F^0 - F^n \tag{5.28}$$

The n-step distribution, by its very definition has a lower free energy than the data distribution. In fact, one way of stipulating an appropriate equilibrium invariant Markov chain operator is by requiring it to descend the gradient of free energy in *distribution space* until we are at the desired equilibrium distribution.

This suggests an approximation in which we replace the 'true' free energies in equation 5.28 with variational approximations. We will replace the 'true' free energy of the data with $F_q^0$. The approximation to $F^n$ is a little more subtle. In place of $F^n$ we will use the free energy, $\widetilde{F^n}$, of a distribution that is obtained by performing gradient descent of free energy in a space of approximating distributions (rather than the space of all distributions), starting at $(\mathbf{v}, \mathbf{h}) \sim p^0(\mathbf{v})q(\mathbf{h}; \mathbf{v})$. This gives us the following variational cost function,

$$\widetilde{CD^n} = F_q^0 - \widetilde{F^n} \tag{5.29}$$

Note, we cannot be sure that $\widetilde{F^n} \leq F^0$, and $\widetilde{CD^n}$ does not give us any sort of bound on $CD^n$. Furthermore, reducing this new cost function is not guaranteed to reduce the true free energy of the data distribution. However, if the variational approximations employed are reasonable then we may hope that the new cost function will still be beneficial with respect to our ultimate goals.

Therefore, subject to having good approximate distributions we can hope to have an effective algorithm by performing approximate[3] gradient descent of the cost function in equation 5.29.

**Mean Field Approximation**

There are several types of approximate distribution $q(\mathbf{h}; \mathbf{v})$ that could be usefully applied to Boltzmann machines. A simple and effective method is to adopt a mean field approach[Peterson and Anderson, 1987], which approximates the true conditional distribution with one that is fully factorised. For example

$$q_{\mathrm{MF}}(\mathbf{h}; \mathbf{m}) = \prod_i m_i^{h_i}(1 - m_i)^{1-h_i} \qquad (5.30)$$

Here the $m_i$ are variational parameters and specify the mean activities of each unit. It turns out that there is a simple form of iterative update that allows us to find optimal values for the $m_i$, i.e. the values that best match the true conditional distribution and hence minimise the free energy of the current approximation within the fully factorised class of distributions. These optimal values are found through a set of recursively defined equations obtained by minimising the KL-divergence between $q_{\mathrm{MF}}(\mathbf{h}; \mathbf{m})$ and $p(\mathbf{h}|\mathbf{v})$ with respect to $\{m_i\}$. I.e we define our approximating posterior distribution as,

$$q(\mathbf{h}; \mathbf{v}) \equiv \arg\min_{\mathbf{m}} KL(q_{\mathrm{MF}}(\mathbf{h}; \mathbf{m})|p(\mathbf{h}|\mathbf{v})) \qquad (5.31)$$

A set of recursive updates that yield this minimisation are given by

$$\forall i \qquad m_i \leftarrow \sigma\left(\sum_{j \neq i} K_{ij} m_j + \mathbf{J}_i^T \mathbf{v}\right) \qquad (5.32)$$

and such updates are guaranteed converge to a fixed point if performed asynchronously. For the more commonly used synchronous update scheme it is necessary to use a small amount of damping to ensure convergence.

We can also perform a similar procedure to find an approximation for the visibles given this factorised distribution for the hiddens, and then given these new visibles we may re-settle the hiddens, and so on. This process can be seen as being analogous to applying an equilibrium invariant Markov chain in the sense that, within the family of fully factored distributions, it performs gradient descent of free energy. If run to absolute convergence then this procedure would give the mean field approximation for the equilibrium distribution. However, for our purposes we need only run this scheme for a one step.

---

[3]As with the original contrastive divergence objective, when computing gradients there are terms that we are unable to calculate (but which we expect to be small) that we ignore. These terms arise as a consequence of the interaction between parameter changes and the distributions reached by the variational approximation and distributions obtained by the n-step procedure.

Mean field approximations to the full equilibrium distribution can often be rather poor because the true equilibrium distribution typically has more than one mode, and this is poorly approximated by a unimodal, fully factorised form. However, the mean field approximations used in approximate CD learning are expected to fare much better since the conditional distribution given a single data point should often be fairly well approximated by a unimodal distribution. Consequently, even though we are now several layers of approximation away from our true objective we still have reason to expect good performance.

Having obtained variational approximations we then use these means, $m_i$, in place of the true distributions when computing the parameter updates. This yields updates of the form

$$\triangle J_{ij} \propto \langle v_i m_i \rangle_0 - \langle \mu_i^n m_j^n \rangle_n \tag{5.33}$$

$$\triangle K_{kl} \propto \langle m_k^0 m_l^0 \rangle_0 - \langle m_l^n m_k^n \rangle_n \tag{5.34}$$

Where $m_i^s$ is mean field approximation for hidden unit $i$ given the current the visible configuration at 'step' $s$, which for $s = 0$ is the data distribution and for $s > 0$ is a mean field approximation. $\mu$ is mean field approximations for the visible distribution given the current mean field hidden distribution.

The modified contrastive divergence algorithm, employing a variational approximation, is summarised in Algorithm 4.

This mean field algorithm was found to have superior performance and execution time than the standard algorithm, Algorithm 3, and was used for all the results reported in this chapter (although for most of the simulations we set $\eta_K$ to 0, i.e. the lateral connections were not learned). Additionally, a value of $n = 1$ in $CD^n$ was used — this was for reasons of both speed and performance.

## 5.4   Experiments: The Model Set-Up

Figure 5.3 depicts our principal model set-up. We have two visible layers, $\mathbf{v}^L$ and $\mathbf{v}^R$, each with $n_v$ units, which we imagine to be the retinal/thalamic activities for signals from the left and right eyes respectively. Additionally we have a single hidden layer, $\mathbf{h}$, containing $n_h$ units, which we interpret as the V1 cortical representation. The activity of a unit in $\mathbf{h}$ might be thought of as analogous to the average activity within a column. The thalamocortical projection strengths are given by $\mathbf{J}^L$ and $\mathbf{J}^R$, whilst $\mathbf{K}$ denotes the pattern of intra-cortical wiring. For many of the simulations reported in this chapter the pattern of cortical wiring was kept fixed as a Mexican-hat/difference-of-Gaussians (DoG) pattern. Our approach does, however, allow us to learn these connection weights and this issue is discussed in section 5.5.3. As an additional constraint, the feed-forward weights $\mathbf{J}$ were forced to take on only

---

**Algorithm 4** Mean Field Variational CD Learning in Boltzmann Machine

---

1. For each data case $\mathbf{v}_d$ clamp the visible units at the data vector and perform mean field settling on the hidden units until convergence (within some tolerance of change between iterations) using synchronous updates of the form,

$$\mathbf{m} \leftarrow [(1 - \lambda)\sigma\,(\mathbf{Jv} + \mathbf{Km}) + \lambda\mathbf{m}]$$

   where $\sigma(\cdot)$ indicates the sigmoid function, and $\lambda$ is a small damping parameter which helps prevent the oscillations that can occur due to our use of synchronous updates. (Asynchronous updates avoid the need for damping, but are implementationally much slower.)

2. Use the visible-conditioned hidden unit fixed points to compute the gradient of the approximate free energy (the 'positive' or 'wake' phase gradient).

3. Fix the hidden units at the posterior means, and then use these real numbers as states when computing the visible means. I.e. Compute the visible updates,

$$\mu \leftarrow \sigma\left[\mathbf{J}^T\mathbf{m}\right]$$

   3b. Sequentially repeat a process of settling the hidden's given the visibles and the visibles given the hiddens for $n - 1$ iterations to implement a CD$^n$ learning procedure.

4. Based upon these mean field values for the visible units, once again settle the hidden units.

5. Use this final pairings of mean field values for the visible, $\mu$, and hiddens, $m$ to compute the gradient of the approximate free energy (the 'negative' or 'sleep' phase gradient).

6. Update the parameters using,

$$\triangle J_{ij} = \frac{\eta_J}{N}\left(\langle v_i m_i\rangle_0 - \langle \mu_i^n m_j^n\rangle_n\right) \qquad (5.35)$$

$$\triangle K_{kl} = \frac{\eta_K}{N}\left(\langle m_k^0 m_l^0\rangle_0 - \langle m_l^n m_k^n\rangle_n\right) \qquad (5.36)$$

   where $\eta_J$ and $\eta_K$ are the learning rates for the visible-hidden (i.e. thalamocortical) weights and hidden-hidden (i.e. lateral or cortico-cortical) weights respectively, and $N$ is the number of samples in each mini-batch of data.

---

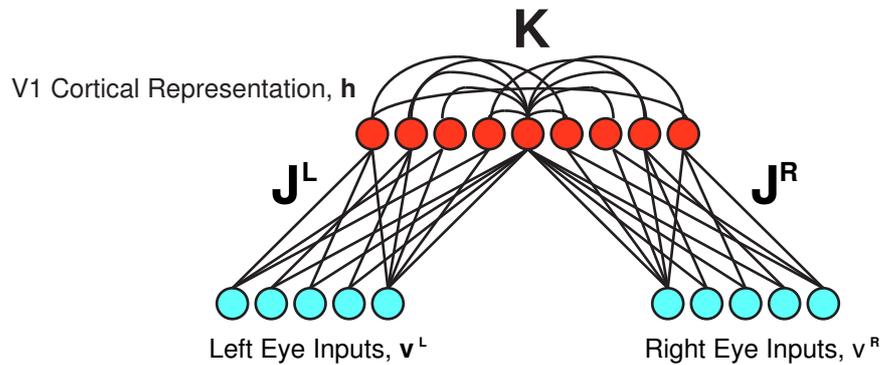

**Figure 5.3:** Model set-up. We have input layers for the left and right eyes — these might be thought to (crudely) reflect activity in the retina or LGN. We also have a fully laterally inter-connected hidden layer, which we consider to be our cortical representation.

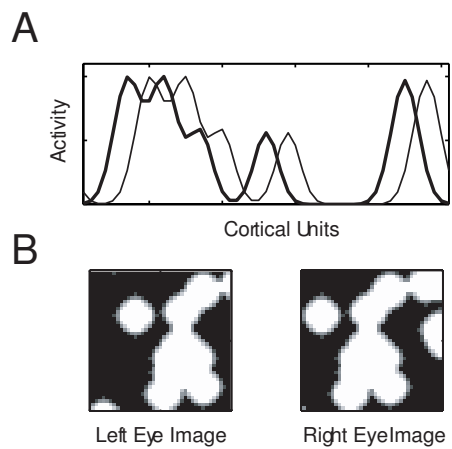

**Figure 5.4:** Examples of some of the synthetic inputs generated. The left eye image is essentially a shifted version of the right eye image, with circular boundary conditions in the 1-dimensional case. The standard deviation of this Gaussian shift distribution is denoted $\Sigma$. (A) Two input profiles of the sort used in the the 1-d simulations. The cortical units are ordered along the abscissa, the activity is plotted as the corresponding ordinate. (B) Images showing two sample inputs of the sort used in the 2-d simulations, the gray scale denotes the activity level.

positive values. This is not a requirement of the our methodology, but was found to improve the appearance of the final weight matrices. Initially, to guide intuition and aid analysis, we consider a 1D topology for which the inputs and representational layer are considered to lie on a ring. Subsequently we consider a 2D topology with inputs and representations lying on a 2D grid.

### 5.4.1 Inputs

Examples of the patterns presented to the models for both 1- and 2-dimensional simulations are illustrated in figure 5.4. The same general process was used to create both types of synthetic data: in the stereo experiments, mono-images were synthesised initially and then processed to make stereo-pairs. To create the mono-images a small number of input locations were seeded with delta functions, then an iterative process of convolution with a Gaussian envelope followed by a thresholding to ensure values remained in the range $[0, 1]$ was applied. The resulting images are not quite true binary patterns, however the real values between zero and one were input as raw data regardless. There are several ways in which this can be justified — we can either view the inputs as themselves being mean pixel values, or we could interpret the whole model as a rate-based Boltzmann machine as in Teh and Hinton [2001]. The rate-based Boltzmann machine relies on the idea that we could replicate $m$ copies of each unit, and scale the weights to each replicated ensemble appropriately by $1/m$. When we set $m$ to be relatively large this gives essentially similar results to just using a single unit and taking its mean activity.

The resulting inputs that we supply to the model are synthetic, but naturalistic, in their content. Our method of input generation was devised with the goal of producing synthetic patterns which possess some of the important properties of real images. The inputs typically have several length scales, and several 'objects' per scene. Furthermore, the resulting scenes are richer than mere templates but are still composed of relatively simple parts.

This choice of training data was made primarily to allow us tighter parametric control over the learning environment but also because it is difficult to format digitised natural scenes in a way that makes them appropriate for the Boltzmann machine — even with a rate-based approach it is difficult to deal with arbitrary real values in the Boltzmann machine framework. One intuition for this is as follows: the shape of the sigmoid activation function makes it is easy for the model to specify tight distributions that are near to 1 or 0 since the upper and lower saturation bounds of the function allow a wide range of inputs to produce the same output; conversely it is difficult to specify tight distributions near $\frac{1}{2}$ since this is where the slope of the function is greatest.

'Stereo pairs' of images (one for each eye) were generated by taking the image for one eye and applying a horizontal translation drawn at random from a zero mean Gaussian distribution before choosing the image for the second eye[4]. In the 1D case the translations were performed with circular boundary conditions; in the 2D case a large synthetic image was generated and then within a prescribed boundary two shifted images were extracted. The degree of correlation between two pixels at corresponding locations in each eye can be varied by altering the standard deviation of the shift distribution. This standard deviation parameter, denoted $\Sigma$, is measured in 'input units' — the spacing of two laterally adjacent input elements is taken to be one such unit.

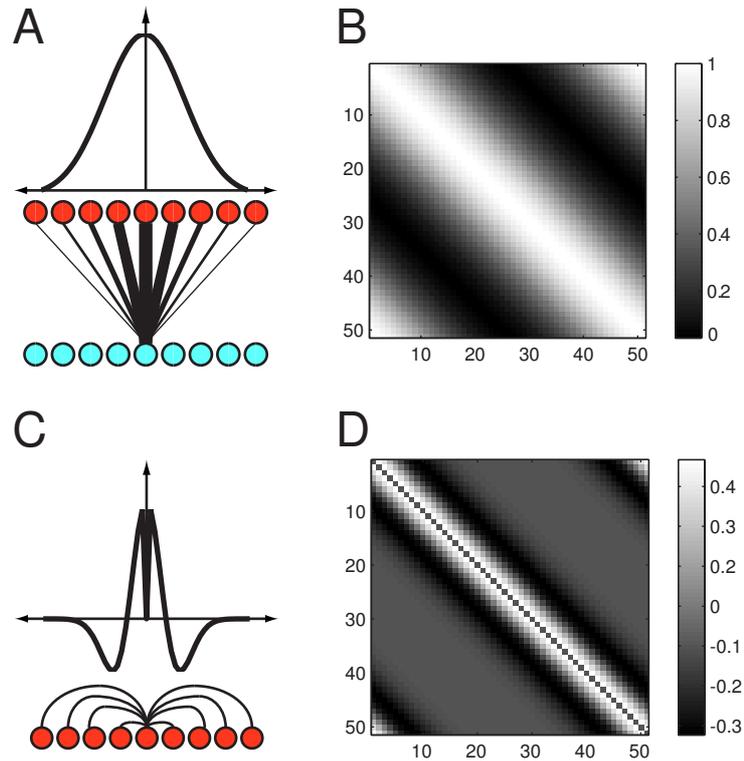

**Figure 5.5:** Examples showing the type of initial conditions applied in our experiments. (A) The connection matrices $\mathbf{J^R}$ and $\mathbf{J^L}$ are initialised such that each unit has a broad Gaussian weight pattern centred on a nominal corresponding retinotopic location. The width of this Gaussian is denoted $\sigma_0$ and the peak strength is denoted $s_0$. (B) Example feedforward weight matrix (for one eye) from a 1-dimensional model. The parameters are: $n_v = n_h = 51$, $\sigma_0 = 13.5$. (C) The lateral connection strength between hidden units is a function of their cortical separation and is given by a difference of two zero mean Gaussians (normalised to unit area), $\mathcal{G}_2 - \mathcal{G}_1$, parameterised by variances $\sigma_1$ and $\sigma_2$. The resulting function is scaled to have peak amplitude $s_m$, before the 'self-connection' is set to zero. The picture schematically shows the lateral connections for one unit. (D) Example lateral connectivity matrix from a 1-dimensional model. The parameters are: $\sigma_1 = 4.77$, $\sigma_2 = 4.66$, and $s_m = .5$.

### 5.4.2   Initial Conditions

Activity independent chemical matching processes are believed to be responsible for initial coarse order in developing projections (e.g.: Tessier-Lavigne and Goodman [1996]). In almost all models such as ours this coarse initial order is essential for the appropriate global order of the final projection. We chose to initialise our feedforward weights as a very broad Gaussian pattern centred retinotopically on a ordering of the hidden units. This ordering gives a spatial location to the hidden units which otherwise, from the point of view of learning, are not embedded in any physical space.

As mentioned previously we maintain the intra-cortical connections as a fixed difference of Gaussians pattern unless otherwise stated; this configuration has short-range excitation and inhibition at longer distances. This ties in with prevalent, although not all, current opinions on the likely pattern of intra-cortical influence as a function of distance from a column. Figure 5.5 illustrates our initial conditions and defines the initialisation parameters $\{(\sigma_0, s_0), (\sigma_1, \sigma_2, s_m)\}$ which set the width and amplitude of the connection weights.

It was necessary to tune the width and scale of the lateral connections and the form of the initial feedforward conditions somewhat in order to produce some of our more beauteous results, and we discuss this issue later.

Finally, the input to each cortical unit from the two eyes were initialised as being the same, up to the precision of a small amount of Gaussian noise added to aid symmetry breaking. There is evidence, at least anatomically, that this is not the case — Crair et al. [1998] show that the contra-lateral eye usually has a strong initial bias, and OD stripes form as the strength of the projection from the ipsilateral eye increases. No model currently addresses this issue, and ours is no exception.

## 5.5   Simulation Results

### 5.5.1   One Dimensional Simulations

**Simple Tests**

As an initial test of the ability of our model/algorithm to learn good representations of our synthetic data, we applied it to a single (as opposed to stereo) one-dimensional input array using inputs of the type shown in figure 5.4.

The initial and final weight patterns are shown in figure 5.6 (A-C) and one can clearly see that the retinotopy has been refined. Panel (D) of the same figure shows example data patterns and the corresponding reconstructions, generated by settling the hidden units to the mean field fixed point of equation 5.32, and using these fixed point activities to reproduce the inputs. The two sets of patterns visibly match each other well, indicating that the hidden units are able to represent the inputs faithfully.

---

[4]We recognise that there are more sophisticated ways of generating pseudo-stereo images, however the procedure described here was deemed adequate for the current purposes

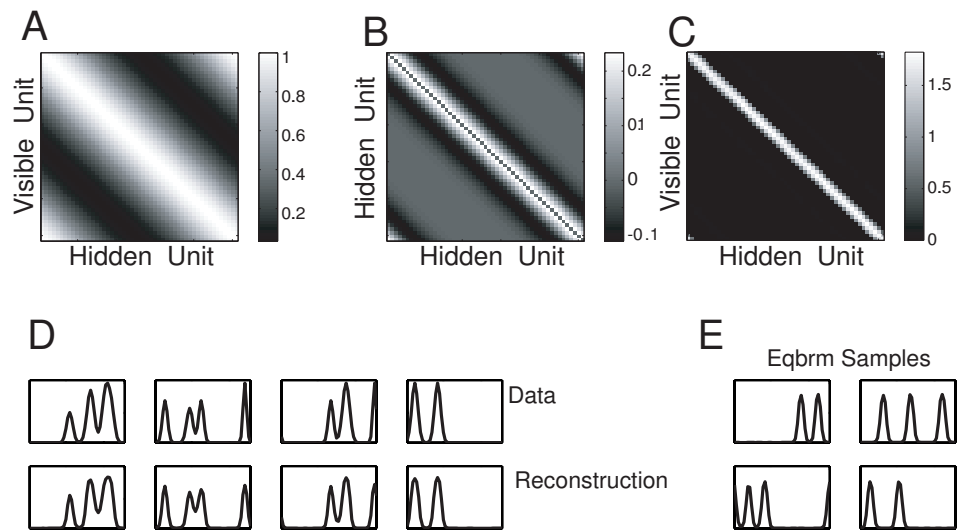

**Figure 5.6:** Simple tests of generative abilities.(A) and (B) show the initial visible-hidden connections, and the fixed lateral connections respectively for a monoscopic model. The initialisation parameters used were: $n_v = n_h = 51$, $(\sigma_0, s_0) = (30, 0.1)$, and $(\sigma_1, \sigma_2, s_m) = (4.77, 4.66, 0.25)$. (C) shows the visible-hidden connections after training. The retinotopy has clearly undergone extensive refinement. (D) Illustrates the fidelity with which the trained model can reconstruct inputs from the induced hidden unit representation. (E) Shows four samples from an approximation of the equilibrium distribution, achieved by Gibbs sampling with annealing.

Turning our attention to the statistical properties of the model, panel (E) in figure 5.6 shows 'fantasy' data generated from the model after a very long period of Gibbs sampling combined with simulated annealing [Kirkpatrick et al., 1983], which should provide samples from a distribution close to that of the equilibrium distribution in equation 5.7. These fantaies do indeed match the inputs well, and show that our model and algorithm can be effective at learning a density model for this simple class of inputs.

**Retinotopic refinement & ocular dominance in 1D**

We now move on to consider a 1-dimensional model with stereoscopic inputs. The following initialisation parameters were used: $n_v = n_h = 51$, $(\sigma_0, s_0) = (30, 0.1)$, $(\sigma_1, s_1) = (4.77, 0.25)$; and $(\sigma_2, s_0) = (4, 66, 0.25)$, and the data has a shift distribution with $\Sigma = 15$.

Figure 5.7 (A) and (B) present the initial feedforward and lateral weight patterns, respectively. The final feedforward weight patterns for left and right eye inputs are shown in 5.7 (C) and (D), with the 'sum mode' and 'difference mode' in figures 5.7 (E) and (F). The results clearly show that the connections have undergone a symmetry breaking process to give rise to a final pattern of alternating eye preferences. In addition, we see that the weight pattern has also undergone extensive retinotopic

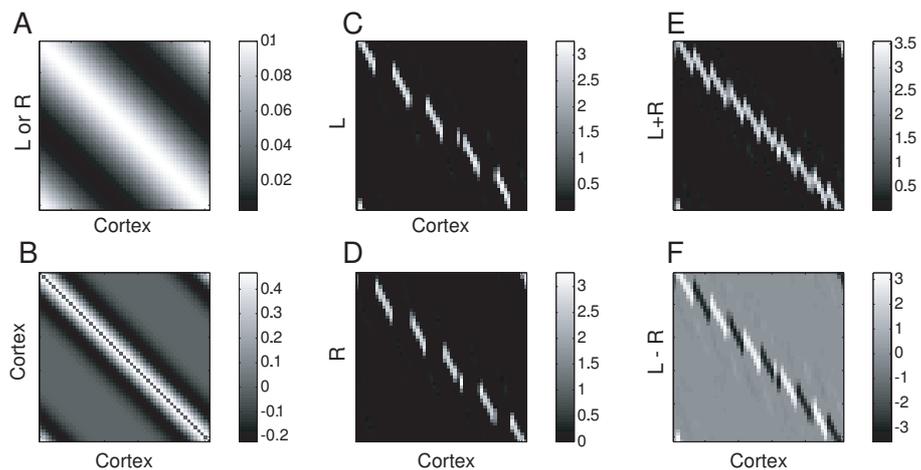

**Figure 5.7:** (A) shows the initial connectivity matrix between visibles to hiddens. This was the same for both eyes. (B) shows the fixed lateral connectivity matrix. (C) and (D) show the weight patterns from the left and right eyes after learning respectively. We note that in addition to retinotopic refinement, there has been a (complete) segregation of ocularity. (E) shows the 'sum mode', $\mathbf{J^R} + \mathbf{J^L}$, the combination of weights from both eyes to a given cortical unit. (F) shows the 'difference mode', $\mathbf{J^R} - \mathbf{J^L}$, which highlights the ocular dominance structure in the connections.

refinement and that, in conjunction with the ocular dominance pattern, a given retinal location is represented roughly equally by both left and right eye-preferring regions. In terms of developmental dynamics we observed that the evolving weight patterns underwent considerable retinotopic refinement *before* the signs of ocular dominance began to develop.

### Pattern dependence on lateral connections and initial conditions

We have noted that the feedforward weight patterns which arise from learning depend rather sensitively on the structure of the lateral connections and on the initial conditions. We now explore this issue in more detail, and study some perturbations about the parameters used in figure 5.7. Representative results of such explorations are shown in figure 5.8.

The following general observations have been made.

**Visible-Hidden Width:** The outcome is fairly robust to changes in the width of the initial visible-hidden weight pattern. Across an order of magnitude, or so, in width the outcome is little affected as shown by figure 5.8 panel (B). However, as we move to very broad initialisation conditions we see considerable disruption in the overall pattern, as evidenced by sub-panel (B-iv). In such cases, the initial bias is insufficient for the system to favour a globally coordinated scheme and whilst there are several patches with locally contiguous retinotopy, there is no global order.

**Visible-Hidden Strength:** The amplitude with which the visible-hidden weights are initialised can have significant effects on the outcome of the final patterns. If the initialisation is very weak then it is common to see hidden units that cease to be active and whose total hidden-visible connectivity essentially falls to zero; the resulting patterns are rather disordered, for instance panel (C-i). Aside from this there seems to be a moderate trend in which larger initial weights tend to lead to narrower ocular dominance bands.

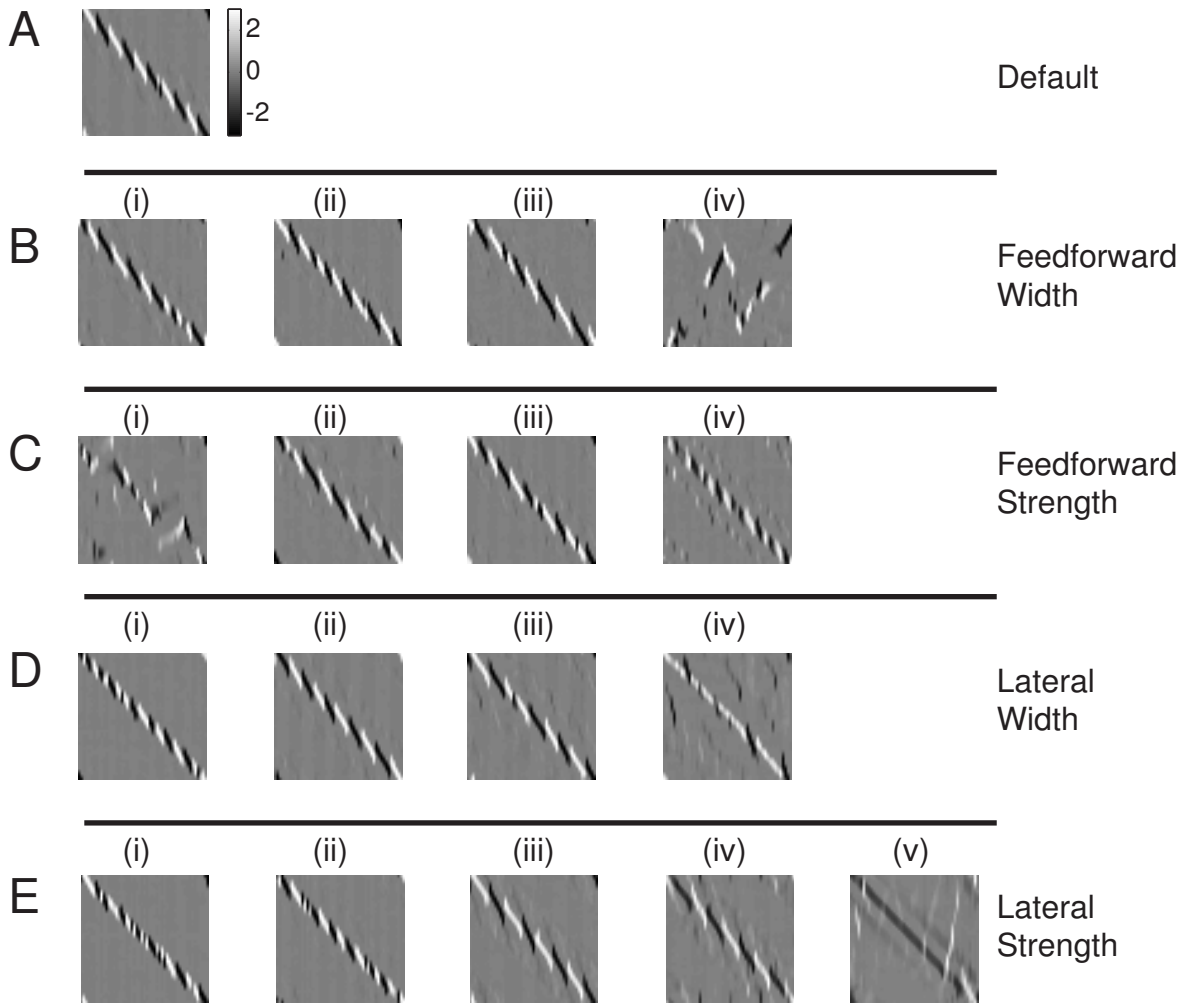

**Figure 5.8:** Figure illustrating some of the dependency on lateral connectivity and initial visible-hidden weights; both can have a significant effect on the final weight patterns generated. The difference mode for various conditions is shown. (A)Default conditions. Parameter settings are:$n_v = n_h = 51$, $(\sigma_0, s_0) = (14, 0.1)$, and $(\sigma_1, \sigma_2, s_m) = (4.77, 4.66, 0.5)$ (B) Effects of altering the width of the initial visible-hidden connections, keeping other parameters fixed. The parameter $\sigma_0$ was scaled, relative to the default by: (i)$\frac{1}{20}\times$, (ii)$\frac{1}{4}\times$, (iii)$2\times$, (iv)$6\times$. (C) Effects of altering the strength of the initial visible-hidden connections, keeping other parameters fixed.The parameter $s_0$ was scaled, relative to the default by: (i)$\frac{1}{8}\times$, (ii)$\frac{1}{4}\times$, (iii)$4\times$, (iv)$64\times$. (D) Effects of altering the width of the lateral connections, keeping other parameters fixed. The parameters $\sigma_1$ and $\sigma_2$ were both scaled, relative to the default by: (i)$\frac{1}{3}\times$, (ii)$\sqrt{1.5}\times$, (iii)$\sqrt{2}\times$, (iv)$2\times$. (E) Effects of altering the strength of the lateral connections, keeping other parameters fixed. The parameter $s_m$ was scaled, relative to the default by: (i)$0\times$ — i.e. zero lateral interaction, (ii)$\frac{1}{2}\times$, (iii)$2\times$, (iv)$3\times$, (v)$4\times$, (vi)$6\times$.

**Lateral Width:** Scaling the width of the DoG lateral connections can have significant effects on the ocular dominance pattern that evolves. Generally speaking, narrower profiles lead to higher frequency ocular dominance patterns whilst broader profiles lead to lower frequency patterns. Very broad profiles can lead to a disordered weight configuration with unilateral dominance, such as in panel (D-iv).

**Lateral Strength:** Panel (E-i) illustrates the type of weight pattern that is obtained in the absence of lateral connections. Provided that the initial visible-hidden connections are not too broad, we still obtain retinotopic refinement. However, the resulting ocular dominance patterns are rather irregular in terms of periodicity. As we increase the strength of the lateral connections, there seems to be a trend towards broader (and hence fewer) ocular dominance bands as we increase the strength of the lateral connections, for example panels (E-ii) to (E-iv). At very large values of lateral connection strength, the learning becomes very slow because the iterative settling in the lateral units takes a long time; the system is rather frustrated at such large weight values. This also impacts the pattern of weights obtained and destabilises periodic ocular dominance patterns. In the example shown in panel (E-v) one eye has practically dominated the whole representational layer.

When trying to optimise a non-convex, highly non-linear system using a gradient based method, it is common to get trapped in local optima; our system is no exception. The particular mode that we end up may depend on several factors, a key one being the initialisation conditions. Panels (B) and (C) of figure 5.8 illustrate this rather clearly. In addition to these somewhat qualitative changes based upon different starting conditions there can also be significant trial-to-trial variation under the same initial conditions. In some regions of the parameter space (for instance the default settings from figure 5.8) the outcome is very consistent, however there are other regions of parameter space (such as that exemplified by figure 5.8 panels (B-iv) or (C-i)), in which there is considerable variation in the solutions obtained. For instance, whilst the majority of runs using the conditions of (C-i) *did* result in a rather disordered pattern, a small number arrived at a solution much more like that shown in panel (A).

We attempted to better understand the developmental behaviour of the system by performing a linear eigenanalysis of the growth of the difference-mode about an (apparent) fixed-point of the sum-mode. Unfortunately, this simplification proved to be insufficient to capture the behaviour of the system and the linear predictions did not match the empirical results particularly well. On a related note, we have been unable to determine the form of globally optimal solutions for a given set of inputs — although we speculate that the oscillatory ocular structure we see will be close to optimal for many settings of the Mexican-hat weight parameters.

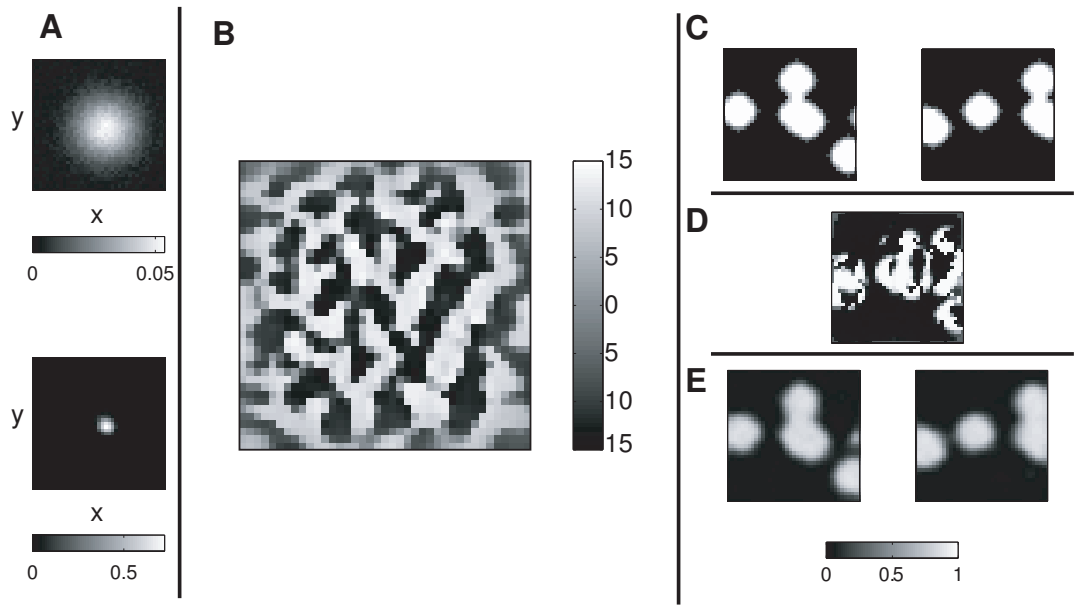

**Figure 5.9:** Retinotopic refinement and ocular dominance in 2D. (A) illustrates retinotopic refinement in the weights to a single hidden unit. The top figure shows the initial visible-hidden weight pattern from one eye, whilst the bottom figure shows the pattern for the same unit after training. (B) illustrates the two-dimensional ocular dominance pattern. Each pixel in the $39 \times 39$ image represents the summed ocularity for the hidden unit at the corresponding position in the 'cortical' layer. The ocularity is simply computed as the sum of the strength of left eye input weights into a unit, minus the sum of the strength of the right eye input weights into the same unit. (C) gives an example of a 'left-eye, right-eye' input pair. (D) shows the representation of this stereo pair in the hidden layer, i.e. the hidden activity pattern after mean field settling. Note the distributed nature of the representation. (E) shows the reconstruction of the original inputs based on the hidden representation in C. This is obtained by taking the settled hidden activities and computing the conditional means of the visible units, given the hiddens.

### 5.5.2 Two Dimensional Simulations

**Retinotopic refinement and ocular dominance in 2D**

We now move on to consider simulations using a square, two-dimensional grid of units in each layer. The following initialisation conditions and parameters we used: $n_v = n_h = 39 \times 39$, $(\sigma_0, s_0) = (30, 0.1)$, and $(\sigma_1, \sigma_2, s_m) = (12, 11, 0.05)$. The standard deviations $\sigma_1$ and $\sigma_2$ now define a circularly symmetric DoG pattern. The data used in training had a shift standard deviation of $\Sigma = 5$.

Figure 5.9 panel (A) shows the refinement of two dimensional retinotopy. In this example we see the rather broad initial projection to the hidden unit has been localised considerably. Panel (B) shows a characteristic pattern of ocularity developed which developed after training. Each pixel in the image represents the summed magnitude of inputs from the one eye to a unit, minus the summed magnitude of inputs from the other eye to that unit. This pattern has many of the hallmarks of the ocular dominance maps found in visual cortex. (We believe the tendency towards low ocularity at the boundaries of the map to be due to edge effects.)

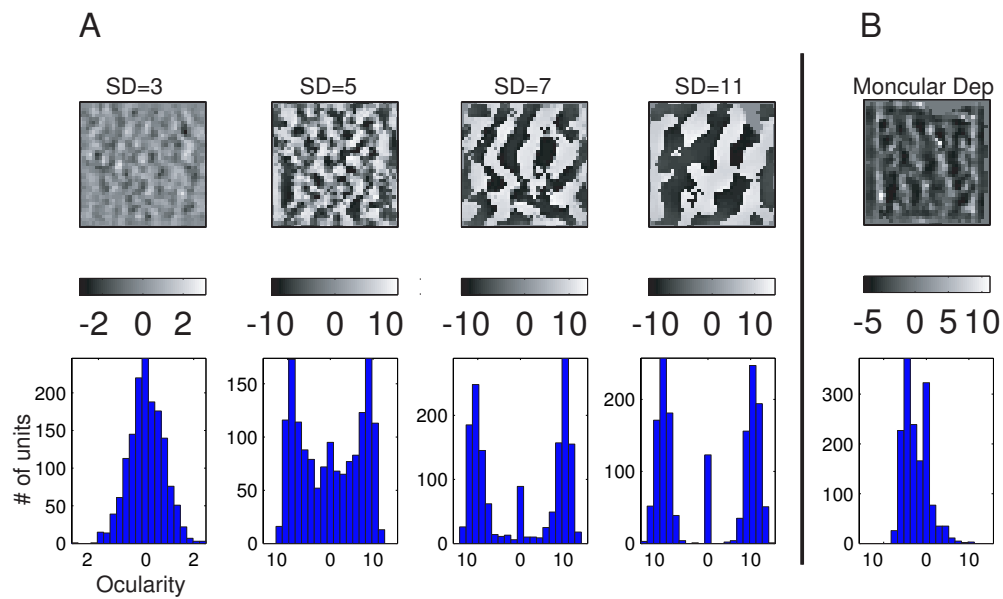

**Figure 5.10:** Representative results for two 'developmental manipulations.'(A) Illustrates the changes in the ocular dominance pattern as we alter the standard deviation of the distribution of shifts between the two eyes. Each column shows a map of ocularity and the histogram of ocularity. The input shift distribution for each column is labelled at the column head.(B) Illustrates the changes in ocular dominance when we reduce the inputs to one eye by 70% during training. Note, in both (A) and (B) the presence of slight boundary artifacts; these are most clearly visible in the histograms (spike at zero.)

**Simulated Strabismus in 2D**

In addition to the effects of initial conditions and lateral connectivity as discussed in the 1-dimensional models, we expect the statistical structure of the inputs to have a significant influence on the structure of the weight patterns learned. This is also a feature of real biological systems, as evidenced by the results of numerous experimental manipulations to normal development. We have implemented some 'equivalent' developmental interventions in our model.

Figure 5.10 (A) shows the effect of altering the standard deviation of the shift distribution used to generate the stereo pairs of inputs. Increasing the standard deviation of the shift distribution is similar to studies of induced strabismus since this intervention effectively decreases the degree of correlation between the inputs to the two eyes. Decreasing the standard deviation of the shift distribution has the opposite effect on correlations. Each sub-figure clearly expresses the characteristic

fingerprint pattern of alternating regions of ocular preference. As we move from left to right, the correlation between the eyes decreases and there is a concomitant increase in the width of the stripes.

In addition to the change in pattern morphology, we also notice an interesting change in the distribution of ocular preferences of the individual units (histograms in figure 5.10 (A). When the correlation between the two eyes is relatively large, the distribution of ocular preferences is unimodal and most units are pretty much binocular. This gradually changes to a bimodal distribution of ocularity as we decrease the correlation between units and eventually gives rise to a situation in which very few units have a binocular response and almost all are strongly dominated by one eye or another. (We note again the presence of artifacts due to edge effects, giving rise to a peak at zero ocularity in the histograms. This peak is absent if we only consider units from a smaller internal region of the map, away from the edges.)

The dependence of stripe frequency on the correlational structure of inputs has been investigated *in vivo* by observing the cortical maps of kittens that have been made strabismic by disrupting the muscles that control gaze direction in one eye Lowel [1994]. The set of patterns shown in figure 5.10 is a good qualitative match to the experimental findings that stripe width increases as inter eye correlations decrease. This finding is also in common with several of the models discussed in Chapter 3, for instance Goodhill [1993].

**Monocular Deprivation in 2D**

Variants of monocular deprivation Hubel et al. [1977], Horton and Hocking [1997], Shatz and Stryker [1978] are also well studied developmental interventions. We simulated this process in our model by reducing the strength of the inputs (by a simple linear scaling) to one eye by 70% during training.

As shown in figure 5.10 panel (B), the resulting OD stripe pattern strongly favors the un-deprived eye which matches very well with experimental findings. Both the width of the stripes and the overall area of coverage devoted to the favoured eye are larger. In keeping with this, the histogram of ocularity also shows a strong skew, as one would expect. What is perhaps surprising, however, is that some of the strongest ocular preferences in the distribution favour the deprived eye (although there are few of them). This perhaps reflects an increased representational burden upon the small number of units that remain to prefer the deprived eye. This is a novel prediction from our model and it might be interesting to see if it holds out *in vivo*.

## 5.5.3   Learning Lateral Connections

Our framework allows us to learn the lateral weights as well as the thalamocortical weights **J**. We have so far restricted ourselves to considering fixed lateral weight

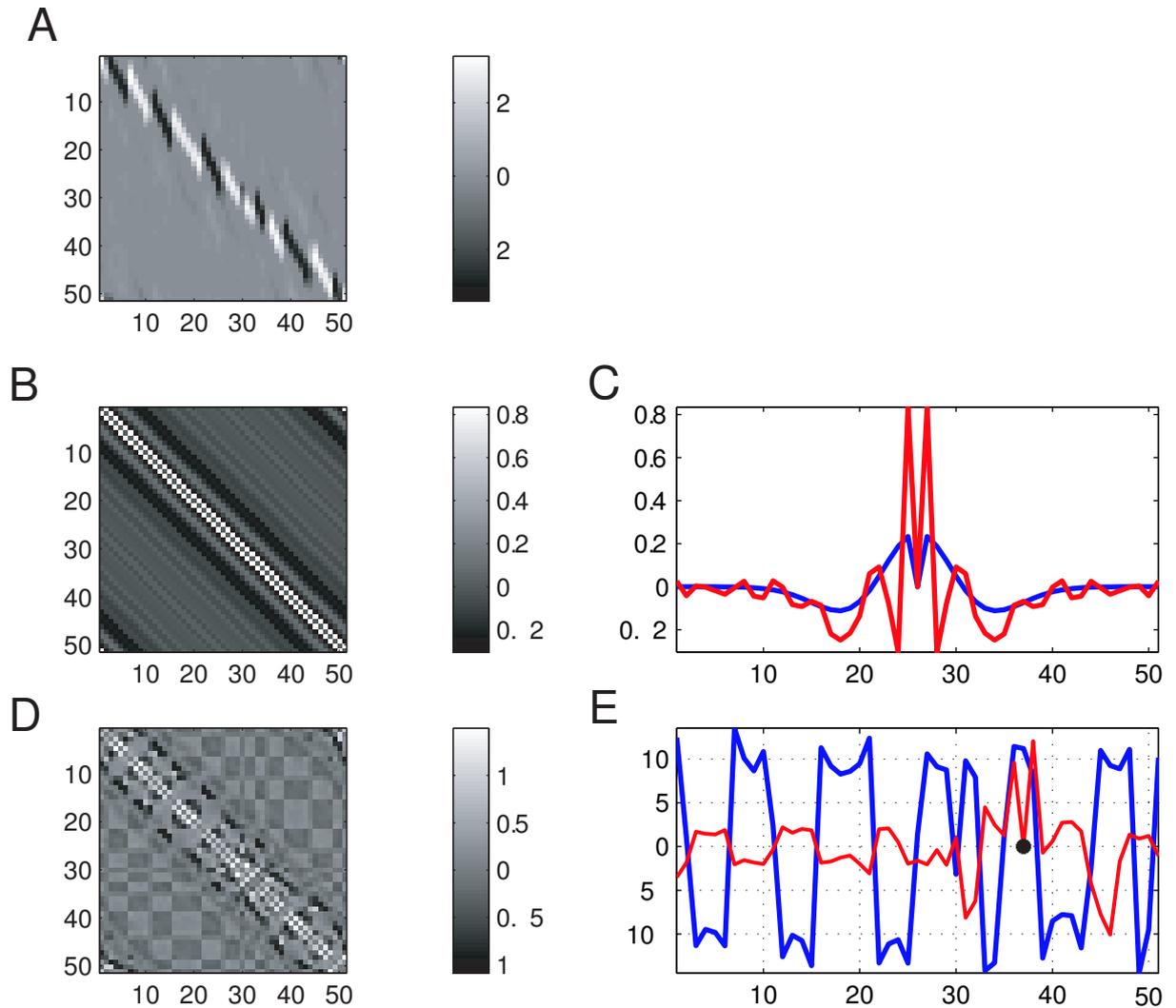

**Figure 5.11:** Figure showing various results on learning of lateral connections in 1d. (A) the difference mode for the thalamocortical weights after they have been learned; this pattern is now kept fixed whilst the lateral weights learn. (B) the lateral weight matrix learnt under the restriction of translation invariance. (C) the lateral weight kernel learnt under the restriction of translation invariance [red], compared with the previous, fixed DoG pattern [blue]. (D) the lateral weight matrix learnt in the absence of restrictions. (E) the blue curve shows the location and ocularity of each unit, the red curve shows the hidden connectivity patter for a weight located at the point indicated by the black dot. This is a typical profile and has the features that: (i) the immediate neighbours have a large excitatory connection; (ii) the other units within the same OD band have a weak connection; (iii) the units in neighbouring OD bands with the same sign have a large inhibitory connection; (iv) beyond this there is a general trend to inhibit regions of the same ocularity and to weakly excite areas of the different ocularity.

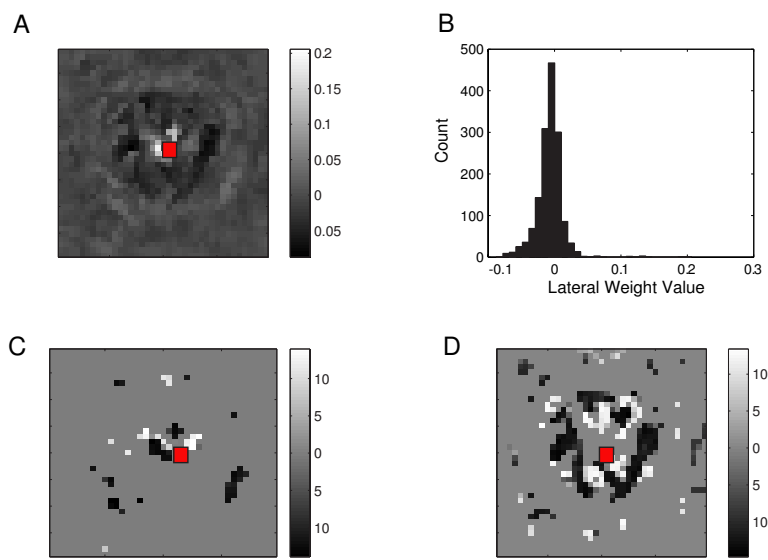

**Figure 5.12:** Figure showing various results on learning of lateral connections in 2d. (A) the hidden connectivity pattern for a single unit, located at the red square. The highlighted unit has a ocularity of −11.4. (B) histogram of the the distribution of hidden weights. (C) shows the location and ocularity of those hidden units to which the labelled unit has a connection of strength greater than 0.02. (D) shows the ocularity of those hidden units to which the labelled unit has a connection of strength less than −0.02.

patterns, **K**. However, we now discuss some of the consequences of relaxing this restriction.

It proved surprisingly difficult to learn **J** and **K** simultaneously. The weight patterns which develop appear very disordered and the model's performance, measured in terms of reconstruction ability, is significantly poorer than if we fix the lateral weights to a DoG pattern. A further problem for many sets of initial conditions is that the strength and configuration of the lateral weights quickly evolve to make the system highly frustrated. Consequently, the mean field settling phase can fail to reach a stable point even after several hundred iterations (compared to $O[10]$ iterations with 'reasonable' lateral weights); this interaction between parameter regimes and our algorithm severely impedes learning. Investigating these problems is difficult because of the sensitive dependence of the outcome of learning on the initial conditions, and also dynamical aspects of learning such as the relative speed with which the lateral connections and thalamocortical connections are changed.

As an alternative to simultaneously learning the two sets of weights, we consider the question of how stable a lateral connection pattern is if (after the usual period of learning the thalamocortical pattern) we then fix **J** and allow **K** to learn. Results which typify the answers to such a question, for 1-dimension, are shown in figure 5.11. We present two situations. In both cases we first learn the visible-hidden weights, as described in section 5.5.1 and using the parameters set out there. Then we fix the visible-hidden weights and allow the lateral connections to learn. In the first scenario, shown in figure 5.11 (B-C), we restrict the lateral weight patterns to be translation invariant; in the second, shown in 5.11 (D-E), they are unconstrained.

In one-dimension the general trends seem to be (i) for immediately neighbouring units to have a positive connection; (ii) for units slightly more distant but within the same OD band to have a zero, or somewhat negative connection (iii) to have negative connection with other units in bands of ocularity of the *same* sign, and a (weaker) positive connection to units in bands with ocularity of the *opposite* sign.

Figure 5.12 shows a qualitatively similar trend in 2-dimensions. Here we see, in panel (D), that the there is a preferential inhibition of other hidden units of the *same* ocularity at a particular horizontal displacements.

Based upon this result we might predict an anisotropy in the extent and sign of longer range lateral influences between ocular dominance columns. It suggests that we might see a bias towards having greater inhibitory interactions with those areas of cortex that, outside of an immediate local region, represent laterally shifted regions of visual space. (Clearly factors such as orientation preference, which are not represented in our simple model, should also play an important role.) Again, this is another novel prediction from our work and it might be interesting to use *in vivo* studies to investigate the possibility of experimental support.

### 5.5.4   Assessing Receptive Fields

In figure 5.9 (A) we showed the retinotopic refinement of the projections to a hidden unit. The form of this projective field will tend to dominate the response of that unit to a given pattern, however the unit will also receive inputs from its laterally connected neighbours.

A more thorough characterisation of the input output mapping, i.e. a deeper characterisation of the 'receptive field', would try to take into account some of these interactions. Although the most complete account of the dependency of a unit on an input pattern is to simply refer to the recursive equations that give the mean field fixed point, this description is not very intuitive. This highlights an interesting issue — even when we have a complete knowledge of a complex system, distilling this knowledge into a palatable form is far from trivial.

We have explored a simple, but informative, characterisation of units from trained models using the following approach. We compute the response of the unit to 100, 000 patterns from the training distribution and then look at averages of the 200 most and least stimulating patterns, a similar method was used in Karklin and Lewicki [2003]. This method is admittedly a little crude, and implicitly involves linearity assumptions which we know to be false. However, the results produced do seem to make sense and fit in with our intuitions — examples of the results of this procedure are given in figure 5.13.

Panel (A) 5.13 of shows the results from a fairly binocular unit taken from the edge of an ocular dominance band, whilst panel (B) shows results from a more monocular unit taken from the centre of a band. Several interesting features are apparent from these plots. Firstly, the plots suggest that the 'effective size' of the receptive field assessed in this way is larger than one would expect given the simple projective field — this is presumably due to the effect of lateral interactions and to the effects spatial correlations in the data. Another interesting point is that a 'centre-surround' like structure seems to have developed. The most effective stimuli have a central 'on' region and surround region which is 'off'; the least effective (i.e. most suppressive) stimuli show the converse.

This characterisation also illuminates a rudimentary sensitivity to disparity. Figure 5.13 (A) suggests a preference for a relative shift between the two eyes in a particular direction. This is most clear in the panel titled 'Difference of Most Excitatory Patterns (L-R)' in which we see bright highlighting on the right and dark shading on the left — this implies that stimuli in which the left eye pattern is a rightwards translation of the right eye pattern are most effective. Figure 5.13 (B) shows a more symmetric preference for the direction of translation. (The vast majority of the units in the model were more like example (B); the example in (A) was selected particularly to highlight the directional preference.)

Lastly, we note that our results do not address the development of other features mapped on V1 such as orientation selectivity and orientation domains, or spatial frequency tuning[5]. We believe this to be partially a consequence of the simplified inputs that were employed, however the incompleteness is somewhat unsatisfactory. It would be desirable to use digitised natural scenes as inputs but this presents difficulties because the Boltzmann machine is really a model for *binary* data. (Although we allow some of our inputs to be real numbers between 0 and 1, the vast majority are very close to being binary; this is not the case for digitised natural scenes.) As well as restricting the types of input data we are able to use, the binary character of the Boltzmann machine also limits the types of representations that may be formed.

## 5.6   Discussion

### Computational issues

Our aim was to build a good statistical model of the input distribution whilst also being informative with regards to the structures seen in visual cortex. We had clear successes in reproducing some of the structures and system behaviour seen *in vivo*, however it is somewhat more difficult to assess the quality of the representations that our model produces. Reconstruction fidelity gives us some indication of how well we are doing, but it is a rather poor proxy for what we are really interested in and could arise simply as a result of poor mixing in our Markov chain. Computing the likelihood of test data sets may be considered a more appealing measure but is not feasible for most Boltzmann machine models of any interest. One alternative approach is to judge the calibre of the generative model by performing long-run Gibbs sampling and qualitatively assessing the 'plausibility' of the samples produced; this is clearly a very subjective method, but our choice is rather Hobsonian.[6]

---

[5]Technically, the apparent centre-surround organisation would allow for very coarse spatial frequency selectivity.

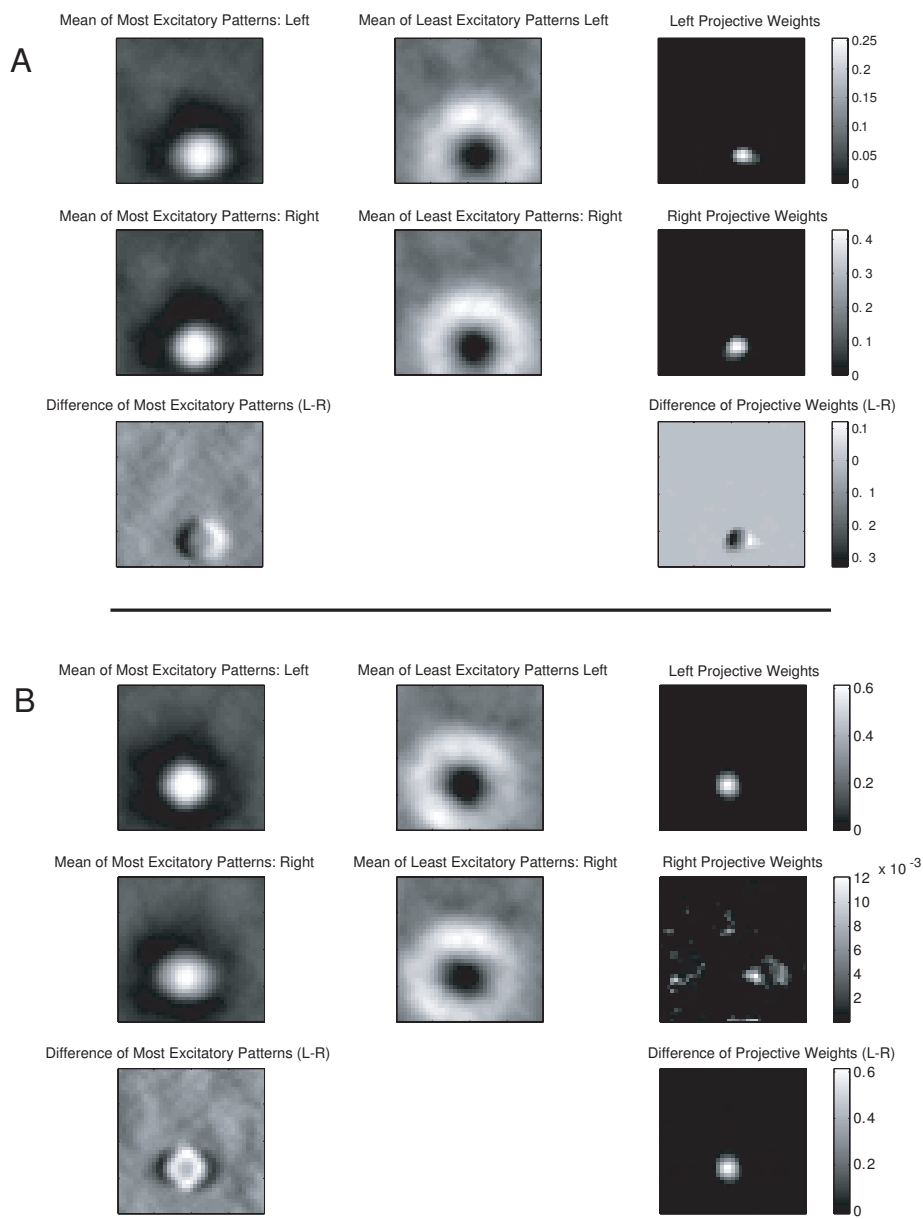

**Figure 5.13:** Simple characterisations of the most and least effective stimuli for two cortical units. The model from which the units are taken has the parameters outlined in section 5.5.2. 100,000 inputs were generated and presented to the network, The leftmost columns show the left-eye and right-eye means for the 200 most excitatory patterns in, along with the mean difference between the left and right eye patterns. The central column shows the mean of the 200 least effective patterns. The rightmost column shows the 'projective fields' from the left and right eyes, along with the difference. (A) shows plots obtained from a fairly binocular unit. (B) shows plots from more a monocular unit. The binocular unit was selected because it nicely illustrates disparity tuning with a shift in a particular direction. However, in general the 'symmetric' disparity tuning, as highlighted by the monocular unit, was more common.

Figures 5.6 (D) and 5.9 (C-E) show that the models are able to form faithful reconstructions. Figure 5.6 (E) shows that we are able to learn distributions for a single eye effectively. The reconstruction of the input pattern in one eye, if we clamp the other eye, are also quite reasonable as illustrated by figure 5.14. The inferred images for the unclamped eye show the same structure as present in the clamped eye, and in 1D at least, there is a clear notion that the left- and right-eye images are laterally shifted.

However, there is clearly some difficulty in learning to represent the full complexity of stereo pairs. Figure 5.15 (A) shows a collection of samples from an approximation to the equilibrium distribution of a stereo model in 1D, figure 5.15 (B) shows samples from a 2D stereo model. (These samples were obtained by performing Gibbs sampling for 100,000 steps in conjunction with simulated annealing.) In our 1 dimensional results, the overall structure of the patterns for each eye are approximate matches to those seen in the data distribution — although the samples have more units active than are typically seen in an input. The stereo nature of the data distribution is also captured, albeit to a lesser extent, with the right eye pattern often being a rough correspondence to a translated version of the left eye pattern. However, in two dimensions the results we show are less convincing; something seems to have gone quite badly wrong and there is a generally excessive level of excitation leading to a single global patch. On the bright side we *are* able to say that the model has captured the constraint that the patterns in the two eyes should be similar. Furthermore, these apparent 'failings' are not as damning as they might initially seem since learning good generative models for stereo is generally recognised to be an *extremely* difficult unsupervised learning problem unless explicit structure is built in.

### Relationship to other approaches

Whilst there are many approaches that are able to reproduce some of the aspects of cortical development that we have shown in this chapter, we feel that our model makes an interesting and significant contribution to the field.

Little work has been done so far in exploring the capability of our energy-based approach as a top-down tool to help understand the processes of receptive field and topographic map formation; this chapter has demonstrated that this approach is viable and can deliver results which can be connected to the neurobiology. Also, we note that there exist few top-down models that can take high-dimensional inputs and learn to produce a statistically motivated, high-dimensional, non-linear, distributed population representation such as that shown in figure 5.9 (D). Finally although some models, such as the LISSOM [Miikkulainen et al., 1997], deftly learn lateral connections most lack this ability, unlike our approach.

---

[6]Although it does not allow us to assess the quality of the final model, comparing the relative energy of data and previously generated negative phase samples provides us with some diagnostic of online learning progress [Hinton, Personal communication].

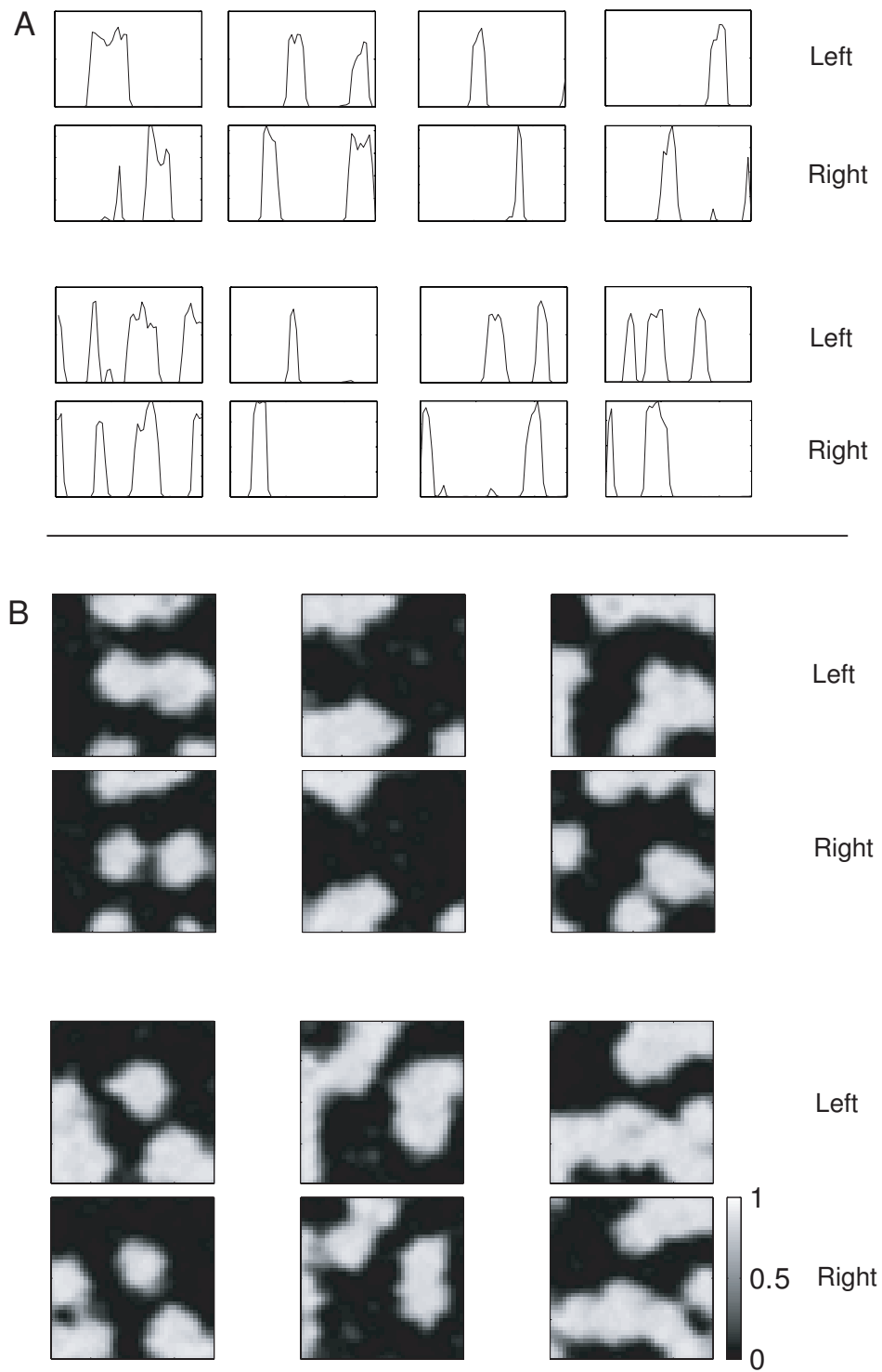

**Figure 5.14:** Examples of stereo pattern completion based upon partial input. (A) Results from 1-dimensional model. (B) Results from 2-dimensional model. The right eye pattern was clamped, and the annealed Gibbs sampling was performed for 1000 steps to 'complete' the pattern for the left eye. We show the reconstructions for both eye after sampling.

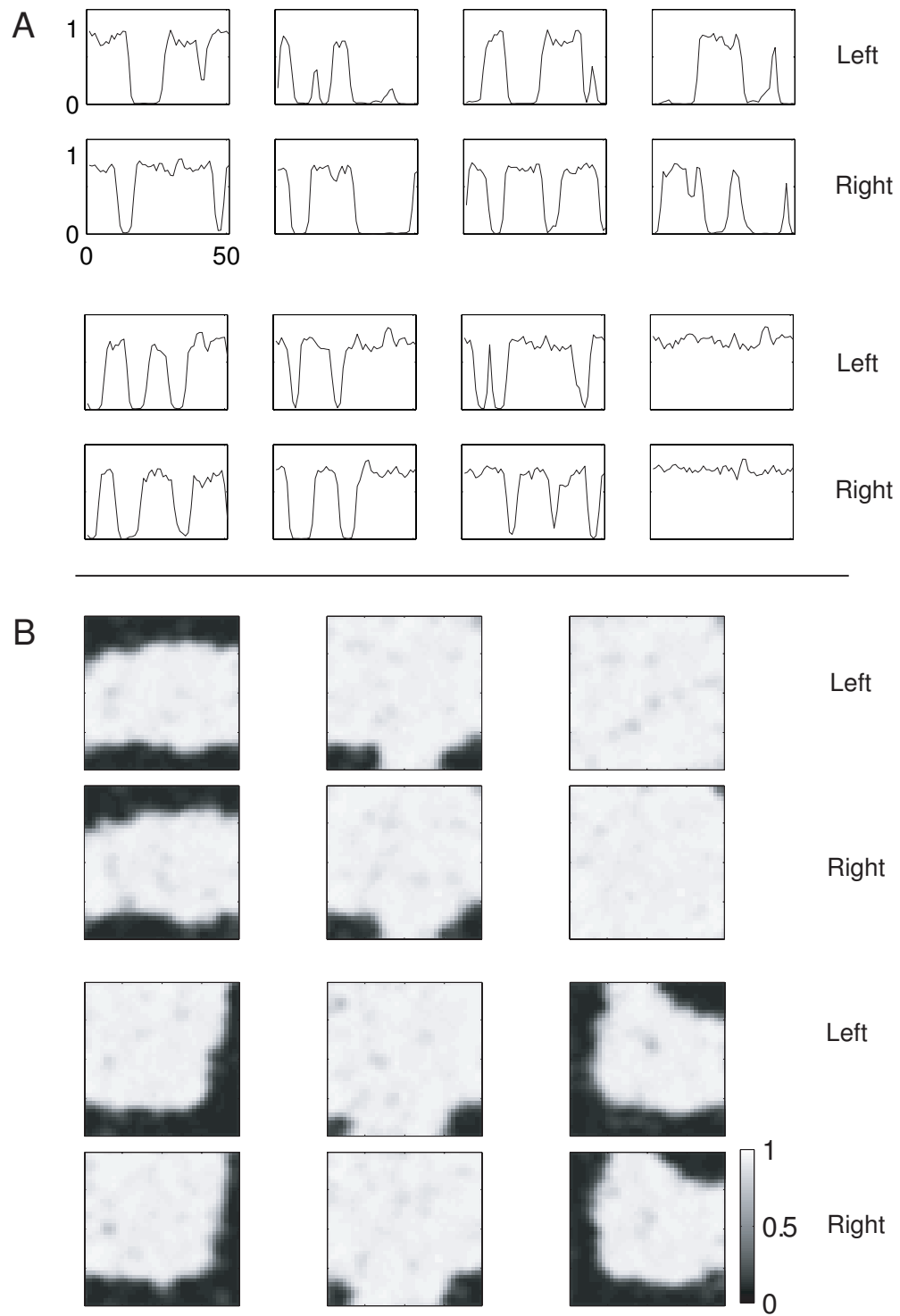

**Figure 5.15:** Samples from (approximations to) the equilibrium distribution of example stereo models. The samples were obtained by a long run of Gibbs sampling (100,000 steps) in conjunction with simulated annealing. (A) shows example patterns from a 1-dimensional model. (B) shows example patterns from a 2-dimensional model.

## Summary & Extensions for future work

We have presented a model for the activity-dependent refinement of retinotopy and for the formation of ocular dominance within the statistical framework of a Boltzmann machine. Development in the model uses two phases of weight adaptation with local Hebbian and anti-Hebbian learning rules. Under certain conditions the model develops ocular dominance stripes for computationally well-motivated reasons — the formation of a probabilistic density model of the input ensemble. The internal representations employed by our approach are non-linear distributed representations and can be used to reconstruct the inputs with high fidelity. Additionally, in the face of experimentally-inspired input manipulations such as strabismus and monocular deprivation, our results show strong qualitative resemblance to experimental observations.

However despite these successes, in its current form our model is obviously highly simplified and suffers several lacunae. There are several intriguing extensions that could be made to the framework which we have presented here.

One such extension is to add multiple layers to the model. In some senses, this is a simple adaptation of what we have already done. Adding a second hierarchical layer is effectively equivalent to adding units to our present hidden layer but forbidding them to contact directly with the input units. We have briefly explored such extensions, but we have not been able to produce anything that obviously captures interesting aspects of biology — although again we are faced with the problem of defining a good success metric, and by the problem of characterising the behaviour of units deep within a multi-layer system

Another direction which one could explore to give the model greater expressive power and flexibility would be to move away from 'neurons' that are essentially binary to ones which have multivalent states and/or different state degeneracies. One might expect that with such modifications, it would be possible to formulate a model that could better handle continuous, real-valued inputs as would be desirable when training on digitised natural scenes.

Finally, the results presented in this chapter mostly come from models that are complete (or even slightly undercomplete) with respect to the dimensionality of the representation in comparison to the dimensionality of the input. We have explored overcomplete models, however within the ranges that we considered they were not noticeably different from complete ones. It is possible that vastly greater degrees of overcompleteness might have lead to interesting new results, but we feel this is unlikely.

Within the context of this thesis these options are left open. The alternative directions pursued in the next chapter appear more fruitful from the point of view of producing a good statistical model of visual inputs.

# Chapter 6

# Sparse Topographic Representations of Natural Scenes

## 6.1  Introduction

The Boltzmann machine models discussed in the previous chapter inherently relied upon binary inputs and binary representations. They were able to deal with continuous values only by bounding the allowed range, and using a 'rate based' interpretation or taking a mean field approach. We now move on to consider a different set of energy-based models which retain many of the Boltzmann machine's virtues but allow for richer, continuous valued inputs and representations. This new set of models, which we refer to as Products of T's (PoT's) for reasons which will become clear later, also have the significant characteristic of being principally formulated using *deterministic* rather than *stochastic* latent variables — this means that inference can be direct and rapid, and that variational approximations are unnecessary.

In sections 6.5 and 6.6 we show that when trained on digitised patches from natural scene photographs our model is able to develop features which can be closely related to simple-cell and complex-cell receptive fields in V1; furthermore we demonstrate that our model can achieve representations of this form in systems that have an impressive degree of overcompleteness. We also show in section 6.7 that, when appropriately constrained, the PoT is able to reproduce some of the patterns associated with the topographic maps of retinotopy, orientation, and spatial frequency, and present some preliminary results in section 6.8.1 demonstrating patterns of ocular dominance and disparity.

Our new approach has interesting relationships to independent components analysis and some extensions thereof [Hyvarinen et al., 2001, Hyvarinen and Hoyer, 2001], and also to Gaussian scale mixtures [Andrews and Mallows, 1974, Wainwright et al., 2000a, Wainwright and Simoncelli, 2000, Wainwright et al., 2000b] and in some ways can be viewed as a novel generalisation of these paradigms. In sections 6.9 and 6.10 we analyse some of these relationships, and also discuss possible relationships to divisive normalisation. Finally, in section 6.11 we discuss some representational properties of our model in terms of population codes, and we contrast overcomplete energy-based model representations with those from overcomplete causal models.

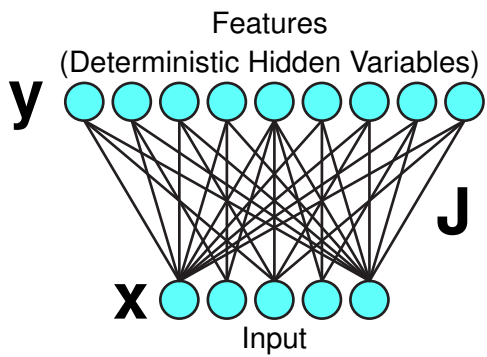

**Figure 6.1:** Our basic model. The input vector **x** is linearly filtered to give our neural representations in the 'features' **y** — deterministic hidden variables.

**Original Contributions**

The main original contributions in this chapter are: (i) the proposal of the product of Student-t (PoT) density model and its hierarchical and topographic extensions; (ii) the derivation of an efficient Gibbs sampling procedure (using auxiliary variables) for such models; (iii) the use of the PoT to simultaneously build a density model for natural image patches whilst reproducing many aspects of the receptive field and topographic map structures experimentally observed in V1, and the achievement of these results in a significantly overcomplete setting; (iv) the derivation of a divisive normalisation procedure based upon our statistical model; (v) the analysis of work by Hyvärinen et al. [2001] showing that the model these authors end up optimising is a restricted example of the framework that we propose; and (vi) the comparison properties of population codes in overcomplete examples of our model and those in overcomplete causal models.

## 6.2   Basic Product of Student-t Model

We begin by considering a simple linear model. Let each input be a continuous valued vector, **x**, denoting the configuration of a set $\{x_i\}$ of input variables. Our representation will be in terms of the values of a set of deterministic hidden variables, **y**, that we will often refer to as 'features', given by the outputs of a set of linear filters applied to the input

$$\mathbf{y} = \mathbf{J}\mathbf{x} \qquad\qquad (6.1)$$

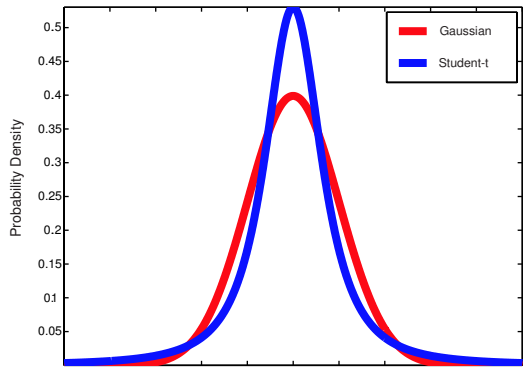

**Figure 6.2:** Comparison of the Gaussian distribution and a Student's t-distribution of order $\frac{3}{2}$. The two distributions shown have the same mean and variance, however note the sharper peak and heavier tails of the Student-t in relation to the Gaussian.

We will define an overall energy that can be written as a sum of functions, $f_i(\cdot)$, of single 'features' $\{y_i\}$

$$
\begin{aligned}
E(\mathbf{x}) &= \sum_i f_i\left(y_i(\mathbf{x})\right) \\
&= \sum_i f_i\left(\mathbf{J}_i^T \mathbf{x}\right)
\end{aligned}
\tag{6.2}
$$

where $\mathbf{J}_i$ is the $i^{\text{th}}$ row of weight matrix $\mathbf{J}$, written as a column vector. Our set-up is shown schematically in figure 6.1. Since the latent variables are deterministic, we will often choose to consider the total energy as being a function of just $\mathbf{x}$.

In particular we will choose to focus upon models having the following functional form

$$
\begin{aligned}
f_i(\mathbf{x}) &= \alpha_i \log\left(1 + \frac{1}{2}y_i(\mathbf{x})^2\right) \\
&= \alpha_i \log\left(1 + \frac{1}{2}(\mathbf{J}_i^T \mathbf{x})^2\right)
\end{aligned}
\tag{6.3}
$$

When this function is exponentiated (and normalised to sum to one) we obtain a one-dimensional Student's-t distribution of order $(2\alpha_i - 1)$, as illustrated in figure 6.2. Consequently, as discussed in chapter 4, if $\dim(\mathbf{y}) = \dim(\mathbf{x})$ then this would be isomorphic to a square, noiseless ICA model with Student-t priors. More generally, the full density model, $p^\infty(\mathbf{x}) = \frac{1}{Z}e^{-E(\mathbf{x})}$ also corresponds to a product of Student's t-distributions — hence our PoT moniker.

A key aspect of our choice of energy function is that it naturally favours sparse, leptokurtotic distributions for the features; consequently the model will assign high probability to those datasets which map, through $\mathbf{J}$, to such distributions. Fitting such a model to data therefore boils down to searching for a set of linear filters that, when applied to the data, result in a peaky, heavy tailed output/latent distribution on the $\{y_i\}$. Several authors have shown that this is particulary suitable description when characterising linear decompositions of natural image data [Field, 1987,

Ruderman, 1994].

Before moving on, we present a further interpretation for this model as inspired by Hinton and Teh [2001]. We can view the PoT as describing an image in terms of soft linear constraints. The $\{y_i\}$ can be thought of as the amount by which a set of linear equality constraints ($\mathbf{J}_i^T\mathbf{x} = 0$) are violated. These constraints are soft in the sense that, if the model fits well, they are 'frequently approximately satisfied' — most of the time the constraints are very close to zero, but occasional large deviations are also observed. Under this interpretation, the representations formed can be thought of as patterns of constraint violations.

When we apply our PoT to natural scene inputs we will interpret the deterministic hidden units as being the activity of neurons forming the internal representation, and the filter matrix as being the set of synaptic weights coupling input to this representation.

## 6.3 Contrastive Divergence Learning In PoT Models

Once again we will appeal to the contrastive divergence algorithm when we come to optimise the parameters of the PoT model. In addition to the hybrid Monte Carlo method described in chapter 4, we have also developed an efficient Gibbs sampling scheme that introduces a new set of auxiliary variables.

### 6.3.1 Gibbs Sampling with Auxiliary Variables

The Hybrid Monte Carlo (HMC) contrastive divergence scheme described in Chapter 4 is very general and works quite well in practice. However, we now present an alternative scheme which exploits the particular mathematical form of the Student's t-distribution and, if applicable, often is preferable to the HMC based algorithm because it tends to be much faster.

One of the key features of the HMC procedure is the use of auxiliary momentum variables. Although it is not immediately obvious from equations 6.2 and 6.3, there is a different way of introducing a different set of auxiliary variables that allow us to set up an efficient Gibbs sampling scheme.

Consider an energy-based model over our original variables, $\mathbf{x}$, and a new set of auxiliary stochastic variables, $\mathbf{u}$. If we take the energy function,

$$E(\mathbf{x}, \mathbf{u}) = \sum_i \left[ u_i \left( 1 + \frac{1}{2}(\mathbf{J}_i^T\mathbf{x})^2 \right) + (1 - \alpha_i) \log u_i \right] \tag{6.4}$$

then the corresponding joint distribution, $p^\infty(\mathbf{x}, \mathbf{u}) = \frac{1}{Z}e^{-E(\mathbf{x}, \mathbf{u})}$, has a marginal distribution over $\mathbf{x}$ that is precisely the one specified by our original PoT model.

The conditional relationships consistent with this joint distribution are,

$$p(\mathbf{u}|\mathbf{x}) = \prod_i \mathcal{G}_{u_i}\left[\alpha_i \; ; \; 1 + \frac{1}{2}(\mathbf{J}_i^T \mathbf{x})^2\right] \tag{6.5}$$

$$p(\mathbf{x}|\mathbf{u}) = \mathcal{N}_{\mathbf{x}}\left[0 \; ; \; (\mathbf{J}^T \mathbf{U} \mathbf{J})^{-1}\right] \qquad \mathbf{U} = \mathbf{Diag}[\mathbf{u}] \tag{6.6}$$

where $\mathcal{G}(a; b)$ denotes a Gamma distribution with shape parameter $a$ and scale parameter $b$, while $\mathcal{N}(\mu; \Sigma)$ denotes a normal distribution with mean $\mu$ and covariance $\Sigma$.[1] The $u_i$ are conditionally independent given $\mathbf{x}$ and are therefore very easy to sample from; the distribution of the $x_i$ given $\mathbf{u}$ is a multivariate Gaussian. A rapidly mixing Markov chain is possible if we can efficiently alternate between sampling the auxiliary variables, $\mathbf{u}$, conditioned on the current visible configuration, $\mathbf{x}$ and vice-versa. The relationship between the variables $\mathbf{x}, \mathbf{y}$ and $\mathbf{u}$ is depicted graphically in figure 6.3.

In the case that we have a complete model (i.e. as many deterministic latent variables as visible variables) then sampling from the conditional Gaussian for $\mathbf{x}$ is easily accomplished. A sample from the conditional $p(\mathbf{x}|\mathbf{u})$ is given by $\mathbf{J}^{-1}\mathbf{U}^{-\frac{1}{2}}\mathbf{n}$ where $\mathbf{n} \sim \mathcal{N}(0, \mathbf{I})$. The diagonal matrix $\mathbf{U}$ is trivially inverted for each data point, and we need only compute the $\mathbf{J}^{-1}$ once per batch. In the case of overcomplete models the situation is made a little more difficult. Here, $\mathbf{J}$ is no longer a square matrix and hence we ought to perform a Cholesky decomposition of the product $\mathbf{J}^T \mathbf{U} \mathbf{J}$ for *every* datum, since the posterior for $\mathbf{u}$ could be different for each datum. We have, however, obtained good results by proceeding as in the complete case and replacing the inverse of the filter matrix with its pseudo-inverse give by $\mathbf{J}^\dagger = (\mathbf{J}^T \mathbf{J})^{-1} \mathbf{J}^T$.

The auxiliary variables $\mathbf{u}$ can be viewed as being analogous to precision parameters (i.e. inverse variances) constraining the latent variables $\{y_i\}$. In this respect we see that the complete version of the PoT model is related to a particular type of 'Gaussian Scale Mixture' (GSM) [Andrews and Mallows, 1974] as we discuss in section 6.9. However, the GSM is best thought of as a causal generative model whereas the PoT is most naturally expressed as an energy-based model.

The 1-step samples for the negative phase are obtained by first sampling a set of $\{u_i\}$ conditioned on the data using equation 6.5, then, conditioned on this sampled vector $\mathbf{u}$, we sample the visible units using equation 6.6. The expressions for the derivative of the energy functions with respect to the parameters, given a visible

---

[1] The Gamma distribution is defined as $\mathcal{G}_u(a; b) = \frac{u^{a-1} e^{-\frac{u}{b}}}{b^a \Gamma(a)}$ where $\Gamma(a)$ denotes the Gamma function.

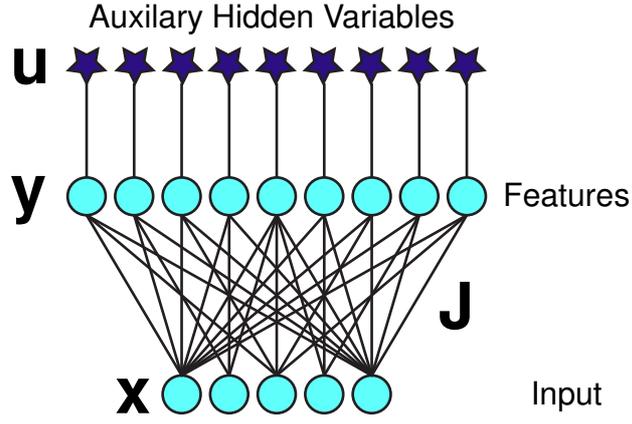

**Figure 6.3:** Relationship between variables $\mathbf{x}, \mathbf{y}$ and $\mathbf{u}$. The $\mathbf{x}$ and $\mathbf{y}$ variables in this diagram are deterministically related to one another by the linear transform $\mathbf{y} = \mathbf{Jx}$. The $u_i$ are Gamma distributed given $\mathbf{x}$ – the exact shape of the Gamma distribution for $u_i$ depends on $\mathbf{x}$ through the corresponding $y_i$. Given $\mathbf{u}$, we have a Gaussian distribution for $\mathbf{x}$. (The distribution for $\mathbf{y}$ given $\mathbf{u}$ is also a Gaussian but is restricted to the subspace that can be reached by projection from $\mathbf{x}$-space to $\mathbf{y}$-space, as discussed in Chapter 4.)

configuration $\mathbf{x}$, are:

$$\frac{\partial E}{\partial J_{ij}} = \frac{\alpha_i (\mathbf{J}_i^T \mathbf{x})}{1 + \frac{1}{2}(\mathbf{J}_i^T \mathbf{x})^2} x_j$$

$$= \frac{\alpha_i y_i}{1 + \frac{1}{2} y_i^2} x_j \tag{6.7}$$

$$\frac{\partial E}{\partial \alpha_{ij}} = \log\left(1 + \frac{1}{2}(\mathbf{J}_i^T \mathbf{x})^2\right)$$

$$= \log\left(1 + \frac{1}{2} y_i^2\right) \tag{6.8}$$

The expression for one-step contrastive divergence parameter updates are given by

$$\triangle J_{ij} = \frac{\eta_J}{N}\left(-\left\langle \frac{\alpha_i y_i}{1 + \frac{1}{2} y_i^2} x_j \right\rangle_0 + \left\langle \frac{\alpha_i y_i}{1 + \frac{1}{2} y_i^2} x_j \right\rangle_1\right) \tag{6.9}$$

$$\triangle \alpha_i = \frac{\eta_J}{N}\left(-\left\langle \log\left(1 + \frac{1}{2} y_i^2\right) \right\rangle_0 + \left\langle \log\left(1 + \frac{1}{2} y_i^2\right) \right\rangle_1\right) \tag{6.10}$$

where $N$ is the number of samples in each mini-batch of data and the $\eta$'s gives the learning rate for the different parameters. (Unless otherwise stated we keep the $\alpha$'s fixed, i.e.: $\eta_\alpha = 0$, with $\alpha = 1.5$ as illustrated in figure 6.2.)

## 6.4   Datasets & Data Pre-Processing

We performed experiments using sets of digitised natural images available from the World Wide Web from Aapo Hyvarinen[2] and Hans van Hateren[3]. An example image from the data set of van Hateren is shown in figure 6.4 (A). The results obtained from the two different datasets were not significantly different, and for the sake of consistency all results reported here are from the van Hateren dataset.

Small, square patches, such as those in figure 6.4 (B), were extracted from randomly chosen locations in the images. As is common for unsupervised learning, these patches were filtered according to computationally well-justified versions of the sort of whitening transformations performed by the retina and LGN [Atick and Redlich, 1992]. First we applied a log transformation to the 'raw' pixel intensities. This procedure somewhat captures the contrast transfer function of the retina. It is not very critical, but for consistency with past work we incorporated it for the results presented here.

The extracted patches were subsequently normalised such that mean pixel intensity for a given pixel across the data-set was zero, and also so that the mean intensity within each patch was zero — effectively removing the DC component from each input. These pre-processing steps also reflect the adaptation of retinal responses to local contrast and overall light levels.

The patches were then whitened, usually in conjunction with dimensionality reduction. Figure 6.5 (A) shows the eigenvectors of the covariance matrix (i.e. the principal components) for a set of $150,000$ patches of size $10 \times 10$. If we then subsequently normalise the retained components by dividing by their variance (i.e. the corresponding eigenvalue) then the second order correlations in the data will have been removed. The centre-surround structure of retinal ganglion cell receptive fields seems to perform an analogous role in real visual systems. The similarity of our whitening operation to biological processing can be made more obvious if we take a filter-set used for whitening and transform it back to the data space — the resulting filters do indeed have a centre-surround structure as shown in figure 6.5 (B). We also give a qualitative impression of the effects of whitening and dimensionality reduction by transforming the processed data back into the image space, shown in figure 6.4 (C).

---

[2]`http://www.cis.hut.fi/projects/ica/data/images/`
[3]`http://hlab.phys.rug.nl/imlib/index.html`

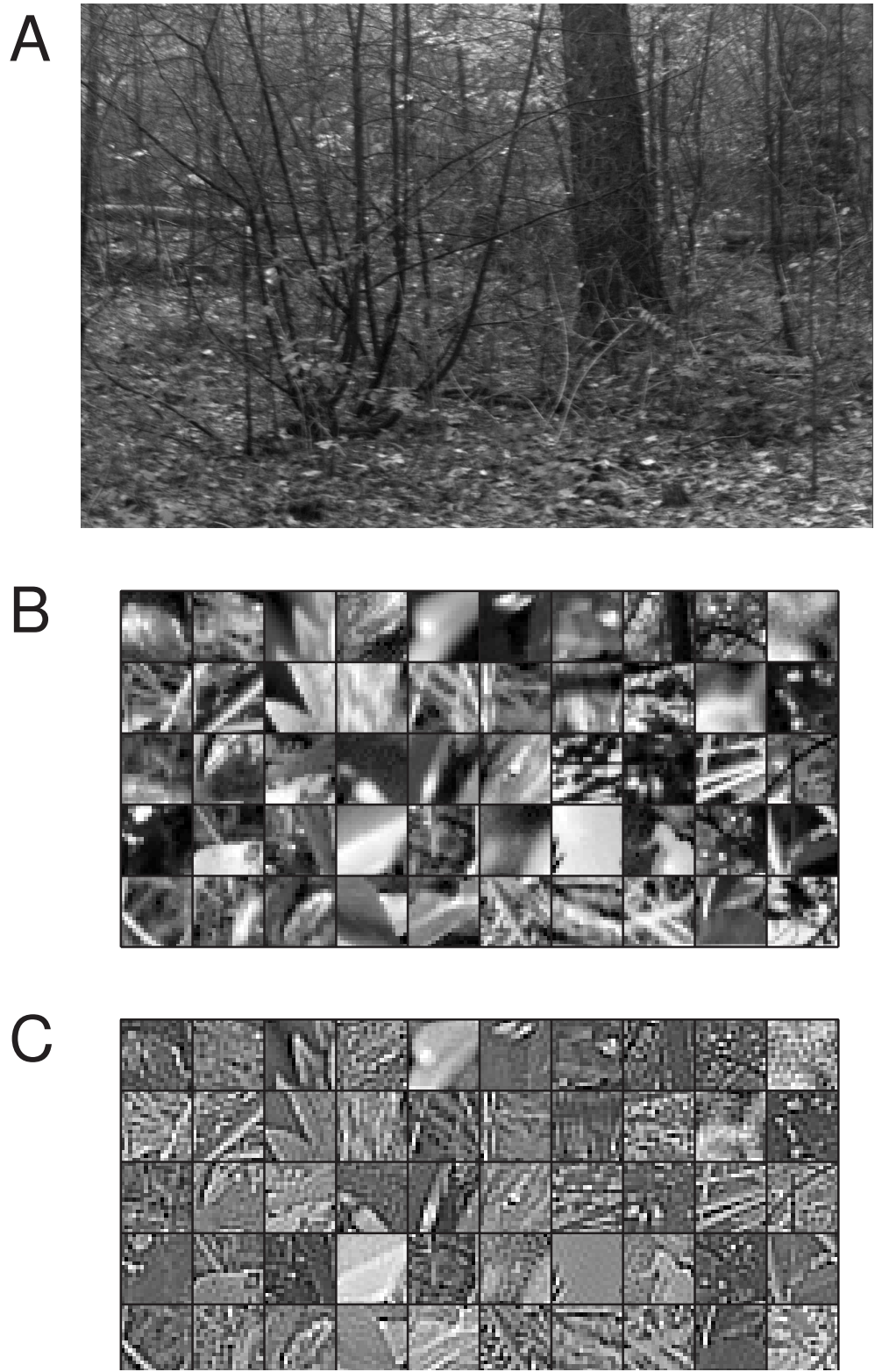

**Figure 6.4:** Example training images and pre-processing. (A) A full image from the dataset of van Hateren[van Hateren and van der Schaaf, 1998]. (B) An example set of 18x18 patches. These would be subjected to sphering and dimensionality reduction before being vectorised and used as inputs to the algorithm. (C) The same set of patches shown in panel B after they have been whitened and reduced to 256 dimensions — the DC component and the subspace corresponding to the 104 smallest eigenvectors had been removed. (The pseudo-inverse of the whitening/dimensionality reducing filter has been used to render them in image space.)

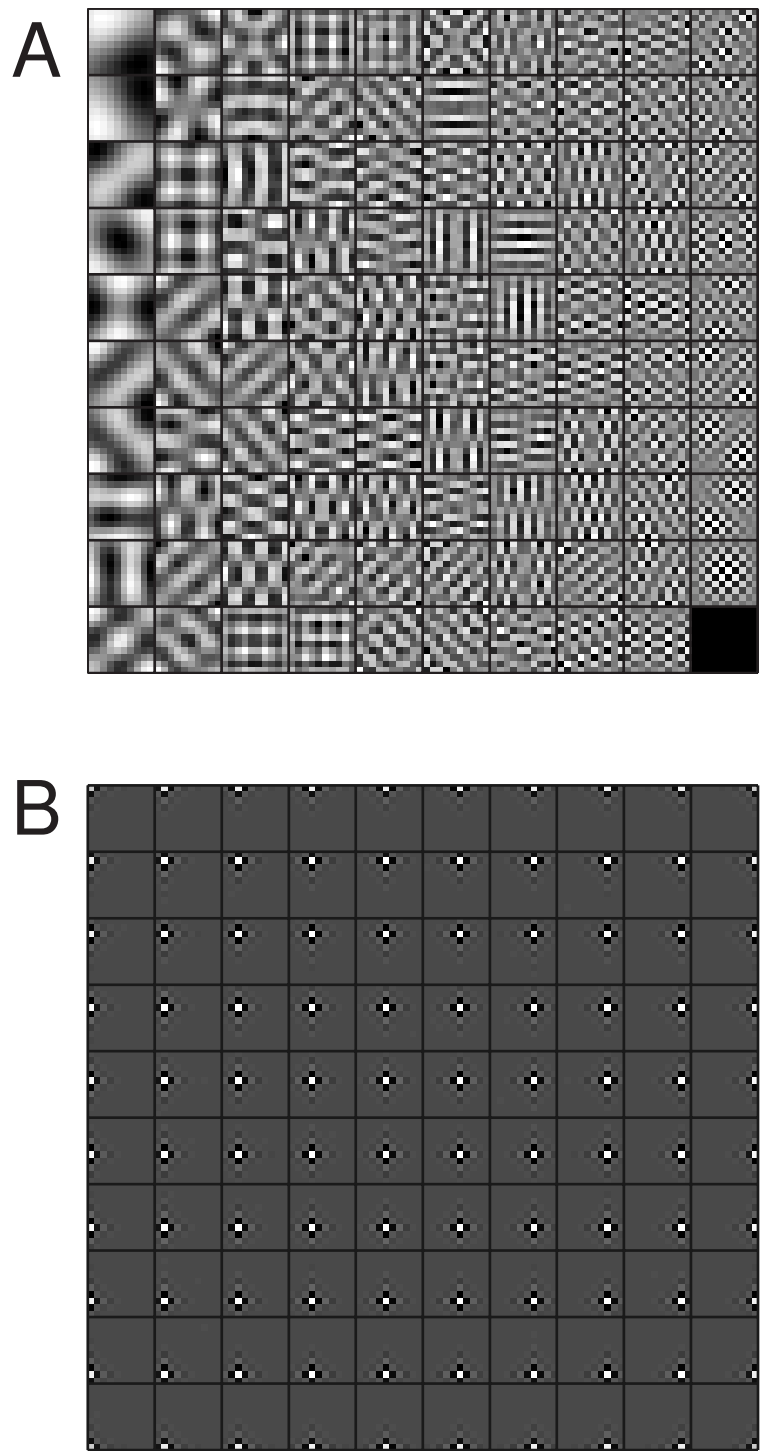

**Figure 6.5:** Example training images and pre-processing. (A) Principal components (i.e. covariance eigenvectors) of an ensemble of $150,000$ $10 \times 10$ image patches. The components are arranged columnwise (top to bottom, left to right) in descending order of eigenvalue. (We chose small patches for this illustration to aid visibility — in the results presented later the patch size is larger, typically $18 \times 18$.) The whitening filter projects an image onto (a subset of) the eigenvectors of the covariance matrix, and then scales each dimension by the inverse of the projection by the corresponding eigenvalue. (B) This panel shows a whitening filter set rotated back into image coordinates. Note the centre-surround structure of these filters.

## 6.5   Single Layer PoT

**Complete Models**

We first present the results of our basic approach in a complete setting, and display a comparison of the filters learnt using our method with a set obtained from an equivalent ICA model learnt using direct gradient ascent in likelihood. The data set data set comprised $150,000$ patches of size $25 \times 25$ that had been reduced to vectors of size 361 by neglecting the 264 components with smallest eigenvalue. We trained both models (learning just $\mathbf{J}$, and keeping $\alpha$ fixed at 1.5[4]) for 200 passes through the entire dataset of $150,000$ patches. The PoT was trained using one-step contrastive divergence as outlined in section 6.3.1 and the ICA model was trained using the exact gradient of the log-likelihood (as in, for instance Bell and Sejnowski [1995]). As expected, at the end of learning the two procedures delivered very similar results, exemplars of which are given in figure 6.6. Furthermore, both the sets of filters bear a strong resemblance to the types of simple cell receptive fields found in V1.

**Overcomplete Models**

We next consider our model in an overcomplete setting. As discussed in Chapter 4, this is no longer equivalent to an ICA model. In EBM's such as the PoT, overcomplete representations are simple generalisations of the complete case and, *unlike* causal generative approaches, the features are conditionally independent and inferring a representation is trivial.

Consequently, it is relatively straightforward to consider highly-overcomplete representations that might seem unmanageable in an ICA model. We trained $1.7\times$ and $2.4\times$ overcomplete models on data from $18 \times 18$ patches than had been whitened and dimensionality reduced to vectors of length 256 (by removing the subspace of the smallest 68 eigenvectors). Additionally, we trained a $100\times$ overcomplete model on data from $10 \times 10$ patches that had been whitened and dimensionality-reduced to vectors of length 81 to demonstrate the power of our method. Although we cannot say that this enormous model necessarily leads to better data modelling (and we do not analyse it further) we note that such a degree of overcompleteness would be difficult to deal with using a causal generative model.

To facilitate learning in the overcomplete setting we have found it beneficial to make two modifications to the basic set-up. Firstly, we set $\alpha_i = \alpha \, \forall i$, and make $\alpha$ a free parameter to be learnt from the data.[5] The learnt value of $\alpha$ is typically less that 1.5 and gets smaller as we increase the degree of overcompleteness. One intuitive way

---

[4]This is the minimum value of $\alpha$ that allows us to have a well behaved density model (in the complete case). As alpha gets smaller than this, the tails of the distribution get heavier and heavier and the variance and eventually mean are no longer well defined.

[5]Note that in an overcomplete setting, depending on the direction of the filters, $\alpha$ may be less than 1.5 and still yield a normalisable distribution overall.

A

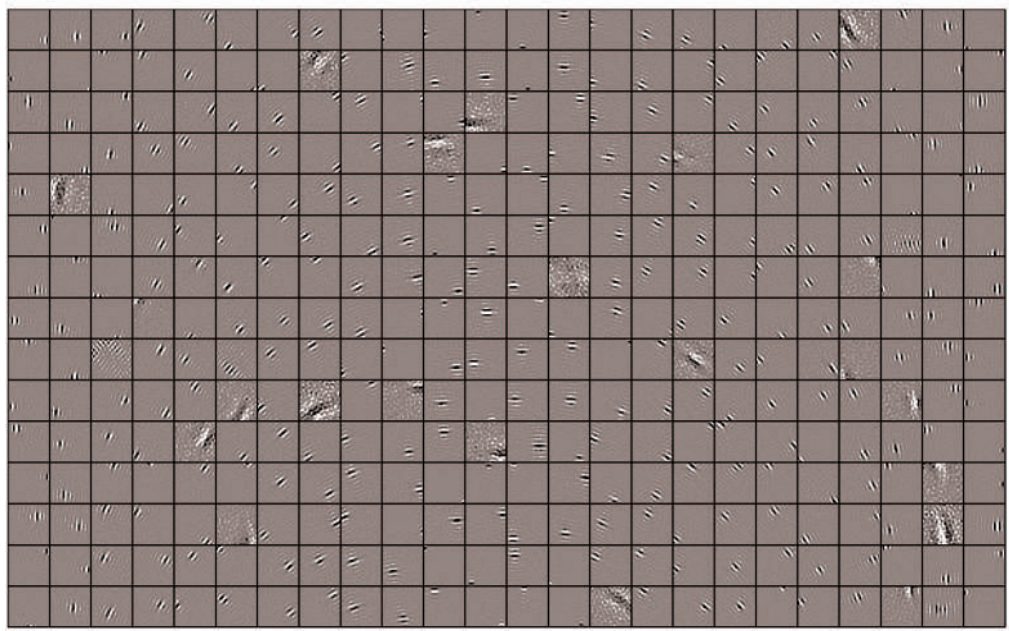

B

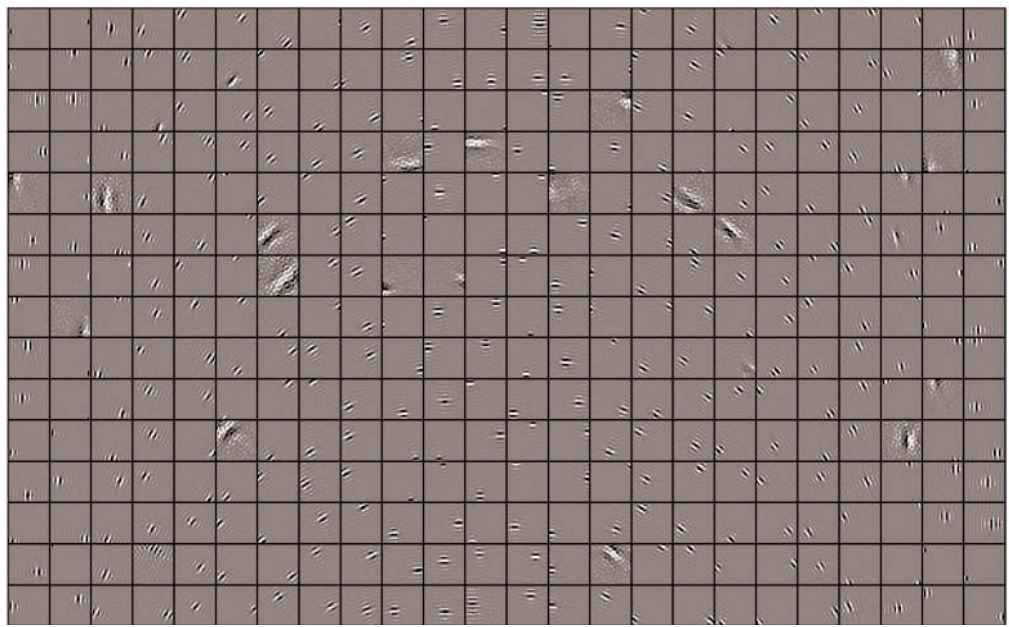

**Figure 6.6:** Learnt filters shown in the raw data space. Each small square represents a filter vector, plotted as an image. The gray scale of each filter square has been (symmetrically) scaled to saturate at the maximum absolute weight value. Additionally, the filters have been ordered, columnwise, by orientation. (A) PoT filters learnt using CD, patch size $25 \times 25$ whitened and reduced to 361 dimensions. (B) Equivalent ICA filters (rows of inverse of generative weight matrix) learned by natural gradient method, patch size $25 \times 25$ whitened and reduced to 361 dimensions.

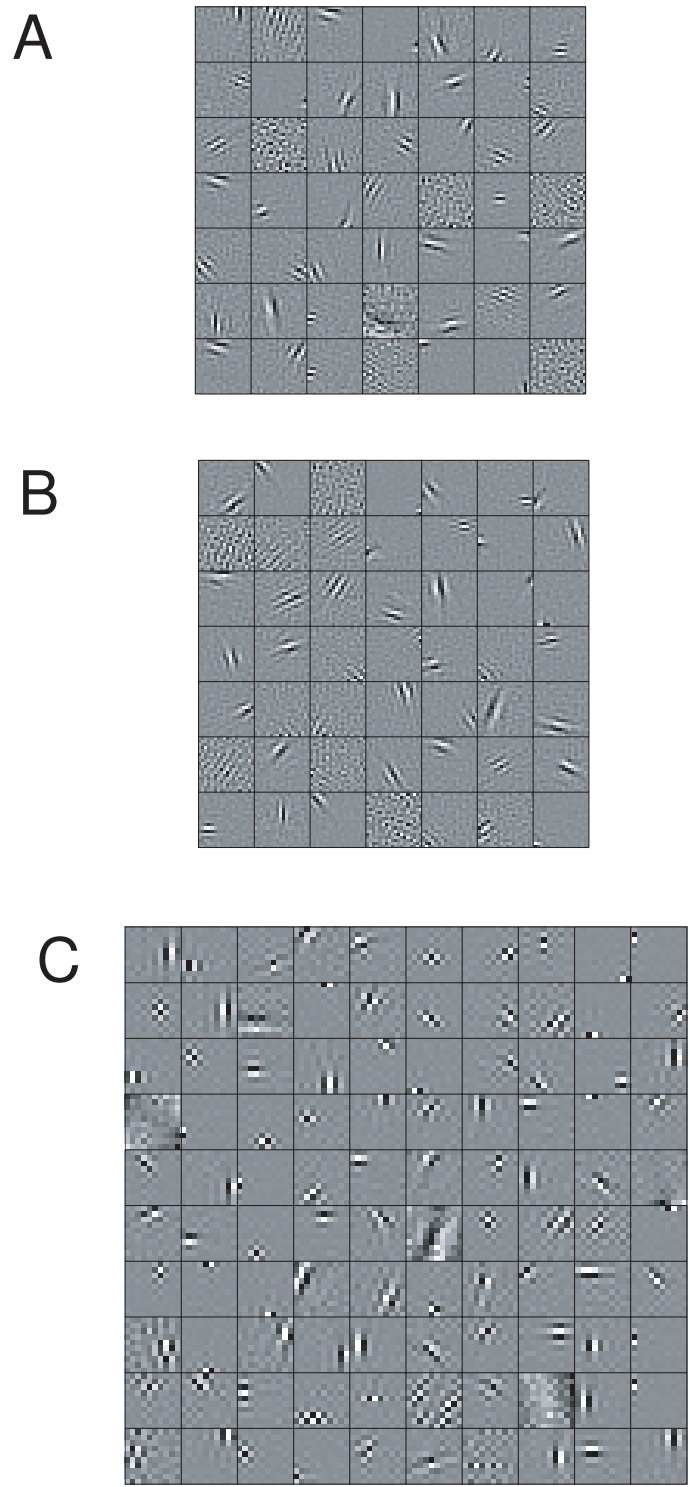

**Figure 6.7:** Collections (randomly chosen) of overcomplete filters at different degrees of overcompleteness. Patch size $18 \times 18$ whitened and reduced to 256 dimensions. The respective degrees of overcompleteness were: (A) 1.7× (B) 2.4× In panel (C) we show a collection of filters learnt from $10 \times 10$ patches, reduced to 81 dimensions, in a representation that was 100× overcomplete.

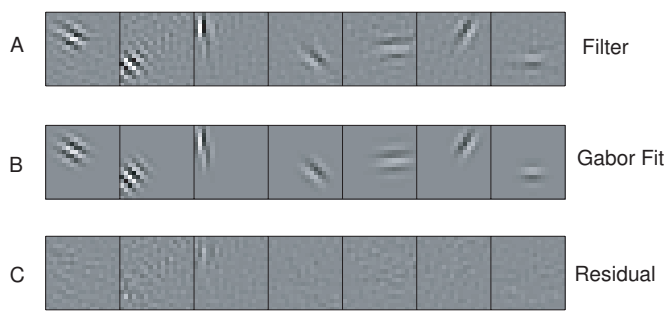

**Figure 6.8:** Comparison of Gabor fits. (A) A random collection of filters from a 1.7× overcomplete model. (B) Gabor function fits. (C) Residual error. All three rows use the same gray scale.

of understanding why this might be expected is the following. Decreasing $\alpha$ reduces the 'energy cost' for violating the constraints specified by each individual feature, however this is counterbalanced by the fact that in the overcomplete setting we expect an input to violate more of the constraints at any given time. If $\alpha$ remained constant as more features were added then the mass in the tails may no longer be sufficient to model the distribution well.

The second modification that we make is to constrain the $L2$ norm of the filters to $l$, making $l$ another a free parameter to be learned. If this modification is not made then there is a tendency for some of the filters to become very small during learning. Once this has happened, it is difficult for them to grow again since the magnitude of the gradient *depends* on the filter output, which in turn depends on the filter length.

The first manipulation simply extends the power of the model, but one could argue that the second manipulation is something of a fudge — in the case that we have sufficient data, a good model and a good algorithm it should be unnecessary to restrict ourselves in this way. There are several counter arguments to this, the principal ones being: (i) we are interested, from a biological point of view, in representational schemes in which the representational units all receive comparable amounts of input; (ii) we can view it as approximate posterior inference under a prior belief that in an effective model, all the units should play a roughly equal part in defining the density and forming the representation. We also note that a similar manipulation is also applied by most practitioners dealing with overcomplete ICA models (eg: [Olshausen and Field, 1996]).

In figure 6.7 we show example filters typical of those learnt in overcomplete simulations. As in the complete case, we note that the majority of learnt filters qualitatively match the linear receptive fields of simple cells found in V1.[6] Like V1 spatial receptive fields, most of learnt filters are well fit by Gabor functions. We analysed in more detail the properties of filter sets produced by different models by fitting a Gabor function to each filter (using a least squares procedure), and then looking at the population properties in terms of Gabor parameters.

---

[6]Approximately 5−10% of the filters failed to localise well in orientation or location — appearing somewhat like noise or checkerboard patterns. These were detected when we fitted with parametric Gabors and were eliminated from subsequent analyses.

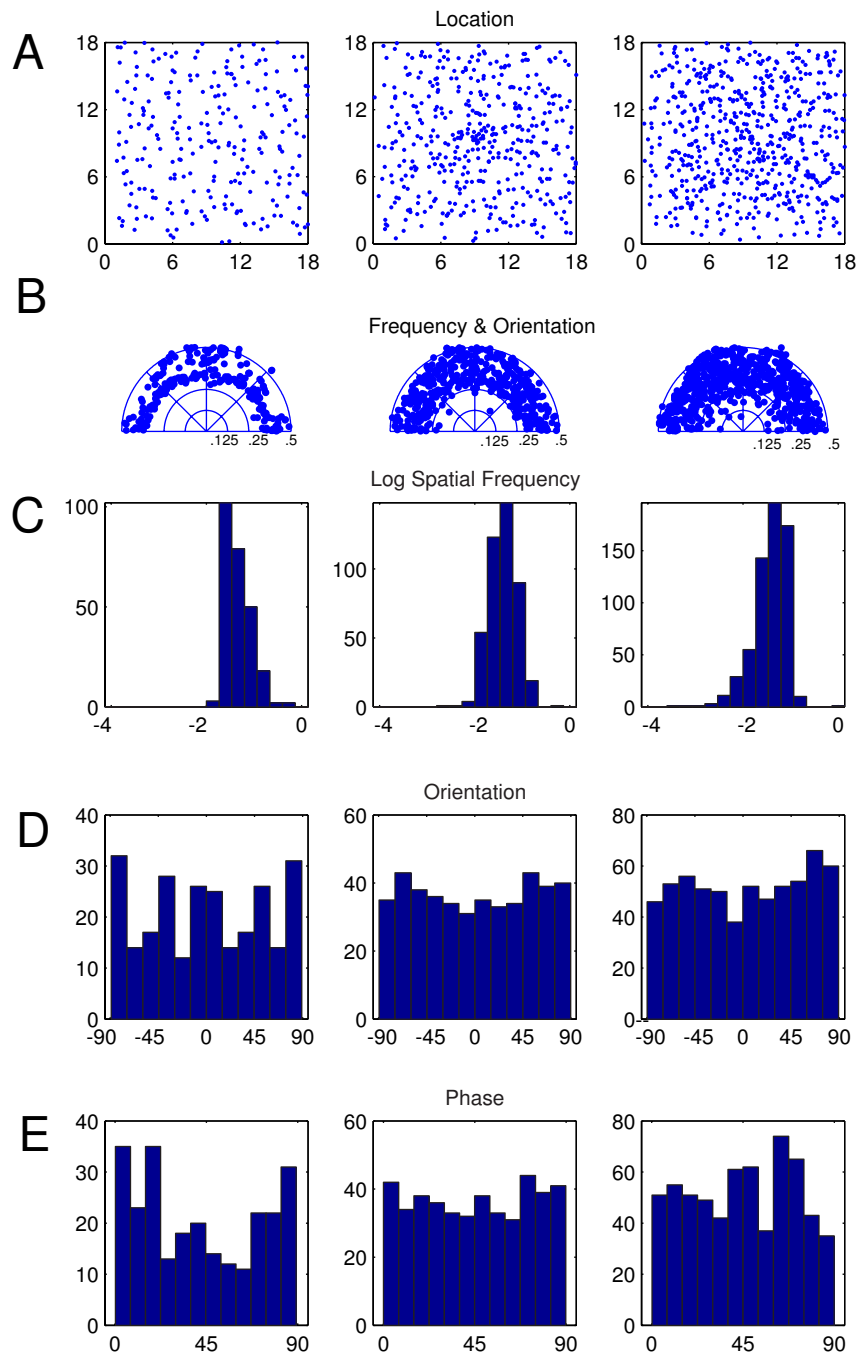

**Figure 6.9:** A summary of the distribution of some parameters used the fit the receptive fields from three models with different representation size. (A) Each dot represents the center location of a fitted Gabor. (B) Plots showing the joint distribution of orientation (azimuthally) and spatial frequency in cycles per pixel(radially). (C) Histogram of (log) spatial frequency of Gabor fit. (D) Histograms of Gabor fit orientation. (E) Histograms of Gabor fit phase (mapped to range 0–90 since we ignore the envelope sign.) The leftmost column is a complete representation, the middle column is 1.7× overcomplete and the rightmost column is 2.4× overcomplete.

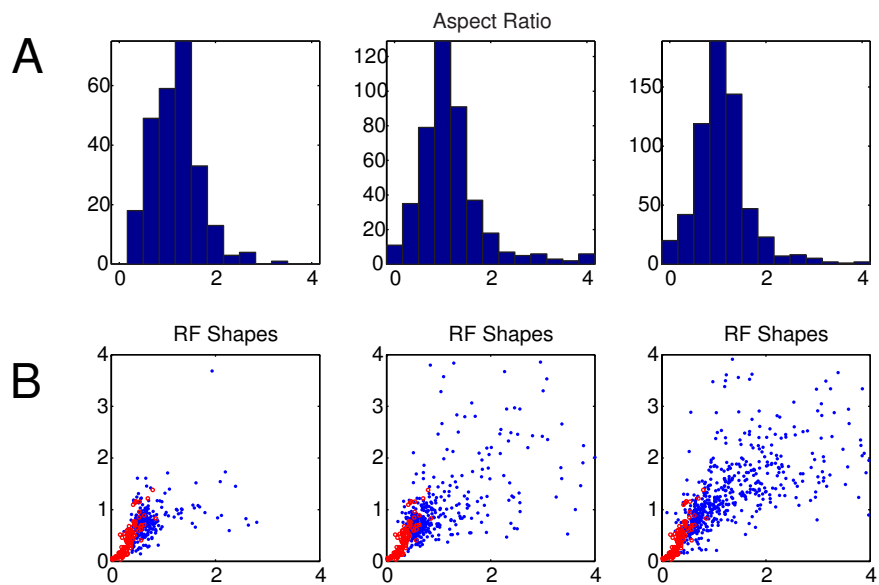

**Figure 6.10:** Characterising the shapes of the learnt receptive fields based upon the fitted parameters. (A) Histograms of the aspect ratio of the Gabor envelope (Length/Width.) (B) The a plot of 'normalised width' versus 'normalised length', c.f. Ringach [2002]. The data from our models is plotted in blue and, for comparison, data from real macaque experiments Ringach [2002] is shown in red. The leftmost column is a complete representation, the middle column is 1.7× overcomplete and the rightmost column is 2.4× overcomplete.

Figure 6.8 shows some example filters along with their corresponding fit — the residuals are barely distinguishable from zero when plotted on the same scale as the filters and Gabor fits. Figures 6.9 and 6.10 show the distribution of parameters obtained by fitting Gabor functions to complete and overcomplete filters. For reference, similar plots for linear spatial receptive fields measured *in vivo* are shown in Chapter 2, figures 2.7 and 2.8.

The plots are all good qualitative matches to those shown for the 'real' V1 receptive shown in Chapter 2. They also help to indicate the effects of representational overcompleteness of the representation. With increasing overcompleteness the coverage in the spaces of location, spatial frequency and orientation becomes denser and more uniform whilst at the same time the distribution of receptive fields shapes remains unchanged. Further, the more overcomplete models give better coverage in lower spatial frequencies that are not directly represented in complete models.

Ringach [2002] reports that the distribution of shapes from ICA/sparse coding can be a poor fit to the data from real cells — the main problem being that there are too few cells near the origin of the plot, which corresponds roughly to cells with smaller aspect ratios and small numbers of cycles in their receptive fields. The results which we present here appear to be a better fit as the figures show; this may be partly due to Ringach's choice of prior. A large proportion of our fitted receptive fields are in the vicinity of the macaque results, although as we become

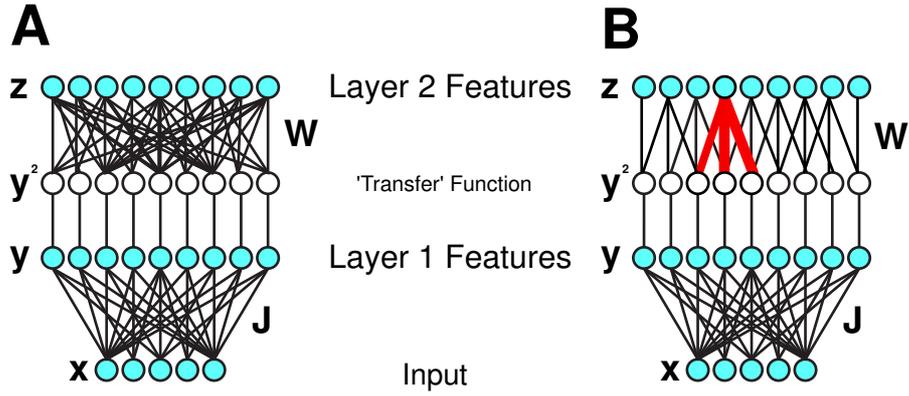

**Figure 6.11:** Graphical depiction of hierarchical extension. We add an additional layer of features/deterministic hidden units. These are obtained by passing the first layer filter outputs through a non-linearity (squaring) and then taking linear combinations. We might think of the squaring operation as embodying some sort of neural transfer function. (A) The general model in which the weights **W** are free to learn. (B) Constrained model in which the **W** are fixed and restricted to enforce a topographic ordering on the hidden units. The thick red lines schematically illustrate the a local topographic neighbourhood for a single top-layer unit. (Although not suggested by the diagram, toroidal boundary conditions were also used.)

more overcomplete we see a spread further away from the origin.

In summary, our results from these single layer PoT models can account for many of the properties of simple cell linear spatial receptive fields in V1.

## 6.6   Hierarchical Extensions

We now present results from a hierarchical extension to our original PoT model — the form of this extension is illustrated schematically in figure 6.11 (A). The extension adds an extra layer of units, which we denote by the elements of vector **z**, connected to the *squared* outputs of the first layer by a synaptic weight matrix **W**.

$$z_i = \sum_j W_{ij} y_j^2 \tag{6.11}$$

We also modify our energy function and choose one which can be defined solely in terms of the top level units. The energy associated with unit $i$ is given by $\alpha_i \log\left(1 + \frac{1}{2}z_i\right)$. (We are able to recover our original model as a special case by setting $W_{ij} = \delta_{ij}$.) In terms of **x**, this energy can be written as,

$$f_i(\mathbf{x}) = \alpha_i \log\left(1 + \frac{1}{2}\mathbf{W}_i^T[\mathbf{y}]^{\circ 2}\right)$$
$$= \alpha_i \log\left(1 + \frac{1}{2}\mathbf{W}_i^T[\mathbf{Jx}]^{\circ 2}\right) \tag{6.12}$$

where. $[\cdot]^{\circ 2}$ indicates element-wise squaring. This model could be learned using HMC but once again we are able to formulate an equivalent model using auxiliary

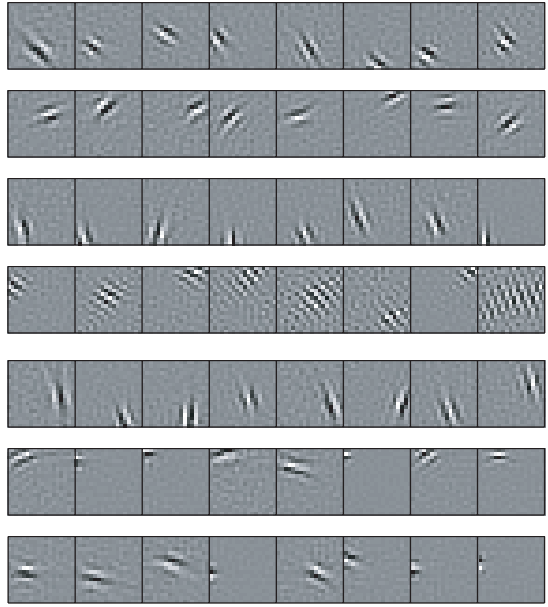

**Figure 6.12:** Each row in this figure illustrates the 'theme' represented by a different top level units. The filters in each row are arranged in descending order, from left to right, of the strength $W_{ij}$ with which they connect to the particular top layer unit.

precision variables. The joint energy function is now,

$$E(\mathbf{x}, \mathbf{u}) = \sum_i \left[ u_i \left( 1 + \frac{1}{2} \mathbf{W}_i^T [\mathbf{J}\mathbf{x}]^{\circ 2} \right) + (1 - \alpha_i) \log u_i \right] \qquad (6.13)$$

and the conditionals are

$$p(\mathbf{u}|\mathbf{x}) = \prod_i \mathcal{G}_{u_i} \left[ \alpha_i \; ; \; 1 + \frac{1}{2} \mathbf{W}_i^T [\mathbf{J}\mathbf{x}]^{\circ 2} \right] \qquad (6.14)$$

$$p(\mathbf{x}|\mathbf{u}) = \mathcal{N}_{\mathbf{x}} \left[ 0 \; ; \; (\mathbf{J}^T \mathbf{V} \mathbf{J})^{-1} \right] \qquad \mathbf{V} = \mathbf{Diag}[\mathbf{W}^T \mathbf{u}] \qquad (6.15)$$

The dimensionality of the $\mathbf{z}$ is allowed to be larger or smaller than that of the $\mathbf{y}$ space, however we will only consider cases in which the dimensionality is the same here. Also note that, even in the complete setting, the $u_i$ are marginally dependent.

In principle we are able to learn both sets of weights, $\mathbf{W}$ and $\mathbf{J}$, simultaneously. However, effective learning in this full system has proved difficult, perhaps due to the large number of degrees of freedom and the consequent existence of many poor local optima. The results which we present in this section were obtained by initialising the $\mathbf{W}$ to the identity matrix and first learning the $\mathbf{J}$, before subsequently releasing the $\mathbf{W}$ weights and then letting the system learn freely. This is therefore equivalent to initially training a single layer PoT and then subsequently introducing a second layer.

When models are trained in this way, the form of the first layer filters remains essentially unchanged from the Gabor receptive fields shown previously. Moreover, we see interesting structure being learnt in the $\mathbf{W}$ weights as illustrated by figure 6.12. The figure is organised to display the filters connected most strongly to a top

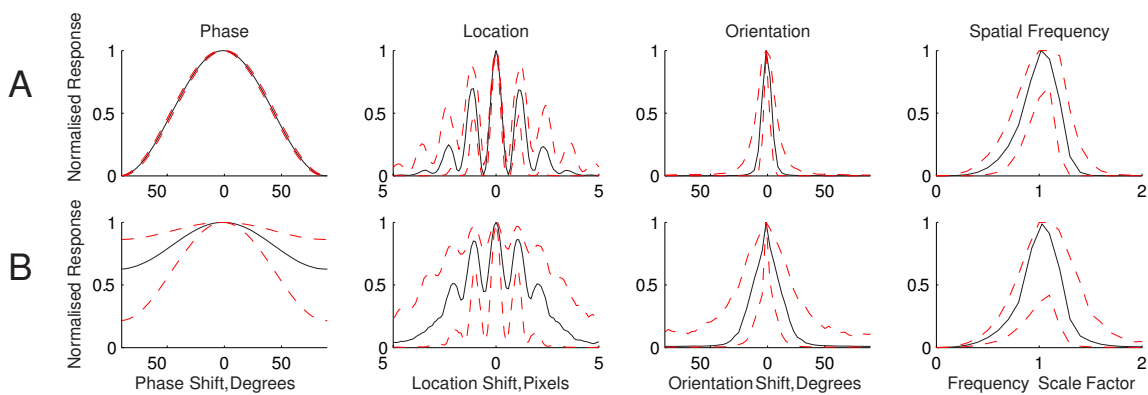

**Figure 6.13:** (A) Tuning curves for 'simple cells', i.e. first layer units. (B) Tuning curves for 'complex cells', i.e. second layer units. The tuning curves for Phase, Orientation and Spatial Frequency were obtained by probing responses using grating stimuli, the curve for location was obtained by probing using a localised Gabor patch stimulus. The optimal stimulus was estimated for each unit, and then one parameter (Phase, Location, Orientation or Spatial Frequency) was then varied and the changes in responses were recorded. The response for each unit was normalised such that the maximum output was 1, before combining the data over the population. The solid line shows the population average (median of 441 units in a 1.7× overcomplete model), whilst the lower and upper dotted lines show the 10% and 90% centiles respectively. (We use a style of display as used in Hyvärinen et al. [2001])

layer unit. There is a strong organisation by what might be termed 'themes' based upon location, orientation and spatial frequency. An intuition for this grouping behaviour follows: some filters will have correlated activity magnitudes, and by having them feed into the same top-level unit the model is able to capture this regularity. For many input images all members of the group will have small combined activity, for few images they will have significant combined activity. This is exactly what the energy function favours, as opposed to a grouping of very different filters which would lead to a rather Gaussian distribution of activity in the top layer.

Interestingly, these collected themes lead to responses in the top layer (if we examine the outputs $z_i = \mathbf{W}_i^T(\mathbf{Jx})^{\circ 2}$) that are rather reminiscent of complex cell receptive fields. As discussed in earlier chapters it can be difficult to accurately describe the response of non-linear units in a network. We choose here a simplification in which we consider the response of the top layer units to test stimuli that are gratings or Gabor patches. The test stimuli were created by finding the grating or Gabor stimulus that was most effective at driving a unit and then perturbing various parameters about this maximum. Typical results from such a characterisation are shown are shown in figure 6.13.

In comparison to the first layer units, the top layer units are considerably more invariant to phase, and somewhat more invariant to position. However, both the sharpness of tuning to orientation and spatial frequency remain roughly unchanged. These results typify the properties that we see when we consider the responses of the second layer in our hierarchical model and are a striking match to the response properties of complex cells.

## 6.7   Sparse Topographic Models

We now move on to consider a constrained form of the hierarchical PoT which we propose in an attempt induce a topographic organisation upon the representations learnt. We have noted in the previous section that the top level units appear to develop responses which are analogous to complex cells and that their weights appear to group filters by themes. We now consider the situation in which the $\mathbf{W}$ weights are fixed and define local, overlapping neighbourhoods with respect to an imposed topology (in this case a square grid with toroidal boundary conditions) as illustrated in figure 6.11 (B). The $\mathbf{J}$ weights are free to learn, and the model is trained as usual.

Representative results from such a simulation are given in figure 6.14. The inputs were patches of size $25 \times 25$, whitened and dimensionality reduced to vectors of size 256; the representation is $1.7\times$ overcomplete.

By simple inspection of the filters in panel 6.14 (A) we see that there is strong local continuity in the receptive field properties of orientation and spatial frequency and location, with little continuity of spatial phase. We can make these feature maps more apparent by plotting the parameters of a Gabor fit to the learnt filters, as shown in figure 6.14 (B). Doing this highlights several features that have been experimentally observed in maps *in vivo* — in particular we see singularities in the orientation map and a low frequency cluster in the spatial frequency map which seems to be somewhat aligned with one of the pinwheels. Whilst the map of retinotopy shows good local structure, there is poor global structure. We suggest that this may be due to the relatively small scale of the model and the use of toroidal boundary conditions (which was necessary to avoid edge effects.)

We now consider how the properties of the developed maps depend on the details of the model — in particular the degree of overcompleteness in the representation and the size of the pooling neighbourhood. Figures 6.15 and 6.16 show a selection of parameter maps representative of different conditions.

Figure 6.15 shows the effects of changing the size of the neighbourhoods over which the weights $\mathbf{W}$ pool. Our general observations are that, as we move to a larger neighbourhood size the length scale of correlations in the orientation map grows and we see greater continuity in the representation of this property. Similarly, the maps for retinotopic location seemed to gain greater continuity as we move to larger neighbourhood sizes. The map of spatial frequency seems largely unaffected by the size of the pooling neighbourhood, although since there is usually only one low frequency 'blob' in our results this observation may be rather vacuous.

As illustrated by figure 6.16, changing the degree of overcompleteness in the representation did not produce any obvious systematic changes in the qualitative appearance of the maps produced. More structural detail is discernable in the feature maps however, with a greater number of pinwheels and low frequency blobs visible.

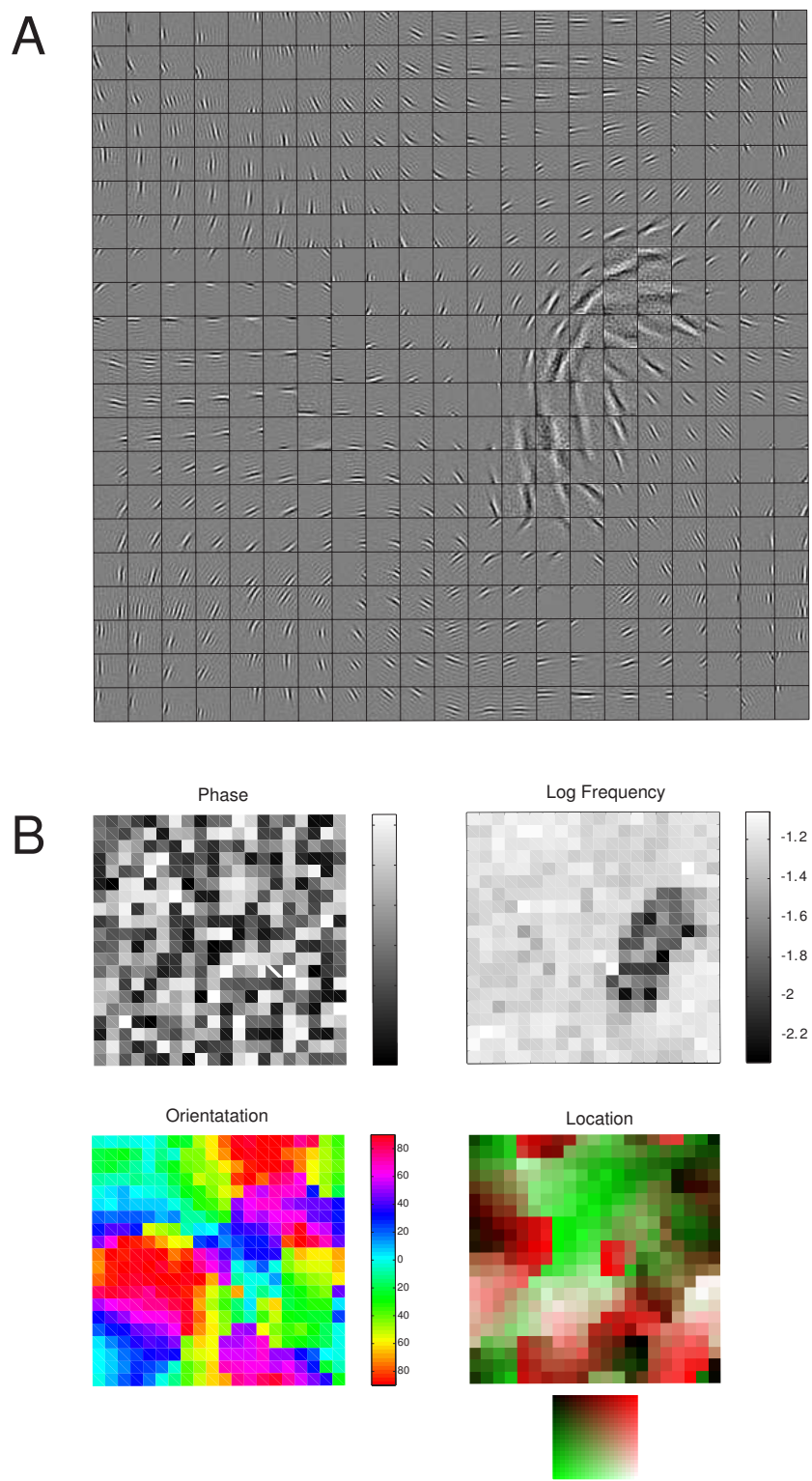

**Figure 6.14:** An example of a filter map, along with fits. (A) a topographically ordered array of learnt filters (gray scale saturating in each cell.) (B) These panels show maps for various parameters of corresponding Gabor fits to the filters. These highlight different aspects of map structure. (The model was trained on $25 \times 25$ patches that had been whitened and dimensionality reduced to 256 dimensions. The representation layer is $1.7 \times$ overcomplete in terms of the inputs. The neighbourhood size was a $3 \times 3$ square (i.e. 8 nearest neighbours.)

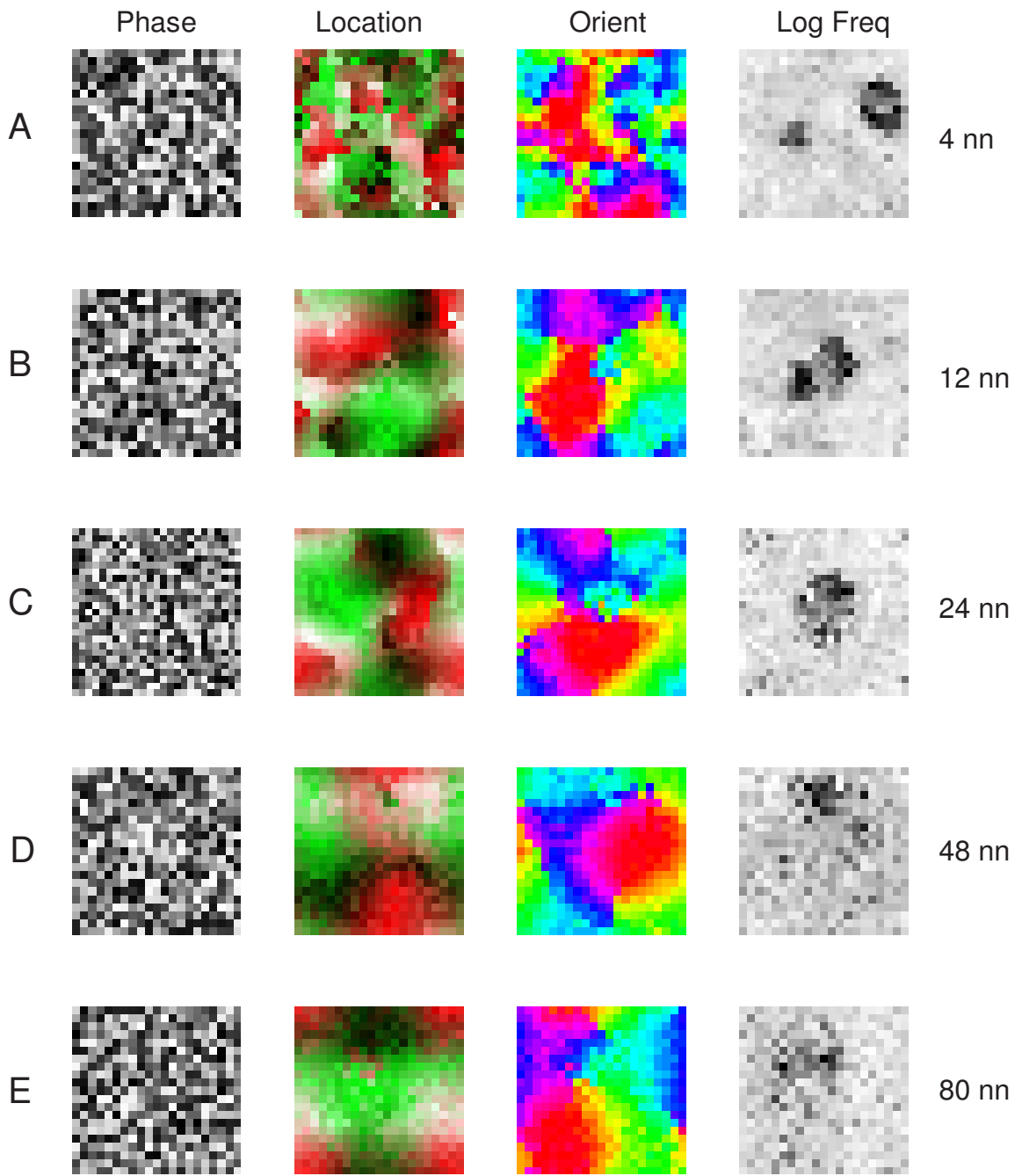

**Figure 6.15:** A selection of representative parameter maps from different neighbourhood sizes. (A) 4 nearest neighbours (Manhattan radius 1). (B) 12 nearest neighbours (Manhattan radius 2). (C) 24 nearest neighbours ($5 \times 5$ square). (D) 48 nearest neighbours ($7 \times 7$ square). (E) 80 nearest neighbours ($9 \times 9$ square). The map illustrated in figure 6.14 has neighbourhood consisting of 8 nearest neighbours ($3 \times 3$ square).

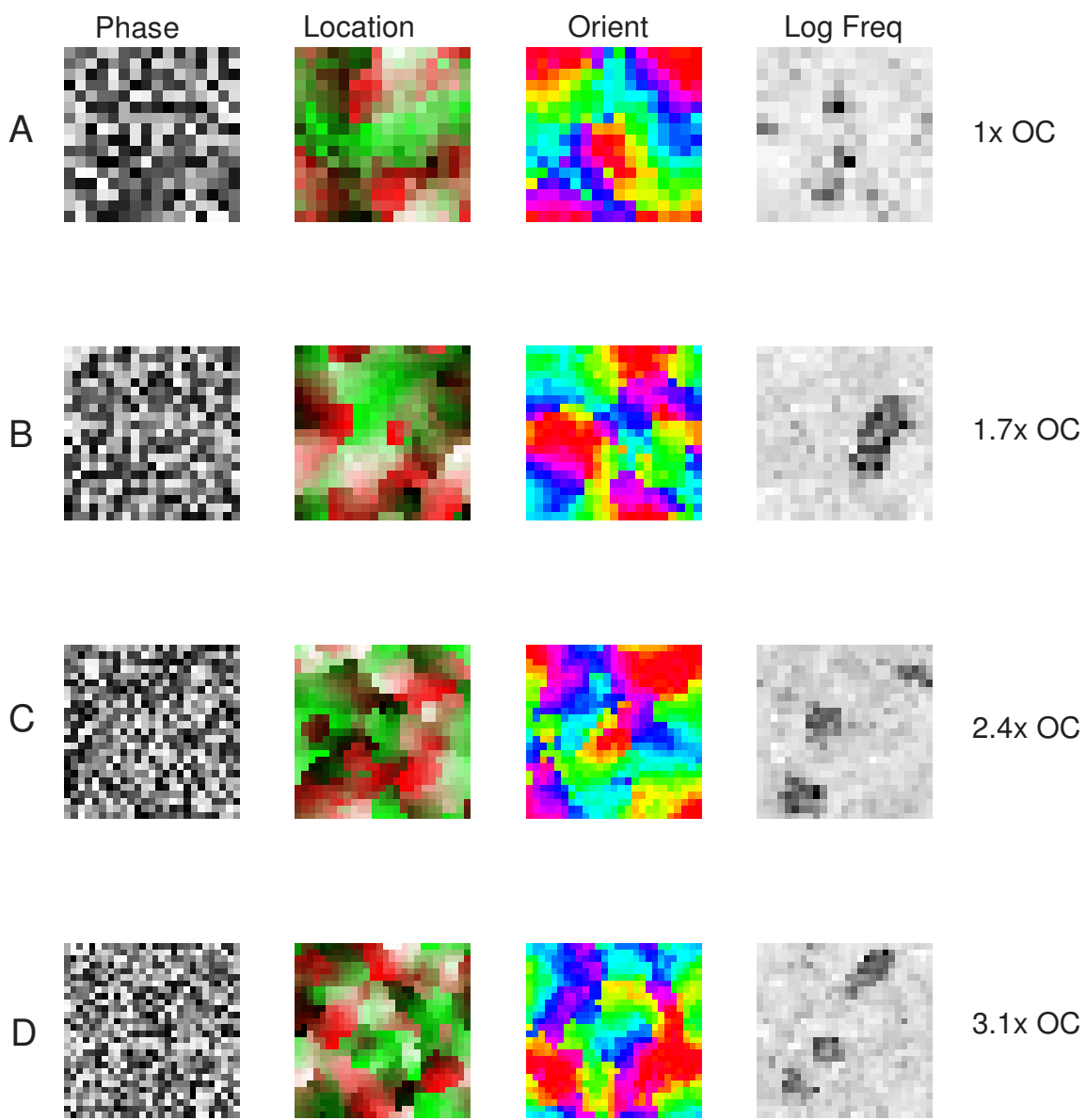

**Figure 6.16:** A selection of representative parameter maps using different degrees of representation, in all cases the neighbourhood size was 8 nearest neighbours (3 × 3 square). (A) Complete. (B) 1.7× overcomplete. (C) 2.4× overcomplete. (D) 3.1× overcomplete.

**A**

**B**

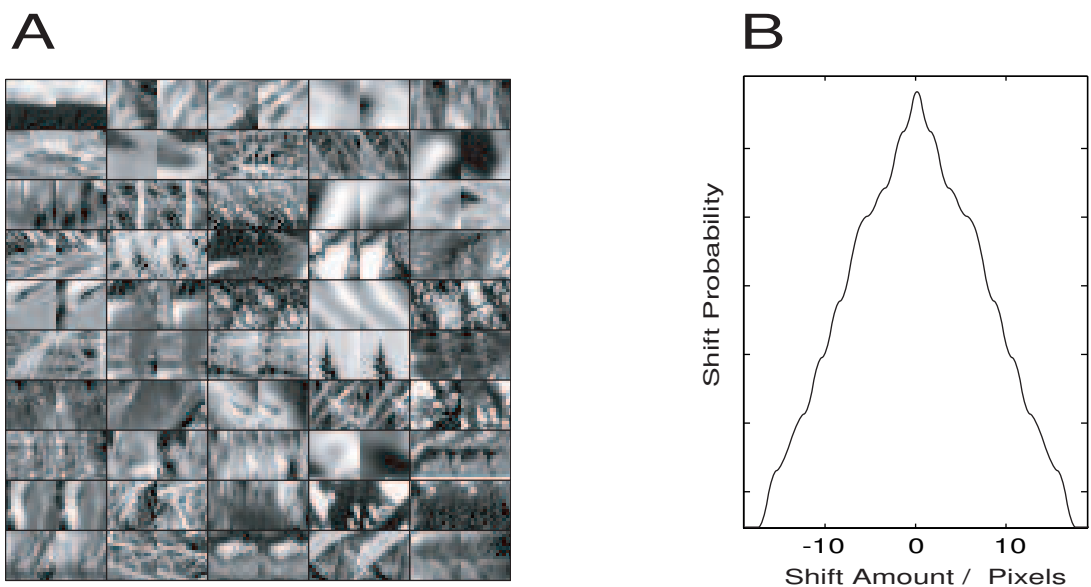

**Figure 6.17:** Synthetic stereo data. (A) Examples of stereo shifted patches. (B) The distribution of shifts used to generate 'synthetic' stereo pairs. (The 'bumps' in the curve are due to interpolation of the discrete shift values used.)

## 6.8   Introducing Stereo

We have also carried out some preliminary investigations using 'synthetic' stereo pairs of natural scenes. There are few high quality datasets of natural-scene stereo-pairs publicly available, so we followed a procedure analogous to that outlined in Chapter 5 and generated 'synthetic pairs' by creating data sets in which laterally shifted pairs of patches were extracted from sets of mono-images. Patches were selected with a distribution of offsets, as illustrated by figure 6.17, and in doing so we hope to replicate the shifts that would be caused on this scale due to disparity between the two eyes.

Before presentation to our algorithm, these patches were centred, and whitened and dimensionality reduced as previously described in section 6.4. In the experiments reported here we performed the whitening and dimensionality reduction in the joint space of left and right eye inputs. Such a process clearly cannot happen *in vivo* since the inputs from the two eyes remain separate until $V1$, however the structure in the inputs is essentially equivalent for our purposes. By coupling some of the redundancy between the two eyes we are able to achieve a higher degree of dimensionality reduction for the same quality of image reconstruction (from the reduced components), which speeds up our computations since we treat smaller input vectors.

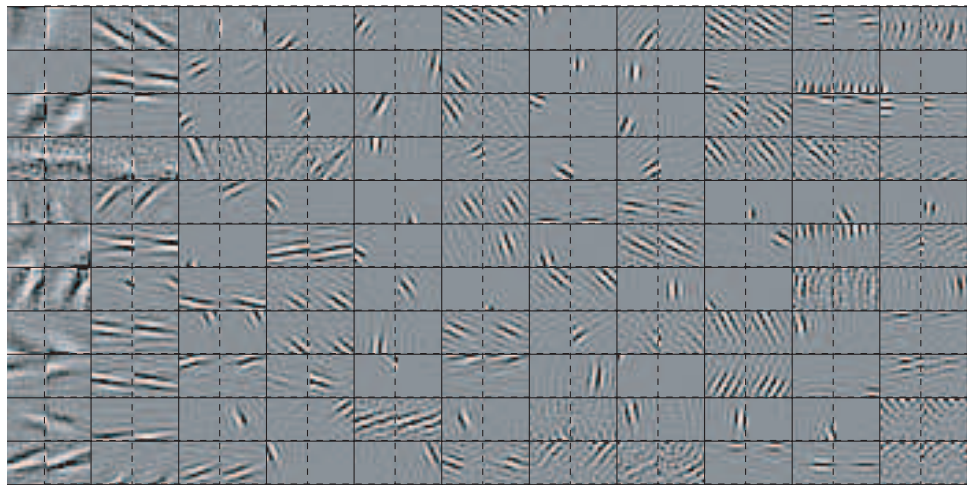

**Figure 6.18:** Stereo receptive fields. A random subset of 121 stereo-pair filters taken from a 1.7× overcomplete PoT model. Each rectangle demarcated by solid lines represents a single unit, the left and right halves of the each rectangle (itself split by a dotted line) show the filter for the left and right eye respectively. The filters have been ordered roughly by spatial frequency.

### 6.8.1   Stereo Results

#### 6.8.1.1   Single Layer

$150,000$ left and right eye patches of size $16 \times 16$, and were jointly whitened and reduced to vectors of length 256 by removing the DC component and the lowest 256 eigenvector subspace. We then trained an overcomplete single layer PoT using contrastive divergence for 200 passes through the entire data set. Figure 6.18 shows a subset of learnt filters, ordered by spatial frequency.

We see that, as in the single eye case, the learnt filters resemble simple cell receptive fields/Gabor patches. The distribution parameters of the Gabors for each eye remains roughly the same as for the single eye case, however we are now able to consider the difference in the learnt receptive fields from the two eyes.

It is immediately clear that whilst most units have comparable inputs from both eyes we see some that are strongly dominated by one eye or the other. Of the cells that are strongly monocular, we note that the majority are of relatively high spatial frequency (as in Li and Atick [1994]) and that they tend to be closer 45° rather than vertical or horizontal in orientation.

In chapter 5 we saw that one of the determining factors for the distribution of ocularity seems to be the correlation between the two eyes, in our case manifested by the distribution of shifts used to generate the data. For narrow shift distributions we observed almost entirely binocular (equal strength from both eyes) units, whilst for independent left and right eye images we saw almost entirely monocular units. Consequently, we must bear in mind that our results depend on the nature of our synthetic data set; it would clearly be desirable to use a 'real' data set of high quality, calibrated stereo images (although there are other complications such as vergence and focus depths).

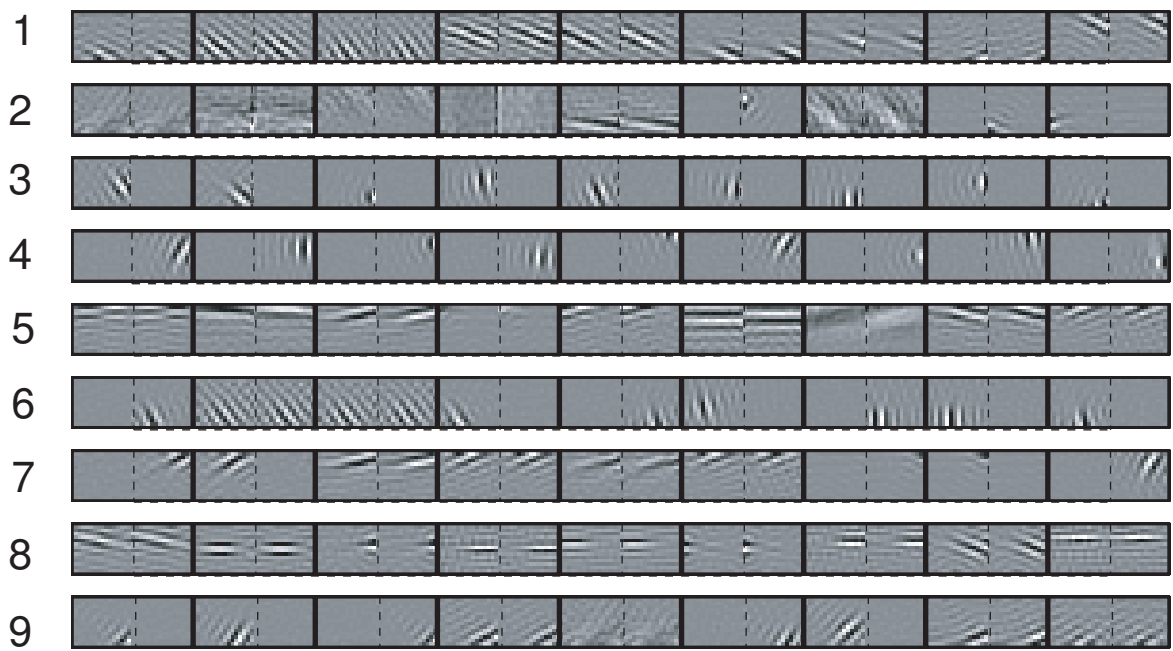

**Figure 6.19:** Each row shows the dominant 9 filter pairs feeding into a top level 'complex cell' unit after training. The solid rectangles depict each unit, with the left and right eyes separated by a dotted line.

### 6.8.1.2   Hierarchical Models & Stereo Tuning Curves

Figure 6.19 illustrates a collection of 'themes' developed by the model if we allow an extra layer of weights and train $\mathbf{W}$ after initially learning the first layer. We see that as before there is a clustering based upon orientation, location and spatial frequency. We also note that some complex cells are predominantly monocular (eg: rows 3 and 4) whilst most are binocular to various degrees.

We now consider the tuning properties of the units in our models after training. The (single eye) tuning properties for orientation, phase, spatial frequency and location are generally the same as those for single eye models, of the type illustrated in figure 6.13. Additionally, we can ask about some properties of stereo tuning — in particular to disparity. However, answering this question is more difficult than it might initially seem. Analytically (i.e. based upon the receptive field fits) there is no wholly satisfactory measure; empirically we are faced with the decision of what to use as a probe stimulus, since this will affect the results which we obtain (for a discussion of some of these issue see Fleet et al. [1996], Zhu and Qian [1996], Anzai et al. [1999a,b,c], Hoyer and Hyvarinen [2000].)

One possibility is to use the 'optimal' stimuli for one eye or the other, but this can lead to misleading confounds arising from, for instance, the periodicity and orientation of the stimulus. Another possibility is to use the same sort of images used in training to probe the disparity tuning. Since our 'stereo pairs' are synthetically generated, we have ready access to naturalistic data in which the offset between left and right eye patterns is known. By averaging over very many such stimuli, one might hope that this method overcomes some of the stimulus dependency effects and gives a 'truer' measure of tuning.

To generate this naturalistic test data we randomly selected a location in the large image and then extract a set of patch pairs, at different lateral shifts. Simple cell 'tuning curves' were obtained by taking the filter output to each image at each shift, applying a half-wave rectification, and then averaging over all such (non-zero) rectified response at a given disparity. (The half-wave rectification is necessary to avoid the cancellation effects that would otherwise be caused by averaging over similar patterns with opposite contrast.) Complex cell tuning curves were similarly obtained.

Examples of our characterisations of disparity tuning curves are given for first-layer units ('simple cells') in figure 6.20, and for second-layer units ('complex cells') in figure 6.21.

We first consider the simple cells. Our examples have been chosen to illustrate the range of tuning behaviours that we have observed. The top row shows a example of a 'tuned excitatory' type profile, with a symmetric preference for zero disparity. The middle row illustrates a 'tuned inhibitory' type profile, with a symmetric preference for the disparity to be different from zero. Whilst the bottom row shows an example of asymmetric preference, 'near' or 'far' tuning depending on the direction of asymmetry.

The plots for the complex cells show a similarly interesting range of behaviour. We see inhibitory and excitatory tuning as in the simple cells and we also see a number of binocular complex cells with a rather broad *invariance* to disparity. However, in our brief analyses, we found few complex cells with striking tuning asymmetries.

In addition to these empirical measures, we also consider some simple analytical measures for simple cells based upon the Gabor fits to the left- and right-eye receptive fields (based upon the analysis of Anzai et al. [1999a]).

We will consider: (i) the phase offset, $d_{\phi p}$, between the left and right eye filters; (ii) the amount of lateral shift spatially, due to phase difference, of the underlying Gabor carriers, $d_{\phi v}$; and also (iii) the lateral shift of the underlying Gabor envelope $d_X$. These quantities are given by,

$$d_{\phi p} = \phi_L - \phi_R \tag{6.16}$$

$$d_{\phi v} = \left(\frac{\phi_L}{2\pi f_L} - \frac{\phi_R}{2\pi f_R}\right)\cos\left(\frac{\theta_L + \theta_R}{2}\right) \tag{6.17}$$

$$d_X = x_L - x_R \tag{6.18}$$

Figure 6.22 shows histograms for these measures from a typical model after training. (We also looked for correlations between these measures and other parameters such as spatial frequency and orientation, however no significant trends could be discerned.) The phase offset, $d_{\phi p}$ is a measure of the similarity of the two receptive fields and we note a slight tendency for pairs to be formed that are in phase or anti-phase, with each other. Both $d_{\phi v}$ and $d_X$ are spatial measures relating to the lateral shift of the left and right eye receptive fields. We note that the range of shifts due to location difference is significantly larger than than for phase shifts, suggesting that in our model position disparity plays a larger role than phase disparity. This goes somewhat against the current understanding of receptive fields *in vivo*, although the debate about the relative significance of the two mechanisms is ongoing (for example Anzai et al. [1999a], Prince et al. [2002]).

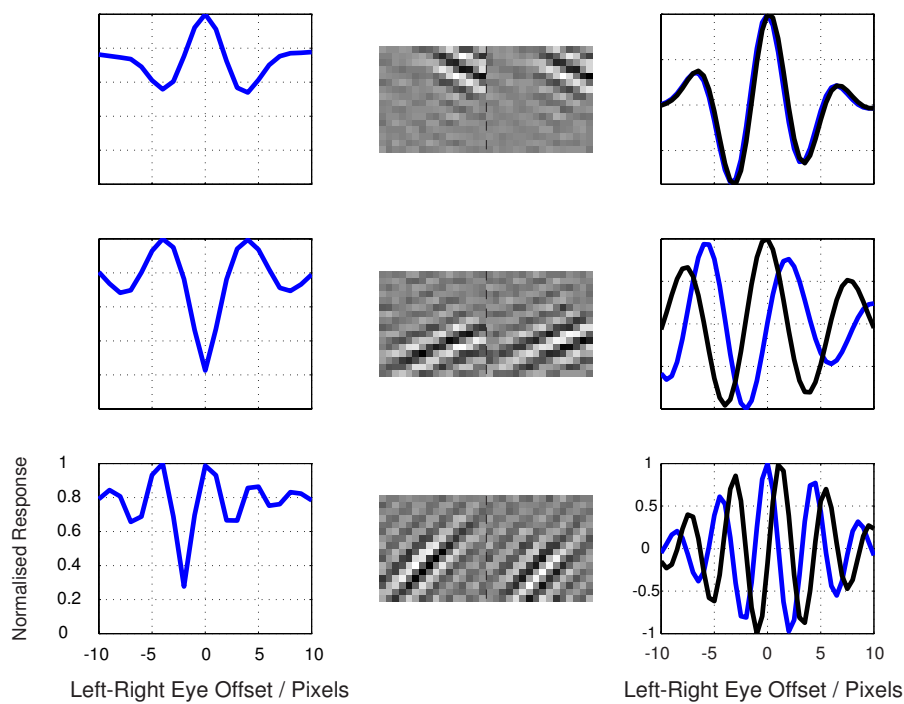

**Figure 6.20:** Disparity tuning examples for 'simple cells'. Each row depicts the results from a single unit. The central column shows the left- and right-eye receptive field pair for that unit. The leftmost column shows the average response to natural 5000 image patches at given disparities. (The output was rectified before taking this average.) The rightmost column shows the response to the left-eye optimal pattern (blue curve) and right eye optimal pattern (black curve). Each curve in this figure has been normalised so that its maximum absolute value is 1.

### 6.8.1.3    Topographic Maps For Stereo Inputs

We have also applied our topographic PoT models to stereo pairs of input patches. Our goal is to incorporate maps of ocular dominance and disparity alongside those for orientation, spatial frequency and retinotopy. We present preliminary results in figures 6.23 and 6.24, which show a map for ocular dominance (ocularity) in addition to those for phase, location, orientation and spatial frequency. We consider the ocularity $\triangle_O$ of a simple cell as being the sum of absolute weight values from the left eye, minus the sum of absolute weight values from the right eye, i.e.

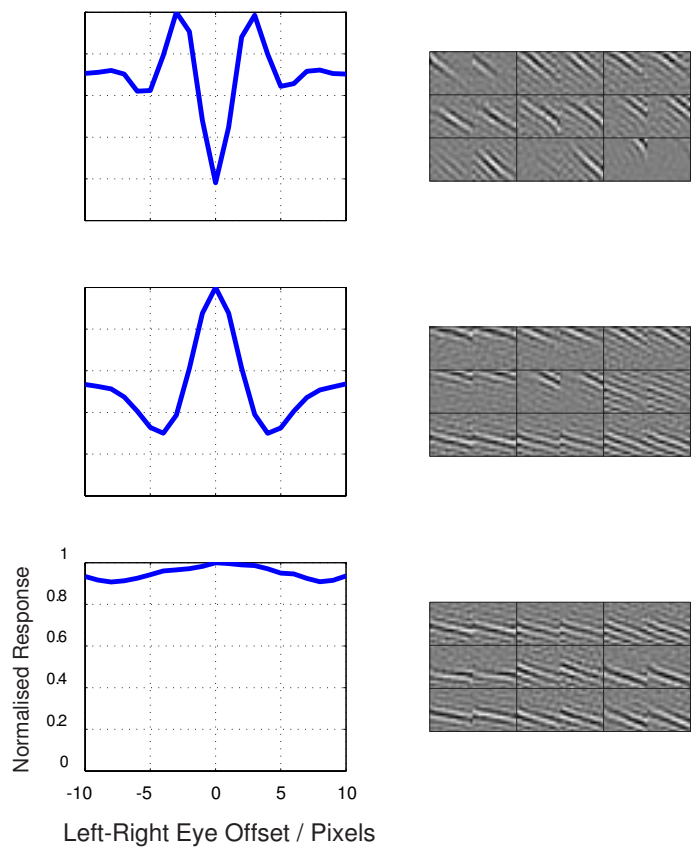

**Figure 6.21:** Disparity tuning examples for 'complex cells'. Each row depicts the results from a single top layer unit. The right hand column shows the left- and right-eye receptive field pairs for the 9 simple cells that are most strongly connected to that unit. The left hand column shows the normalised, average response to 5000 image patches at different disparities.

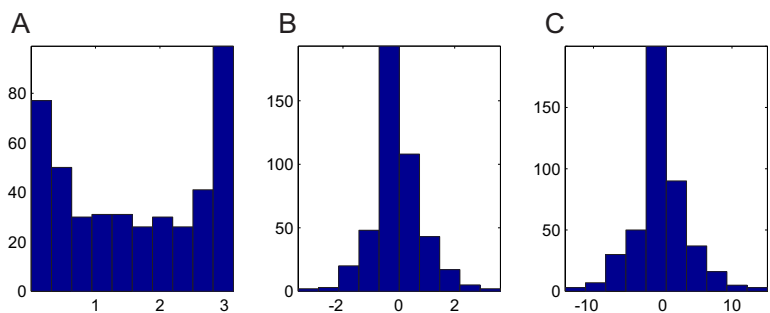

**Figure 6.22:** Typical histograms of different measures of simple cell disparity from a trained model, based upon Gabor fits to left and right eye receptive fields. (A) Phase offset (modulo $pi$) $d_{\phi p}$. (B) Spatial disparity due to phase difference $d_{\phi v}$. (C) Spatial disparity due to Gabor envelope offset $d_X$.

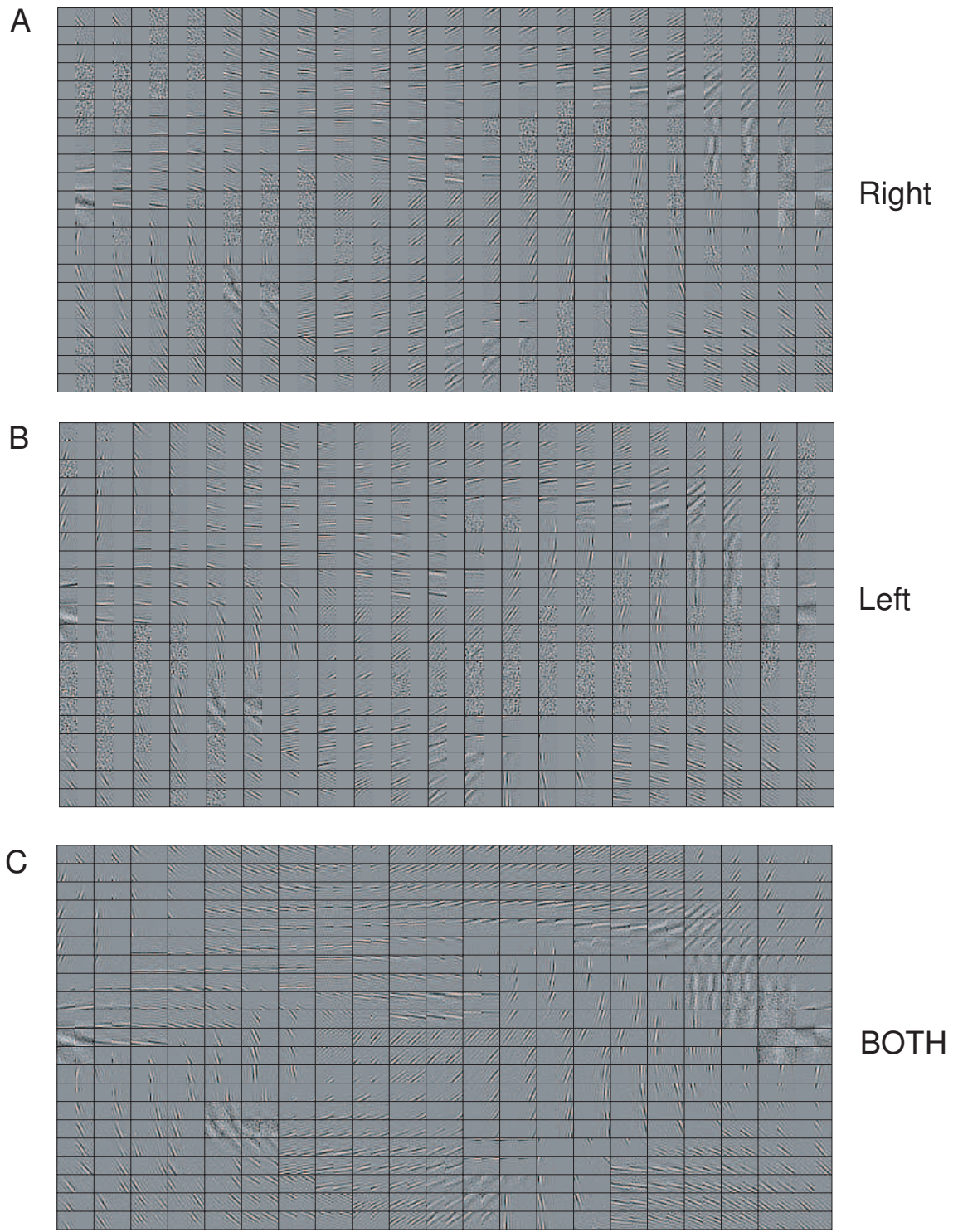

**Figure 6.23:** (A) Map showing right eye filters only. Left eye filters have been set to zero for this plot. (B) Map showing left eye filters only. Right eye filters have been set to zero for this plot. (C) Joint map showing topographically ordered filter pairs for both eyes. Not that each unit is plotted independently normalised to fill the full gray scale in each plot; consequently, monocular regions preferring the 'other eye' appear as noisy patches in panels B and C.

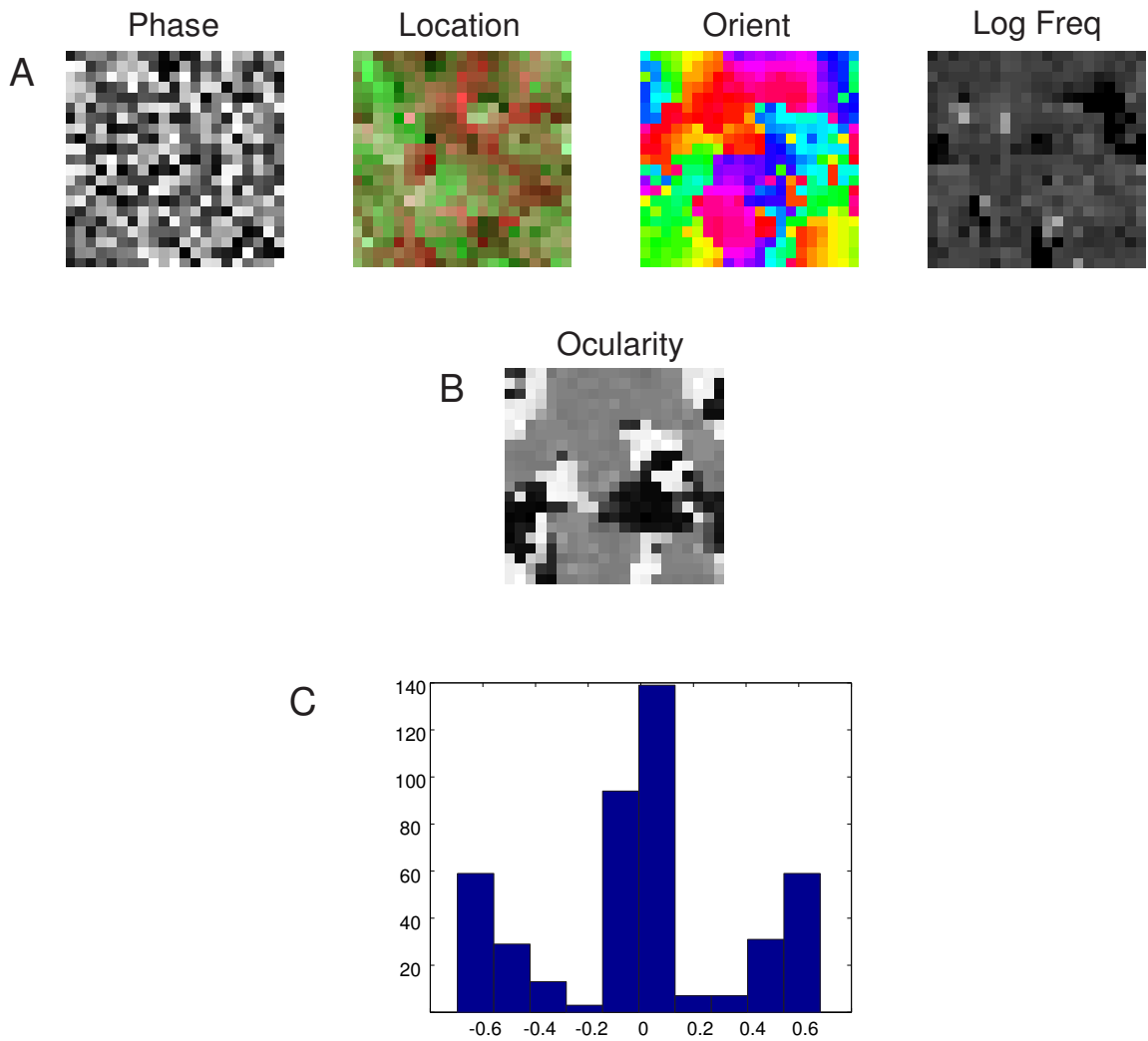

**Figure 6.24:** (A) Feature maps for phase, location, orientation and log frequency. (B) Feature map for ocular dominance. (C) Histogram of ocularity.

$$(\triangle_O)_i = \sum_j \left( |J_{ij}^L| - |J_{ij}^R| \right) \tag{6.19}$$

We see there there is some patterned ocular dominance structure concomitant with the map properties already outlined. Although this organisation is rather weaker and much more patchy, there does seem to be a tendency towards interdigitated left and right eye preferring domains as is found *in vivo*. We speculate that in yet larger models, using genuine stereo inputs, perhaps with a larger factor of overcompleteness, and appropriate construction of neighbourhood interactions that such maps may be better defined although we leave this possibility open at present.

## 6.9   Gaussian Scale Mixtures & Divisive Normalisation

As mentioned in the introduction to this chapter we can consider the complete version of our model as a Gaussian scale mixture [Andrews and Mallows, 1974, Wainwright and Simoncelli, 2000, Wainwright et al., 2000a,b] with a particular (complicated) form of scaling function.[7]

Wainwright and Simoncelli [2000] present the form for a GSM density for a variable, **g**, as follows,

$$p_{\mathrm{GSM}}(\mathbf{g}) = \int_\infty^\infty \frac{1}{(2\pi)^{\frac{N}{2}} |c^2 \mathbf{Q}|^{\frac{1}{2}}} \exp\left( -\frac{\mathbf{g}^T (c\mathbf{Q})^{-1} \mathbf{g}}{2} \right) \phi_c(c) dc \tag{6.20}$$

where $c$ is a non-negative scalar variate and and $\mathbf{Q}$ is a positive definite covariance matrix. This is the distribution that results if we draw $c$ from $\phi_c(c)$ and variable **v** from a multi-dimensional Gaussian $\mathcal{N}_\mathbf{V}(0, \mathbf{Q})$ and then take $\mathbf{g} = \sqrt{c}\mathbf{v}$.

Wainwright et al. [2000a,b] also discuss a model in which the distributions of co-efficients in a wavelet decomposition for images are described by a GSM which has a separate scaling variable, $c_i$, for each co-efficient. The $c_i$ have a Markov dependency structure based on the multi-resolution tree which underlies the wavelet decomposition.

In the complete setting of our hierarchical PoT we can consider columns of the matrix **A**, given by $\mathbf{A} = \mathbf{J}^{-1}$, i.e. the inverse of our filter matrix, as a basis vector associated with each deterministic hidden variable at the first level, i.e. the features $y_i$. If we then integrate the distribution given by the energy function in equation 6.13 with respect to **x**, we obtain the following form for the marginal distribution of

---

[7]In simple terms a GSM density is one that can be written as a (possibly infinite) mixture of Gaussians that differ only in the scale of their covariance structure. A wide range of distributions can be expressed in this manner.

the auxiliary variables $\mathbf{u}$

$$p(\mathbf{u}) = \frac{1}{Z_u} \prod_i e^{-u_i} u_i^{\alpha_i - 1} \prod_k \left( \sum_j W_{jk} u_j \right)^{-\frac{1}{2}} \qquad (6.21)$$

where the partition function $Z_u$ ensures normalisation. (We are unfortunately unable to provide a simple intuition for what this distribution 'looks like'.)

Conditioned upon $\mathbf{u}$, the features $\{y_i\}$ of the hierarchical PoT are independent Gaussians (in the complete case). Thus we see that the marginal distribution of each of the $y_i$ is then a Gaussian scale mixture in which the scaling variate for $y_i$ is given by $c_i(\mathbf{u}) = \left( \sum_j W_{ji} u_j \right)^{-1}$.

The neighbourhoods defined by $\mathbf{W}$ in our model play an analogous role to the tree structure Markov process in Wainwright et al. [2000a,b], and determine the correlations in scaling between different co-efficients.

The overcomplete version of our PoT is not so easily interpreted as a GSM because the $\{y_i\}$ are no longer independent given $\mathbf{u}$, nor is the distribution over $\mathbf{x}$ a simple GSM due to the way in which $\mathbf{u}$ is incorporated into the covariance matrix (see equation 6.15). However, much of the flavour of a GSM seems to remain.

Wainwright et al. [2000a,b] also show that Gaussian scale mixtures, with appropriate choices of the distribution of scaling variables, are able to capture a particular form of higher order structure found in images. This structure is evident in the correlations between (square) rectified outputs of linear filters applied to natural scene ensembles. Even when the raw filter outputs themselves are uncorrelated, there is often dependency between their rectified outputs. This can be demonstrated neatly by so-called bow-tie plots in which the distribution for the output of one filter *conditioned* on the output of another filter is plotted. An example of such bow-tie plots for the filters learnt using a square implementation of our PoT is given in figure 6.25 (A) & (B).

This 'bow-tie' phenomenon has also been given prominence in the context of understanding 'divisive normalisation' of V1 responses [Simoncelli and Schwartz, 1999, Wainwright et al., 2001, Cavanaugh et al., 2002a,b]. In these approaches the authors suggest a divisive normalisation of the form given in equation 6.22 might help to remove higher order dependency from the simple cell responses.

$$R_j^* = \frac{L_j^2}{\sigma_j^2 + \sum_j \omega_{jk} L_k^2} \qquad (6.22)$$

where $\sigma_j^2$ and $\omega_{jk}$ are free parameters and $L_k$ are the outputs of linear filters applied to an image. Their parameters are chosen to minimise the expectation of $\left( \log \left( R_j^* \right) \right)^2$ over a set of input patterns. This can also be somewhat related to the GSM framework.

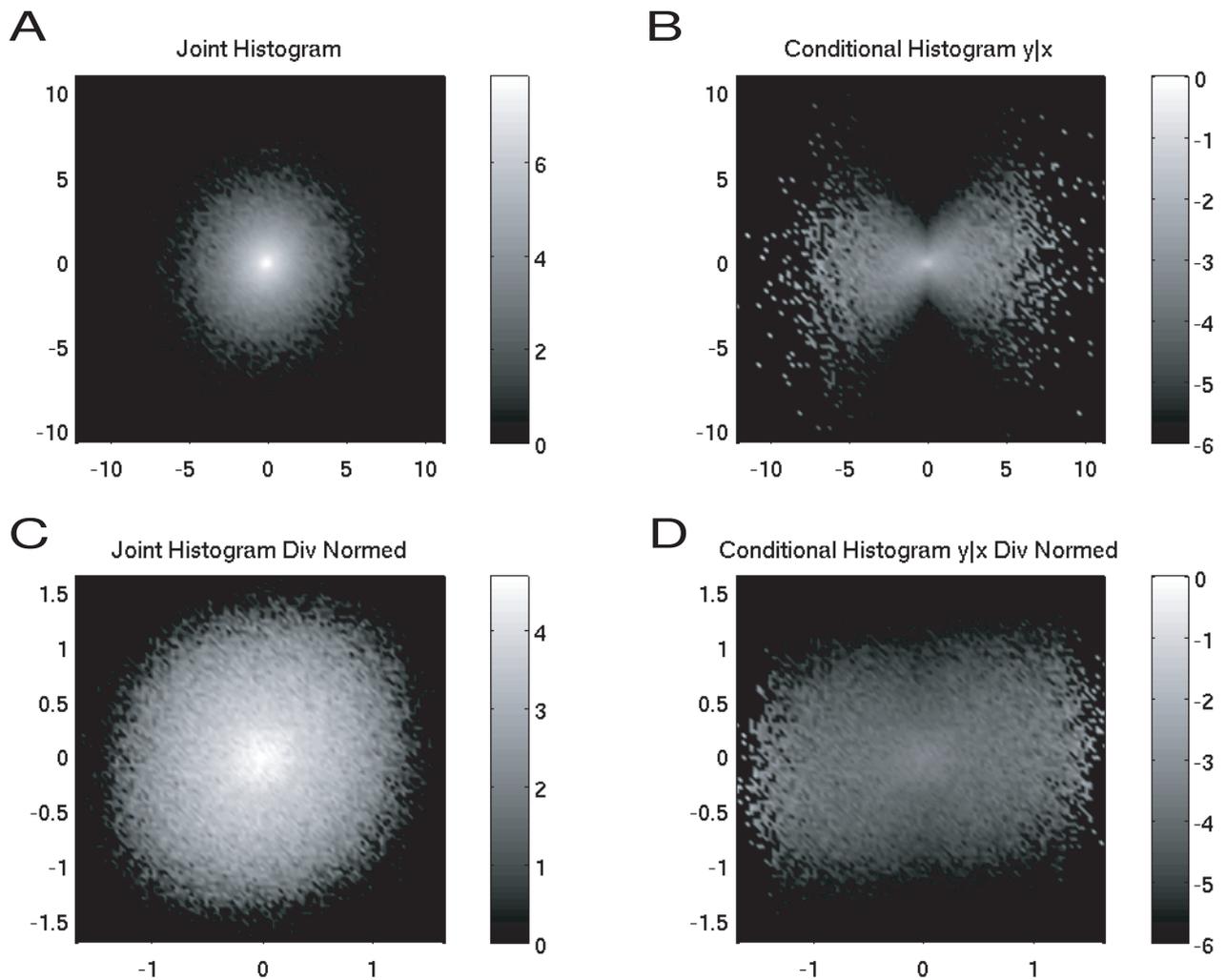

**Figure 6.25:** (A) Joint histogram of responses for two neighbouring filters from a topographic model. (B) Conditional histogram for response of one filter given the response of the second. Note the characteristic 'bow tie' shape — whilst the linear responses are uncorrelated, there is a magnitude of the two filter responses. Filter 2 is much more likely to respond strongly when filter 1 is also responding strongly. (C) Divisively normalised joint response histogram. (D) Conditional histogram of divisively normalised responses. There is a much greater independence as compared with panel (C).

We note that the divisive interaction in equation 6.22 is extremely similar to the form that would arise in our framework if we inferred the MAP estimate of $\mathbf{u}$, given an input pattern, and then used these $\{u_i\}$ to recover 'underlying' Gaussian variates. So, it is natural to investigate how the learnt weights $\mathbf{W}$ perform at removing this higher order redundancy. We perform normalisation as,

$$y_i^* = y_i \left( \sum_j \frac{W_{ji}(\alpha_j - 1)}{1 + \frac{1}{2}z_j} \right) \tag{6.23}$$

where $\mathbf{y}$ are the first layer deterministic hidden units and $\mathbf{z}$ are the second layer.

Figure 6.25 (C) & (D) shows an example of the results of such an operation. We see that there is much greater independence between the normalised responses and the raw responses. Furthermore, we note that the information required to perform the normalisation is present within the activities of the top layer activities, $\mathbf{z}$ — our 'complex cells'.

Combining this with earlier results, it is tempting to postulate the existence of cortical micro-circuitry in which complex cell responses feed back (via interneurons) to divisively normalise simple cell responses — although this is complicated by the fact that complex cells themselves show divisive normalisation effects.

## 6.10 Relationships to Topographic ICA

We will now show that, in the complete case, the sparse topographic models outlined in section 6.7 are isomorphic to the model optimised (but not the model initially proposed) by Hyvarinen et al. [2001] in their work on topographic ICA. These authors define an ICA generative model in which the components/sources are not completely independent but have a dependency that is defined with relation to some topology, such as a toroidal grid — components close to one another in this topology are have greater co-dependence than those that are distantly separated.

Their generative model is shown schematically in figure 6.26. The first layer takes a linear combination of 'variance-generating' variables, $\mathbf{t}$, and then passes them through some non-linearity, $\phi(\cdot)$, to give positive scaling variates, $\sigma$. These are then used to set the variance of the sources, $\mathbf{s}$, and conditioned on these scaling variates, the components in the second layer are independent. These sources are then linearly mixed to give the observables, $\mathbf{x}$. The joint probability $p(\mathbf{s}, \mathbf{t})$ in this model is given by

$$p(\mathbf{s}, \mathbf{t}) = \prod_i p_{s_i} \left( \frac{s_i}{\phi\left(\mathbf{H}_i^T \mathbf{t}\right)} \right) \frac{p_{t_i}(t_i)}{\phi\left(\mathbf{H}_i^T \mathbf{t}\right)} \tag{6.24}$$

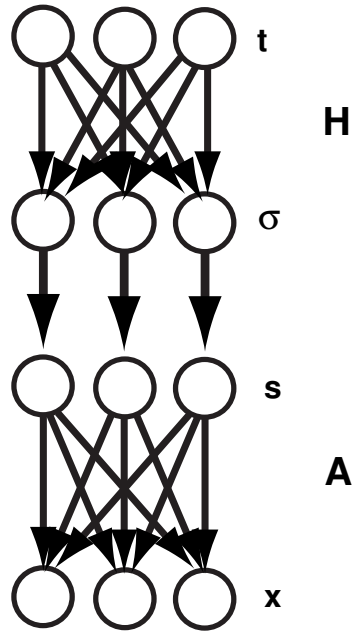

**Figure 6.26:** Graphical model for topographic ICA [Hyvärinen et al., 2001]. First the variance 'generating variables', $t_i$, are generated independently from their prior. They are then linearly mixed inside their topographic neighbourhoods through the non-negative matrix $\mathbf{H}$, before being non-linearly transformed using function $\phi(\cdot)$ to give the variances, $\sigma_i = \phi(\mathbf{H}_i^T \mathbf{t})$, for each of the sources, $i$. Values for these sources, $s_i$, are then then generated from independent zero mean Gaussians with variances $\sigma_i$, before being linearly mixed through matrix $\mathbf{A}$ to give observables $x_i$.

and the log likelihood of data given their parameters is

$$\mathcal{L}(\mathbf{B}) = \sum_{\text{data } \mathbf{x}} \int \prod_i p_{s_i} \left( \frac{\mathbf{B}_i^T \mathbf{x}}{\phi \left( \mathbf{H}_i^T \mathbf{t} \right)} \right) \frac{p_{t_i}(t_i)}{\phi \left( \mathbf{H}_i^T \mathbf{t} \right)} |\det \mathbf{B}| d\mathbf{t} \qquad (6.25)$$

where $\mathbf{B} = \mathbf{A}^{-1}$, $p_{t_i}(\cdot)$ are the marginal densities of $t_i$ and $p_{s_i}(\cdot)$ are the densities of $s_i$ when the variance is set to unity.

As noted in their paper, the data likelihood is intractable to compute because of the integral over possible states of $\mathbf{t}$. This prompts the authors to derive an approach that makes various simplifications and approximations to give a lower bound on the likelihood.

Firstly, they restrict the form of the base density for $s$ to be gaussian[8] and $\phi(\cdot)$ is taken to be $(\cdot)^{-\frac{1}{2}}$. This yields the following expression for the marginal density of $\mathbf{s}$,

$$p(\mathbf{s}) = \int \frac{1}{(2\pi)^{\frac{d}{2}}} \exp\left( -\frac{1}{2} \sum_k t_k \left[ \sum_i h(i,k) s_i^2 \right] \right) \prod_i p_{t_i}(t_i) \sqrt{\mathbf{H}_i^T \mathbf{t}} d\mathbf{t} \qquad (6.26)$$

This expression is then simplified by the approximation,

$$\sqrt{\mathbf{H}_i^T \mathbf{t}} \approx \sqrt{h(i,i) t_k} \qquad (6.27)$$

---

[8]Their model can therefore be considered as type of GSM, although the authors do not comment on this.

Whilst this approximation may not always be a good one, it is a strict lower bound on the true quantity and thus allows for a lower bound on the likelihood as well. Their final approximate likelihood objective is then given by,

$$\widetilde{\mathcal{L}(\mathbf{B})} = \sum_{data} \left( \sum_{j=1}^{d} G\left( \sum_{i=1}^{d} h(i,j)(\mathbf{B}_i^T \mathbf{x})^2 \right) + \log|\det(\mathbf{B})| \right) \qquad (6.28)$$

where the form of the scalar function $G$ is given by,

$$G(\tau) = \log \int \frac{1}{\sqrt{2\pi}} \exp\left( \frac{1}{2} t\tau \right) p_t(t) \sqrt{h(i,i)} dt \qquad (6.29)$$

The results obtained by these authors [Hyvarinen and Hoyer, 2001, Hyvarinen et al., 2001] are very similar to those presented here in section 6.7, and these authors also noted the similarity between elements of their model and the response properties of simple and complex cells in V1.

Interestingly, the optimisation problem that they *actually* solve (maximisation of the equation 6.28), rather than the one they originally propose, can be mapped directly onto the optimisation problem for a square, topographic PoT model if we take: $\mathbf{B} \equiv \mathbf{J}^{\mathrm{PoT}}$, $h(i,j) = \mathbf{W}_{ij}^{\mathrm{PoT}}$ and $G(\tau) = \log(1 + \frac{1}{2}\tau)$. More generally, we can construct an equivalent, square energy-based model whose likelihood optimisation corresponds exactly to the optimisation of their 'approximate' objective function. In this sense, we feel that our perspective has some advantages. Firstly, in that we have a more accurate picture of what model we are *actually* (trying) to optimise. Secondly, in that we are able to move more easily to overcomplete representations. If Hyvarinen et al. were to make their model overcomplete there would no longer be a deterministic relationship between their sources $\mathbf{s}$ and $\mathbf{x}$ — this additional complication makes the already difficult problems of inference and learning significantly harder. As with standard ICA, however, when we extend our PoT to an overcomplete setting there is no simple interpretation as a causal graphical model.

## 6.11  Overcomplete Population Coding: EBM's vs Causal Models

As well as specifying different probabilistic models, overcomplete energy-based models such as the PoT differ from overcomplete causal models in the types of representation they (implicitly) entail. This has interesting consequences when we consider the 'population codes' suggested by the two types of model. We focus on the representation in the first layer ('simple cells'), although similar arguments might be made as we proceed hierarchically.

In an overcomplete causal model, many configurations of the sources are com-patible with a configuration of the input.[9] For a given input, a posterior distribution is induced over the sources in which the inferred values for different sources are conditionally dependent. As a result, although linear in the generative direction, the formation of representation in overcomplete causal models is essentially *non-linear* and moreover it is *non-local* due to the lack of conditional independence. A corollary of this is that although we can specify the basis functions associated with a unit, it is harder to specify any kind of feed-forward receptive field. The issue of how such a posterior distribution could be encoded in a representation remains open; a common postulate (made on the grounds of efficient coding) is that a maximum a posteriori (MAP) representation should be used.

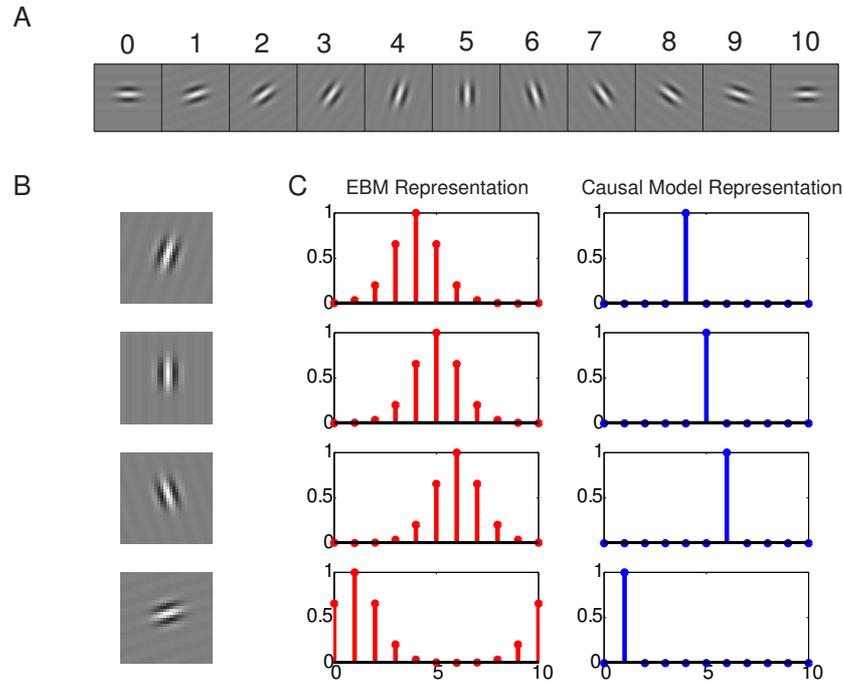

**Figure 6.27:** Representational differences between overcomplete causal models and overcomplete deterministic EBM's. (A) The 11 image vectors in this panel should be considered illustrative of the vectors associated with a subset of representational units in either an overcomplete EBM or an overcomplete causal model. In the EBM they would be the feed-forward filter vectors; in the causal model they would be basis functions. (B) Probe stimuli — these are the same vectors as those associated with units 4,5,6, & 2. (C) The left-hand column shows the normalised responses in an EBM model of the 11 units assuming they are filters. The right-hand column shows the normalised response from the units assuming that they are basis functions in a causal model, and that we have formed a representation by taking the MAP configuration for the source units.

Conversely, in overcomplete EBM's with deterministic hidden units such as we have presented in this chapter, the mapping from inputs to representations remains simple and requires only local information.

In figure 6.27 we try to schematically illustrate a salient consequence of this difference between EBM's and causal models. Panel (A) shows an array of image vectors which should be understood to be either a subset of the basis functions in an overcomplete causal model, or a subset of the filters in overcomplete PoT model. In panel (B) we illustrate the way in which four example inputs are represented; these inputs have been chosen to be four of the vectors shown in panel (A). The left-hand column of panel (C) shows the responses of the units in an EBM-style model, whilst the right-hand column shows the MAP responses from an overcomplete causal model.

---

[9]In fact, strictly speaking there is a subspace of of source configurations of the same dimensionality as the input space.

This is admittedly an extreme case, but it provides an good illustration of the point we wish to make. More generally, although representations in an overcomplete PoT are sparse there is also some redundancy; the PoT representation generally has lower population sparseness [Willmore and Tolhurst, 2001] than an 'equivalent' causal model.

Interpreting the two models as a description of neural coding, one might expect the EBM representation to be more robust to the influences of neural noise as compared with the representation suggested from a causal approach. Furthermore, the EBM style representation is shiftable — it has the property that for small changes in the input there are small changes in the representation. This property would not necessarily hold for a highly overcomplete causal model. Such discontinuous representation might make subsequent computations difficult and non-robust, and it also seems somewhat at odds with the neurobiological data — however proper comparison is difficult since there is no real account of dynamic stimuli or spiking in either model. At present, it remains unclear which type of model — causal or energy based — provides the more appropriate description of coding in the visual system, especially since there are many aspects that neither approach captures.

## 6.12    Discussion

**Whither PoT?**

The choice of the Student-t as an underlying component for our model was largely due to the fact that we could devise an efficient Gibbs sampler to assist learning. Mathematical conveniences aside, we feel that many other sparse, heavy tailed distributions (such as generalised exponential distributions, corresponding to energy functions of the form $f_i(s_i) \sim |s/\sigma|^\beta$) would have been equally appropriate choices. Preliminary studies using other such functions suggest that the results would have be quite similar.

**Future Directions**

There are a number of possible directions in which models presented in this chapter might fruitfully be extended, but there are also several problematic issues that need to be addressed.

Some augmentations that one might consider include the following: (i) adding a temporal aspect to allow the representation of spatio-temporal stimuli; (ii) allowing for a (small) fixed number of iterative lateral interactions; (iii) allowing for deeper hierarchies — with design aimed at capturing properties of area V2; (iv) broader exploration of different transfer functions and energy functions; (v) further analysis of population coding and representational properties.

To deal with temporal stimuli and recurrent interactions, some form of back-propagation through time (BPTT) could be employed [Rumelhart et al., 1986, Werbos, 1990] (and this is why we stipulate a fixed number of iterative interactions.) Such modifications and the extensions to deeper hierarchies would likely prevent us from using a rapidly mixing Gibbs sampler during learning, however a Hybrid Monte Carlo sampling scheme would still be feasible. But a consequence of this, and as a consequence of the general increase in computational complexity, learning would be slowed down and the size of the systems that could be considered might be reduced.

Another issue regarding extension to deeper hierarchies is the choice of transfer functions to generate features, and energy functions based on these features. These functional forms will affect the model and results, but it is difficult to get a good sense of which forms are most appropriate — especially since the EBM framework is flexible enough to cope with a very broad range of possibilities. A potentially interesting option would be to somehow explicitly incorporate divisive normalisation into the density model. We might, for instance, consider taking linear combinations of outputs from the first layer that have been divisively normalised, as in section 6.9, to construct a new layer of features. Finally, we note that even if models can be learned, it can be hard to interpret the representation roles that units play in a deterministic EBM with deep, recurrent interactions; designing better ways of performing such characterisation (both for real and synthetic systems) remains an active area of consideration.

## Summary

We have presented a hierarchical energy-based density model for natural scene images that is able to provide an interpretational account for many aspects of receptive field and topographic map structure within primary cortex, and which also develops sensible high-dimensional population codes.

Deterministic hidden variables or 'features' within our model play a key role in defining the density of a given image patch, and we are able to make a close relationship between these features and the responses of simple cells and complex cells in V1. Furthermore, by constraining our model to interact locally within its 'cortical representation' we are able to provide some computational motivation for the forms of maps for retinotopy, phase, spatial frequency and orientation. Also, preliminary results suggest that stereo response properties and the maps for ocular dominance might be similarly explained. We are not aware of any other high-dimensional, statistically-motivated models that have been used to give accounts for this range of topographic maps along with simple and complex cell responses in an overcomplete setting.

# Acknowledgements

The auxiliary Gibbs sampler was developed in collaboration with Max Welling, and much of the work in this chapter benefited greatly from interactions with him, Geoff Hinton and Yee Whye Teh. The Gibbs sampling algorithm, and some results on topographic maps also appears in Welling et al. [2002] co-authored with Max Welling and Geoff Hinton.

# Chapter 7

# Applications of EBM's

## 7.1 Introduction

In this section we explore the performance of our energy-based framework on two simple applications. As well as helping us to gauge the quality of our modelling assumptions and methods, these experiments are also interesting in their own right.

We begin by applying the PoT model for natural images as a prior in the task of statistical de-noising. If our model is truly doing a good job of capturing the statistical structure of natural images then we ought to be able to use a trained model to help us infer the most likely noise free image given one that has been corrupted by noise.

In a second application, we use the PoT to perform unsupervised feature extraction of sparse feature sets from the NIST database of handwritten digits and the FERET database of human faces.

**Original Contributions**

The main original contributions in this chapter are: (i) the derivation and evaluation of a denoising algorithm based upon the PoT model for statistical structure in natural image patches; and (ii) the exposition of the PoT model and contrastive divergence algorithm as a tool for performing feature extraction from two real world data sets (digits and human faces.)

## 7.2 Denoising Natural Images

As discussed in Chapter 4, the intractability of the partition function makes it very difficult to assess the quality of an energy-based model since we are unable to compute likelihoods.[1] In this section we use the performance on a denoising task as a proxy for assessing how well various models have managed to capture the statistical structure of natural images.

---

[1] We can, however, compute likelihood ratios or equivalently log likelihoods up to an unknown constant — the log partition function. These quantities may be useful for density based discrimination or classification tasks, although we do not pursue that here

We will take a maximum likelihood approach to denoising under additive Gaussian noise of known variance, and will use our learnt model as a prior to infer the most likely noise free image given a corrupted one. We will also show that following this procedure using the PoT can yield a denoising process that is rather like an iteratively adapting Wiener filter.

### 7.2.1   MAP Denoising

We will consider images that have been corrupted by additive Gaussian white noise. That is noisy images given by,

$$\Upsilon_{\text{noise}} = \Upsilon_{\text{true}} + \mathbf{n} \tag{7.1}$$

where $\mathbf{n} \sim \mathcal{N}_{\mathbf{n}}(0, \Sigma)$ is a noise vector. Given a noise free image, $\Upsilon_{\text{true}}$, the distribution over possible noisy images, $\Upsilon_{\text{noise}}$, is a Gaussian

$$p(\Upsilon_{\text{noise}}|\Upsilon_{\text{true}}) = \mathcal{N}(\Upsilon_{\text{true}}, \Sigma) \tag{7.2}$$

Given a corrupted image and a known noise distribution, we can simply apply Bayes theorem to give us the posterior distribution for the true noise free image.

$$p(\Upsilon_{\text{true}}|\Upsilon_{\text{noise}}) \propto p(\Upsilon_{\text{noise}}|\Upsilon_{\text{true}})p(\Upsilon_{\text{true}}) \tag{7.3}$$

where $p(\Upsilon_{\text{true}})$ is our prior probability for the distribution of images. By maximising the expression in equation 7.3 we can infer the true image most likely to have generated the observed noisy image.[2]

### 7.2.2   Iterative Wiener Filtering

We propose to perform denoising using a trained PoT density model as the prior over images. The MAP estimate, $\widehat{\Upsilon}_{\text{true}}$, of the noise free image, $\Upsilon_{\text{true}}$, is then given by maximising the (log) probability in equation 7.3. For the PoT model this reduces

---

[2]Note that for real world applications we may wish to use a different noise model (perhaps poisson noise on each pixel) and further we may wish not to simply maximise the posterior probability of our de-noised image, $\widehat{\Upsilon}_{\text{true}}$. Rather we may wish to set up a loss function $R(\Upsilon_{\text{true}}, \widehat{\Upsilon}_{\text{true}})$ which measures the 'cost' of the true image being $\Upsilon_{\text{true}}$ and us assuming it was $\widehat{\Upsilon}_{\text{true}}$. In this case we choose $\widehat{\Upsilon}_{\text{true}}$ to minimise the expected loss over the posterior distribution. Obviously, however, such increased sophistication comes at increased computational cost.

to,

$$\widehat{\Upsilon}_{\text{true}} = \arg\min_{\Upsilon_{\text{true}}} \left[ \frac{1}{2} (\Upsilon_{\text{true}} - \Upsilon_{\text{noise}})^T \Sigma^{-1} (\Upsilon_{\text{true}} - \Upsilon_{\text{noise}}) \right.$$

$$\left. + \sum_i \alpha_i \log \left( 1 + \frac{1}{2} \sum W_{ij} (\mathbf{J}_j^T \Upsilon_{\text{true}})^{.2} \right) \right] \qquad (7.4)$$

where $\mathbf{J}$, $\mathbf{W}$, and $\alpha_i$ are parameters of a PoT model that have been learnt from data and $\Sigma$ is the (known) covariance of the image noise on $\Upsilon_{\text{noise}}$.

This function could be minimised by gradient descent, however we propose a more efficient, sequential variational procedure in which we upper bound the logarithm terms using the following general inequality for the log function,

$$\log \xi \leq \gamma - \xi \log \gamma - 1 \qquad \forall \xi, \gamma \geq 0 \qquad (7.5)$$

This bound is saturated when $\gamma = 1/\xi$. Applying this to every logarithm in the summation of equation 7.4 and iteratively minimising this bound over $\xi$, (which is given by $\xi_i = \left(1 + \frac{1}{2} \sum W_{ij} (\mathbf{J}_j^T \Upsilon_{\text{true}})^{.2}\right)$), and $\gamma_i$ we obtain the following pair of recursive update equations, which we run until convergence[3],

$$1/\gamma_i \leftarrow 1 + \frac{1}{2} \sum_j W_{ij} (\mathbf{J}_j^T \widehat{\Upsilon}_{\text{true}})^{.2} \qquad (7.6)$$

$$\widehat{\Upsilon}_{\text{true}} \leftarrow \left( \Sigma^{-1} + \mathbf{J}^T \mathbf{D} \mathbf{J} \right)^{-1} \Sigma^{-1} \Upsilon_{\text{noise}} \qquad (7.7)$$

where

$$\mathbf{D} = \text{diag}[\mathbf{W}^T (\alpha \odot \gamma)] \qquad (7.8)$$

with $\odot$ denoting element wise multiplication. Equation 7.7 has the form of a Wiener filter under noise covariance $\Sigma$ and a Gaussian image prior with covariance $(\mathbf{J}^T \mathbf{D} \mathbf{J})^{-1}$, consequently we have named the method 'Iterative Wiener Filtering.' At each iteration of equations 7.6 and 7.7 we are applying a Wiener filter and then using the results to re-estimate the filter settings.

Our method is preferable to simple gradient descent since it is computationally more efficient and allows us to make larger changes in the estimate at each step, and also does not require us to set a step size. However, note that the function that we seek to minimise is non-convex in certain parameter/noise regimes, and consequently we are not guaranteed to to reach a global optimum. Nevertheless, we hope to find good solutions regardless.

As an aside, we mention that when the filters are orthonormal, the noise covariance isotropic and the weight matrix $\mathbf{w}$ is the identity, the minimisation in 7.4 decouples into $M$ separate minimisations over the transformed variables $\lambda_i = \mathbf{J}_i^T \widehat{\Upsilon}_{\text{true}}$.

---

[3]To within some small tolerance.

Defining $\nu_i = \mathbf{J}_i^T \Upsilon_{\text{noise}}$ we can derive that $\lambda_i$ is the solution of a cubic equation,

$$\lambda_i^3 - \nu_i \lambda_i^2 + 2\left(1 + \Sigma_{ii}\alpha_i\right)\lambda_i - 2\nu = 0 \tag{7.9}$$

However, the constraint of orthonormality is a rather severe restriction if the data is not pre-whitened but, on the other hand, if we decide to work with whitened data then isotropic noise assumption seems unrealistic. Nevertheless, several authors (for example Hyvarinen [1999]) apply a 'shrinkage' technique that is based on precisely these assumptions, and they seem to obtain reasonable results.

Finally, we mention that our approach is also related to denoising approaches based upon using GSM's and wavelet decompositions, such as Wainwright et al. [2000a].

### 7.2.3   Experiments & Results

We tested the performance of the iterated Wiener filters using parameters extracted from several 'flavours' of PoT model. For comparison, we also apply a standard Wiener filter (`wiener2.m` from Matlab[4]). Our models were trained using contrastive divergence, as described in Chapter 6.[5]

Figure 7.1 (A) and (B) shows an example of a noise free image and the corresponding noisy image (obtained by adding isotropic Gaussian noise to each pixel). Since the filters that we learn only apply to small patches, we must use a windowed approach when denoising a full image. We have not taken measures to optimise the speed of this process; it might be possible to greatly improve efficiency by using some sort of fast convolution/FFT based algorithm. However, at present we take a very naive approach and cut the image up into blocks of the appropriate patch size and perform denoising on each patch separately. The denoised blocks are then recombined to yield a final estimate for the noise free image.

Figure 7.1 (C) demonstrates the effects of our denoising procedure alongside results from the optimal[6] `wiener2.m` Wiener filter in figure 7.1 (D). The peak signal-to-noise ratios (PSNR's) for each image are shown alongside each panel. Our method clearly outperforms simple Wiener filtering although admittedly it does take considerably longer to run in the present implementation.

Table 7.1 summarises the performance of several different flavours of PoT on a selection of 11 full images of size $768 \times 512$.

We see that the IWF approach consistently outperforms the basic Wiener filter approach (this is true on individual images as well as the averages shown in the table), and that this difference becomes particularly pronounced at higher noise levels.

---

[4]http://www.mathworks.com/access/helpdesk/help/toolbox/images/wiener2.html

[5]The images used for testing were the same as those from which the training patches were extracted, however we do not expect this to make a significant difference to our results.

[6]We used multiple neighbourhood sizes and always chose the best one.

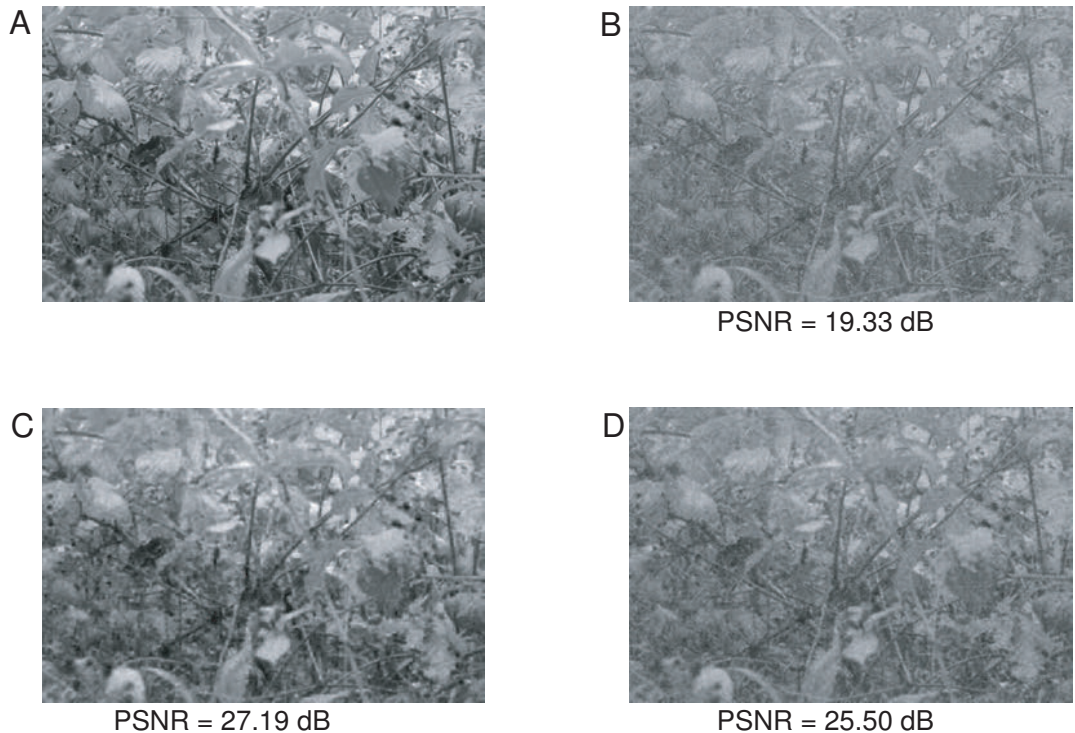

**Figure 7.1:** (A) Original image. (B) Image after being corrupted with white Gaussian noise. (C) Denoised image produce by a IWF using a trained hierarchical PoT model. (D) Denoised image produced by 'optimal' Matlab Wiener filter. The effects are somewhat difficult to see in this print reproduction, but the IWF generally seems to suffer from less blurring and has retained more of the 'sharp' edges structure.

| METHOD | CONDITION (PSNR IN NOISY IMAGE) | | | |
|---|---|---|---|---|
| | Noise level 1 | Noise level 2 | Noise level 3 | Noise level 4 |
| | 18.0 dB | 12 dB | 8.5 dB | 6 dB |
| Wiener2.m | $24.93 \pm 2.21$ | $21.58 \pm 1.98$ | $19.71 \pm 1.68$ | $17.13 \pm 1.09$ |
| S–PoT | $\mathbf{25.04 \pm 2.21}$ | $22.29 \pm 2.20$ | $20.85 \pm 2.00$ | $18.95 \pm 1.64$ |
| HS–PoT | $24.85 \pm 2.75$ | $\mathbf{22.43 \pm 2.39}$ | $\mathbf{21.16 \pm 2.15}$ | $19.16 \pm 1.75$ |
| OC–PoT | $23.55 \pm 2.75$ | $21.43 \pm 2.36$ | $20.56 \pm 2.13$ | $19.24 \pm 1.77$ |
| HOC–PoT | $24.98 \pm 3.81$ | $21.72 \pm 2.38$ | $20.86 \pm 2.13$ | $\mathbf{19.27 \pm 1.77}$ |

**Table 7.1:** Table showing denoising results. Each row shows the results for a different method and/or parameter set. S–PoT: Square PoT. HS–PoT: Hierarchical Square PoT, the first level filters were fixed to be the same as S–PoT and the second layer weights were subsequently trained. OC–PoT: Overcomplete (1.96×) PoT. HOC–PoT: Hierarchical and overcomplete PoT, the first level filters were fixed to be the same as OC–PoT and the second layer weights were subsequently trained. The entries in the different columns give the average PSNR (Peak Signal to Noise Ratio) and it's standard deviation over a set of 11 images. The column headers show mean signal to noise ratio present in the corrupted images. PSNR is defined as $20 \log_{10} \frac{S_{\mathrm{MAX}}}{\mathrm{MSE}}$, where $S_{\mathrm{MAX}}$ is the maximum signal strength and MSE is the mean square error between a probe image and the true image.

Amongst the different PoT models we see only a moderate performance gain from having a hierarchical model or being overcomplete; the more significant gains again being in the high noise regime. Whilst these results are encouraging in that they suggest that the models *are* capturing statistical regularities, they are somewhat disappointing. One possible explanation for the relatively poor performance boost seen in more complicated models could be the procedure used to implement the IWF. As discussed earlier, the posterior may not be convex and so convergence to local, rather than global, optima is a potential problem. This hypothesis is somewhat bourn out by the fact that the best *relative* performance occurs at high noise levels, at which we expect the posterior to be more usually convex. It remains to be seen whether an efficient procedure to find global optima in overcomplete and/or hierarchical models can be developed, and we leave this as an open issue.

## 7.3   Feature Extraction

We now consider the general problem of unsupervised feature extraction. In addition to describing the probability density of a dataset, models such as ours can often yield useful and interpretable 'features' of the data. We saw in Chapter 6 that we can relate these features in a PoT model trained on natural scenes to simple cell and complex cell receptive fields in V1. However, being able to extract such features may have other uses, particularly since they are well motivated by the statistical foundations of the data they are trained upon. For instance, one might explore their use in classification or data visualisation tasks.

In this section we present preliminary results using a simple PoT model to perform unsupervised feature extraction on a database of faces and on a database of images. As mentioned in Chapter 6 we expect the PoT energy function to deliver features that are sparse and that might be interpreted as frequently-satisfied soft-constraints.

### 7.3.1   Features of Digits

We used the digit set of size $16 \times 16$ real values from the "br" set on the CEDAR cdrom #1. There are 11000 digits available, divided equally into 10 classes. The mean image from the entire dataset was subtracted from each datum, and the digits were whitened with ZCA. (This is PCA whitening, followed by a rotation of the axes to align back with the image space.) An overcomplete, single layer PoT with 361 features was trained using the contrastive divergence.

The entire set of learned filters is shown in figure 7.2. We note the superficial similarity between these filters and those found from the natural scene experiments. However, in addition to straight 'edge-like' filters we also see several curved filters. We might interpret the results as a set of 'stroke' detectors, modelling a space of strokes that gives rise to the full digit set.

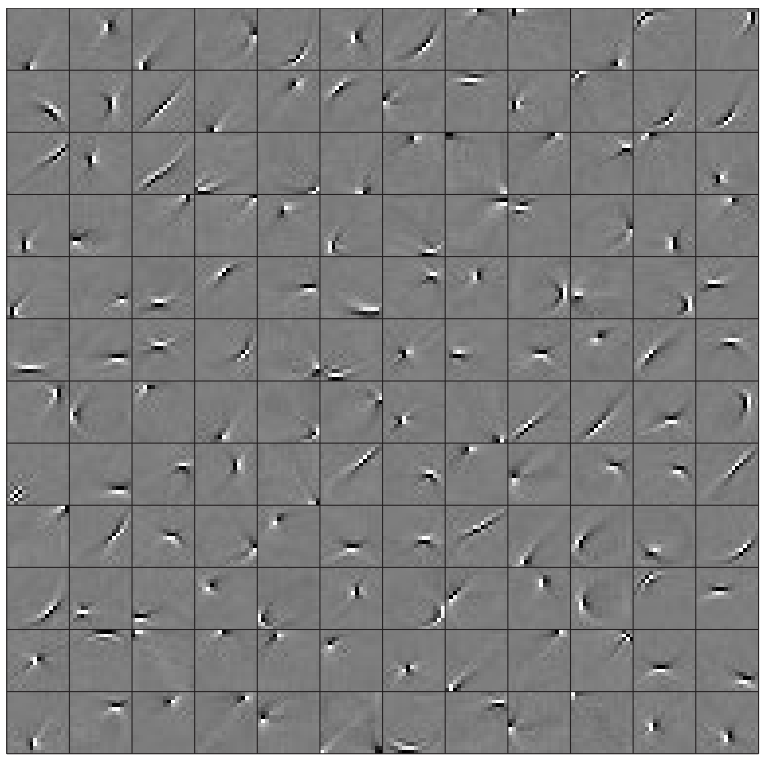

**Figure 7.2:** Random sample of 144 out of 361 learned filters for CEDAR digits. The gray scale saturates in each cell. Also, to make it easier to discern the structure of the learnt filters, we present them in the ZCA whitened domain rather than in raw pixel space.

We also looked at the features learned by a hierarchical model, and we show a simple characterisation of some examples in figure 7.3. This figure shows the dominant 36 first-layer filters feeding into a given top layer unit, along with a collection of the most and least excitatory input patterns for that unit. The most excitatory patterns are somewhat uninformative since most are zeros or ones — due to the way in which the original handwriting has been digitised and normalised, these digits simply tend to have much more 'ink'. Nevertheless we do see some examples of structure here. The least excitatory patterns are perhaps more interesting to consider since these do seem to have captured richer structure in the 'classes' of digits. If we take a soft-constraint-based perspective of the PoT energy function then these patterns illustrate inputs that satisfy the constraint specified by a particular unit very well. This nicely reminds us that a population representation can be informative by what is *not* actively signalled, as well as by what is.

## 7.3.2   Features of Faces

We used the full NIST FERET database of frontal face images[7]. The data was first pre-processed in the standard manner to align the faces, normalise the pixel inten-

---

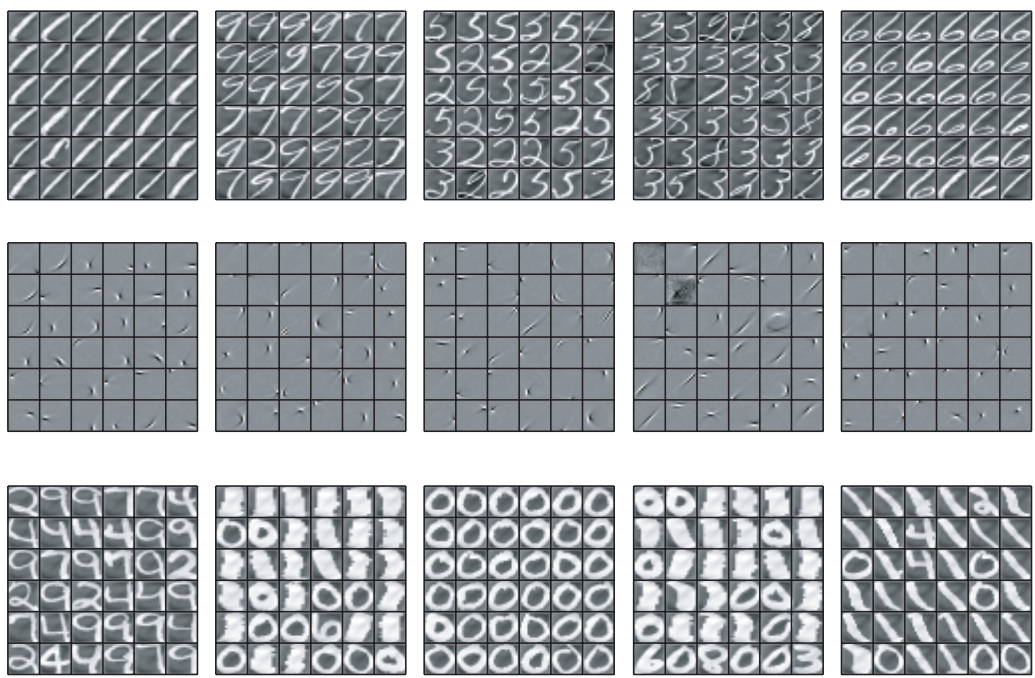

**Figure 7.3:** Examples of hierarchical features learnt from digits. Each row represents a separate top level unit. The top row shows the 36 least excitatory stimuli for that unit (out of 10,000), bottom row shows a collection of the *number* most excitatory stimuli for that unit. The middle row shows the dominant first layer filters that feed into the unit. The rankings descend columnwise in each square, starting at the top left.

sities and crop a central oval shaped region[8]. Then as an additional pre-processing step we centered the data and performed (substantial) PCA whitening, retaining the projections onto the leading 256 out of 17154 eigenvectors as the input dimensions to algorithm. (Due to the redundancy in the structure of human appearance and the standardisation of the image alignments, these 256 components actually capture almost all the variance of the full dataset.) We then trained a single layer PoT with 361 features, and both $\alpha$ as well as $\mathbf{J}$ were unconstrained and free to learn.

Figure 7.4 (A) shows the 64 leading eigenvalues plotted as face images, and figures 7.4 (B) and (C) show a subset of 64 filters as learnt by our algorithm. In Bartlett et al. [2002] two kinds of complete ICA were applied to a lower resolution version of the FERET database. In relation to their work, the filters that we learn are closer to 'type II' ICA (each face is an input) rather than 'type I' ICA (the value of a pixel across all faces is an input) but there are notable differences. As with their 'type II' results, many of the filters that we learn are somewhat global in the sense that most pixels having a non-zero weight. However, in addition to these global features and in contradistinction to their results, we also develop features with most of their weight concentrated in localised sub-regions – for instance focussing on glasses, eyes, smiling mouths, moustaches, etc. Furthermore, as well as global features that can perhaps be described as 'archetypical faces' we also see global features which appear to mainly capture structure in the illumination of a face.

---

[8]http://www.cs.coloradostate.edu/evalfacerec/index.htm

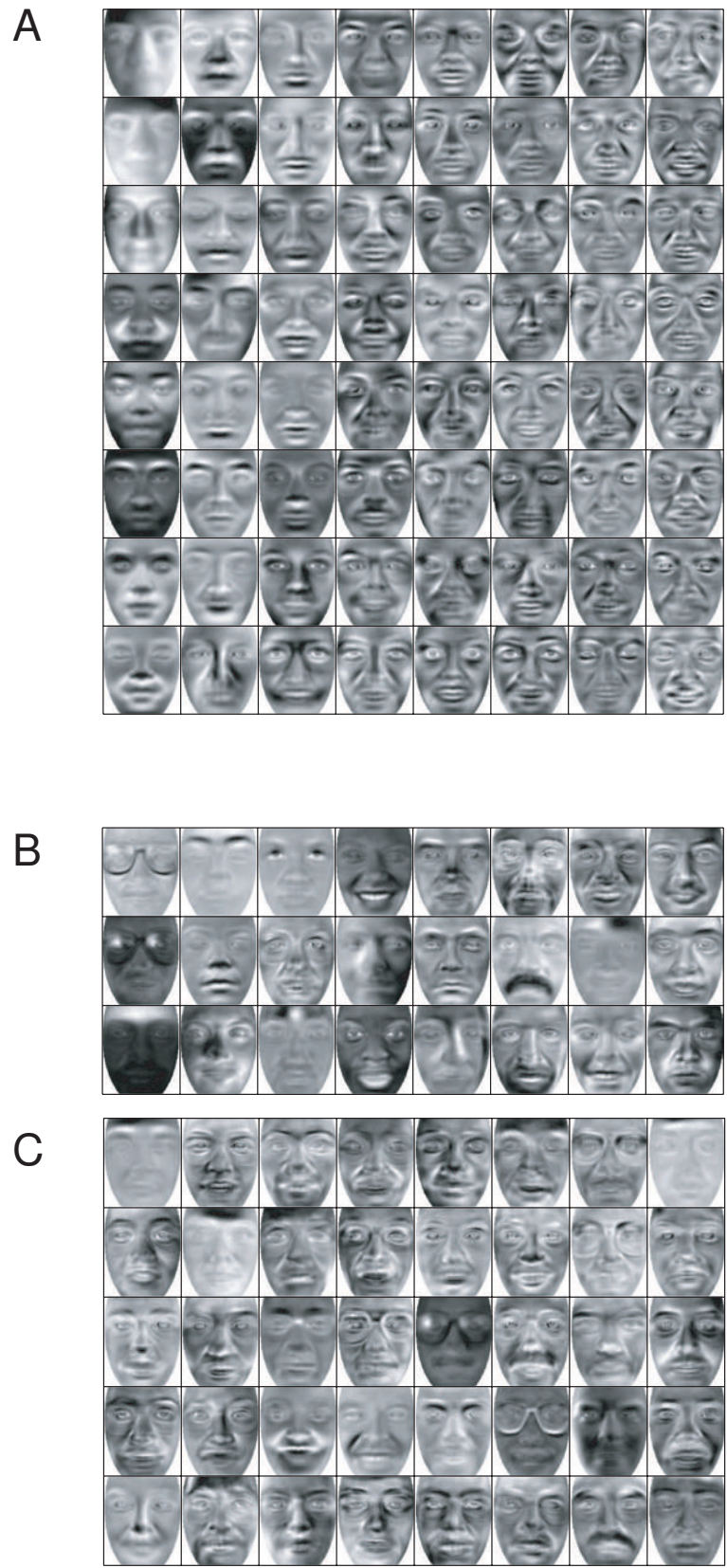

**Figure 7.4:** (A) 64 eigenfaces with largest eigenvalues plotted columnwise in descending eigenvalue order. (B) Handpicked subset of 24 and (C) Random subset of 24 feature vectors from set of 361 learnt using our algorithm.

Such features could perhaps be fed into a classifier, as is done in Bartlett et al. [2002] with features obtained from PCA and ICA respectively, although we have not explored that here. Future work may show whether the representations that this model, or hierarchical extensions/generalisations thereof, can be usefully employed in a face or expression recognition system.

## 7.4  Summary

We have shown successful performance on a range of practical tasks, thereby highlighting the competence and utility of our approach.

We have demonstrated that the PoT model, trained using a contrastive divergence algorithm, is able to produce statistical models that do reflect the structure in natural images. Our method outperforms basic Wiener filtering and at higher noise levels our advantage ($\sim$ 2 dB) is comparable to other advanced methods tailored for natural images [Hyvarinen, 1999, Wainwright et al., 2000a]. Hierarchical and/or overcomplete models seem to give noticeable, but small, improvements on this task over the basic approach. This is somewhat surprising and a little disappointing, however it is possible that the implementation of the denoising procedure, rather than the models themselves, is to blame.

In a second task we have shown effective and intelligible feature extraction from two real world data sets, and have also demonstrated hierarchical feature extraction for digits.

We feel that these are promising preliminary results which could be improved upon with further work.

## 7.5  Acknowledgements

The development of the Iterated Wiener filter was made in collaboration with Max Welling and is presented in Welling et al. [2002]. Our results on feature extraction have also appeared in Teh et al. [2003]. We are grateful to Hans van Hateren for making his image data publicly available.

# Chapter 8

# Conclusion

## 8.1 Introduction

In this thesis we set out: (i) to develop and explore models and algorithms for energy-based approaches to unsupervised machine-learning and density estimation; and (ii) to explore the potential utility of these models as theoretical tools to to help improve our understanding of receptive field development, topographic map development and coding in sensory systems.

The main contributions of this thesis are:

- In chapters 2 and 3 we reviewed and discussed a broad selection of literature, covering early developments through to state-of-the-art results, in relevant topics from neurobiology, theoretical neuroscience and machine learning/computer science.

- In chapter 4 we presented the theoretical development for a flexible density-estimation framework, along with algorithms and methods that render learning in such frameworks tractable. This framework is then used as a basis for modelling in subsequent chapters, but should have broader applicability to a range of other domains.

- In chapter 5 we presented experimental simulations using a Boltzmann machine. We first provided approximate algorithms for learning, and then used these algorithms to learn the probability density of simple, naturalistic inputs.

  The trained models exhibited components that could be related to aspects of development and structure in visual cortex. In particular, our model gave a good qualitative reproduction of retinotopic refinement in receptive fields, along with topographic maps for retinotopy and ocular dominance. Our model also demonstrated sparse, distributed non-linear population coding that could represent inputs with good fidelity. Also, in a distinct advance on previous models, we demonstrated a limited ability to learn computationally appropriate *lateral* connections as well as feed-forward connections.

- In chapter 6 we presented a different model framework that builds upon some of the ideas in chapter 5, overcoming some of the difficulties that we encountered there. Our product of t's (PoT) model uses the output of deterministic feedforward filters to compute energies which can then be used to specify a distribution over real-valued image vectors.

  By developing extensions to this basic set-up we were able to provide a neurobiologically revealing hierarchical density model that was able to capture higher-order statistics of digitised natural visual scenes. In particular, our model was able to give an account for the basic response properties of simple cells and complex cells in V1, as well as for some of the structure seen in the feature preference maps for spatial location, spatial phase, spatial frequency, orientation and (partially) ocular dominance.

  Other very recent approaches have delivered somewhat similar results; however we believe ours to be the first to incorporate (pseudo)-stereo inputs and, more importantly, to deal with overcomplete representations. We discussed in some detail how our framework can be formally related to some of these other approaches, and also highlighted some of the important ways in which our model differs in terms of the representations formed.

- Finally, in chapter 7 we briefly explored some practical applications of the image models developed in chapter 6. In the first application we demonstrated the utility of the learnt density model for natural scenes as a prior for Bayesian image denoising. In the second application we demonstrated the capability of the PoT model to perform sparse feature extraction from two different image databases.

Despite these successes, there are various important outstanding issues that lead to suggestions and ideas for extensions and future work.

## 8.2   Outstanding Issues

The outstanding issues can be separated into two categories: those relating to technical aspects of the models and algorithms; and those relating to the use of our modelling framework as an interpretational tool for understanding visual coding and development.

### 8.2.1   Machine Learning

A key component in all of our work has been the contrastive divergence algorithm [Hinton, 2000, 2002]. As discussed in chapter 4 there are several aspects of this learning procedure that are somewhat difficult to control. There are no guarantees that it will converge (although it always seems to), but moreover the conditions under which it converges to good maximum likelihood solutions (which is generally the goal) are still somewhat poorly understood. It is clear, however, that these conditions depend on the nature of Markov chain used to provide negative phase samples and the way in which this Markov chain interacts with a given model. It would be desirable to have a better understanding of what conditions favour proper convergence.

On a somewhat related note it is hard to evaluate the quality of a model quantitatively once it has been learned. Since it is intractable to calculate the partition function, using off-sample likelihood seems infeasible. However, it might be possible to find ways of bounding the partition function and thereby help to obtain an indicator of performance.

Having a better handle on the partition function would also help with elevating learning from maximum likelihood to a fully Bayesian approach. Doing so at present is not feasible, since the posterior probabilities for different models require global partition functions in their calculation. Of course, for our modelling purposes, maximum likelihood actually seems more appropriate since we wish to compare a particular set of learnt parameters to a neurobiological system.

### 8.2.2 Theoretical Neuroscience

Though we have successfully reproduced a broad range of the developmental and response properties in V1 from computationally well motivated considerations , our approach suffers from the same problems as its many fellow travellers.

From a Marrian perspective, we have concentrated on the computational and algorithmic levels, and recognise that there is much work to do regarding implementational aspects. The choice of implementational details that we have honored, such as local learning rules, could be considered a little arbitrary. Also some key issues, such as the timing of development compared with the timing of different input statistics, are unclear. Nevertheless, there is much to understand at our computational level, being informed, though not rigidly constrained, by implementational details.

Similar issues concern our stimuli. Whilst small patches of digitised natural scenes seem to capture much that is important about an organisms sensory input they are lacking in two key areas. Firstly (and foremost) they do not take into account the temporal nature of the world and it may be that the conclusions drawn from static stimuli will cease being true given time-varying stimuli. However, results such as Olshausen [2001] suggest that many of the results from static analyses do carry over when we consider temporal variations. Secondly, there are issues of global context; as one moves further up the visual processing stream, receptive fields cover larger areas of space and show greater spatial invariance. It will be necessary to entertain considerably larger input vectors if one wishes to retain a high-dimensional approach that captures these aspects.

A similar issue of modelling scale highlights a drawback in our approach to understanding cortical maps. Whilst our results *do* replicate impressive amounts of the structure that has been observed experimentally the maps we produce aren't (as yet) large enough to facilitate some of the comparisons and analyses one would like to make — especially with respect to global structure.

On a somewhat brighter note, it seems that for many of these issues the limiting factor is computational power as opposed to a lack of theoretical inspiration — we might hope for greater success in these areas as Moore's law helps hardware catch up with our modelling desires.

## 8.3    Alternative Directions

We draw this thesis to a close with some brief suggestions for general areas of further work; specific extensions for particular models were given previously in the relevant chapters.

### Conditional Energy-Based Models (CEBM's)

We have concentrated on unsupervised learning of marginal densities and it would be interesting to apply the general, flexible energy-based framework put forward in chapter 4 for what might be termed 'conditional density estimation' or associative learning.

Such a model could be constructed in a fairly straightforward manner using a network whose energy functions depend parametrically both on the conditioning input as well as the target/conditioned output. Learning would proceed in essentially the same way as described in chapter 4, except that one would clamp the conditioning data partition during *both* the positive and negative phases, whilst the target pattern would undergo negative phase sampling as in normal contrastive divergence learning.

Such models have potential application in many different areas, and they seem particularly well suited to tasks like sequence-labelling. For instance, learning structural models for parts-of-speech tagging, or perhaps nucleotide/amino acid sequence annotation for bioinformatics. Indeed, there are several examples of very similar approaches to ours delivering promising results in this area — in particular conditional random fields and maximum entropy Markov models. Compared to these methods, an energy-based approach might hold advantages in allowing for greater flexibility and having greater expressive power.

**Contrastive Divergence in Chain Graphs**

A second possible innovation is to use some of the contrastive divergence techniques and energy-based formalism discussed in this thesis to learn 'chain graphs' [Cowell et al., 1999] — that is probabilistic graphical models that have both causal links as well as acausal (energy-based) ones. Such models have, so far, received relatively little attention from either the machine learning or theoretical neuroscience communities. Interesting in their own right, these models might also be better suited to building interpretive models of neural development and representation than either causal models or energy-based models alone.